\documentclass[9pt,twocolumn,twoside]{pnas-new}

\templatetype{pnasresearcharticle}
\usepackage{amsmath}
\usepackage{amssymb}
\usepackage{bm}
\usepackage{placeins}
\usepackage{graphicx}
\usepackage{listings}
\usepackage{mathrsfs}
\usepackage[numbers]{natbib}
\usepackage{relsize}
\usepackage{subcaption}
\usepackage{url}
\usepackage{multirow}
\usepackage{xcolor}
\makeatletter
\long\def\XR@test#1#2#3#4\XR@{%
  \let\XR@next\@gobbletwo
  \ifx#1\newlabel
    \let\XR@next\@firstoftwo%
  \else\ifx#1\@input
     \let\XR@next\@secondoftwo
  \fi\fi
   \XR@next{\newlabel{\XR@prefix#2}{#3}}{\edef\XR@list{\XR@list#2\relax}}%
  \ifeof\@inputcheck\expandafter\XR@aux
  \else\expandafter\XR@read\fi}
\makeatother

\setboolean{displaywatermark}{false}
\makeatletter
\fancypagestyle{firststyle}{
   \fancyfoot[R]{\footerfont\textbf{\thepage\textendash\pageref{LastContentPage}}}
   \fancyfoot[L]{\footerfont\@ifundefined{@doi}{}{\@doi}}
}

\makeatother
\usepackage{tikz}

\usetikzlibrary{shapes,arrows}
\usetikzlibrary{decorations.pathreplacing, decorations.pathmorphing, snakes, calligraphy}

\tikzstyle{block} = [rectangle, draw, thick, align=center, rounded corners]
\tikzstyle{boundingbox} = [thick, lightgray]
\tikzstyle{dashblock} = [rectangle, draw, thick, align=center, dashed]
\tikzstyle{conc} = [ellipse, draw, thick, dashed, align=center]
\tikzstyle{netnode} = [circle, draw, very thick, inner sep=0pt, minimum size=0.75cm]
\tikzstyle{relunode} = [rectangle, draw, very thick, inner sep=0pt, minimum size=0.5cm]
\tikzstyle{line} = [draw, very thick, -latex']
\tikzstyle{arrow} = [draw, ->, thick]
\tikzstyle{mapsto} = [draw, |->, thick]

\def\checkmark{\tikz\fill[scale=0.4](0,.35) -- (.25,0) -- (1,.7) -- (.25,.15) -- cycle;}
\def\bigcheckmark{\tikz\fill[scale=0.66](0,.35) -- (.25,0) -- (1,.7) -- (.25,.15) -- cycle;}

\definecolor{bpurp}{HTML}{984ea3}
\definecolor{bblue}{HTML}{377eb8}
\definecolor{bgreen}{HTML}{4daf4a}
\definecolor{borange}{HTML}{ff7f00}
\definecolor{bred}{HTML}{a50f15}

\graphicspath{{./figures/}}
\title{Transforming task representations to perform novel tasks}
\author[a,1]{Andrew K. Lampinen}
\author[a]{James L. McClelland}

\affil[a]{Department of Psychology, Stanford University, 450 Jane Stanford Way, Stanford CA 94043}

\leadauthor{Lampinen}

\correspondingauthor{\textsuperscript{1}To whom correspondence should be addressed. E-mail: andrewlampinen@gmail.com}


\begin{abstract} %% TODO: update
An important aspect of intelligence is the ability to adapt to a novel task without any direct experience (zero-shot), based on its relationship to previous tasks. Humans can exhibit this cognitive flexibility. By contrast, models that achieve superhuman performance in specific tasks often fail to adapt to even slight task alterations. To address this, we propose a general computational framework for adapting to novel tasks based on their relationship to prior tasks. We begin by learning vector representations of tasks. To adapt to new tasks, we propose \emph{meta-mappings}, higher-order tasks that transform basic task representations. We demonstrate the effectiveness of this framework across a wide variety of tasks and computational paradigms, ranging from regression to image classification and reinforcement learning. We compare to both human adaptability and language-based approaches to zero-shot learning. Across these domains, meta-mapping is successful, often achieving 80-90\% performance, without any data, on a novel task, even when the new task directly contradicts prior experience. We further show that meta-mapping can not only generalize to new tasks via learned relationships, but can also generalize using novel relationships unseen during training. Finally, using meta-mapping as a starting point can dramatically accelerate later learning on a new task, and reduce learning time and cumulative error substantially. Our results provide insight into a possible computational basis of intelligent adaptability and offer a possible framework for modeling cognitive flexibility and building more flexible artificial intelligence systems. 
\end{abstract}

\dates{This manuscript was compiled on \today}
\begin{document}

\maketitle
\thispagestyle{firststyle}
\ifthenelse{\boolean{shortarticle}}{\ifthenelse{\boolean{singlecolumn}}{\abscontentformatted}{\abscontent}}{}

%\section*{Introduction}

%\dropcap{I}ntelligent systems should be able to transform their behavior on a task, in accordance with a change in goals. Adaptability is a key feature of biological intelligence --- adaptation is necessary for a system to efficiently handle all the vagaries of its environment \citep{Siegelmann2013}. Humans often exhibit substantial adaptability \citep{Lake2016}.
Adaptability is a key feature of biological intelligence --- adaptation is necessary for a system to efficiently handle all the vagaries of its environment \citep{Siegelmann2013}. An advantage of neural networks over ordinary computer programs is that they can adapt by learning from training examples. Yet this is only a limited form of adaptability. An intelligent system should be able to transform its behavior on a task in accordance with a change in goals, and humans often exhibit this form of adaptability \citep{Lake2016}. 
For example, if we are told to try to lose at poker, we can perform quite well on our first try, even if we have always tried to win previously. If we are shown an object, and told to find the same object in a new color or texture, we can do so. By contrast, this type of first-try adaptation is quite difficult for standard deep-learning models \citep{Lake2016,Marcus2018,Russin2020}. How could models reuse their knowledge more flexibly? 

We suggest that this ability to adapt can arise from exploiting the relationship between the adapted version of the task and the original. In this work, we propose a computational model of adaptation based on task relationships and demonstrate its success across a variety of domains, ranging from regression to classification to reinforcement learning. Our approach could provide insights into the flexibility of human cognition and allow for more flexible artificial intelligence systems.   

Our model incorporates several key cognitive insights. First, in order to perform different tasks, it is useful for the system to constrain its behavior by an internal task representation \citep[e.g.][]{Cohen1990}. Prior work in machine learning and cognitive science has constructed task representations from a natural language instruction \citep{Larochelle2008, Hermann2017, Hill2019a}, or by learning to infer task representations from examples, a procedure called meta-learning \citep[e.g.][]{Vinyals2016, Rusu2019}. %We build on these ideas, but extend them, by allowing the model to accomodate task alterations by transforming its representation of a task. 
We extend these ideas, proposing that the model can adapt to a novel task by transforming its representation for a prior task into a representation for the new task, thereby % allowing it to perform the new task.  %% TODO: update?
exploiting the task relationship to perform the new task.

We refer to these transformations of task representations as \textbf{meta-mappings}. That is, meta-mappings are higher-order functions over tasks --- functions that take a task as input and transform it to produce an adapted version of that task. Meta-mappings allow the model to adapt to a new task \emph{zero-shot} (i.e. without requiring any data from that new task), based on the relationship between the new task and prior tasks. %We propose that meta-mapping can offer a useful inductive bias that complements, and in some cases exceeds, other zero-shot learning approaches, by allowing more effective use of knowledge about the prior task.  %% TODO: update
We propose that meta-mapping is a powerful way to promote adaptation, because the task relationships it exploits are the fundamental conceptual structure on which systematic generalization can be predicated.

As a concrete example, our model is able to switch to losing at poker on its first try. To do so, it constructs a representation of poker from experience with trying to win the game. It then infers a ``try-to-lose'' meta-mapping, either from language or from examples of winning and losing at other games, such as blackjack. It then applies this meta-mapping to transform its representation of poker, thereby yielding a representation for losing at poker. This adapted task representation can then be used to perform the task of trying to lose at poker zero-shot --- that is, without any prior experience of losing at poker.

Our main contributions are: 
\vspace{-5pt}
\begin{itemize} \itemsep -4pt
\item To propose meta-mapping as a computational framework for zero-shot adaptation to novel tasks. 
\item To provide a parsimonious architecture for meta-mapping. 
\end{itemize}
\vspace{-5pt}

We demonstrate the success of meta-mapping across a variety of task domains, ranging from visual classification to reinforcement learning, and show that the model can even adapt using new meta-mappings not encountered during training. We further show that adapting by meta-mapping provides a useful starting point for later learning. To our knowledge, this is the first work that proposes transforming a task representation in order to adapt zero-shot. We consider related work and implications for cognitive science and artificial intelligence in the discussion.

\section*{Task transformation via meta-mappings} \label{sec:HoMM_metamappings}

\subsection*{Basic tasks are input-output mappings} We take as a starting point the construal of basic tasks as mappings (functions) from inputs to outputs. For example, poker can be seen as a mapping from hands to bets (Fig. \ref{fig:HoMM_architecture:basic_task}), chess as a mapping of board positions to moves, and object recognition as a mapping from images to labels. This perspective is common in machine learning approaches, which generally try to infer a task mapping from many input/output examples, or meta-learn how to infer it from fewer examples. We use the phrase ``basic task'' to refer to any elementary task a system performs (e.g. any card game), including both standard tasks (``poker'') and variants that can be produced by a transformation (``lose at poker'').
%In our work, we infer these mappings in three different ways: from examples, from natural language instructions, or by transforming a prior task mapping. We view all three approaches as important; the third is the novel component of our work, and therefore the main focus. 

\subsection*{Tasks can be transformed via meta-mappings} 
\begin{figure*}[!htb]
\centering
\resizebox{\textwidth}{!}{%
\begin{tikzpicture}[auto]

%% a) basic task 
\begin{scope}[shift={(-10.45,0)}]
\draw[boundingbox, draw=gray, fill=white] (-2.25, -2.5) rectangle (2.25, 2.5);
\node[gray, text width=2cm, align=center] at (0, 2.1) {Poker task};
\begin{scope}[shift={(0,-0.5)}]
\node[text width=0.5cm] at (-1, -1.3) {\includegraphics[width=0.5cm]{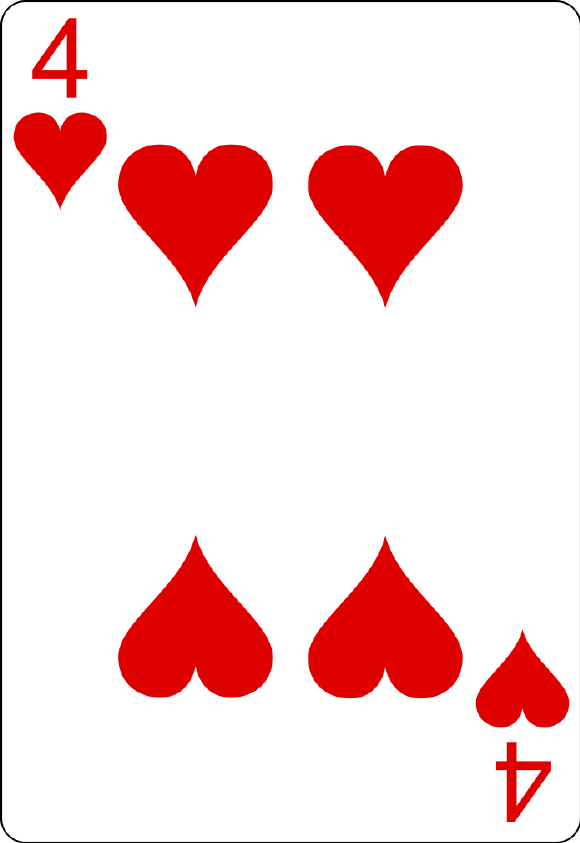}};
\node[text width=0.5cm] at (-0.85, -1.25) (pokerin0) {\includegraphics[width=0.5cm]{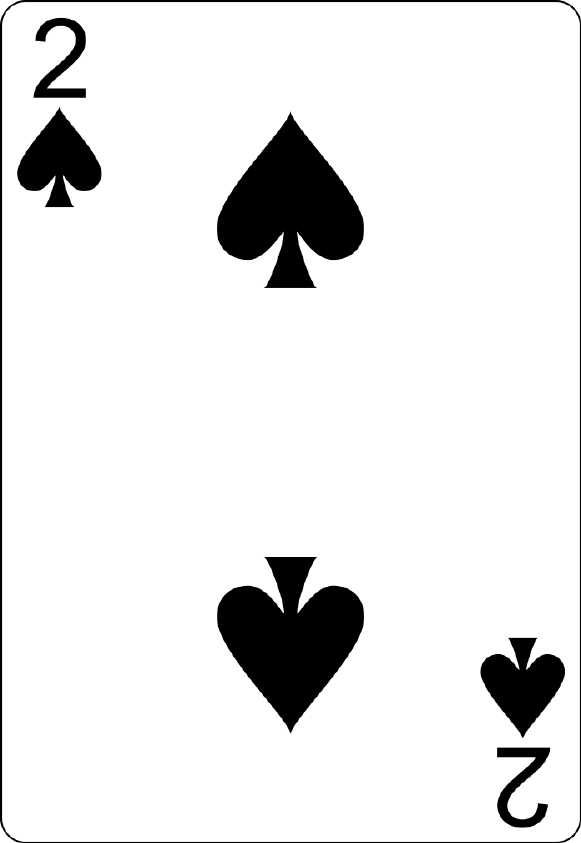}};
\node[text width=0.5cm] at (0.85, -1.25) (pokerout0) {\bf ?};
\path[mapsto] (pokerin0) to (pokerout0);
\end{scope}
\begin{scope}[shift={(0,0.5)}]
\node[text width=0.5cm] at (-1, -1.3) {\includegraphics[width=0.5cm]{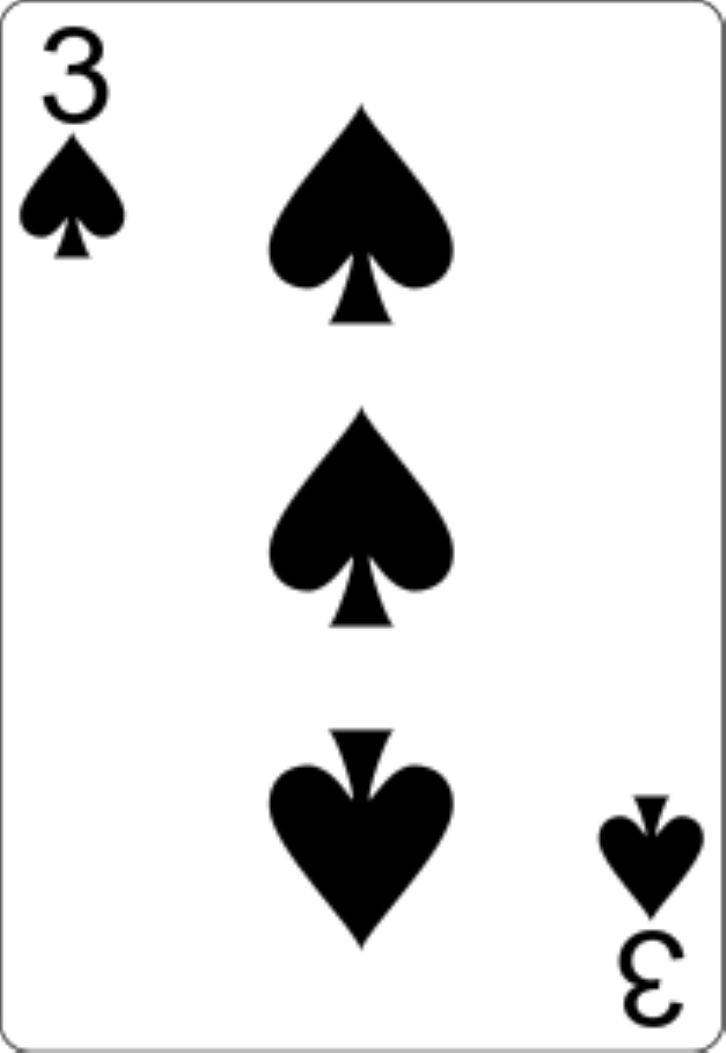}};
\node[text width=0.5cm] at (-0.85, -1.25) (pokerin0) {\includegraphics[width=0.5cm]{figures/4_of_hearts.pdf}};
\node[text width=0.5cm] at (0.85, -1.25) (pokerout0) {\bf \$1};
\path[mapsto] (pokerin0) to (pokerout0);
\end{scope}
\begin{scope}[shift={(0,1.5)}]
\node[text width=0.5cm] at (-1, -1.3) {\includegraphics[width=0.5cm]{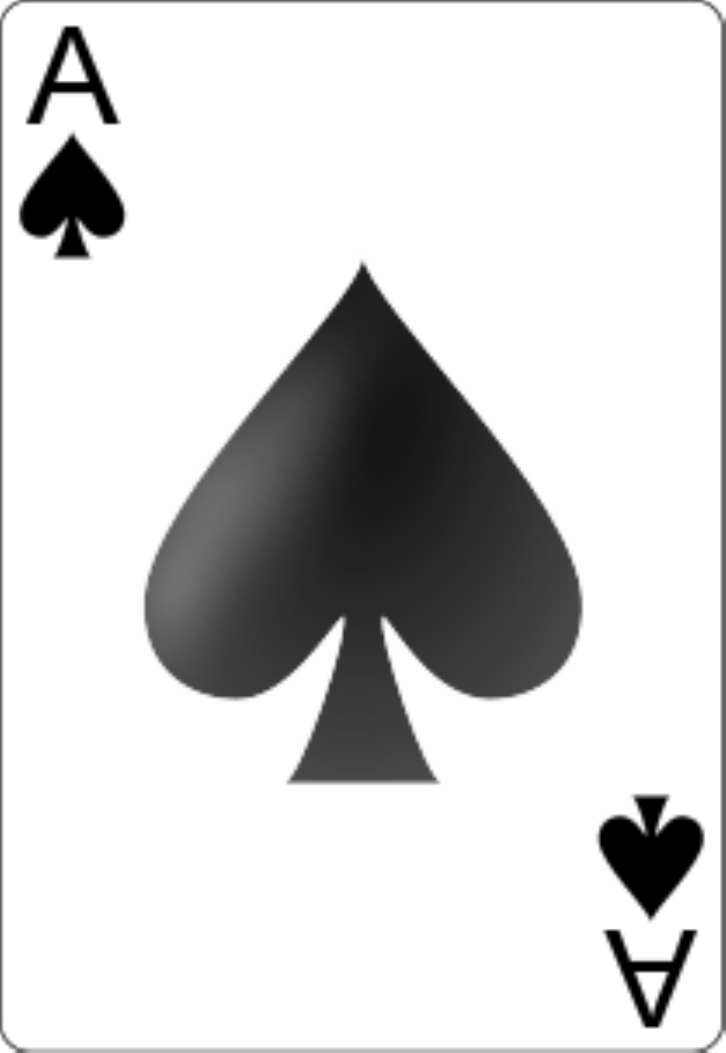}};
\node[text width=0.5cm] at (-0.85, -1.25) (pokerin0) {\includegraphics[width=0.5cm]{figures/2_of_spades.pdf}};
\node[text width=0.5cm] at (0.85, -1.25) (pokerout0) {\bf \$2};
\path[mapsto] (pokerin0) to (pokerout0);
\end{scope}
\begin{scope}[shift={(0,2.5)}]
\node[text width=0.5cm] at (-1, -1.3) {\includegraphics[width=0.5cm]{figures/4_of_hearts.pdf}};
\node[text width=0.5cm] at (-0.85, -1.25) (pokerin0) {\includegraphics[width=0.5cm]{figures/ace_of_spades.pdf}};
\node[text width=0.5cm] at (0.85, -1.25) (pokerout0) {\bf \$0};
\path[mapsto] (pokerin0) to (pokerout0);
\end{scope}

\end{scope}

\begin{scope}[shift={(0.4,0)}]
%% b) constructing
\draw[boundingbox, draw=gray, fill=white] (-8.5, -2.5) rectangle (-2, 2.5);

%% from language
\node[gray] at (-7, 2.1) {Instructions};
\node at (-7, 1.25) (language) {``Play poker.''};

\node[gray, text width=2cm, align=center] at (-4.1, 2.1) {Language network};
\node[block] at (-4.1, 1.25) (languagenet) {\(\mathcal{L}\)}; 
\path[arrow] (language.east) -- ([xshift=-3]languagenet.west);

\node[bpurp, text width=2cm] at (-2, 1.25) (languagetaskrep) {\(z_{task}\)}; 
\path[arrow] ([xshift=3]languagenet.east) -- (languagetaskrep.west);

%% from examples

\node[gray, text width=2.5cm, align=center] at (-7, -0.2) {Task examples (encoded)};
\node at (-7, -1.25) (examples) {
\(\left\{
\begin{matrix}
({\color{bgreen}z_{hand_{1}}}, {\color{bgreen}z_{win_{1}}})\\
$\vdots$
\end{matrix}\right\}\)};

\node[gray, text width=2cm, align=center] at (-4.1, -0.4) {Example network};
\node[block] at (-4.1, -1.25) (examplenet) {\(\mathcal{E}\)}; 
\path[arrow] (examples.east) -- ([xshift=-3]examplenet.west);

\node[bpurp, text width=2cm] at (-2, -1.25) (examplestaskrep) {\(z_{task}\)}; 
\path[arrow] ([xshift=3]examplenet.east) -- (examplestaskrep.west);

\end{scope}

%% c) performing 
\draw[boundingbox, draw=gray, fill=white] (-1.5, -2.5) rectangle (8.5, 2.5);
\node[bpurp] at (-0.75, 1.25) (taskrep) {\(z_{task}\)};

\node[gray, text width=2cm, align=center] at (0.75, 2.1) {Hyper network};
\node[block] at (0.75, 1.25) (hypernet) {\(\mathcal{H}\)}; 
\path[arrow] (taskrep.east) -- ([xshift=-3]hypernet.west);

\node[text width=0.5cm] at (-1, -1.3) {\includegraphics[width=0.5cm]{figures/4_of_hearts.pdf}};
\node[text width=0.5cm] at (-0.85, -1.25) (inputs) {\includegraphics[width=0.5cm]{figures/2_of_spades.pdf}};

\node[gray, text width=2cm, align=center] at (0.6, -2) {Perception network};
\node[block] at (0.6, -1.25) (perceptionnet) {\(\mathcal{P}\)}; 
\path[arrow] (inputs.east) -- ([xshift=-3]perceptionnet.west);

\node[bgreen] at (2.05, -1.25) (handrep) {\(z_{hand}\)};
\path[arrow] ([xshift=3]perceptionnet.east) -- (handrep.west);

\node[bblue, block, dashed] at (3.5, -1.25) (tasknet) {\(\mathcal{T}\)}; 
\node[bblue, text width=1.5cm, align=center] at (3.5, -2) {Task network};
\path[arrow] (handrep.east) -- ([xshift=-3]tasknet.west);
\path[arrow, out=0, in=90] ([xshift=3]hypernet.east) to ([yshift=3]tasknet.north);

\node[bgreen] at (4.85, -1.25) (betrep) {\(z_{bet}\)};
\path[arrow] ([xshift=3]tasknet.east) -- (betrep.west);

\node[gray, text width=3cm, align=center] at (6.2, -2) {Output decoding\\network};
\node[block] at (6.2, -1.25) (actionnet) {\(\mathcal{O}_d\)}; 
\path[arrow] (betrep.east) -- ([xshift=-3]actionnet.west);

\node at (7.4, -1.25) (output) {\bf \$};
\path[arrow] ([xshift=3]actionnet.east) -- (output.west);

%% d subpanel) task network

\draw[boundingbox, draw=gray, fill=white] (4.25, 0) rectangle (8.5, 2.5);
\node[gray] at (6.35, 2.25) {\(\mathcal{H}\) adapts all weights in \(\mathcal{T}\)};

\node at (4.6, 1.45) (pseudohyper) {};

\node[bblue, text width=1cm] at (4.9, 1.45) (Wprime) {\scriptsize \(W'\!, b'\) (from \(\mathcal{H}\))};
\node[bblue] at (6.05, 0.25) (W0) {\scriptsize Default \(W^{0}\!, b^{0}\)};
\node[netnode, bblue, thick] at (4.75, 0.5) (tnode1) {I};

\node[netnode, bblue, thick] at (7.35, 0.5) (tnode2) {O};
%\node[bblue] at (7.35, 1.4) {\scriptsize\(O\!=\!(W^{0}\!+\!W') I\)};
%\node[bblue] at (7.4, 1.1) {\scriptsize\(+(b^{0}\!+\!b')\)};
\node[bblue] at (7, 1.65) {\tiny\(\!W^{0}\!+\!W'\)};
\draw[pen colour={bblue},decoration={calligraphic brace, amplitude=0.1cm},decorate,line width=0.25mm] (7.4, 1.55) -- (6.5, 1.55);
\node[bblue] at (8.1, 1.65) {\tiny\(\!b^{0}\!+\!b'\)};
\draw[pen colour={bblue},decoration={calligraphic brace, amplitude=0.1cm},decorate,line width=0.25mm] (8.35, 1.55) -- (7.75, 1.55);
\node[bblue] at (7.35, 1.1) {\scriptsize\(O\!=\!W^{t}I\!+\!b^{t}\)};

\path[arrow, line width=0.5pt, bblue] (6.95, 1.45) to (7.15, 1.25);
\path[arrow, line width=0.5pt, bblue] (8.05, 1.45) to (7.95, 1.25);

\path[arrow, bblue, out=0, in=90] (pseudohyper) to ([yshift=10]W0);
\path[arrow, bblue, dashed] (tnode1) to (tnode2);

%% connecting to subpanel

\path[draw, gray, thick] (3.75, -0.95) to (4.25, 2.5);
\path[draw, gray, thick] (3.75, -0.95) to (8.5, 0);
\end{tikzpicture}
}
\begin{subfigure}{0.21\textwidth}%%Haaaaaacky
\caption{A basic task.}\label{fig:HoMM_architecture:basic_task}
\end{subfigure}%
\begin{subfigure}{0.31\textwidth}
\caption{Constructing a basic-task representation.}\label{fig:HoMM_architecture:constructing_basic}
\end{subfigure}%
\begin{subfigure}{0.48\textwidth}
\caption{Performing a basic task from its representation.}\label{fig:HoMM_architecture:performing_basic}
\end{subfigure}
\resizebox{\textwidth}{!}{%
\begin{tikzpicture}[auto]
\begin{scope}[shift={(-10.45,0)}]
\draw[boundingbox, draw=gray, fill=white] (-2.25, -2.5) rectangle (2.25, 2.5);

\node[gray, align=center] at (0, 2.1) {Lose meta-mapping};
\begin{scope}[shift={(0,-0.5)}]
\node[text width=1.4cm, align=center] at (-1.15, -1.25) (metain0) {Poker};
\node[text width=1.4cm, align=center] at (1.15, -1.25) (metaout0) {?};
\path[mapsto] (metain0) to (metaout0);
\end{scope}

\begin{scope}[shift={(0,0.5)}]
\node[text width=1.4cm, align=center] at (-1.15, -1.25) (metain0) {Hearts};
\node[text width=1.4cm, align=center] at (1.15, -1.25) (metaout0) {Lose hearts};
\path[mapsto] (metain0) to (metaout0);
\end{scope}

\begin{scope}[shift={(0,1.5)}]
\node[text width=1.4cm, align=center] at (-1.15, -1.25) (metain0) {Rummy};
\node[text width=1.4cm, align=center] at (1.15, -1.25) (metaout0) {Lose rummy};
\path[mapsto] (metain0) to (metaout0);
\end{scope}

\begin{scope}[shift={(0,2.5)}]
\node[text width=1.4cm, align=center] at (-1.15, -1.25) (metain0) {Blackjack};
\node[text width=1.4cm, align=center] at (1.15, -1.25) (metaout0) {Lose blackjack};
\path[mapsto] (metain0) to (metaout0);
\end{scope}

\end{scope}
\begin{scope}[shift={(0.4,0)}]
%% a) constructing
\draw[boundingbox, draw=gray, fill=white] (-8.5, -2.5) rectangle (-2, 2.5);

%% from language
\node[gray] at (-6.7, 2.1) {Instructions};
\node at (-6.7, 1.25) (language) {``Try to lose.''};

\node[gray, text width=2cm, align=center] at (-4.1, 2.1) {Language network};
\node[block] at (-4.1, 1.25) (languagenet) {\(\mathcal{L}\)}; 
\path[arrow] (language.east) -- ([xshift=-3]languagenet.west);

\node[borange, text width=2cm] at (-2, 1.25) (languagetaskrep) {\(z_{meta}\)}; 
\path[arrow] ([xshift=3]languagenet.east) -- (languagetaskrep.west);

%% from examples

\node[gray, text width=3.5cm, align=center] at (-6.7, -0.2) {Meta-mapping examples (input/output tasks)};
\node at (-6.7, -1.25) (examples) {
\(\left\{
\begin{matrix}
({\color{bpurp}z_{hearts}}, {\color{bpurp}z_{lose hearts}})\\
$\vdots$
\end{matrix}\right\}\)};

\node[gray, text width=2cm, align=center] at (-4.1, -0.4) {Example network};
\node[block] at (-4.1, -1.25) (examplenet) {\(\mathcal{E}\)}; 
\path[arrow] (examples.east) -- ([xshift=-3]examplenet.west);

\node[borange, text width=2cm] at (-2, -1.25) (examplestaskrep) {\(z_{meta}\)}; 
\path[arrow] ([xshift=3]examplenet.east) -- (examplestaskrep.west);
\end{scope}

%% d) performing 
\draw[boundingbox, draw=gray, fill=white] (-1.5, -2.5) rectangle (8.5, 2.5);
\node[borange] at (-0.75, 1.25) (taskrep) {\(z_{meta}\)};

\node[gray, text width=2cm, align=center] at (0.75, 2.1) {Hyper network};
\node[block] at (0.75, 1.25) (hypernet) {\(\mathcal{H}\)}; 
\path[arrow] (taskrep.east) -- ([xshift=-3]hypernet.west);

\node[bpurp] at (1.66, -1.25) (handrep) {\(z_{poker}\)};

\node[bblue, block, dashed] at (3.5, -1.25) (tasknet) {\(\mathcal{T}\)}; 
\node[bblue, text width=1.5cm, align=center] at (3.5, -2) {Task network};
\path[arrow] (handrep.east) -- ([xshift=-3]tasknet.west);
\path[arrow, out=0, in=90] ([xshift=3]hypernet.east) to ([yshift=3]tasknet.north);

\node[bpurp] at (5.5, -1.25) (output) {\(z_{lose poker}\)};
\path[arrow] ([xshift=3]tasknet.east) -- (output.west);

%% d subpanel) performing basic from meta-mapped

\draw[boundingbox, draw=gray, fill=white] (4.25, 0) rectangle (8.5, 2.5);
\node[gray] at (6.35, 2.25) {Performing the new task};
\begin{scope}[scale=0.5, shift={(8.5,2.5)}, every node/.append style={transform shape}]

\node[gray, text width=2cm, align=center] at (2, 1.1) {Hyper network};
\node[block, semithick, rounded corners=2] at (2, 0.25) (hypernet2) {\(\mathcal{H}\)}; 

\node[text width=0.5cm] at (0.55, -1.3) {\includegraphics[width=0.5cm]{figures/4_of_hearts.pdf}};
\node[text width=0.5cm] at (0.7, -1.25) (inputs2) {\includegraphics[width=0.5cm]{figures/2_of_spades.pdf}};

\node[gray, text width=2cm, align=center] at (1.9, -2) {Perception network};
\node[block, semithick, rounded corners=2] at (1.9, -1.25) (perceptionnet2) {\(\mathcal{P}\)}; 
\path[arrow, semithick] (inputs2.east) -- ([xshift=-3]perceptionnet2.west);

\node[bgreen] at (3.2, -1.25) (handrep2) {\(z_{hand}\)};
\path[arrow, semithick] ([xshift=3]perceptionnet2.east) -- (handrep2.west);

\node[bblue, block, semithick, rounded corners=2, dash pattern=on 2pt off 2pt] at (4.5, -1.25) (tasknet2) {\(\mathcal{T}\)}; 
\node[bblue, text width=1.5cm, align=center] at (4.5, -2) {Task network};
\path[arrow, semithick] (handrep2.east) -- ([xshift=-3]tasknet2.west);
\path[arrow, semithick, out=0, in=90] ([xshift=3]hypernet2.east) to ([yshift=3]tasknet2.north);

\node[bgreen] at (5.65, -1.25) (betrep2) {\(z_{bet}\)};
\path[arrow, semithick] ([xshift=3]tasknet2.east) -- (betrep2.west);

\node[gray, text width=3cm, align=center] at (6.8, -2) {Output decoding\\network};
\node[block, semithick, rounded corners=2] at (6.8, -1.25) (actionnet2) {\(\mathcal{O}_{d}\)}; 
\path[arrow, semithick] (betrep2.east) -- ([xshift=-3]actionnet2.west);

\node at (7.8, -1.25) (output2) {\bf \$};
\path[arrow] ([xshift=3]actionnet2.east) -- (output2.west);
\end{scope}

%% connecting to subpanel

\node[inner sep=0, outer sep=0] at (6.8, -0.8) (c1) {};
\node[inner sep=0, outer sep=0] at (4.5, -0.3) (c2) {};
\node[inner sep=0, outer sep=0] at (4, 0.5) (c3) {};
\path[draw, gray, very thick, out=0, in=-90, dash pattern=on 2pt off 2pt] (output.east) to (c1);
\path[draw, gray, very thick, out=90, in=0, dash pattern=on 2pt off 2pt] (c1) to (c2);
\path[draw, gray, very thick, out=180, in=-90, dash pattern=on 2pt off 2pt] (c2) to (c3);
\path[arrow, gray, very thick, out=90, in=180, dash pattern=on 2pt off 2pt] (c3) to ([xshift=-3]hypernet2.west);

\end{tikzpicture}
}
\begin{subfigure}{0.21\textwidth}%%Haaaaaacky
\caption{A meta-mapping.}\label{fig:HoMM_architecture:meta_mapping}
\end{subfigure}%
\begin{subfigure}{0.31\textwidth}
\caption{Constructing a meta-mapping representation.}\label{fig:HoMM_architecture:constructing_meta}
\end{subfigure}%
\begin{subfigure}{0.48\textwidth}
\caption{Transforming a task via a meta-mapping.}\label{fig:HoMM_architecture:performing_meta}
\end{subfigure}
\vspace{-1.7em}
\caption{Performing and transforming tasks with a meta-mapping architecture. (\subref{fig:HoMM_architecture:basic_task}) Basic tasks are mappings from inputs to outputs, which can be generalized from examples. (\subref{fig:HoMM_architecture:meta_mapping}) Meta-mappings are mappings from tasks to other tasks, which can be generalized from examples. (\subref{fig:HoMM_architecture:constructing_basic},\subref{fig:HoMM_architecture:constructing_meta}) The HoMM architecture performs basic tasks and meta-mappings from a task representation, which can be constructed from a language cue or examples. (\subref{fig:HoMM_architecture:performing_basic}) The task representation is used to alter the parameters of a task network (see detail) which executes the appropriate task mapping. (\subref{fig:HoMM_architecture:performing_meta}) The meta-mapping representation is used to parameterize the task network to transform a task representation. The transformed representation can then be used to perform the new task zero-shot (see detail). Our architecture exploits a deep analogy between basic tasks and meta-mappings --- both can be seen as mappings of inputs to outputs. This analogy is reflected in the parallels between the top and bottom rows of the figure.} \label{fig:HoMM_architecture}
\end{figure*}

We propose meta-mappings as a computational approach to the problem of transforming a basic task mapping. A meta-mapping is a higher-order task, which takes a task representation as input, and outputs a representation of the transformed version of the task. For example, we might have a ``lose'' meta-mapping (Fig. \ref{fig:HoMM_architecture:meta_mapping}), that would transform the representation of poker into a representation of losing at poker.% If given a representation of poker as input, the lose meta-mapping would output a representation which allows the system to play a losing variation of poker. %If given poker as an input, the lose meta-mapping would output a losing variation of poker. If we have such a meta-mapping, we can use it to transform a task in order to perform the transformed version of the task. This allows a model to adapt to a transformed task without having any data on it, just as humans are able to easily switch to trying to lose at a game they have only tried to win in the past. 

How can a meta-mapping be performed? We exploit an analogy between meta-mappings and basic task mappings -- both are simply functions from inputs to outputs. Thus to perform a meta-mapping we use approaches analogous to those we use for basic tasks. We infer a meta-mapping from examples (e.g. winning and losing at a set of example games), or natural language (e.g. ``try to switch to losing''). We can then apply this meta-mapping to other basic tasks, in order to infer losing variations of those tasks. Importantly, the system can generalize to new meta-mappings --- task transformations never seen in training --- as well as to new basic tasks. %For example, if it experienced meta-mappings which altered the rank of some cards (e.g. ''ace is high rather than low''), it could generalize to switching the rank of other cards, either from examples or an instruction. 

\section*{A meta-mapping architecture} \label{sec:HoMM_architecture}

We propose a class of architectures that can both perform basic tasks and adapt to task alterations via meta-mappings. In this section, we describe the general features of our architectures and their training. See SI \ref{supp_sec:methods} for details, including a formal model description (SI \ref{supp_sec:methods:model_formalism}), all hyperparameters (SI \ref{supp_sec:model_details}), etc.

\subsection*{Constructing a task representation (Fig. \ref{fig:HoMM_architecture:constructing_basic})} When humans perform a task, we need to know what the task is. In our model, we specify the task using a task representation, which we derive from language, from supporting examples of appropriate behavior, or from meta-mapping. To construct a task representation from language, we process the language through a deep recurrent network (LSTM), as in other work \citep[e.g.][]{Hermann2017,Oh2017a,Hill2019a}. To construct a task representation from examples, as in other work \citep[e.g.][]{Garnelo2018}, we process each example (i.e. an input and its corresponding target) to construct an appropriate representation of the example, and then aggregate across those representations by taking an element-wise maximum, to combine examples in a nonlinear but order-invariant way. This aggregated representation then receives further processing to produce the task representation. 

%\looseness=-1
\subsection*{Performing a task from its representation (Fig. \ref{fig:HoMM_architecture:performing_basic})} Once we have a task representation, we use it to perform the task. We allow a large part of the input processing (perception) and output decoding (action) to be shared across the tasks within each domain we consider,\footnote{Of course, with different input types, this type of processing will be different. While the core model components are similar across experiments, the input and output systems can therefore differ.} so that the task-specific computations can be relatively simple and abstract. For example, if a human is playing card games, the cards will be identical whether the game is poker or bridge, and the task-specific computations will be performed over abstract features such as suit and rank relationships. We thus allow the system to learn a general basis of perceptual features over all tasks within a domain.%This idea is related to the long-standing notion that deep networks (both artificial and biological) construct disentangled representations of the task relevant features in deeper layers \citep{Dicarlo2007, Erhan2010}, and is often used in meta-learning \citep[e.g.]{Vinyals2016}. 

The system then uses these features in a task-specific way to perform task-appropriate behavior. Specifically, the model uses a HyperNetwork \citep{Ha2016,McClelland1985} which takes as input the representation of a task. This network adapts the values of learned ``default'' connection weights, to make the network task-sensitive (Fig. \ref{fig:HoMM_architecture:performing_basic} detail). The adapted network transforms the perceptual features into task-appropriate output features, which can then be decoded to outputs via a shared output decoding network. The whole model (including the construction of the task representations) can be trained end-to-end, just as a standard meta-learning system would be. (Our approach outperforms an alternative architecture, in which the task representation is provided as another input to a feed-forward task network, see Fig. \ref{supp_fig:HoMM_arch_cond_vs_hyper}.) 

\subsection*{Transforming task representations via meta-mappings (Fig. \ref{fig:HoMM_architecture:constructing_meta}-\subref{fig:HoMM_architecture:performing_meta})} We defined a meta-mapping to be a higher-order task, which takes as input a task representation, and outputs a transformed task representation. Thus, we need a way of transforming the task representations constructed above. To do so, we exploit the functional analogy between basic-tasks and meta-mappings. We infer a representation for a meta-mapping from examples of that meta-mapping, or from a language description, just like we infer a basic task representation from examples or language. We use this meta-mapping representation to adapt the parameters of the task network to transform other task representations. This approach is analogous to how we used a representation of a basic task to adapt the task network to perform that task. (See SI \ref{supp:HoMM_vector_analogies_inadequate} for proof that a simpler vector-analogy approach to meta-mapping is inadequate.) 

\subsection*{Homoiconicity}
Our architectures exploit the analogy between basic tasks and meta-mappings by using exactly the same networks (with exactly the same parameters) to infer and perform a meta-mapping as for inferring and performing a basic task. To allow this, the system embeds individual data points, task representations, and meta-mapping representations into a shared representational space. This means that all task- or meta-mapping-specific computations can be seen as operations on objects in this shared space, and can be inferred using the same processes regardless of object type. (Note that sharing of the space is only enforced implicitly in that the same networks are processing different entities.) % e.g. data representations and task representations are both used as inputs to the same example network.) 
This approach is in keeping with the idea that humans have a single mind that implements computations of all types. Our approach is also inspired by the computational notion of \emph{homoiconicity}. In a homoiconic programming language programs can be manipulated just as data can. Our task representations are like programs that perform tasks, and our implementation is thus homoiconic in the sense that it operates on data and tasks in the same way.  

Homoiconicity is parsimonious, in that it does not require adding new networks for each new type of computation. Furthermore, in many cases, functions have some common structure with the entities they act over. For example, both numbers and functions can have inverses. For another example, the set of linear maps over a vector-space is itself a vector space. If the different levels of abstraction share structural features, sharing computation should improve generalization. Homoiconicity could also support the ability to build abstractions recursively on top of prior abstractions, as humans do in mathematical cognition \citep{Wilensky1991, Hazzan1999, Lampinen2017b}. Although homoiconicity is not a \emph{necessary} part of meta-mapping, we suggest that homoiconic approaches will be beneficial (and verify this empirically, see below). 

\subsection*{Classifying task representations} In all of the domains discussed below, except the RL domain, we also trained the model to classify task representations by relevant attributes (for example, whether a game was a variation of poker), again using the same architectural components. See SI \ref{supp_model_training_details} for details. This may improve generalization by helping the model learn the structure of the task space (but is not essential, see Fig. \ref{supp_fig:HoMM_metaclass_lesion}). 

\subsection*{Training the model} 
We train the system in epochs, during which it receives one training step on each trained basic task, and one training step on each trained meta-mapping, interleaved in a random order. To train the system to perform the basic tasks, we compute a task-appropriate loss at the output of the output decoding network, and then minimize this loss with respect to the parameters in all networks. This includes the networks used to construct the task representation, and even the representations of the examples or language input. That is, we train the system end-to-end to perform the basic tasks. 

When constructing a task representation from examples, we do not allow the example network to see every example in the training batch. This forces the model to generalize in a standard meta-learning fashion. Specifically, we separate the batch of examples into a \emph{support set} which are provided to the example network, and a \emph{probe set} which are only passed through the task network to compute an output, from which a loss can be computed against a task target. For example, in a card game the system will have to construct a task representation from the support hands that will be useful for playing the probe hands. This approach encourages the task representations to capture the task structure, rather than just memorizing examples. We randomly split the training examples into support and probe sets on each step, so that over the course of training every training example would fill both roles. In this approach, the task representation is constructed anew at each training step. However, to stabilize learning in difficult domains, it can be useful to maintain a persistent task representation which updates slowly with each new set of examples (see SI \ref{supp_model_training_details}).

Training the system to construct basic task representations from language is similar, except that a language description (e.g. ``play poker'') is provided rather than examples. Thus no support set is needed, so all examples can be used as probes.

To train the system to perform meta-mappings from examples, we start with a training set of example task representation pairs, where each pair consists of a source task representation and the corresponding transformed task representation. Again, on each training step, a subset of these examples is used as a support set to construct a meta-mapping representation. The remaining examples are used as a probe set to train the system to transform the source representation for each pair to its corresponding target.
%To train the system to perform meta-mappings, we first instantiate the meta-mapping representation using either examples or language, and use that representation to parameterize the task network. We then process a set of tasks representations (that were not used as examples) through the task network, and try to match the output task representations to known target task embeddings. 
Specifically, we present the source task embedding as input to the task network, and minimize an \(\ell_2\) loss on the difference between the output embedding the task network produces and the task representation for the target transformed task. For example, suppose the system has been trained to play winning and losing variations of blackjack, hearts, and rummy. We might use the representations of winning and losing hearts and rummy as support-set examples to instantiate the meta-mapping, then input the task representation for winning blackjack as a probe, and try to match the output to the task representation for losing blackjack. Again, we randomly chose which examples were used as support or probes on each training step. On the next training step, we might use hearts and blackjack as examples, and train the meta-mapping to generalize to losing at rummy. 

Training the system to perform meta-mappings from language is similar, except that again a language description (e.g. ``switch to losing'') is provided rather than examples of the transformation. Thus, as when using language rather than examples to perform basic tasks, all pairs can be used as probes.  

\subsection*{Evaluating base-task \& meta-mapping performance} After training, we can evaluate the model's base-task performance using held-out examples unseen during training. To test generalization of a meta-mapping (e.g. ``try-to-lose''), we can pass in the representation for a task that has never been used for any training on this meta-mapping (either as a support example or a probe for generalization), for example, poker. We construct a meta-mapping representation using \emph{all} the training examples of the lose meta-mapping as a support set. We then apply the lose meta-mapping to the task representation of poker (i.e. pass it through the task network parameterized by the lose meta-mapping representation) to produce a transformed representation. We then actually perform the losing variation of poker with this transformed representation. Meta-mapping performance is always evaluated by zero-shot performance on held-out tasks that the system has never performed during training.

In meta-mapping, generalization is possible at different levels of abstraction. The paragraph above refers to basic generalization --- applying a meta-mapping seen during training to a basic task that meta-mapping has not been applied to during training, in order to perform a held-out transformed version of that task. However, if the system has experienced sufficiently many meta-mappings during training, we can also test its ability to generalize to \textbf{held-out meta-mappings}. For example, if the system has been trained to switch various pairs of colors in a classification task (red for blue, green for yellow, etc.), it should be able to generalize to switching held-out pairs (red for yellow, green for blue, etc.) from an appropriate cue (examples or instructions). That is, even if a meta-mapping has never been encountered during training, we can construct a representation for it by providing a support set of transformation examples, or a language instruction that is systematically related to those used for trained meta-mappings. We view this as an important part of intelligent adaptability --- the system should not only be able to adapt to tasks via meta-mappings that it has directly experienced, but also to infer and use novel meta-mappings based on specific instructions or examples. We demonstrate this ability in the subset of our experimental domains where we can instantiate sufficiently many meta-mappings. 

%\textbf{Comparing to language-based generalization:} Natural language instructions are key to human adaptation to new tasks, and prior work on zero-shot performance has often assumed a description of the task as input \citep[e.g.]{Larochelle2008}. For example, a system that has learned to behave in accordance with instructions like ``win at poker,'' ``win at blackjack,'' and ``lose at blackjack,'' should be able to generalize to ``lose at blackjack,'' given sufficiently many training tasks. This, too, does not require data on the novel task. However, transforming the task representation via a meta-mapping may be a useful inductive bias that allows the system to better use prior task knowledge. To verify this, we compare our meta-mapping approach to an approach that simply constructs task representations from language. We show below that meta-mapping results in better performance on the new tasks, especially when the space of tasks is sparsely sampled or generalization is challenging. 
\vspace{-0.5em}
\section*{Experiments}

Meta-mapping is an extremely general framework. Because the assumptions are simply that the basic tasks are mappings from inputs to outputs, and that meta-mappings transform basic tasks, the approach can be applied to most paradigms of machine learning with minor modifications. We demonstrate our results in four experimental domains. We summarize the contributions of each domain in Table \ref{table:HoMM_experiment_summary}. 
\begin{table}
\centering
\begin{tabular}{|p{1.2cm}|ccp{1cm}p{0.8cm}p{1.35cm}|}
\hline
\textbf{Domain} & \begin{minipage}[t]{1.05cm}\centering\textbf{Held-out MMs}\end{minipage} & \begin{minipage}[t]{0.7cm}\centering\textbf{Lang. Comp.}\end{minipage} & \textbf{Type} & \textbf{Input} & \textbf{Output}\\[1.3em]
\hline
Polynomials  & \checkmark & & Regression & Vector (\(\mathbb{R}^4\)) & Scalar (\(\mathbb{R}\))\\%[0.5em] 
Cards  & & \checkmark & Regression & Several-hot & Bet values (\(\mathbb{R}^3\)) \\%[0.5em]
Visual\phantom{bl} concepts &  \checkmark & \checkmark & Classific-ation & \(50\!\times\!50\) image & Label (\(\{0, 1\}\))\\%[0.5em]
RL &  & \checkmark & RL & \(91 \times 91\) image & Action Q- values (\(\mathbb{R}^4\))\\%[0.5em]
\hline
\end{tabular}
\vspace{0.2em}
\caption{The contributions of our four experimental domains. Our results span various computational paradigms and data types. (Note: ``Lang. Comp.'' refers to a comparison to language alone, see below.)} \label{table:HoMM_experiment_summary}
\vspace{-1.5em}
\end{table}

\subsection*{Polynomials}
\begin{figure}[thb]
\vspace{-0.5em}
%\begin{subfigure}[t]{0.5\textwidth}
\centering
\resizebox{0.5\textwidth}{!}{
\begin{tikzpicture}[auto]
%% Basic
\node[align=left, text width=1.65cm] at (-4.8, 1.5) {Basic\\tasks};
\draw[boundingbox, draw=gray, fill=white] (-4, 2.7) rectangle (-0.25, -0.2);
\draw[boundingbox, draw=gray, fill=white] (-4, 2.7) rectangle (-0.25, 1.7);
\node at (-3.5, 2.45) {\it Task:};
\node[align=center] at (-2.2, 2) {\(f(w,x,y,z) = x^2 + 1\)};
\node at (-2.4, 1.45) {\it Input-output pairs:};
\node[align=center] at (-2.2, 0.5) {%
    \(\begin{aligned}
    (0, 0, 0, 0) & \mapsto 1\\[-0.2em] 
    (1.5, -1, 3.1, 0) & \mapsto 2\\[-0.75em] 
    & \;\;\vdots
    \end{aligned}\)};

\begin{scope}[shift={(4, 0)}]
\draw[boundingbox, draw=gray, fill=white] (-4, 2.7) rectangle (-0.25, -0.2);
\draw[boundingbox, draw=gray, fill=white] (-4, 2.7) rectangle (-0.25, 1.7);
\node at (-3.5, 2.45) {\it Task:};
\node[align=center] at (-2.2, 2) {\(f(w,x,y,z) = 3w + yz\)};
\node at (-2.4, 1.45) {\it Input-output pairs:};
\node[align=center] at (-2.2, 0.5) {%
    \(\begin{aligned}
    (0.5, 0, 1, 2) & \mapsto 3.5\\[-0.2em] 
    (1, 0.2, -1, 0.5) & \mapsto 2.5\\[-0.75em] 
    & \;\;\vdots
    \end{aligned}\)};
\end{scope}

%% meta-mapping
\node[align=left, text width=1.65cm] at (-4.8, -2) {Meta\\mappings};

\begin{scope}[shift={(0, -3)}]
\draw[boundingbox, draw=gray, fill=white] (-4, 2.7) rectangle (-0.25, -0.2);
\draw[boundingbox, draw=gray, fill=white] (-4, 2.7) rectangle (-0.25, 1.7);
\node at (-2.75, 2.45) {\it Meta-mapping:};
\node[align=center] at (-2.8, 2) {Multiply by 3.};
\node at (-2.4, 1.45) {\it Input-output pairs:};
\node[align=center] at (-2.2, 0.5) {%
    \(\begin{aligned}
    x^2 + 1 & \mapsto 3x^2 + 3\\[-0.2em] 
    3w + yz & \mapsto 9w + 3yz\\[-0.75em] 
    & \;\;\vdots
    \end{aligned}\)};
\end{scope}

\begin{scope}[shift={(4, -3)}]
\draw[boundingbox, draw=gray, fill=white] (-4, 2.7) rectangle (-0.25, -0.2);
\draw[boundingbox, draw=gray, fill=white] (-4, 2.7) rectangle (-0.25, 1.7);
\node at (-2.75, 2.45) {\it Meta-mapping:};
\node[align=center] at (-2.4, 2) {Permute \((w, z, x, y)\)};
\node at (-2.4, 1.45) {\it Input-output pairs:};
\node[align=center] at (-2.2, 0.5) {%
    \(\begin{aligned}
    x^2 + 1 & \mapsto z^2 + 1\\[-0.2em] 
    3w + yz & \mapsto 3w + xy\\[-0.75em] 
    & \;\;\vdots
    \end{aligned}\)};
\end{scope}
\end{tikzpicture}
}%
%\begin{tikzpicture}[overlay]
%\node at (-7.5, 6.5) {\small \sf \bf (a)};
%\end{tikzpicture}
%\phantomcaption\label{fig:HoMM_polynomials:tasks}
\vspace{-0.5em}
\caption{The polynomial task domain. A basic polynomial task consists of regressing a single polynomial, i.e. the inputs are points in \(\mathbb{R}^4\) and the outputs are the value of the polynomial at that point. These basic regression tasks can be transformed by various meta-mappings, such as multiplying by a constant, or permuting their variables.}\label{fig:HoMM_polynomials:tasks}
%\caption{Polynomial tasks and meta-mappings}\label{fig:HoMM_polynomials:tasks}
%\end{subfigure}%
\vspace{-0.5em}
\end{figure}

\begin{figure}[tbh]
%\begin{subfigure}[t]{0.5\textwidth}
\centering
\includegraphics[width=0.45\textwidth]{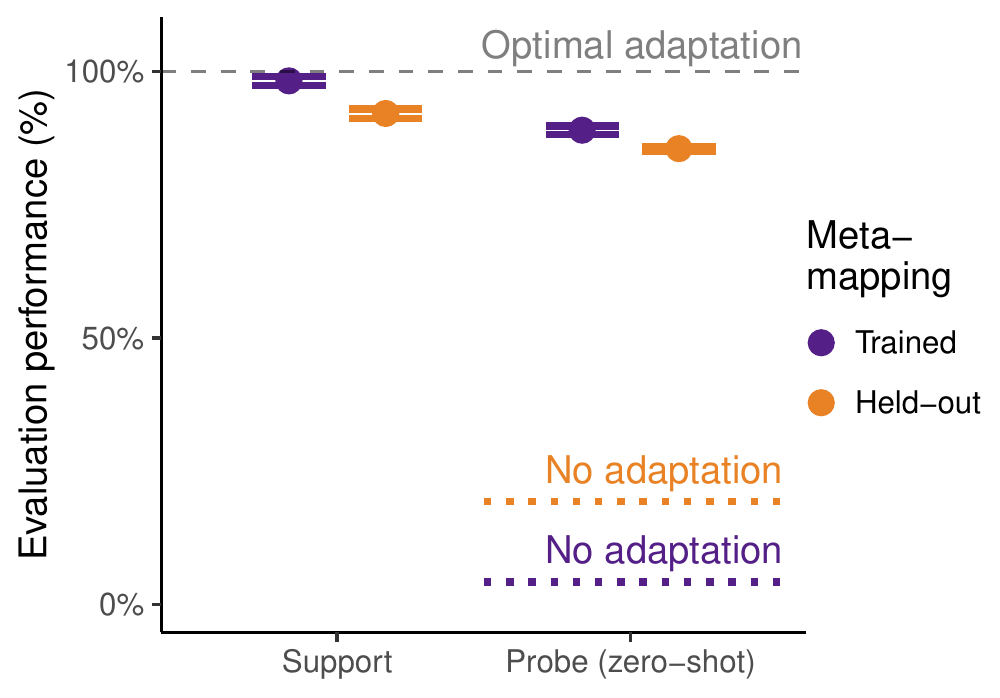}%
%\includegraphics[width=0.45\textwidth]{figures/polynomials_adaptation.pdf}%
%\caption{Meta-mapping results.}\label{fig:HoMM_polynomials:results}
%\begin{tikzpicture}[overlay]
%\node at (-9, 6.5) {\small \sf \bf (b)};
%\end{tikzpicture}
%\phantomcaption\label{fig:HoMM_polynomials:results}
%\end{subfigure}
\vspace{-0.5em}
\caption{Meta-mapping can adapt to a new polynomial zero-shot, based on its relationship to prior polynomials. We plot performance (normalized, see text) on transformed polynomials via meta-mappings. The system performs well on support-set target tasks after adaptation. More importantly, it can perform probe target polynomials that it has never encountered before zero-shot, and does so substantially better than if it did not adapt (dotted lines). It generalizes well both on trained meta-mappings (purple), and on held-out meta-mappings (orange). (Dots show mean and lines show bootstrap 95\%-CI across 5 runs.)}\label{fig:HoMM_polynomials:results}
%\caption{Meta-mapping allows zero-shot generalization in the polynomials domain. We plot performance (normalized, see text) on new tasks via meta-mappings. The system not only generalizes trained meta-mappings to polynomials it has never seen before (purple), but also generalizes to held-out meta-mappings from examples (orange), and does both substantially better than if it did not adapt (dotted lines). Meta-mapping can adapt to a new polynomial without any data from that polynomial, based on its the relationship to prior polynomials. (Mean and bootstrap 95\%-CI across 5 runs.)}\label{fig:HoMM_polynomials:results}
\end{figure}

As a proof of concept %, and to demonstrate the benefits of a homoiconic approach,
we first apply meta-mapping to polynomial regression (see Fig. \ref{fig:HoMM_polynomials:tasks}). We construct basic tasks that are polynomial functions (of degree \(\leq 2\)) in four variables (i.e. from \(\mathbb{R}^4 \rightarrow \mathbb{R}\)). These polynomials can be inferred from a support set of (input, output) examples, where the input is a point in \(\mathbb{R}^4\) and the output is the evaluation of that polynomial at that point. For details, and to see that the system performs this simple meta-learning regression problem extremely well, see SI \ref{meth_data_poly} and Fig. \ref{supp_fig:HoMM:polynomials_basic_meta_learning}. 

%These tasks/polynomials can then be transformed by various meta-mappings --- we considered squaring a polynomial, permuting its variables, or adding or multiplying by a constant. We trained the model on 100 basic polynomials, and we train mapped versions of 60 of these for each meta-mapping. We can evaluate the performance of that meta-mapping on the remaining 40 target tasks (corresponding to the 40 other basic polynomials) that the model has never experienced before. We also held out some of these meta-mappings to evaluate the ability of our method to generalize at the meta-mapping level (see above). For example, we trained the model to adapt to a subset of the input variable permutations, and then evaluated its ability to adapt to a held-out permutation based on examples of that permutation. In total, we trained on 20 meta-mappings, and held-out 16, corresponding to 2260 (\(=100 + 60 \times 36 \)) trained basic tasks, and 1440 held-out for evaluation. 

These basic tasks/polynomials can be transformed by various meta-mappings --- we considered squaring a polynomial, permuting its variables, or adding or multiplying by a constant. We considered 36 meta-mappings in total, of which we trained the model to perform 20, and held out the remaining 16 to evaluate the model's ability to generalize to held-out meta-mappings (see above). The held-out meta-mappings included some of the possible permutation, addition, and multiplication transformations. We used 60 example (source polynomial, transformed polynomial) mapping pairs as a training set for each meta-mapping, and held-out another 40 transformed polynomials per meta-mapping for evaluation. The source and transformed polynomials for all 60 example pairs were trained for each trained or held-out meta-mapping. This results in a total of 2260 polynomials trained, and 1440 held-out for evaluation. For the 20 trained meta-mappings, the 60 trained (source polynomial, transformed polynomial) pairs were used to train the meta-mapping, and as the support set for evaluation. For the 16 held-out meta-mappings, these pairs were only used as the support set for evaluation. See SI \ref{meth_data_poly} for further details.

In Fig. \ref{fig:HoMM_polynomials:results}, we show the success of our meta-mapping approach in this setting. We plot a normalized performance measure, \(100\% (1 - \text{loss}/c)\), where \(c\) is the loss for a baseline model that always outputs zero. This measure is 0\% for a model which outputs all zeros, and 100\% if the system performs perfectly. % Specifically, we measure performance as  %\footnote{This measure is closely related to the variance explained, except that the square meta-mapping skews the mean of the output polynomials slightly away from zero.}
See Table \ref{supp_tab:analyses:poly:loss_perf} for raw losses. Meta-mapping achieves good performance on the support set examples that are used to instantiate the mapping, with 98.3\% performance (bootstrap 95\%-CI across runs [97.3, 99.0]) on trained meta-mappings and 92.1\% [91.3, 93.0] on held-out meta-mappings. More importantly, on polynomials never experienced during training, meta-mapping achieves 89.0\% [89.3, 89.8] zero-shot performance on average based on a trained meta-mapping, and 85.5\% [85.1, 85.9] performance based on a held-out meta-mapping. We also show the performance the model obtains when it is scored on the new task using the untransformed source task representation (no adaptation). This baseline yields only 4.3\% and 19.3\% performance, respectively. In summary, meta-mapping is able to achieve good performance on a new task without any data, based only on its relationship to prior tasks.

\begin{figure*}
\begin{subfigure}[b]{0.458\textwidth}
\includegraphics[width=\textwidth]{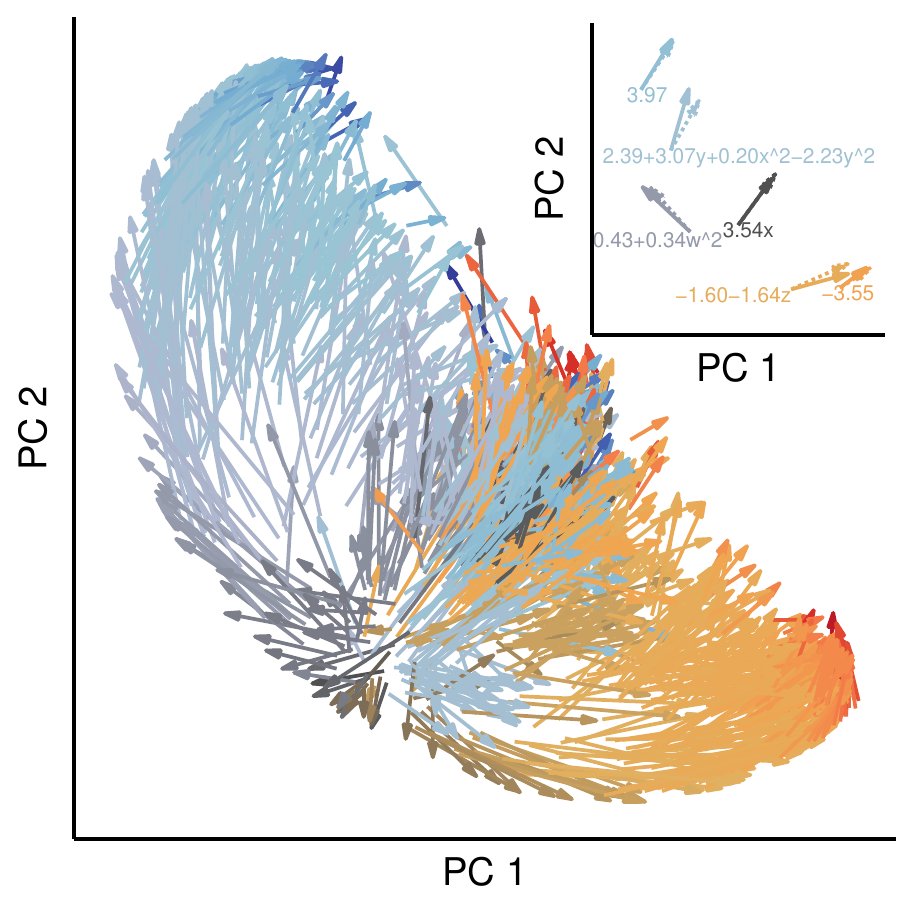}
\begin{tikzpicture}[overlay]
\node at (0.25, 8.15) {\small \sf \bf (a)};
\end{tikzpicture}
%\caption{Meta-mapping: multiply by 3.}
\phantomcaption\label{fig:HoMM_polynomial_transforms:mult}
\end{subfigure}%
\begin{subfigure}[b]{0.541\textwidth}
\includegraphics[width=\textwidth]{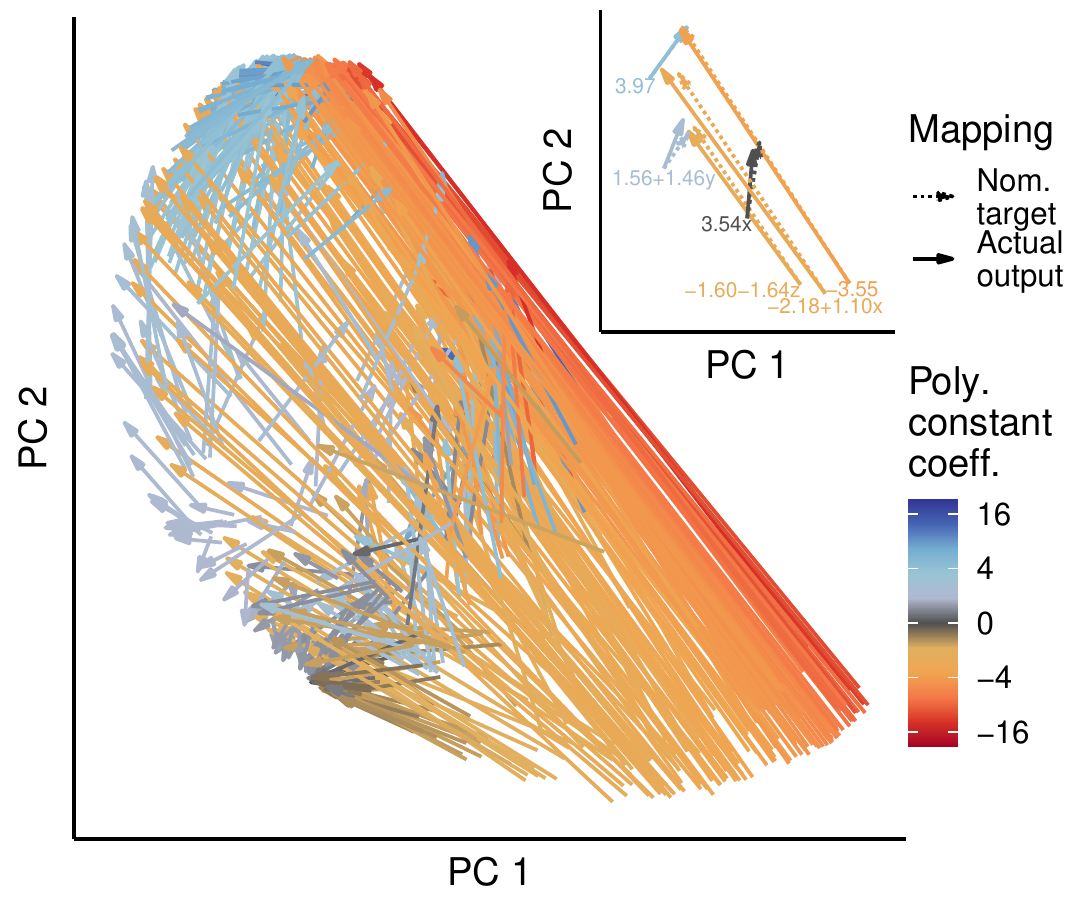}
\begin{tikzpicture}[overlay]
\node at (0.25, 8.15) {\small \sf \bf (b)};
\end{tikzpicture}
%\caption{Meta-mapping: square.}
\phantomcaption\label{fig:HoMM_polynomial_transforms:square}
\end{subfigure}
\vspace{-3em}
\caption{Visualizing how meta-mappings systematically transform the model's polynomial representations. The panels show two meta-mappings: (\subref{fig:HoMM_polynomial_transforms:mult}) multiplying by 3 and (\subref{fig:HoMM_polynomial_transforms:square}) squaring. Each arrow shows how a single polynomial's representation transforms under the meta-mapping. The arrows are colored by the constant term of the polynomial, and the representations are generally organized so that constant polynomials are around the outside, with constant value increasing clockwise, and the polynomials involving more variables are closer to the center (see SI \ref{supp_sec:analyses:polynomial_reps}). (\subref{fig:HoMM_polynomial_transforms:mult}) Multiplying by 3 pushes polynomials away from the center, with the negative constant polynomials rotating counterclockwise as they become more negative, and the positive constant polynomials rotating clockwise as they become more positive. The non-constant polynomials extend in similar directions, but their trajectories are more complicated. (\subref{fig:HoMM_polynomial_transforms:square}) Squaring polynomials results in both rotation of the representation space, and folding as the negative constants flip to being positive. The inset panels show that polynomial transformations align closely to their nominal targets (that is, the model's representation of the target task). (Plots show the top two principal components. Note that only 60 of the 1200 polynomials shown were used for training each mapping. See SI \ref{supp_sec:analyses:polynomial_reps} for further representation analysis.)}\label{fig:HoMM_polynomial_transforms}
\end{figure*}

This success is consistent across all the meta-mapping types we evaluated, see Fig. \ref{supp_fig:HoMM_polynomials_results_by_mapping}. The model is reasonably sample-efficient at inferring both polynomials and meta-mappings (Figs. \ref{supp_fig:HoMM:polynomials_varying_mbs_base}, \ref{supp_fig:HoMM_polynomials_varying_mbs_meta}). Further, we show in Fig. \ref{fig:HoMM_polynomial_transforms} (and SI \ref{supp_sec:analyses:polynomial_reps}) that polynomial and meta-mapping representations are systematically organized and transform in systematic ways. In general, the transformed representations are close to the nominal targets where targets are known. (Note that even missing the nominal target does not necessarily mean the model is incorrect; just as we could write \(2(x+1)\) instead of \(2x + 2\), the model may have different representations for the same function.) 

Finally, our homoiconic approach significantly outperforms a non-homoiconic baseline, which differs from the homoiconic architecture only in that separate example- and hyper-networks are used for the basic tasks and meta-mappings (Fig. \ref{supp_fig:HoMM:nonhomoiconic_baseline}), suggesting that sharing these networks improves generalization. Why is homoiconicity beneficial? We %, but why? We %analyzed the task representations, and 
show that there is non-trivial overlap between the basic-task and meta-mapping representations (Figs. \ref{supp_fig:HoMM_polynomials:meta_base_rep_simil}, \ref{supp_fig:HoMM_polynomials:meta_base_PC_alignments}, \ref{supp_fig:HoMM_polynomials:meta_base_PC_variance}), and that some of this overlap reflects structural isomorphisms (Fig. \ref{supp_fig:HoMM_polynomials:meta_mult_base_const_alignment}). While this may not fully explain the benefits of homoiconicity, it suggests that the model may be exploiting the shared structure between basic tasks and meta-mappings. By contrast, there is little alignment between the representations of numerical polynomial inputs and task representations, potentially because there are fewer constraints encouraging such an alignment (see SI \ref{supp_sec:analyses:polynomial_reps}).

\subsection*{Card games}
%\begin{figure*}
%\begin{subfigure}[b]{0.5\textwidth}
%\includegraphics[width=\textwidth]{figures/cards_screenshot.pdf}
%\caption{A trial from the participant's perspective.}\label{fig:HoMM_cards:domain}
%\end{subfigure}%
%\begin{subfigure}[b]{0.5\textwidth}
\begin{figure}[tbh]
\includegraphics[width=0.45\textwidth]{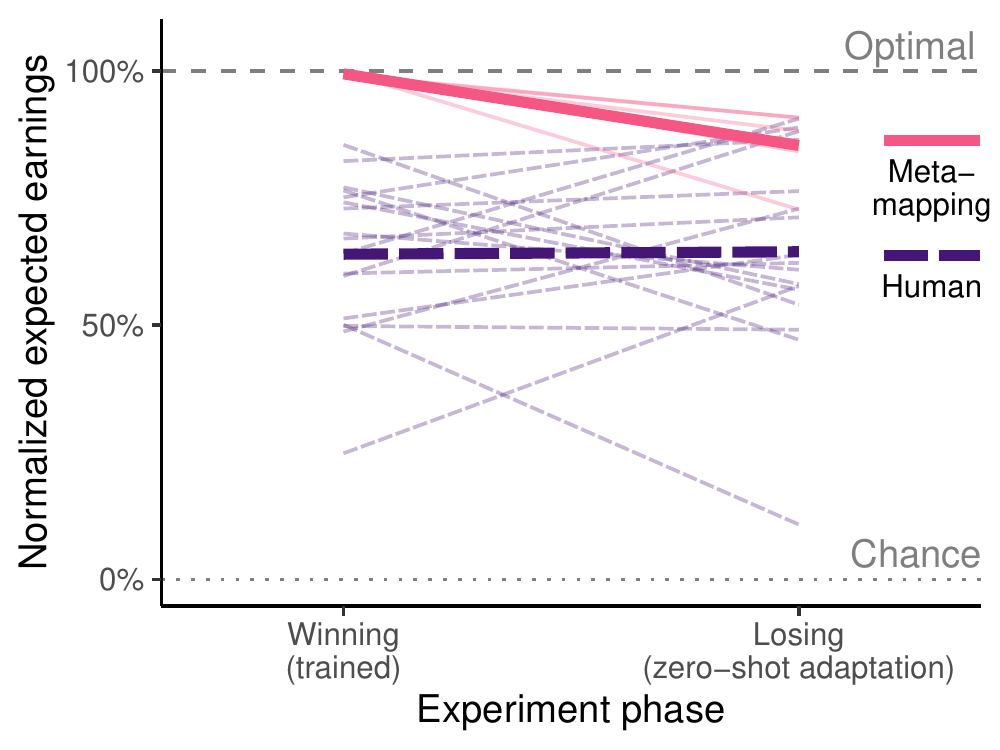}
%\end{subfigure}
%\caption{Comparing HoMM to human adaptation in simple card games. (\subref{fig:HoMM_cards:domain}) The participants were asked to make a bet of 0, 5, or 10 cents after seeing their hand. Partway through the experiment, they were told to switch to trying to lose. (\subref{fig:HoMM_cards:results}) Performance in the two phases of the experiment: baseline testing on the basic game, and after adapting to losing zero-shot. Human participants are behaving suboptimally on average, but are adapting near-optimally, although their is substantial inter- and intra-subject variability. The language-based model and HoMM are both performing near-optimally at baseline, but the language is not able to generalize from the small set of tasks it has experienced to the losing variation of the new game. By contrast, HoMM is achieving around 90\% performance at the new game. (We plot performance as expected earnings of the actions taken, as a percentage of the earnings of an optimal policy. Thick lines are averages, thin lines are 5 runs of each model, and 19 individual participants who passed attention checks.)} \label{fig:HoMM_cards}
\vspace{-0.5em}
\caption{Comparing meta-mapping to human adaptation in simple card games. This plot shows performance in the two phases of the experiment: baseline testing on the basic game, and after adapting to losing zero-shot. Human participants are behaving sub-optimally on average, but are achieving similar performance after adaptation, although there is substantial inter- and intra-subject variability. The model performs near-optimally at baseline, and by meta-mapping achieves around 90\% performance at the new game. (We plot performance as expected earnings from the bets made, as a percentage of the expected earnings of an optimal policy. Thick lines are averages, thin lines are 5 runs of the model, and 19 individual participants.)} \label{fig:HoMM_cards}
\end{figure}
%\end{figure*}

We motivated our work in part by observations about human flexibility, so we next compare our model to human adaptation in a simple card game. The basic tasks consist of receiving a hand of two cards, and making a bet. The human (or model) plays against an opponent, and wins (or loses) their bet if their hand beats (loses to) the opponent's.

We trained human participants to play one poker-like game with two-card hands (card rank 1-4, suit red or black). We evaluated their ability to play that game, and then to switch strategy when told to try to lose. We evaluated on multiple trials without feedback, to get multiple ``zero-shot'' measurements from each participant. (See SI \ref{supp_sec:behavioral} for experimental details.)

We compare human adaptation to that of a meta-mapping model trained on poker and four other card games. The specific rules vary from game to game. We created eight variations of each game, by applying any subset of three transformations, each of which could be learned as a meta-mapping (see SI \ref{meth_data_cards} for details). The most dramatic transformation is switching from trying to win to trying to lose. This variation requires completely inverting the strategy. We trained the network on 36 of the 40 basic tasks; all losing variations of poker were held out. We used the learned task representations to train meta-mappings for each of the three transformations. Two of the meta-mappings were trained using all five games, but the lose meta-mapping was trained only on the games other than poker. 

After training, the lose meta-mapping is applied to the task representation of poker, to transform it into a a task representation of losing at poker. This representation is then used to play the losing variation of poker. This evaluation exactly matches the evaluation of the human participants. 

For these tasks and the RL tasks (below), we must alter the representation of basic task examples, since rewards are observed only for the action taken. Instead of (input, target) examples we use (state, (action, reward)) examples (SI \ref{supp_sec:model_RL_modifications}). %The meta-mapping idea is quite general, and alterations like this do not substantially change it. %% TODO: could delete last line

See Fig. \ref{fig:HoMM_cards} for the results. Human subjects are not optimal at the game (mean performance 64\%, bootstrap 95\%-CI \([0.57, 0.70]\)), but are adapting well, at least in the sense that performance is similar in the losing variation on average (losing phase mean performance 64\%, bootstrap 95\%-CI \([0.55, 0.72]\)). However, there is substantial inter-subject variability in base task performance and adaptation. The evaluation hands were sampled in a stratified way in each phase, so this variability in adaptation is either due to randomness in participants' behavior (e.g. because they are probability-matching rather than optimizing bets), or in the way that their behavior changes between winning and losing phases. The meta-mapping model performs near optimally at the trained task, and adapts quite well (mean 85\%, 95\%-CI [79, 90]). In summary, the model performed differently than the human participants, but both the model and humans were able to switch from winning to losing zero-shot. See SI \ref{supp_sec:analyses:cards} for further analyses. 

%Both the language-based model and HoMM perform nearly optimally on the trained task. However, .\footnote{Language generalization was also similar in a non-hyper-network based architecture, see Fig. \ref{supp_fig:human_cards_lang_tcnh_vs_hyper}.} Intriguingly, this corresponds to behaving approximately randomly; performance would be worse if it did not adapt at all. By contrast, the HoMM model adapts quite well (mean 85\%, 95\%-CI \([79, 90]\)).  

\subsection*{Visual concepts}

\begin{figure*}[htb]
\centering
\begin{tikzpicture}[auto, scale=0.8, every node/.style={scale=0.8}]
\draw[boundingbox, draw=gray, fill=white] (-9.4, 2.7) rectangle (-0.1, 0);
\node[align=center] at (-4.7, 2.45) {\large Visual concept (basic task)};
\node at (-9, 1.23) {\Huge \color{bgreen} \bigcheckmark};
\draw[decoration={calligraphic brace, amplitude=0.4cm},decorate,line width=1mm] (-8.1, 0.2) -- (-8.1, 2.2);
\node at (-6.6, 1.21) {\includegraphics[width=1.5cm]{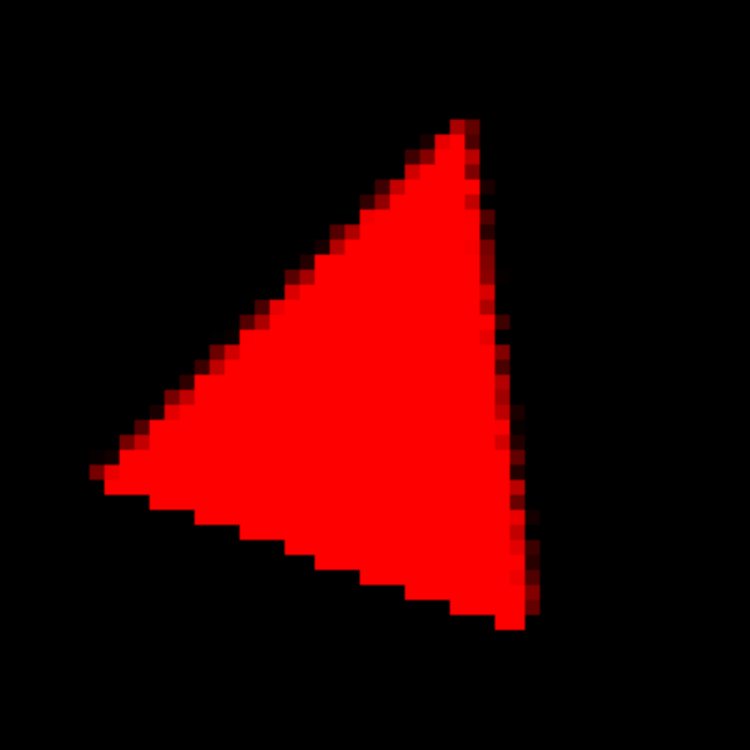} \includegraphics[width=1.5cm]{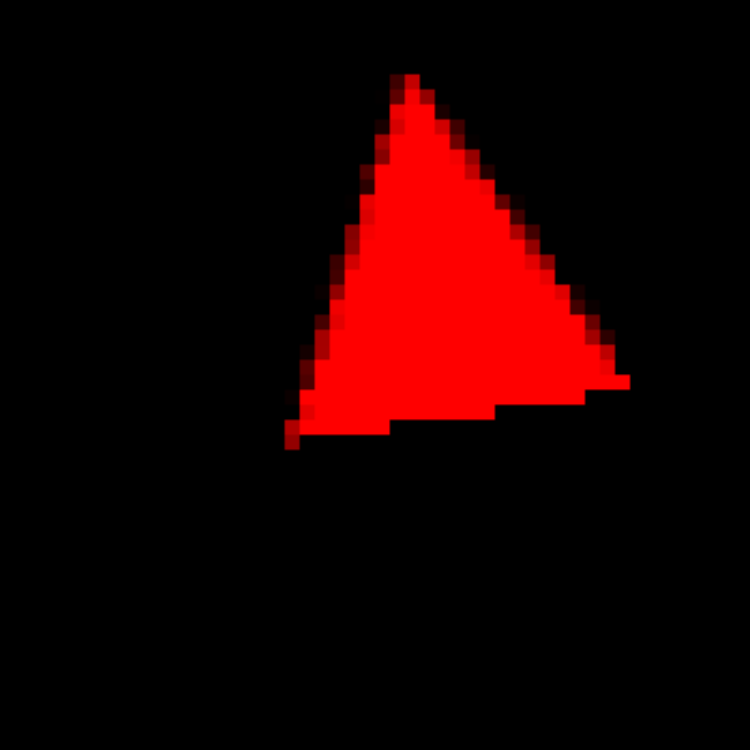}};

\node at (-4.3, 1.21) {\Huge \color{red}\(\bm \times\)};
\draw[decoration={calligraphic brace, amplitude=0.4cm},decorate,line width=1mm] (-3.4, 0.2) -- (-3.4, 2.2);
\node at (-1.9, 1.21) {\includegraphics[width=1.5cm]{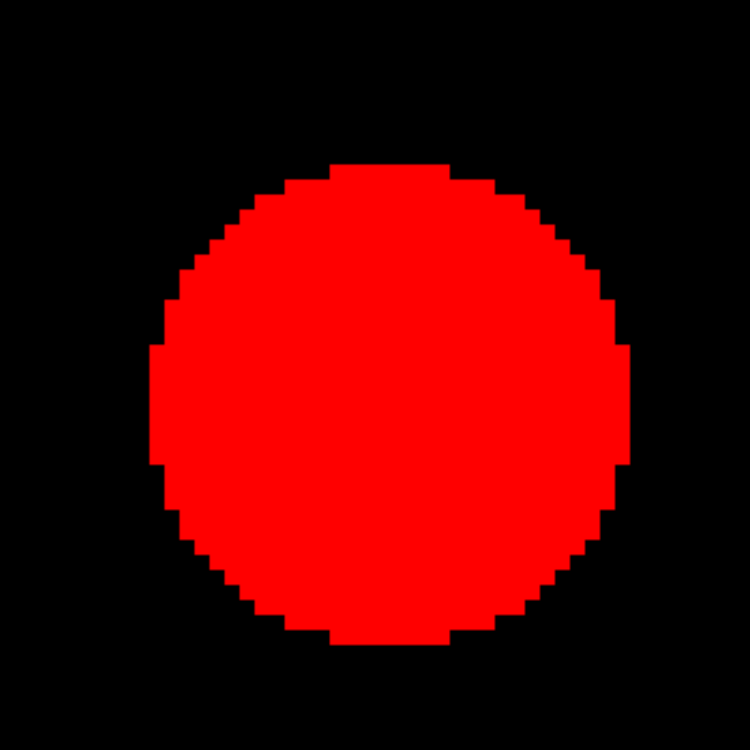} \includegraphics[width=1.5cm]{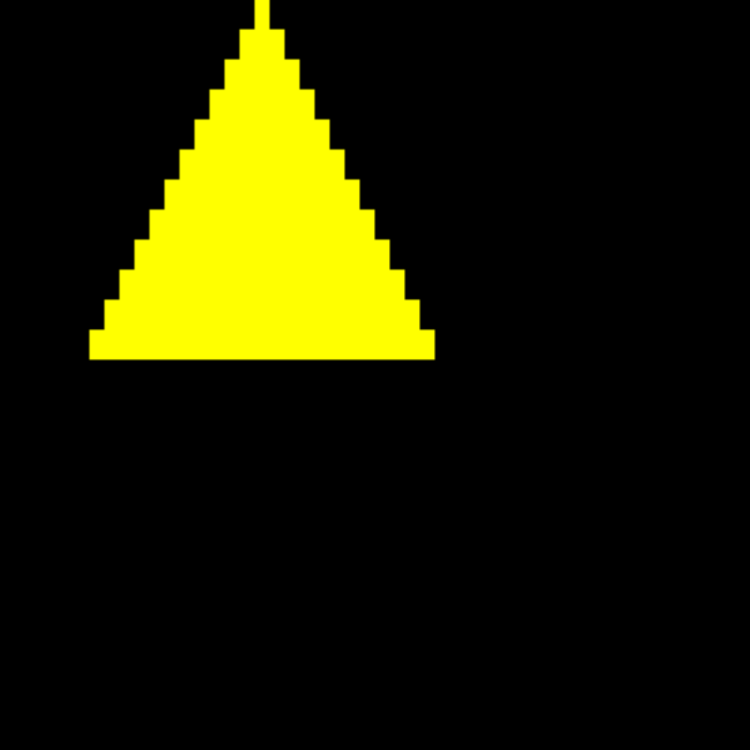}};

\begin{scope}[shift={(12.3, 0)}]
\draw[boundingbox, draw=gray, fill=white] (-9.4, 2.7) rectangle (-0.1, 0);
\node[align=center] at (-4.7, 2.45) {\large Transformed concept};
\node at (-9, 1.23) {\Huge \color{bgreen} \bigcheckmark};
\draw[decoration={calligraphic brace, amplitude=0.4cm},decorate,line width=1mm] (-8.1, 0.2) -- (-8.1, 2.2);
\node at (-6.6, 1.21) {\includegraphics[width=1.5cm]{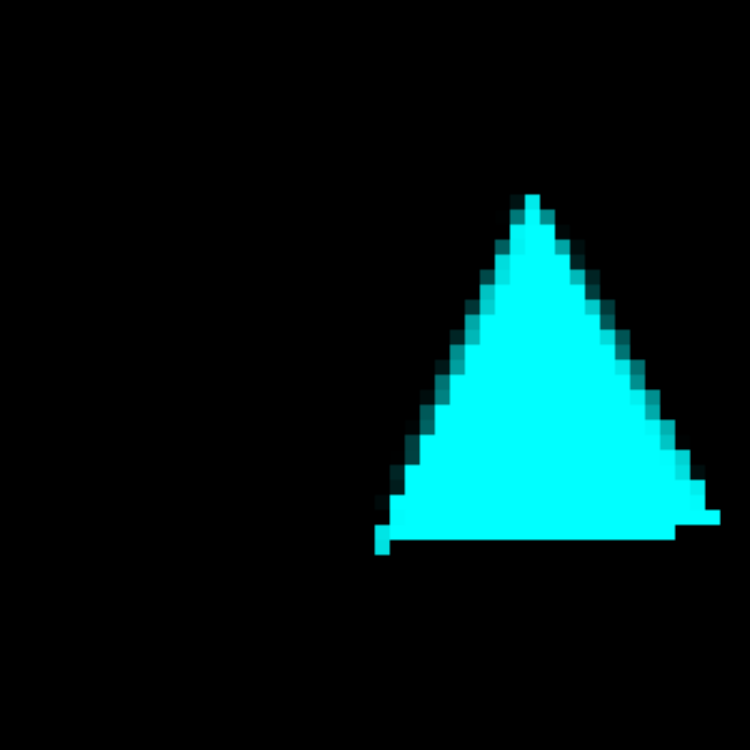} \includegraphics[width=1.5cm]{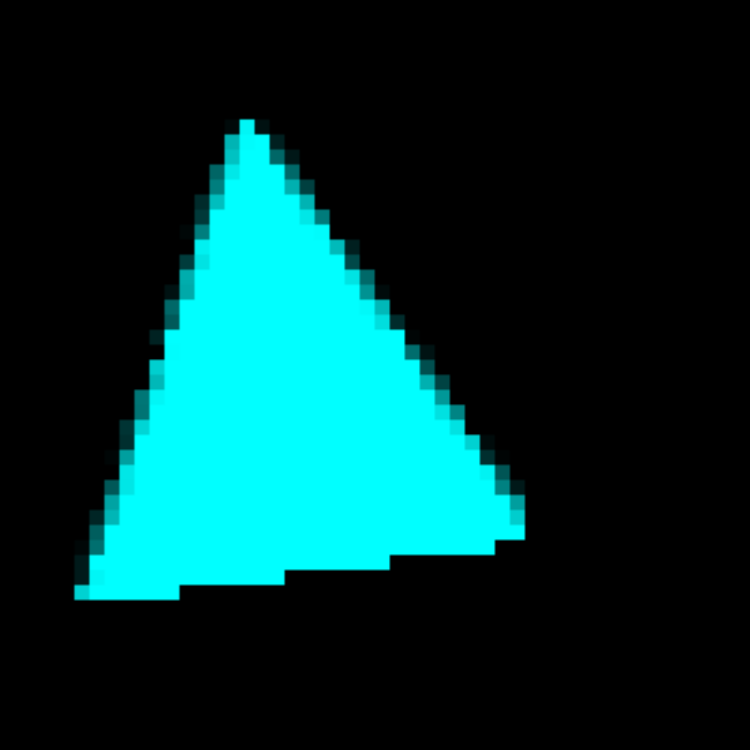}};

\node at (-4.3, 1.21) {\Huge \color{red}\(\bm \times\)};
\draw[decoration={calligraphic brace, amplitude=0.4cm},decorate,line width=1mm] (-3.4, 0.2) -- (-3.4, 2.2);
\node at (-1.9, 1.21) {\includegraphics[width=1.5cm]{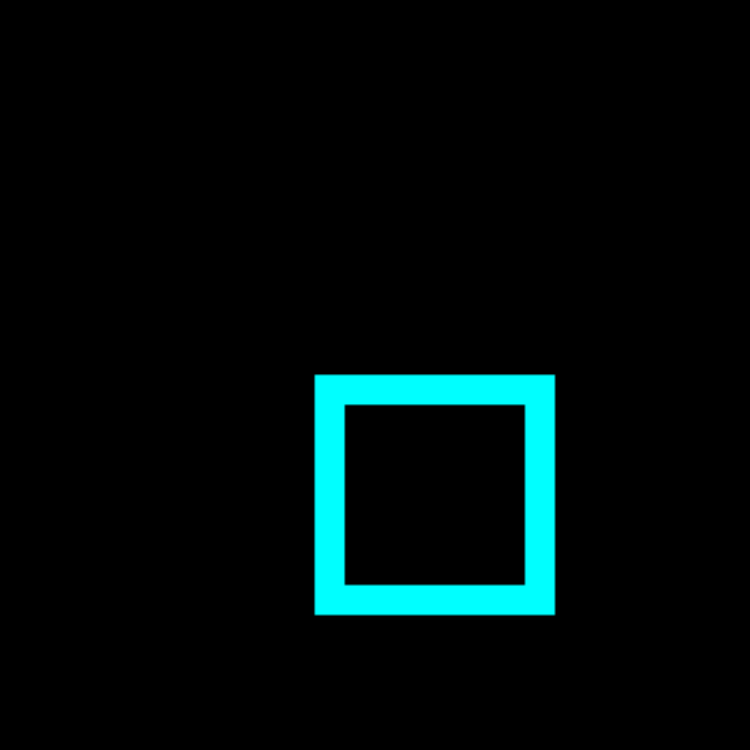} \includegraphics[width=1.5cm]{figures/categorization_stimuli/32_red_triangle_1.pdf}};
\end{scope}

\node at (-0.1, 1.21) (in) {};
\node at (2.9, 1.21) (out) {};
\draw[arrow, very thick, decorate, decoration={snake,aspect=0,amplitude=.1cm,segment length=1cm,post length=0.05cm}] (in) -- (out); 
\node[align=center] at (1.45, 2.2) {Meta-mapping};
\node[align=center] at (1.45, 1.7) {Switch red\(\rightarrow\)cyan};
\end{tikzpicture}
\caption{The visual concepts domain. Concepts consist of mappings from images to binary labels, e.g. 1 for images that are red AND triangle, 0 otherwise. These concepts can be transformed by meta-mappings that alter their attributes, such as switching red to cyan.} \label{fig:HoMM_visual_results:domain}
\vspace{-0.5em}
\end{figure*}

\begin{figure}[htb]
\centering
\includegraphics[width=0.45\textwidth]{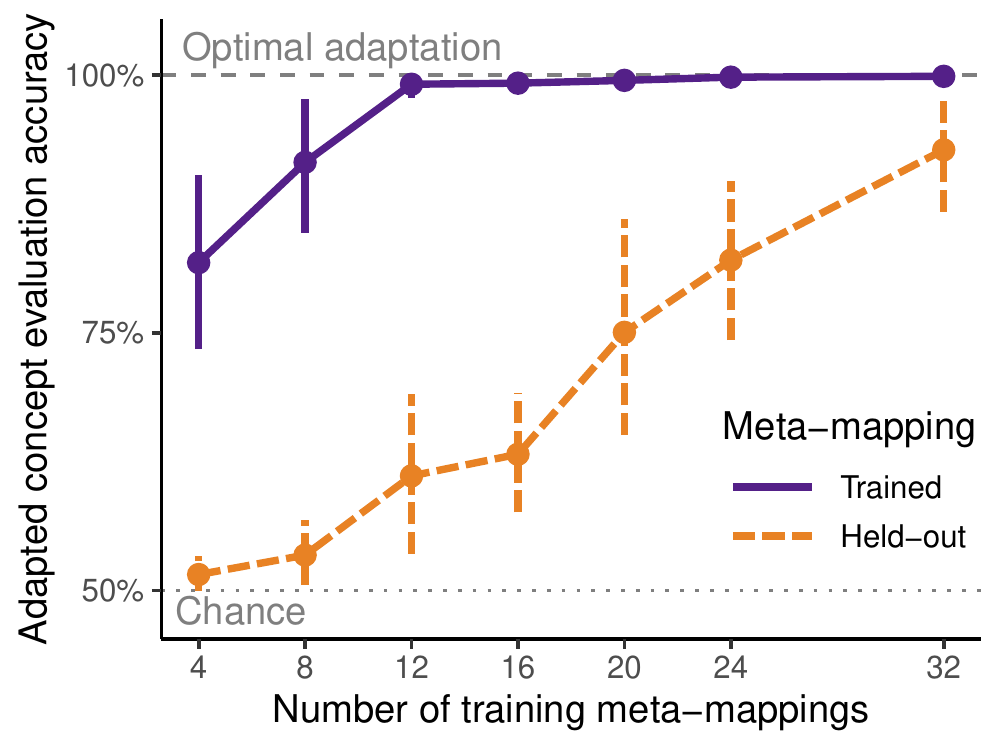}
\vspace{-0.5em}
\caption{Applying meta-mapping to visual concepts, after training the model on various numbers of training meta-mappings. The model is able to generalize trained meta-mappings to perform new tasks zero-shot. Furthermore, it can generalize to new meta-mappings once it experiences sufficiently many training meta-mappings. (Results are from 10 runs with each training set size. Error bars are bootstrap 95\%-CIs across runs.)}\label{fig:HoMM_visual_results:accuracy}
%\end{subfigure}%
%\begin{subfigure}{0.5\textwidth}
%\includegraphics[width=\textwidth]{figures/concepts_adaptation_sweeping_eval.pdf}
%\caption{Performance on held-out meta-mappings.}\label{fig:HoMM_visual_results:accuracy_eval}
%\end{subfigure}
%\caption{Applying HoMM to visual concepts (\subref{fig:HoMM_visual_results:domain}). Results are shown after training the model on various numbers of training meta-mappings (or the equivalent set of training concepts). HoMM and language generalization are well above chance (50\% with our balanced evaluation sets). (\subref{fig:HoMM_visual_results:accuracy}) HoMM is able to generalize trained meta-mappings to perform new tasks zero-shot, though both systems perform well. (\subref{fig:HoMM_visual_results:accuracy_eval}) HoMM is also able to generalize to adaptation based on held-out meta-mappings, once it experiences sufficiently many training meta-mappings. (Results are from 10 runs of each model with each training set size. Errorbars are bootstrap 95\%-CIs across runs.)} \label{fig:HoMM_visual_results}
\end{figure}

We next applied meta-mapping to visual concepts, a long-standing cognitive paradigm \citep[e.g.]{Bourne1970}. Past work has focused almost entirely on learning a concept from examples. However, adult humans can also understand some novel concepts without any examples at all. If you learn that ``blickets'' are red triangles, and then are told that ``zipfs are cyan blickets,'' you will instantly be able to recognize a zipf without ever having seen an example. This zero-shot performance can be understood as applying a ``switch-red-to-cyan'' meta-mapping to the ``blicket'' classification function (Fig. \ref{fig:HoMM_visual_results:domain}). To capture this ability, we applied meta-mapping. 

We constructed stimuli by selecting from 8 shapes, 8 colors, and 3 sizes. We rendered each item at a random position and rotation within a \(50 \times 50\) pixel image. We defined the basic concepts (basic tasks) as binary classifications of images (i.e. functions from images to \(\{0, 1\}\)). We trained the system on all uni-dimensional concepts (i.e. one-vs.-all classification of each shape, color, and size) as basic tasks, so that it could learn all the basic attributes. We also constructed composite basic tasks based on conjunctions, disjunctions, and exclusive-disjunctions (XOR) of these attributes. For example, one composite concept might be ``red AND triangle.''

For each concept, we chose balanced datasets of examples (that is, there was a 50\% chance that each stimulus was a member of the category), both during training and evaluation. We only included negative examples that were one alteration away from being a category member. These careful contrasts can encourage neural networks to extract more general concepts \citep{Hill2019}. %They also make the task more challenging, and the evaluation more informative. %% TODO: could delete last line

In this domain we constructed both the basic task and meta-mapping representations from language rather than examples (see Fig. \ref{fig:HoMM_architecture:constructing_basic},\subref{fig:HoMM_architecture:constructing_meta}), to show that meta-mapping can use this human-relevant cue. That is, there is no example network, instead a language network processes descriptions of tasks and meta-mappings to construct task and meta-mapping representations. 

We trained the system on meta-mappings that switched one shape for another, or one color for another. We sampled 6 composite concept transformation pairs that supported each mapping, and another 6 with held-out targets for evaluation. However, our task sampling meant that each held-out example had a closely matched trained example, unlike the other experimental domains. See SI \ref{meth_data_visual} for details of sampling.

We varied the number of meta-mappings trained, and evaluated the system on its ability to apply meta-mappings to trained source concepts in order to recognize the held-out target concepts. (Note: we exclude disjunctions from evaluation, because not adapting works fairly well on them.) Because there are many meta-mappings available, we were able to hold out one shape meta-mapping and one color meta-mapping for evaluation. The same basic concepts instantiating a held-out meta-mapping were trained as would be for a trained mapping, but the meta-mapping itself was not. This reduces possible confounds when evaluating meta-mapping generalization. 

%In this setting the HoMM and the language-generalization model perform comparably (Fig. \ref{fig:HoMM_visual_results:accuracy}). In a mixed linear model, language generalization results in very slightly worse generalization at moderate numbers of training mappings (\(-1.50\)\%, \(t(2612) = -2.775\), \(p =0.006\)), and a small interaction with number of training meta-mappings (\(-0.25\)\% per trained meta-mapping,  \(t(2617) = -4.26\), \(p < 0.001\)). (Effect of one additional trained mapping for HoMM \(1.00\)\%, \(t(6828) = -5.52\), \(p < 0.001\).) This comparable performance may be due in part to the fact that our task sampling guaranteed a training task close to each evaluation task in this setting, see the discussion and Fig. \ref{supp_fig:HoMM_concepts_random_tasks}. Both models show near-perfectly systematic generalization in many runs (Figs. \ref{supp_fig:HoMM_concepts_perfect}, \ref{supp_fig:HoMM_concepts_all_runs}). 

The model generalizes well (Fig. \ref{fig:HoMM_visual_results:accuracy}). On trained meta-mappings, its performance reaches close to ceiling around 12 training mappings. Furthermore, given enough training meta-mappings it is able to generalize well to held-out meta-mappings from a language description of that meta-mapping. This generalization improves rapidly as the number of meta-mappings trained increases. Although the average held-out meta-mappings performance is not perfect even at 32 training meta-mappings, it is perfect in 40\% of the runs (Fig \ref{supp_fig:HoMM_concepts_perfect}). %We find these results impressive, given that the model experiences at most 32 training meta-mappings --- performing held-out meta-mappings from language is at least as complicated as performing a held-out task from language, and we show below that language generalization on base tasks is often at chance with \(\sim\!30\) examples (e.g. with 36 card tasks). 

\subsection*{Reinforcement learning}

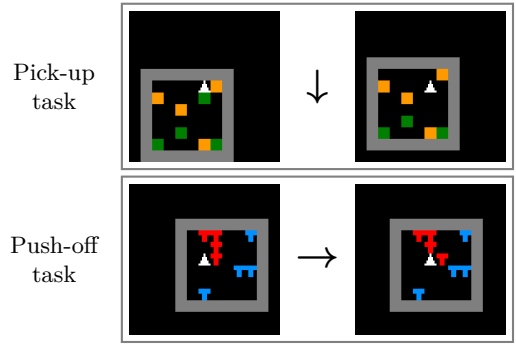
\begin{figure}[t]
\centering
\begin{tikzpicture}[auto]%, scale=0.8, every node/.style={scale=0.8}]
\draw[boundingbox, draw=gray, fill=white] (-7.1, 1.65) rectangle (-1.9, -0.55);
\node[align=center] at (-8, 0.55) {Pick-up\\task};
\node at (-6, 0.55) (pu1) {\includegraphics[width=2cm]{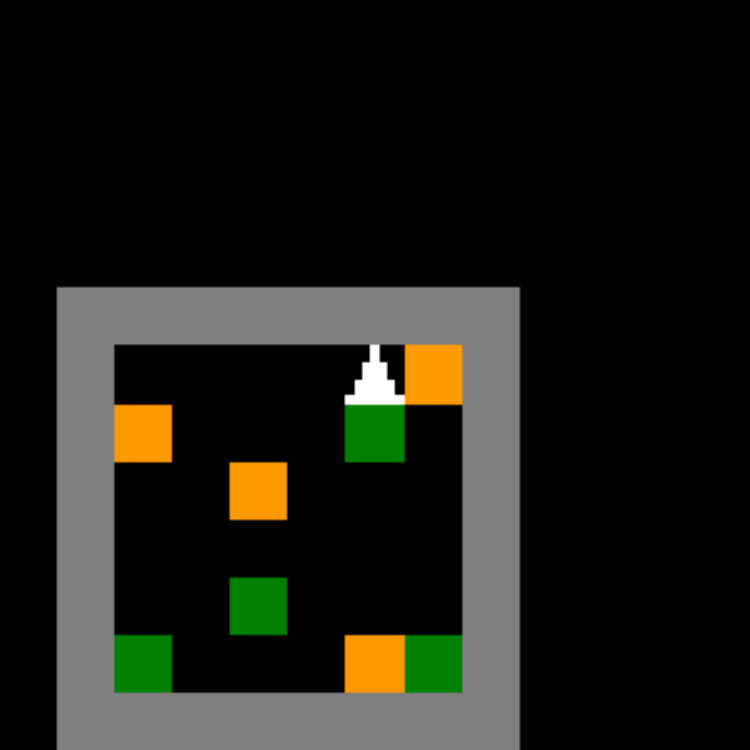}};
\node at (-3, 0.55) (pu2) {\includegraphics[width=2cm]{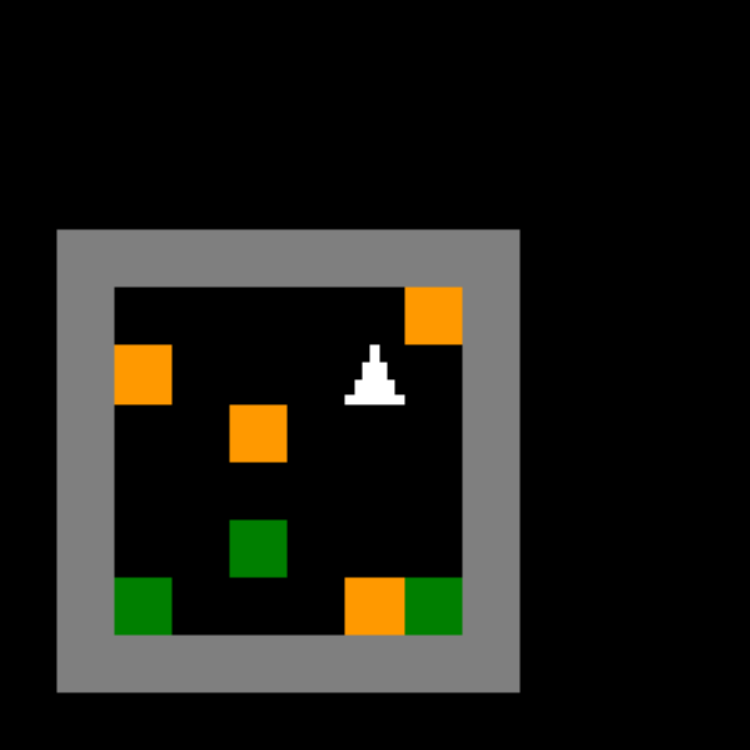}};
\node at (-4.5, 0.55) {\huge \(\bm \downarrow\)};
%\draw[arrow, very thick, gray] (pu1) -- node[black] {\huge``\(\bm \downarrow\)''} (pu2);

\begin{scope}[shift={(0, -2.3)}]
\draw[boundingbox, draw=gray, fill=white] (-7.1, 1.65) rectangle (-1.9, -0.55);
\node[align=center] at (-8, 0.55) {Push-off\\task};
\node at (-6, 0.55) (pu1) {\includegraphics[width=2cm]{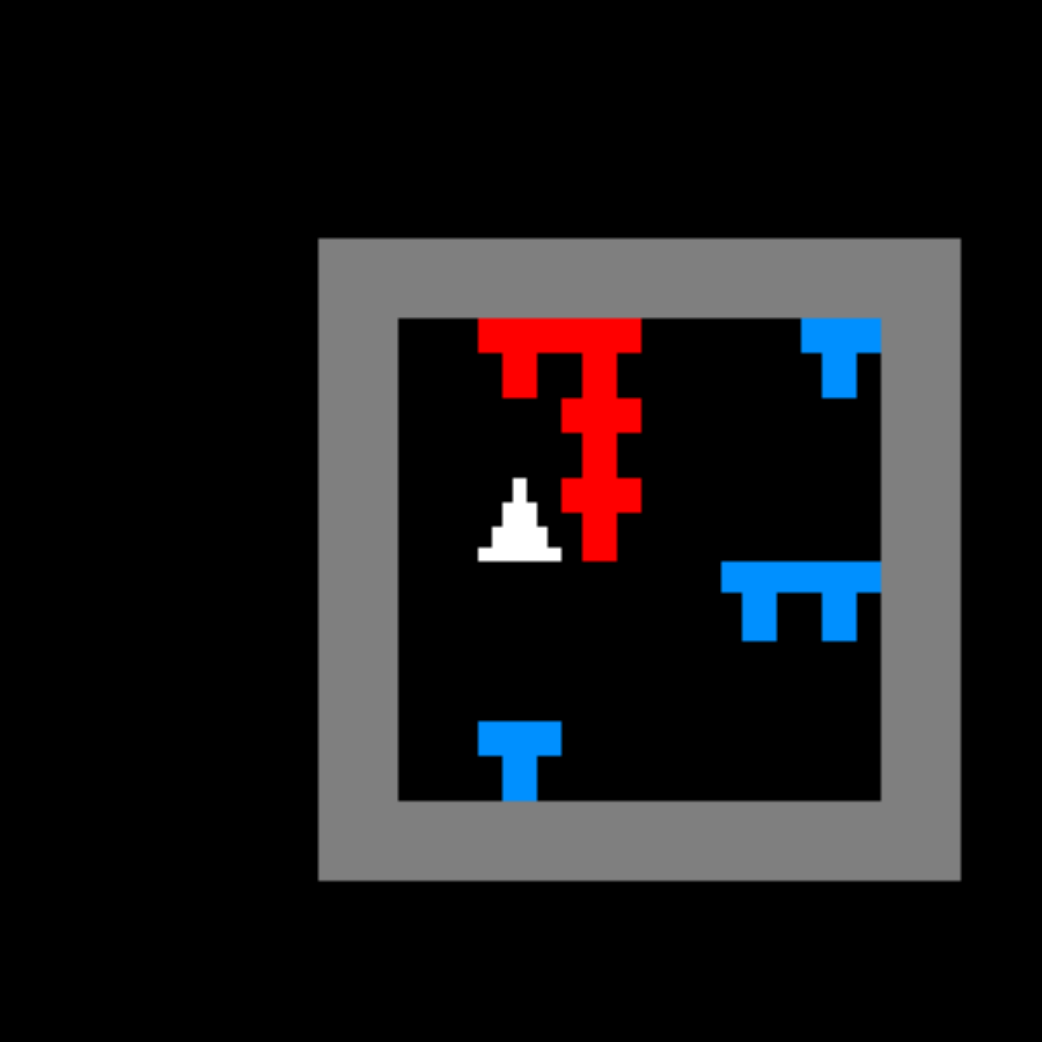}};
\node at (-3, 0.55) (pu2) {\includegraphics[width=2cm]{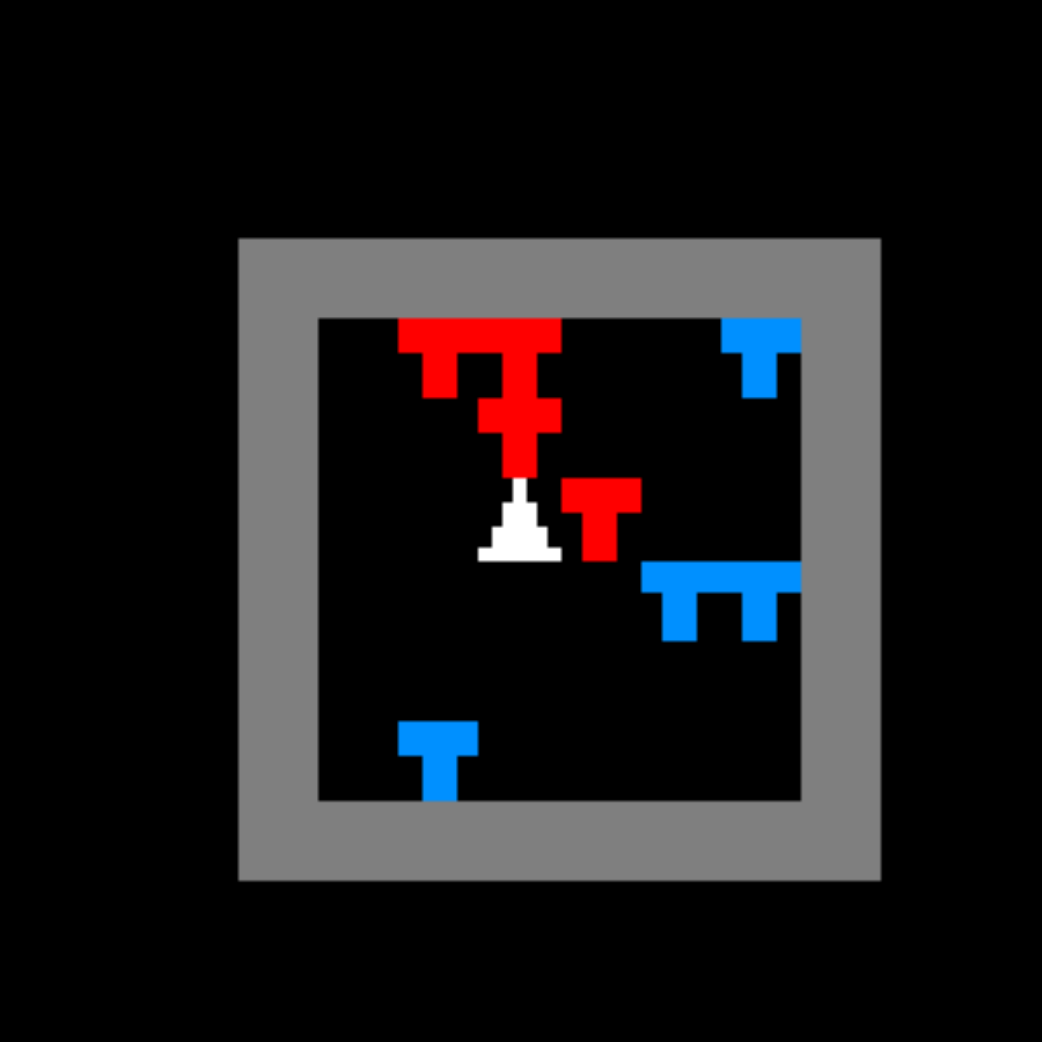}};
\node at (-4.5, 0.55) {\huge \(\bm \rightarrow\)};
%\draw[arrow, very thick, gray] (pu1) -- node[black] {\huge``\(\bm \downarrow\)''} (pu2);
\end{scope}
\end{tikzpicture}
\vspace{-0.5em}
\caption{Illustrative RL task state transitions. In the pick-up example (top), the agent moves down and picks up the green object. In the push-off example (right), the agent moves right and pushes the red object. Each image is precisely the visual input the agent would receive. Note that the agent is always centered (egocentric perspective). } \label{fig:HoMM_RL_domain}
\end{figure}

We next apply our approach to reinforcement learning (RL). RL-like computations relate to neural activity \citep{Niv2009, Dabney2020}, and RL has driven recent AI achievements in complex tasks like Go and StarCraft \citep{Silver2016, Vinyals2019}. Furthermore, RL requires sophisticated adaptation, since actions have lasting consequences. Thus, RL is an important testing domain for meta-mapping.  

Our RL tasks consist of simple 2D games (Fig. \ref{fig:HoMM_RL_domain}), which take place in a \(6 \times 6\) room with an additional impassable barrier of \(1\) square on each side. This grid is rendered at a resolution of 7 pixels per square to provide visual input to the agent. The agent receives egocentric input, i.e. its view is always centered on its position. This improves generalization \citep{Hill2019a}. The agent can take four actions, corresponding to moving in the four cardinal directions. Invalid actions, such as trying to move onto the edge of the board, do not change the state. 

The tasks the agent must perform relate to objects that are placed in the room. The objects can appear in 10 different colors. In any given task, the room only has two colors of objects in it. Each color of objects only appears with one other color, so there are in total 5 possible color pairs that can appear. In any given task, one of the present colors is ``good,'' and the other is ``bad.'' On some tasks, the good and bad colors in a pair are switched.

There are two types of tasks, a ``pick-up'' task, and a ``push-off'' task. In the pick-up task, the agent is rewarded for moving to the grid location of each good object, which then disappears, and is negatively rewarded for moving to the location of bad objects. In the push-off task, the agent is able to push an adjacent object by moving toward it, if there is no other object behind it. The agent is rewarded for pushing the good-colored objects off the edges of the board, and negatively rewarded for pushing the bad colored objects off. The two types of tasks (``pick-up'' and ``push-off'') are visually distinguishable, because the shape of the objects used for them are different. However, which color is good or bad is not visually discernible, and must be inferred from the example (state, (action, reward)) tuples used to construct the task representation.  

There are in total (2 task types) \(\times\) (5 color pairs) \(\times\) (binary switching of good and bad colors) \(= 20\) tasks. (See SI \ref{meth_data_RL} for further details of the task domain.) We trained the system on 18 tasks, holding out the switched color combinations of (red, blue) in both task types. That is, during training the agent was always positively rewarded for interacting with red objects and negatively rewarded for interacting with blue objects. We trained the system on the ``switch-good-and-bad-colors'' meta-mapping using the remaining four color pairs in both task types, and then evaluated its ability to perform the held-out tasks zero-shot based on this mapping. This evaluation is a difficult challenge, since the model was always negatively rewarded during training for interacting with the objects that it must interact with in the evaluation tasks. 

We evaluate the model for each task by requiring the training accuracy to be above a threshold, and selecting an optimal stopping time when the other task is performed well. We also used two minor model modifications to stabilize learning: persistent task representations (discussed above) and weight normalization. See SI \ref{supp_sec:model_RL_modifications} for details. Despite the challenging setting, the model adapts well, achieving 88.0\% of optimal rewards (mean, bootstrap 95\%-CI [75.0-99.0]) on the held-out pick-up task, and 71.7\% (mean, bootstrap 95\%-CI [42.0, 94.6]) on the held-out push-off task. The results are plotted in Fig. \ref{fig:HoMM_RL_results}, along with the results from the comparison models from the next section. (Intriguingly, the model also takes longer to complete generalization episodes, see Fig. \ref{supp_fig:HoMM_RL_behavioral_uncertainty}; perhaps humans, too, might be more hesitant in novel situations.) 

In SI \ref{supp_sec:analyses:RL_color_shape}, we show that meta-mapping is able to extrapolate meta-mappings beyond the dimensions it has been trained on, to transform new dimensions. Specifically, when trained with the switch-good-and-bad meta-mapping applied to colors, it can generalize to switching shapes. This is further evidence for the flexibility and systematicity of meta-mapping. 

\subsection*{Language \& meta-mapping}
\begin{figure}
\centering
\includegraphics[width=0.45\textwidth]{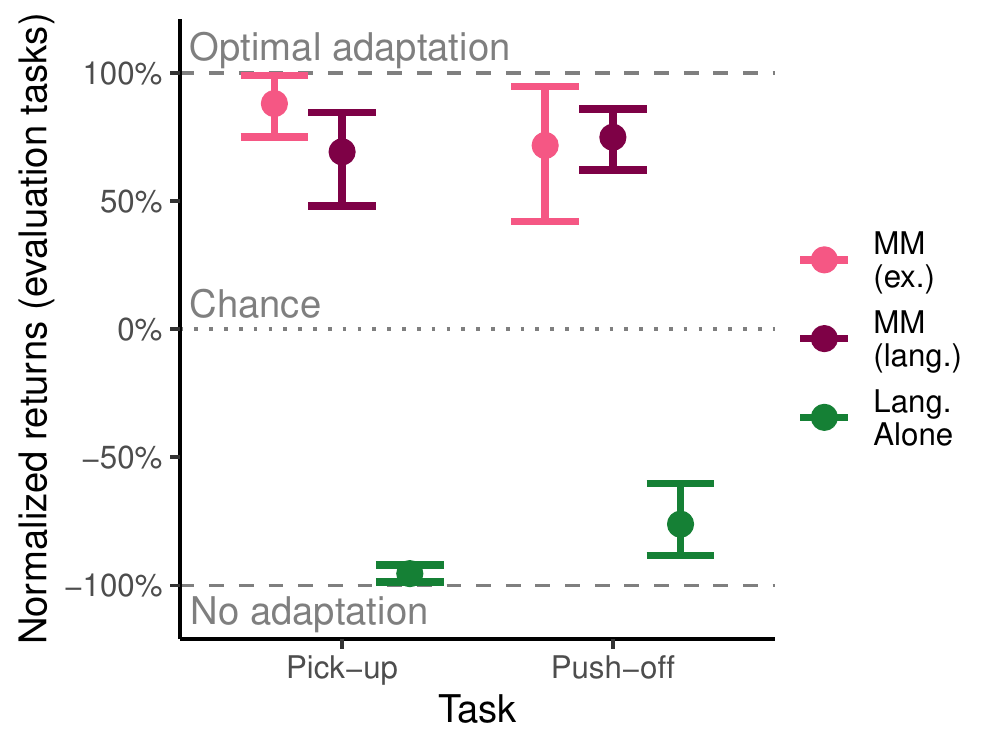}
\vspace{-0.75em}
\caption{Comparing RL adaptation performance when meta-mapping with task representations constructed from examples, when meta-mapping with task representations constructed from language, or when generalizing from language alone. Meta-mapping generalizes well with either type of task representation, while language alone generalizes poorly. (Chance refers to taking random actions.)}\label{fig:HoMM_RL_results}
\end{figure}

Language is often key to human adaptation, and prior work on zero-shot performance has often used a task description as input \citep[e.g.][]{Larochelle2008,Hermann2017,Hill2019a}. We showed in the visual concepts domain that language provides a suitable cue for basic tasks and meta-mappings; in this section we explore the relationship between language, examples, and meta-mapping further. We compare three approaches to zero-shot task performance in the RL domain: meta-mapping from examples (shown in the previous section), meta-mapping from language, and generalization from language alone.

First, we consider meta-mapping from language. We use language input both to generate task representations (e.g. ``pick-up, red, blue, first'' to indicate picking up objects, where the first color, red, is good) and as a cue for meta-mapping (``switch colors''). Applying this approach to the same training and hold-out setup used above for meta-mapping from examples yields comparable performance: 69.2\% (mean, bootstrap 95\%-CI [49.5, 84.5]) on the pick-up task and 74.9\% [60.9, 85.5] on the push-off task (Fig. \ref{fig:HoMM_RL_results}). This shows (as with the visual concepts) that generating task representations from examples is not essential --- language can support meta-mapping. 

However, a model that generates task representations from language offers an alternative approach to performing a new task zero-shot. If language descriptions systematically relate to tasks, the model should be able to generalize to new tasks from their description alone. If the system learns that ``green, yellow, first'' means that the objects will be green and yellow, and the first color (green) is good; that ``green, yellow, second'' means that yellow will be good; and that ``red, blue, first'' means that red will be good and blue bad; it could in principle generalize appropriately to ``red, blue, second.'' Indeed, this approach to zero-shot task performance has been demonstrated in prior work \citep{Hermann2017, Hill2019a}. However, we find that transforming the task representation via a meta-mapping can provide a stronger basis for adapting, compared to systematic language alone.

To demonstrate this, we compare the example- and language-based meta-mapping approaches to generalizing from language alone, again using the same basic tasks to train the network to perform tasks from language, but without meta-mapping training (Fig. \ref{fig:HoMM_RL_results}). Performing the new tasks from language alone results in very poor generalization performance: -92.8\% (mean, bootstrap 95\%-CI [-96.3, -88.4]) on the pick-up task and -79.7\% [-92.8, -59.1] on the pusher task. Meta-mapping provides much better generalization. 

The direct comparison between language-based meta-mapping and language alone shows that meta-mapping is beneficial, but there are two mechanisms by which it could help. Meta-mapping at test time could be key to generalization, or meta-mapping training could simply improve the learning of the basic task representations, such that even language alone would allow good generalization \emph{in a meta-mapping trained model}. However, language-alone generalization is not significantly improved even in the language-based meta-mapping model (see SI \ref{supp_sec:analyses:language} for results and discussion), suggesting that meta-mapping at test-time is key to the benefits we observe. %See the \textbf{Discussion} for a possible explanation of why meta-mapping is beneficial. %% TODO: update

We also compared meta-mapping to language alone in the cards and visual concepts domains. We summarize the results here, see SI \ref{supp_sec:analyses:language} for details. In the cards domain, the language based model was not able to generalize well to the losing game, instead degrading to chance-level performance (Fig. \ref{supp_fig:lang_cards}). In the visual concepts domain, by contrast, the language model generalizes comparably to meta-mapping (Fig. \ref{supp_fig:lang:visual}). This may be due to the concept sampling --- each evaluation concept had several closely-related training concepts, unlike the other domains. Indeed, meta-mapping shows a greater advantage when new concepts are less similar to trained ones (Fig. \ref{supp_fig:HoMM_concepts_random_tasks}).

In summary, meta-mapping (from examples or language) outperforms or equals language alone in all our experiments. Meta-mapping is especially beneficial when the task space is sparsely sampled or generalization is challenging. We consider the advantage of meta-mapping further in the \textbf{Discussion}. 

\subsection*{Meta-mapping as a starting point for later learning}

%Why is zero-shot adaptation useful? One reason is that it provides a starting point for later learning. Although meta-mapping may result in only an approximation of the correct behavior, this makes learning much easier than starting from scratch. 
Zero-shot adaptation by meta-mapping allows a model to perform a new task without any direct experience. However, as we have seen, zero-shot performance is not always as good as the ultimate performance after training on the task. Here, we show that even if zero-shot performance is not completely optimal, it makes learning much faster than starting from scratch. We also show that this learning can be done in a way that avoids interference with performance on prior tasks.

We return to the polynomials domain to demonstrate this. We reinstate a trained model, and consider how it could learn on the held-out tasks once it encounters them. To do so, we optimize the representations of the new tasks in order to improve performance on those tasks, without allowing any of the network weights to change (see SI \ref{supp_sec:model_optimizing_details}). This approach can improve performance on the new tasks without the possibility of interfering with prior knowledge \citetext{\citealp{Reed2015}; c.f. \citealp{Rogers2004, Lampinen2018a}}. Thus it provides a useful approach to learning after zero-shot adaptation, once the system is actually performing the new tasks.

\begin{figure}
\centering
\includegraphics[width=0.45\textwidth]{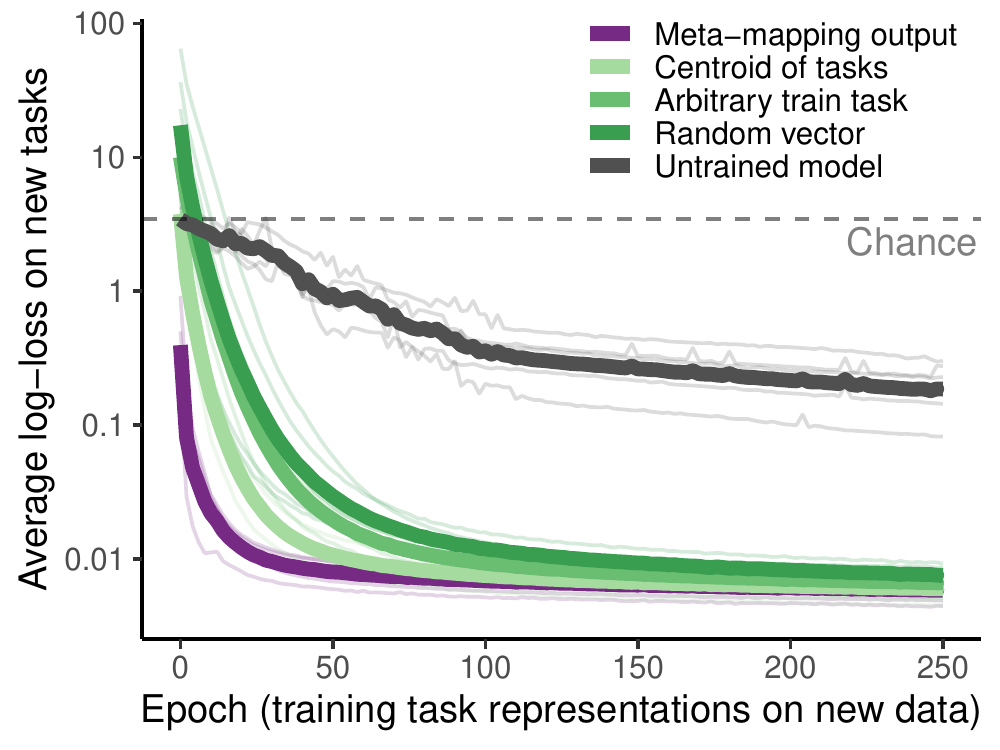}
\vspace{-0.75em}
\caption{Meta-mapping provides a good starting point for later learning. This figure shows learning curves (mean log-loss) while optimizing task representations on new tasks in the polynomials domain. Using meta-mapping as a starting point offers much lower initial loss, and results in faster learning than alternative initializations. (Light curves are 5 individual runs, thick curves are averages.)} \label{fig:HoMM_timescales_results}
\end{figure}

We evaluate a variety of starting points for initializing the new task representations. We compare initializing via meta-mappings to a variety of reasonable alternatives, such as small random values (the standard in machine learning), the embedding of an arbitrary trained task, and the centroid of all trained task representations. We plot learning curves from these different initializations in Fig. \ref{fig:HoMM_timescales_results}. Producing an initial task representation by meta-mapping results in much lower initial loss and faster learning than any other method.

To quantify this, we consider the cumulative loss over learning, i.e. the integral of the learning curves. This measures how much loss the model had to suffer in order to reach perfect behavior on the new tasks. Starting from a meta-mapping results in almost an order of magnitude less cumulative error (mean \(= 24.58\), bootstrap 95\%-CI \([17.71, 32.08]\)) than the next best initialization (centroid of trained task representations, mean \(= 192.89\), bootstrap 95\%-CI \([151.98, 234.53]\)). Meta-mapping provides a valuable starting point for future learning. (We also show this in the visual concepts domain, in Fig. \ref{supp_fig:optimizing_curves_categories}, and show that a hypernetwork architecture is essential, Figs. \ref{supp_fig:timescales_polynomial_optimization_tcnh_curves}, \ref{supp_fig:timescales_polynomial_optimization_tcnh_regret}.) 

\section*{Discussion}\label{sec:HoMM_discussion}

We have proposed meta-mappings as a computational mechanism for performing a novel task zero-shot --- without any direct experience on the task --- based on the relationship between the novel task and prior tasks. We have shown that our approach performs well across a wide range of settings, often achieving 80-90\% performance on a new task with no data on that task at all. With enough experience, as in the visual classification settings with enough training tasks, it can adapt perfectly. It can also adapt using novel relationships (held-out meta-mappings) that it has never encountered during training.

As noted in the introduction, there are computational benefits to adaptivity. Its potential contributions to biological intelligence have been highlighted by Siegelmann \citep{Siegelmann2013}, who proposes that there is a ``hierarchy of computational powers'' and that a particular system's location in that hierarchy depends on its ``particular level of richness and adaptability.'' Because our work offers a new perspective on adaptation, it would be interesting to explore the theoretical computational power of meta-mapping under different input and representation regimes.

As Siegelmann notes, for a model to be able to adapt, it must first be capable of performing a variety of related tasks \citep{Siegelmann2013}. Thus, instead of learning parameters that execute a single task, our model learns to construct task representations from examples or language, and to use those representations to perform appropriate behaviors. The key insight of this work is that those task representations are then available for transformation, and that transforming task representations by meta-mappings can allow effective adaptation. 

In our experiments, directly exploiting task relationships by meta-mapping allowed more systematic adaptation than indirectly exploiting them by generalizing through compositional language alone. Even when language alone generalized poorly, as in the RL domain, meta-mapping with language-based task representations resulted in strong generalization. This illustrates the value of a transformation-oriented perspective.

Why is transforming tasks according to task relationships so effective? We suggest that this is because meta-mapping constructs and uses an explicit cognitive operation that captures what is systematic in the task relationships. For example, “trying to lose” is systematic precisely insofar as the relationship between winning and losing is similar across different games. The meta-mapping approach gives primacy to these relationships. It thus directly exploits systematic structure where it exists in the cognitively-meaningful relationships between tasks. % and the concepts that underly cognitively meaningful task specifications.

We also highlight the results showing that meta-mapping provides a useful starting point for later learning. While meta-learning approaches \citep[e.g.][]{Finn2017a} can construct a good starting point for learning new tasks, they do not use task relationships to offer a uniquely appropriate starting point for each novel task. Our results show that using a task relationship to adapt a prior task can substantially reduce the errors made along the way to mastering the new task. This could make deep learning more efficient. It could also be useful in settings like robotics, where mistakes during learning can be extremely costly \citep{Turchetta2016}. 

Our results should not be taken as a suggestion that meta-mapping is the only possible mechanism for adaptation. We see intelligence as multi-faceted, and any single model is a simplification. Meta-mapping may be useful as one tool for building models with greater flexibility.

Meta-mapping increases the adaptability of our models, although our present work has limitations that we discuss below. Our models can perform tasks from examples, from natural language, and from meta-mappings, which we have shown are an effective way to adapt zero-shot. Thus our work has many potential applications in machine learning and cognitive science.

\subsection*{Related work in machine learning}

To allow zero-shot adaptation, we built on ideas from several areas of machine learning. First, there is a large body of prior work on allowing models to learn to behave more flexibly, for example by meta-learning, that is, learning-to-learn from examples \citep[e.g.][]{Vinyals2016, Finn2017a, Ravichandran2019}. Our approach to inferring tasks from examples draws on recent ideas like aggregating examples in a permutation-invariant way to produce a task representation \citep{Garnelo2018}.  

Second, a range of work has highlighted the idea of different timescales of weight adaptation --- that is, even if some parameters of a network may need to be learned slowly, it may be useful to alter others much more rapidly \citep{Hinton1982}. We have drawn particularly on the idea that the parameters of a network could be specified by another network, in a single forward inference \citep{McClelland1985,Ha2016}. This approach has shown success in meta-learning recently \citep[e.g.][]{Li2019a, Rusu2019}, and improved our model's adaptation (Fig. \ref{supp_fig:HoMM:nonhomoiconic_baseline}).  

There has been a variety of other work on zero-shot task performance. We compared to the zero-shot task performance from language alone. The idea of performing tasks from descriptions was proposed by Larochelle et al. \citep{Larochelle2008}. More recent work has considered zero-shot classification using language \citep{Socher2013,Xian2018}, or performing tasks from language in RL \citep{Hermann2017, Hill2019a}. Some of this latter work has even exploited relationships between tasks as a learning signal \citep{Oh2017a}, but without transforming task representations. As discussed above, transforming task representations with meta-mappings directly exploits systematic relationships, allowing meta-mapping to outperform language alone in our experiments. To our knowledge none of the prior work has proposed task transformations to adapt to new tasks.  

Other prior work has used similarity between tasks to help generate representations for a new task \citep{Pal2019}. Again, meta-mapping may be a stronger approach, since it can specify \emph{along which dimensions} two tasks are related, and the specific ways in which they differ, which a scalar similarity measure cannot.

%Our work is also related to the rapidly-growing literature on meta-learning \citep[e.g.][]{Vinyals2016, Santoro2016, Finn2017a, Ravichandran2019}. Our architecture builds off of HyperNetworks \citep{Ha2016} and other recent applications thereof . 
%In particular, recent work in natural language processing has shown that having contextually generated parameters can allow for zero-shot task performance, assuming that a good representation for the novel task is given \citep{Platanios2017} -- in their work this representation was evident from the factorial structure of translating between many pairs of languages. 
%It is also related to work on different time-scales of weight adaptation \citep{Hinton1982}, and its recent applications in meta-learning \citep[e.g.][]{Ba2016,Garnelo2018}. % and continual learning \citep[e.g.]{Hu2019}. It is more abstractly related to work on learning to propose architectures \citep[e.g.][]{Zoph2016, Cao2019}, and to models that learn to select and compose skills to apply to new tasks \citep[e.g.][]{Andreas, Andreas2016, Tessler2016, Reed2015, Chang2019a}. 
%Some work in visual question answering has explored building a classifier conditioned on a question \citep{Andreas}, which is related to our visual-categorization approach.

Aspects of zero-shot adaptation have also been explored in model-based reinforcement learning. Work in model-based RL has partly addressed how to transfer knowledge between different reward functions \citep[e.g.][]{Laroche2017}. Meta-mapping can potentially be applied to this form of transfer as well; indeed, our RL experiments show that meta-mapping can offer a model-free alternative to model-based adaptation. Meta-mapping may also offer advantages that could complement model-based methods. Meta-mapping provides a principled way to infer a new reward estimator by transforming a prior one. It could also transform a transition function used in the planning model in response to environmental changes. Thus, exploring the relationship and synergies between meta-mapping and model-based RL methods provides an exciting direction for future work.

There has also been other recent interest in task representations. Achille et al. \citep{Achille2019} proposed computing embeddings for visual tasks from the Fisher information of a task-tuned model. They show that this captures some interesting properties of the tasks, including some semantic relationships, and can help identify models that can perform well on a task. Other recent work has tried to learn representations for skills \citep[e.g.][]{Eysenbach2019} or tasks \citep[e.g.][]{Hsu2019} for exploration and representation learning, but without exploring zero-shot transformation of these skills.

\subsection*{Related work in cognitive science}
Our work is related to several streams of research in cognitive science. Prior work has suggested that analogical transfer between structurally isomorphic domains may be a key component of ``what makes us smart'' \citep{Gentner2003}. Analogical transfer is a kind of zero-shot mapping, and has been demonstrated across various cognitive domains \citep[e.g.][]{Bourne1970, Gick1980}. We hope our work stimulates further exploration of the conditions under which humans can adapt to task transformations zero-shot. Different types of task relationships might be made accessible through culture or education --- ``relational concepts are not simply given in the natural world: they are culturally and linguistically shaped'' \citep{Gentner2003}.  

Our work also touches on complex issues of compositionality, productivity, and systematicity. Fodor and others have advocated that cognition must use compositional representations in order to exhibit systematic and productive generalization \citep[e.g.][]{Fodor2001, Fodor2008lot2, Lake2017}. We see our work as part of an alternative approach to this issue, exploring how systematic, structured generalization can instead emerge from the structure of learning experience, without needing to be built in \citep{McClelland2010, Hansen2017}. By focusing on task relationships, rather than building in compositional representations of tasks, our model can learn to exploit the shared structure in the concept of ``losing'' across a few card games to achieve 85\% performance in losing a game it has never tried to lose before.

Crucially, the question of whether the model adapts according to compositional task structure is distinct from the question of whether the model's representations exhibit compositional structure. Because the mapping from task representations to behavior is highly non-linear, it is difficult to craft a definition of compositional representations that is either necessary or sufficient for generalization. For example, if ``compositional'' is taken to mean that Euclidean vector addition of the representations of two constant polynomials results in the representation of their sum, this is clearly untrue for our model (e.g. Fig. \ref{supp_fig:HoMM_polynomials:reps_const_poly_PCA}). However, the non-linear mapping from representations to behavior can allow for systematic generalization from non-linear structure. Indeed, it appears that the constant polynomial representations may be approximately systematically arranged in a compressed polar coordinate system. This may support generalization better than a more intuitively compositional representational structure.

Furthermore, there are a number of potential benefits to letting systematic behavior emerge, rather than attempting to build in compositional representations. First, the structure does not need to be hand-engineered separately for each domain. Our system required no special knowledge about the domains beyond the basic tasks and the existence of relationships between them. The fact that some of these relationships corresponded to e.g. permutations of variables in the polynomial domain did not need to be hard-coded; instead, the model was able to discover the nature of this transformation from the data (in that it was able to generalize well to held-out permutations). Emergence may also allow for novel decompositions at test time. The ability of our model  to perform well on held-out meta-mappings indicates that it has some promise in this regard. Future work should assess this capability of the model more fully. 

We also believe that our approach can capture some of the recursive processing that Fodor and others have emphasized \citep[e.g.][]{Fodor2008lot2}. We have also been influenced by ideas in mathematical cognition about how concepts build upon more basic concepts \citep{Wilensky1991, Hazzan1999, Lampinen2017b}. This recursive construction reflects the way that meta-mappings transform basic tasks --- complex transformations are built upon simpler ones. If humans can handle an indefinite number of levels of abstraction, the advantage of using a shared representational space for all levels increases, since it eliminates the need to create a new space for each level. Relatedly, our shared workspace for data points, tasks, and meta-mappings connects to ideas like the Global Workspace Theory of consciousness \citep{Baars2005}. The ability to reason about and explain concepts at different levels of abstraction can be explained parsimoniously by assuming a shared representational space. Exploring these connections would be an exciting future direction. 

We found particular inspiration in Karmiloff-Smith's work on re-representing knowledge \citep{Karmiloff-Smith1986, Clark1993}. It would be interesting to explore modeling the phenomena she considered, which she argued required that representations be ``objects for further manipulation,'' as task representations are in meta-mapping.

Our work also relates to Fodor's ideas about the modularity of the mind. %Two examples are his view of mental processes as ``transforming internal representations,'' and his argument that what is accessible about the stimulus is only ``what is given in [...] its \emph{proximal} representations'' \citep[][pp. 200-201]{fodor1975language}. 
Indeed, our division of the architecture into input and output systems, with the flexible, task-specific computations in the middle, may seem very reminiscent of the modularity that he advocated \citep{fodor1983modularity}. However, we chose this implementation for simplicity--- we believe that in reality processes such as perception can be influenced by the task, as well as contextual constraints \citep{McClelland2014}.

Reciprocally, we believe that higher-level computations are influenced and constrained by the modalities in which they are supported. This computational feature can emerge in our model; despite the fact that different types of data and tasks are embedded in a shared latent space, the model generally learns to organize distinct types of inputs into somewhat distinct regions of this space. This means that the task-specific processing can potentially exploit domain-specific features of the input, as for example humans do when they use gestures to think and learn in spatial contexts like mathematical reasoning \citep{Goldin-Meadow1999}. At the same time, the shared space can allow a graded overlap in the structure that is shared across different entities, insofar as they are related to each other. For example, in the polynomial domain there is more overlap between polynomial representations and meta-mapping representations than between either type of representations and the representations of numerical inputs. Using a shared space allows the model to discover what should be shared and what should be separated --- that is, ``modularity may not be built in [but] may result from the relationship among representations'' \citep{Tanenhaus1987}.

Finally, our approach relates to earlier work on cognitive control \citep{Cohen1990}. %A failure to meta-map perfectly could capture some failures of control. %, as could contamination of task examples with other tasks. 
The ``default'' task-network weights could be used to model more automatic processing. This processing can be overridden by task-specific constraints set by the HyperNetwork, when conditioned on an appropriate task representation. We provide a simple implementation of these ideas in SI \ref{supp:HoMM_cognitive_control}. Meta-mapping itself could also be relevant, for example an imperfect meta-mapping might capture some failures of control.

\subsection*{Limitations \& future directions}

Although we believe our approach is promising, the present work has limitations. We have explored meta-mapping within a limited range of settings. While we used one particular model, meta-mapping could potentially be useful in any meta-learning approach that uses task representations \citep[e.g.][]{Rusu2019}. Furthermore, we have only demonstrated our model within relatively simple, small domains. The model adapts quite well, but does not always achieve perfect fidelity of adaptation. One factor that may contribute is the relatively limited range of experience of the model -- our models lack the rich lifetime of experience that our human participants have. Furthermore, recent work shows that more realistic and embodied environments can improve generalization \citep{Hill2019a}. %Both of these restrictions on the model's experience affect our comparisons as well; language generalization would presumably also improve with more experience and grounding. 
Thus, evaluating our approach in richer, more realistic settings, will be an important future direction.%In addition, it would be useful for future work to explore human cognitive flexibility in greater detail, to evaluate the circumstances under which humans can effectively adapt. 

Another important limitation is that our approach requires the imposition of structured training to provide the network with experience of the relationships between tasks. However, we suggest that identifying task relationships is useful for building more flexible intelligent systems, and that exposure to task relationships is an important part of human experience. A long-term goal would be to create a system that learns to identify task relationships for itself from such experience.   %% TODO: update

Our work suggests many other possibilities. For simplicity we considered using language, examples, and meta-mapping to infer task representations in this work. However, it would likely be beneficial to use multiple constraints to both infer and adapt task representations. Furthermore, we considered language as input, but producing language as output (in the form of explanations) can improve understanding and generalization in both humans \citep{Chi1994} and neural networks \citep{Mu2019}. %While classifying task representations may provide a small part of this benefit,
Adding language output would likely improve performance and better capture the structure of human behavior.

In addition, we did not thoroughly explore robustness and the effect of noise. We showed that our model is reasonably robust to sample-size variability (Figs. \ref{supp_fig:HoMM:polynomials_varying_mbs_base}, \ref{supp_fig:HoMM_polynomials_varying_mbs_meta}), but future work should explore the robustness of our approach more thoroughly. For example, how would input noise affect the computations? How would errors compound if multiple meta-mappings were applied sequentially?

%A number of architectural choices could also be altered. We explored a few possibilities, such as showing that a simpler task-performance architecture without a hyper network was deleterious. However, many other alterations could be explored in future work. 
%We also noted in the visual categories domain that linear task networks seemed to improve meta-mapping, while nonlinear ones seemed to result in better basic task performance (Fig. \ref{supp_fig:HoMM_concepts_lang_arch}) --- thus it might be reasonable to consider a deep, nonlinear task network, but with a linear skip-connection from beginning to end. 
Our model architecture also has limitations; cognitive tasks often require more complex processing than our model allows. Replacing the feed-forward task network with a recurrent or attentional network --- or a network with external memory \citep[e.g.][]{Graves2016} --- would increase the flexibility of the model. It will be important to incorporate these ideas in future work. 

\subsection*{Conclusions} %% TODO: update

An intelligent system should be able to adapt to novel tasks zero-shot, based on the relationship between the novel task and prior tasks. We have proposed a computational implementation of this ability. Our approach is based on constructing task representations, and learning to transform those task representations through meta-mappings. We have also proposed a homogeneous implementation that reuses the same architectures for both basic tasks and meta-mappings. We see our proposal as a logical development from the fundamental idea of meta-learning --- that tasks themselves can be seen as data points in a higher-order task of learning-to-learn. This insight leads to the idea of transforming task representations just like we transform data. 

Meta-mapping is an extremely general approach --- we have shown that it performs well across several domains and computational paradigms, with task representations constructed from either examples or language. Meta-mapping is able to perform well at new tasks zero-shot, even when the new task directly contradicts prior learning. It is generally able to adapt more effectively after experiencing fewer tasks than approaches relying on language alone and sometimes seems to exhibit more systematic behavior. We suggest that this is because task relationships better capture the underlying conceptual structure. Meta-mapping provides a valuable starting point for later learning, one that can substantially reduce both time to learn a new task and cumulative errors made in learning. Our results thus provide a possible mechanism for an advanced form of cognitive adaptability, and illustrate the role it may play in future learning. We hope our work will lead to a better understanding of human cognitive flexibility, and the development of artificial intelligence systems that can learn and adapt more flexibly.

%\matmethods{We present some details of the tasks and models here. Full model implementation details are described in SI \ref{supp_sec:methods}, which also includes links to repositories containing all code for the experiments.
%
%\subsection**{Polynomials}
%We sampled the polynomials by first uniformly sampling a number of variables to be included sampling a subset with that number, and including each possible monomial from the subset with probability 0.5, with coefficients sampled from \(\mathcal{N}(0, 2.5)\). We restrict the input range to \([-1, 1]\) to avoid extreme values. 
%
%\subsection**{RL}
%The model maintained persistent representations for each trained task, and performed the task with a random convex combination of the persistent task representation and one generated from the present examples, while also trying to match the two via an \(\ell_2\) loss. The persistence helps the model overcome conflicting signals from switched-color tasks. Second, we found that incorporating weight normalization \citep{Salimans2016} in the task network improved stability. It is likely that neither modification was strictly necessary, but they made training the model easier and faster. 
%
%\subsection**{}\vspace{-1em} % this fixes spacing in last paragraph of mat-methods
%}

\showmatmethods{} % Display the Materials and Methods section

\acknow{AKL was supported by a National Science Foundation Graduate Research Fellowship. % and Ric Weiland Graduate Fellowship.
The authors appreciate helpful suggestions from Noah Goodman, Surya Ganguli, Felix Hill, Steven Hansen, Erin Bennett, Katherine Hermann, Arianna Yuan, Andrew Nam, Effie Li, and the anonymous reviewers.}
\label{LastContentPage} %% for preprint page count

\showacknow{} % Display the acknowledgments section

\bibliography{arrr}

\begin{thebibliography}{10}

\bibitem{Siegelmann2013}
HT Siegelmann, {Turing on Super-Turing and adaptivity}.
\newblock {\em\protect\JournalTitle{Progress in Biophysics and Molecular
  Biology}} \textbf{113}, 117--126 (2013).

\bibitem{Lake2016}
BM Lake, TD Ullman, JB Tenenbaum, SJ Gershman, {Building Machines that learn
  and think like people}.
\newblock {\em\protect\JournalTitle{Behavioral and Brain Sciences}} (2017).

\bibitem{Marcus2018}
G Marcus, {Deep Learning: A Critical Appraisal}.
\newblock {\em\protect\JournalTitle{arXiv preprint}}, 1--27 (2018).

\bibitem{Russin2020}
J Russin, RC O'Reilly, Y Bengio, {Deep learning needs a pre-frontal cortex} in
  {\em ICLR Workshop on Bridging AI and Cognitive Science}.
\newblock (2020).

\bibitem{Cohen1990}
JD Cohen, K Dunbar, JL McClelland, {On the control of automatic processes: A
  parallel distributed processing account of the stroop effect}.
\newblock {\em\protect\JournalTitle{Psychological Review}} \textbf{97},
  332--361 (1990).

\bibitem{Larochelle2008}
H Larochelle, D Erhan, Y Bengio, {Zero-data learning of new tasks} in {\em
  Proceedings of the Twenty-Third AAAI Conference on Artificial Intelligence}.
\newblock (2008).

\bibitem{Hermann2017}
KM Hermann, et~al., {Grounded Language Learning in a Simulated 3D World}.
\newblock {\em\protect\JournalTitle{arXiv preprint}} (2017).

\bibitem{Hill2019a}
F Hill, et~al., {Environmental drivers of generalization in a situated agent}
  in {\em Proceedings of the 8th International Conference on Learning
  Representations}.
\newblock (2020).

\bibitem{Vinyals2016}
O Vinyals, C Blundell, T Lillicrap, K Kavukcuoglu, D Wierstra, {Matching
  Networks for One Shot Learning}.
\newblock {\em\protect\JournalTitle{Advances in Neural Information Processing
  Systems}} (2016).

\bibitem{Rusu2019}
AA Rusu, et~al., {Meta-Learning with Latent Embedding Optimization} in {\em
  Proceedings of the 7th International Conference on Learning Representations}.
\newblock (2019).

\bibitem{Oh2017a}
J Oh, S Singh, H Lee, P Kohli, {Zero-Shot Task Generalization with Multi-Task
  Deep Reinforcement Learning} in {\em Proceedings of the 34th International
  Conference on Machine Learning}.
\newblock (2017).

\bibitem{Garnelo2018}
M Garnelo, et~al., {Conditional Neural Processes} in {\em Proceedings of the
  35th International Conference on Machine Learning}.
\newblock (2018).

\bibitem{Ha2016}
D Ha, A Dai, QV Le, {HyperNetworks}.
\newblock {\em\protect\JournalTitle{arXiv preprint arXiv:1609.09106}} (2016).

\bibitem{McClelland1985}
JL McClelland, {Putting knowledge in its place : A scheme for programming
  parallel processing structures on the fly}.
\newblock {\em\protect\JournalTitle{Cognition}} \textbf{146}, 113--146 (1985).

\bibitem{Wilensky1991}
U Wilensky, {Abstract Meditations on the Concrete and Concrete Implications for
  Mathematics Education} in {\em Constructionism}, eds.{} I Harel, S Papert.
\newblock (Ablex Publishing), (1991).

\bibitem{Hazzan1999}
O Hazzan, {Reducing Abstraction Level When Learning Abstract Algebra Concepts}.
\newblock {\em\protect\JournalTitle{Educational Studies in Mathematics}}
  \textbf{40}, 71--90 (1999).

\bibitem{Lampinen2017b}
AK Lampinen, JL McClelland, {Different Presentations of a Mathematical Concept
  Can Support Learning in Complementary Ways.}
\newblock {\em\protect\JournalTitle{Journal of Educational Psychology}} (2018).

\bibitem{Bourne1970}
LE Bourne, {Knowing and using concepts}.
\newblock {\em\protect\JournalTitle{Psychological Review}} \textbf{77},
  546--556 (1970).

\bibitem{Hill2019}
F Hill, A Santoro, D Barrett, A Morcos, T Lillicrap, {Learning to make
  analogies by contrasting abstract relational structure} in {\em Proceedings
  of the International Conference on Learning Representations}.
\newblock (2019).

\bibitem{Niv2009}
Y Niv, {Reinforcement learning in the brain}.
\newblock {\em\protect\JournalTitle{Journal of Mathematical Psychology}}
  (2009).

\bibitem{Dabney2020}
W Dabney, et~al., {A distributional code for value in dopamine-based
  reinforcement learning}.
\newblock {\em\protect\JournalTitle{Nature}} \textbf{577}, 671--675 (2020).

\bibitem{Silver2016}
D Silver, et~al., {Mastering the game of Go with deep neural networks and tree
  search}.
\newblock {\em\protect\JournalTitle{Nature}} \textbf{529}, 484--489 (2016).

\bibitem{Vinyals2019}
O Vinyals, et~al., {Grandmaster level in StarCraft II using multi-agent
  reinforcement learning}.
\newblock {\em\protect\JournalTitle{Nature}} \textbf{575}, 350--354 (2019).

\bibitem{Reed2015}
S Reed, N de~Freitas, {Neural Programmer-Interpreters} in {\em Proceedings of
  the International Conference on Learning Representations}.
\newblock (2016).

\bibitem{Rogers2004}
TT Rogers, JL McClelland, {\em {Semantic Cognition: A Parallel Distributed
  Processing Approach}}.
\newblock (MIT Press), (2004).

\bibitem{Lampinen2018a}
AK Lampinen, JL McClelland, {One-shot and few-shot learning of word
  embeddings}.
\newblock {\em\protect\JournalTitle{arXiv preprint arXiv:1710.10280}} (2017).

\bibitem{Finn2017a}
C Finn, P Abbeel, S Levine, {Model-Agnostic Meta-Learning for Fast Adaptation
  of Deep Networks} in {\em Proceedings of the 34th Annual Conference on
  Machine Learning}.
\newblock (2017).

\bibitem{Turchetta2016}
M Turchetta, F Berkenkamp, A Krause, {Safe exploration in finite Markov
  decision processes with Gaussian processes}.
\newblock {\em\protect\JournalTitle{Advances in Neural Information Processing
  Systems}}, 4312--4320 (2016).

\bibitem{Ravichandran2019}
A Ravichandran, R Bhotika, S Soatto, {Few-Shot Learning with Embedded Class
  Models and Shot-Free Meta Training}.
\newblock {\em\protect\JournalTitle{arXiv preprint}} (2019).

\bibitem{Hinton1982}
GE Hinton, DC Plaut, {Using Fast Weights to Deblur Old Memories} in {\em
  Proceedings of the 9th Annual Conference of the Cognitive Science Society}.
\newblock No.{} 1987, (1982).

\bibitem{Li2019a}
H Li, et~al., {LGM-Net: Learning to Generate Matching Networks for Few-Shot
  Learning} in {\em Proceedings of the 36th International Conference on Machine
  Learning}.
\newblock (2019).

\bibitem{Socher2013}
R Socher, M Ganjoo, CD Manning, AY Ng, {Zero-shot learning through cross-modal
  transfer}.
\newblock {\em\protect\JournalTitle{Advances in Neural Information Processing
  Systems}} (2013).

\bibitem{Xian2018}
Y Xian, CH Lampert, B Schiele, Z Akata, {Zero-Shot Learning - A Comprehensive
  Evaluation of the Good, the Bad and the Ugly}.
\newblock {\em\protect\JournalTitle{IEEE Trans. on Pattern Anal. and Machine
  Intelligence}} (2018).

\bibitem{Pal2019}
A Pal, VN Balasubramanian, {Zero-Shot Task Transfer} in {\em Proceedings of the
  IEEE Conference on Computer Vision and Pattern Recognition}.
\newblock (2019).

\bibitem{Laroche2017}
R Laroche, M Barlier, {Transfer Reinforcement Learning with Shared Dynamics} in
  {\em Proceedings of the Thirty First AAAI Conference on Artificial
  Intelligence}.
\newblock pp. 2147--2153 (2017).

\bibitem{Achille2019}
A Achille, et~al., {Task2Vec: Task Embedding for Meta-Learning}.
\newblock {\em\protect\JournalTitle{arXiv preprint}} (2019).

\bibitem{Eysenbach2019}
B Eysenbach, A Gupta, J Ibarz, S Levine, {Diversity is all you need: learning
  skills without a reward function} in {\em Proceedings of the International
  Conference on Learning Representations}.
\newblock (2019).

\bibitem{Hsu2019}
K Hsu, S Levine, C Finn, {Unsupervised Learning Via Meta-Learning} in {\em
  Proceedings of the International Conference on Learning Representations}.
\newblock (2019).

\bibitem{Gentner2003}
D Gentner, {Why We're So Smart} in {\em Language in mind: Advances in the study
  of language and thought.}
\newblock pp. 195--235 (2003).

\bibitem{Gick1980}
ML Gick, KJ Holyoak, {Analogical Problem Solving}.
\newblock {\em\protect\JournalTitle{Cognitive P}} \textbf{12}, 306--355 (1980).

\bibitem{Fodor2001}
JA Fodor, {Language, thought and compositionality}.
\newblock {\em\protect\JournalTitle{Mind and Language}} \textbf{16}, 1--15
  (2001).

\bibitem{Fodor2008lot2}
JA Fodor, {\em {LOT 2: The language of thought revisited}}.
\newblock (Oxford University Press), (2008).

\bibitem{Lake2017}
BM Lake, M Baroni, {Generalization without systematicity: On the compositional
  skills of sequence-to-sequence recurrent networks} in {\em Proceedings of the
  International Conference on Machine Learning}.
\newblock (2018).

\bibitem{McClelland2010}
JL McClelland, et~al., {Letting structure emerge: connectionist and dynamical
  systems approaches to cognition}.
\newblock {\em\protect\JournalTitle{Trends in Cognitive Sciences}} \textbf{14},
  348--356 (2010).

\bibitem{Hansen2017}
SS Hansen, A Lampinen, G Suri, JL McClelland, {Building on prior knowledge
  without building it in}.
\newblock {\em\protect\JournalTitle{Behavioral and Brain Sciences}} \textbf{40}
  (2017).

\bibitem{Baars2005}
BJ Baars, {Global workspace theory of consciousness: Toward a cognitive
  neuroscience of human experience}.
\newblock {\em\protect\JournalTitle{Progress in Brain Research}} \textbf{150},
  45--53 (2005).

\bibitem{Karmiloff-Smith1986}
A Karmiloff-Smith, {From meta-processes to conscious access: Evidence from
  children's metalinguistic and repair data}.
\newblock {\em\protect\JournalTitle{Cognition}} \textbf{23}, 95--147 (1986).

\bibitem{Clark1993}
A Clark, A Karmiloff-Smith, {The Cognizer's Innards: A Psychological and
  Philosophical Perspective on the Development of Thought}.
\newblock {\em\protect\JournalTitle{Mind {\&} Language}} \textbf{8}, 487--519
  (1993).

\bibitem{fodor1983modularity}
JA Fodor, {\em {The modularity of mind}}.
\newblock (MIT press), (1983).

\bibitem{McClelland2014}
JL McClelland, D Mirman, DJ Bolger, P Khaitan, {Interactive activation and
  mutual constraint satisfaction in perception and cognition}.
\newblock {\em\protect\JournalTitle{Cognitive Science}} \textbf{38}, 1139--1189
  (2014).

\bibitem{Goldin-Meadow1999}
S Goldin-Meadow, {The role of gesture in communication and thinking}.
\newblock {\em\protect\JournalTitle{Trends in Cognitive Sciences}} \textbf{3},
  419--429 (1999).

\bibitem{Tanenhaus1987}
MK Tanenhaus, MM Lucas, {Context effects in lexical processing}.
\newblock {\em\protect\JournalTitle{Cognition}} \textbf{25}, 213--234 (1987).

\bibitem{Chi1994}
MT Chi, N {De Leeuw}, MH Chiu, C Lavancher, {Eliciting self-explanations
  improves understanding}.
\newblock {\em\protect\JournalTitle{Cognitive Science}} \textbf{18}, 439--477
  (1994).

\bibitem{Mu2019}
J Mu, P Liang, N Goodman, {Shaping visual representations with language for
  few-shot classification} in {\em Visually Grounded Interaction and Language
  Workshop, NeurIPS}.
\newblock (2019).

\bibitem{Graves2016}
A Graves, et~al., {Hybrid computing using a neural network with dynamic
  external memory}.
\newblock {\em\protect\JournalTitle{Nature Publishing Group}} \textbf{538},
  471--476 (2016).

\bibitem{Salimans2016}
T Salimans, DP Kingma, {Weight Normalization: A Simple Reparameterization to
  Accelerate Training of Deep Neural Networks}.
\newblock {\em\protect\JournalTitle{Advances in Neural Information Processing
  Systems}} (2016).

\bibitem{Xu2015a}
B Xu, N Wang, T Chen, {Empirical evaluation of rectified activations in
  convolution network}.
\newblock {\em\protect\JournalTitle{arXiv preprint arXiv:1505.00853}} (2015).

\bibitem{Glorot2010}
X Glorot, Y Bengio, {Understanding the difficulty of training deep feedforward
  neural networks}.
\newblock {\em\protect\JournalTitle{Proceedings of the 13th International
  Conference on Artificial Intelligence and Statistics (AISTATS)}} \textbf{9},
  249--256 (2010).

\bibitem{Saxe2013}
AM Saxe, JL McClelland, S Ganguli, {Exact solutions to the nonlinear dynamics
  of learning in deep linear neural networks}.
\newblock {\em\protect\JournalTitle{Advances in Neural Information Processing
  Systems}}, 1--9 (2013).

\bibitem{Li2019b}
Y Li, C Wei, T Ma, {Towards Explaining the Regularization Effect of Initial
  Large Learning Rate in Training Neural Networks}.
\newblock {\em\protect\JournalTitle{Advances in Neural Information Processing
  Systems}}, 1--49 (2019).

\bibitem{Mnih2015}
V Mnih, et~al., {Human-level control through deep reinforcement learning}.
\newblock {\em\protect\JournalTitle{Nature}} \textbf{518}, 529--533 (2015).

\bibitem{lampinen2020computational}
AK Lampinen, {\em {A Computational Framework for Learning and Transforming Task
  Representations}}.
\newblock (PhD Dissertation, Stanford University,
  https://stacks.stanford.edu/file/druid:xj689nb3522/dissertation-augmented.pdf),
  (2020).

\bibitem{harrower2003colorbrewer}
M Harrower, CA Brewer, {ColorBrewer. org: an online tool for selecting colour
  schemes for maps}.
\newblock {\em\protect\JournalTitle{The Cartographic Journal}} \textbf{40},
  27--37 (2003).

\bibitem{Hermann2020}
KL Hermann, AK Lampinen, {What shapes feature representations? Exploring
  datasets, architectures, and training}.
\newblock {\em\protect\JournalTitle{arXiv preprint}} (2020).

\bibitem{Oswald2020}
JV Oswald, C Henning, J Sacramento, BF Grewe, {Continual learning with
  hypernetworks}.
\newblock {\em\protect\JournalTitle{International Conference on Learning
  Representations}}, 1--25 (2020).

\bibitem{Monsell2003}
S Monsell, {Task switching}.
\newblock {\em\protect\JournalTitle{Trends in Cognitive Sciences}} \textbf{7},
  134--140 (2003).

\bibitem{Mikolov2013}
T Mikolov, Wt Yih, G Zweig, {Linguistic regularities in continuous space word
  representations}.
\newblock {\em\protect\JournalTitle{Proceedings of NAACL-HLT}}, 746--751
  (2013).

\bibitem{Siegelman1992}
HT Siegelman, ED Sontag, {On the computational power of neural nets}.
\newblock {\em\protect\JournalTitle{Proceedings of the fifth annual workshop on
  computational learning theory}} (1992).

\end{thebibliography}

\FloatBarrier
\clearpage
\onecolumn
\section*{Supporting Information (SI)}
The Supporting Information is organized as follows: in Section \ref{supp_sec:methods}, we describe the details of the model, including providing a mathematical formulation, diagram of gradient flow, and architectural and hyperparameters for all experiments. In Section \ref{supp_sec:methods_datasets_tasks} we describe the different task domains and dataset sizes for our experiments. In Section \ref{supp_sec:behavioral} we describe the behavioral experiment that we performed on human adaptibity. In Section \ref{supp_sec:github} we provide links to the repositories containing the code for all experiments and analyses. In Section \ref{supp_sec:analyses} we show supplemental analyses, and in Section \ref{supp_sec:proofs} we provide a proof that a simpler vector-analogy approach is insufficient for meta-mapping. 

\subsection{Model details, training, and methods} \label{supp_sec:methods}

This section is organized as follows: in Section \ref{supp_sec:methods:model_formalism} we give a formal (mathematical) description of the model, In Section \ref{supp_sec:model_details} we describe the architectural details and hyperparameters, and provide motivation for some of them. In Section \ref{supp_model_training_details} we provide further details of model training and evaluation. In Section \ref{supp_sec:model_RL_modifications} we provide details of the model modifications for the Cards and RL domains. Finally, in Section \ref{supp_sec:model_optimizing_details} we provide details about the optimization of task representations for the \textbf{Meta-mapping as a starting point for later learning} experiments. 

\subsubsection{Mathematical formulation of the model} \label{supp_sec:methods:model_formalism}
In this section we describe each of the networks in the system mathematically and give functional representations of each computation used in the model. 
\begin{table*}[htbp]
\begin{subtable}{\textwidth}
\centering
\begin{tabular}{|c||c|p{8cm}|}
\hline
Symbol & Characterization & Description\\
\hline
input & Varies. & The input space for the base tasks, e.g \(\mathbb{R}^4\) for polynomials, or RGB images for visual concepts.\\[0.25em]
output & Varies. & The output space for the base tasks, e.g \(\mathbb{R}\) for polynomials, 4 action \(Q\)-values for the RL domain.\\[0.25em]
language & Varies. & A sentence of words from a discrete vocabulary.\\[0.25em]
\(Z\) & \(\mathbb{R}^n\) & The shared representational space used for representing inputs, tasks, etc.\\[0.25em]
\(\Theta\) & \(\underbrace{\underbrace{\mathbb{R}^{l_0 \times l_1}}_{\text{weights}} \times \underbrace{\mathbb{R}^{l_1}}_{\text{biases}}}_{\text{one layer's parameters}} \times \cdots\) & The parameter space of the task-network \(\mathcal{T}\), that is, the set of matrices (and vectors) representing the weights (and biases) of each layer of the MLP. \\[3.5em]
\hline
\end{tabular}
\caption{Representation spaces.} \label{supp_table:notation_guide:spaces}
\end{subtable}

\begin{subtable}{\textwidth}
\centering
\begin{tabular}{|c||c|p{8cm}|}
\hline
Symbol & Characterization & Description\\
\hline
\(z_{input}\) & \multirow{13}{*}{Representation \(\in Z\)}  & The representation of a base-task input (e.g. \(z_{hand}\) for a hand of cards), after it is processed by the perception network \(\mathcal{P}\). \\[0.25em]
\(z_{output}\) &  & The representation of a base-task output (e.g. \(z_{bet}\) for a bet in the card game). This is processed by the output decoder \(\mathcal{O}_d\) to produce the task output. \\[0.25em]
\(z_{target}\) &  & The representation of a base-task target output (e.g. a ground-truth classification of an image for a visual concept). This is processed by the target output encoder \(\mathcal{O}_e\) to produce a target embedding for the input processor. Note that in the case of the cards and RL domains, the ``target'' is actually an (action, reward) tuple (see below). \\[0.25em]
\(z_{task}\) & & Representation of a task. These are used to perform the task, and as inputs and outputs (and targets) of meta-mappings.\\[0.25em]
\(z_{meta}\) & & Representation of a meta-mapping, used to perform that meta-mapping.\\[0.25em]
\hline
\end{tabular}
\caption{Different types of representations in \(Z\).} \label{supp_table:notation_guide:z_elements}
\end{subtable}

\begin{subtable}{\textwidth}
\centering
\begin{tabular}{|c||c|p{8cm}|}
\hline
Symbol & Characterization & Description\\
\hline
\(\mathcal{P}\) & \(\text{input} \rightarrow Z\) & The perception network, an MLP (for polynomial and card tasks) or CNN (for visual concepts and RL tasks), which processes inputs into the shared representational space.\\[0.25em]
\(\mathcal{O}_e\) & \(\text{target output} \rightarrow Z\) & The target output encoder network, an MLP, which processes base-task example targets into the shared representational space. Note that in the case of the cards and RL models, these are not in fact outputs, but are rather (action, reward) tuples, see below.\\[0.25em]
\(\mathcal{L}\) & \(\text{language} \rightarrow Z\) & The language network, a multi-layer LSTM, which processes langauge into the shared representational space.\\[0.25em]
\(\mathcal{E}\) & \(\{Z^2\} \rightarrow Z\) & The example network, which processes a support set of tuples of (embedding of input, embedding of target output), and outputs a task representation in the shared representation space. This network consists of 1) parallel application of an MLP to each of the (input, output) tuples to produce a representation for each, 2) followed by max-pooling across that set of representations to produce a single representation, 3) followed by another MLP to produce the task representation.\\[0.25em]
\(\mathcal{H}\) & \(Z \rightarrow \Theta\) & The hyper network, an MLP, which maps a task representation to a set of parameters for the network \(\mathcal{T}\).\\[0.25em]
\(\mathcal{T}\) & \(\Theta \times Z \rightarrow Z\) & The task network, which is an MLP parameterized by the parameter-space \(\Theta\). Once the parameters are specified, it serves as an MLP mapping \(Z \rightarrow Z\).\\[0.25em]
\(\mathcal{O}_d\) & \(Z \rightarrow \text{output}\) & The output decoder, an MLP mapping from the representational space \(Z\) to the output space for the task (e.g. \(\mathbb{R}\) for the polynomials, and Q-values for the actions for the RL tasks).\\[0.25em]
\hline
\end{tabular}
\caption{Networks.} \label{supp_table:notation_guide:networks}
\end{subtable}
\caption{Notation used for the (\subref{supp_table:notation_guide:spaces}) representation spaces, (\subref{supp_table:notation_guide:z_elements}) types of representations in the shared space \(Z\), (\subref{supp_table:notation_guide:networks}) and networks in the paper.} \label{supp_table:notation_guide}
\end{table*}

First, in Table \ref{supp_table:notation_guide} we remind the reader of the notation we use, and provide a mathematical characterization of each component, as well as its description. Given this notation, we next describe the computations of the model mathematically, with annotations indicating the meaning of key elements of each equation.

\textbf{Constructing a basic task representation (from examples):} Given a support set of (input, target output) examples tuples \(\{({input}_0, {target_0}), ({input}_1, {target_1})...\}\), the representation would be computed as 
\[z_{task} = \mathcal{E}(\underbrace{\{\underbrace{(\underbrace{\mathcal{P}(input_0)}_{\text{input embedding} \in Z}, \underbrace{\mathcal{O}(target_0)}_{\text{target embedding}\in Z})}_{\text{(input embedding, target embedding) tuple} \,\in Z^2}, \cdots\}}_{\text{set containing encoded tuple for each support-set example}})\]

\textbf{Constructing a basic task representation (from language):} The representation would be computed as 
\[z_{task} = \mathcal{L}(\underbrace{language}_{\text{description of task}})\]

\textbf{Performing a task from a representation:} Given a task representation denoted by \(z_{task}\), and an input, the output embedding (\(z_{out} \in Z\)) would be computed as: 
\[z_{out} = \mathcal{T}(\underbrace{\mathcal{H}(z_{task})}_{\text{parameters} \in \Theta}, \underbrace{\mathcal{P}(input)}_{\text{input embedding}\in Z})\]
and the output would be computed as:
\[\text{output} = \mathcal{O}_d(z_{out})\]

\textbf{Constructing a meta-mapping representation (from examples):} Given a support set of (input task, target task) example tuples, the representation would be computed as follows:
\[z_{meta} = \mathcal{E}(\underbrace{\{\underbrace{(\underbrace{z_{input task_0}}_{\text{task embedding} \in Z}, \underbrace{z_{target task_0}}_{\text{task embedding}\in Z})}_{\text{(input embedding, target embedding) tuple} \,\in Z^2}, \cdots\}}_{\text{set containing tuple for each mapping example}})\]

\textbf{Constructing a meta-mapping representation (from language):} The representation would be computed as 
\[z_{meta} = \mathcal{L}(\underbrace{language}_{\text{description of meta-mapping}})\]

\textbf{Performing a meta-mapping from a representation:} Given a meta-mapping representation \(z_{meta}\) and an input task representation \(z_{task}\), the transformed task representation would be computed as: 
\[z_{transformed\,\,task}  = \mathcal{T}(\underbrace{\mathcal{H}(z_{meta})}_{\text{parameters} \in \Theta}, z_{task})\]

\subsubsection{Model architecture \& hyperparameters} \label{supp_sec:model_details}
\begin{figure}[h]
\begin{subfigure}{\textwidth}
\begin{tikzpicture}[auto]
%% from examples
\node[text width=0.5cm] at (-6.8, -0.5) (inputs0) {\includegraphics[width=0.5cm]{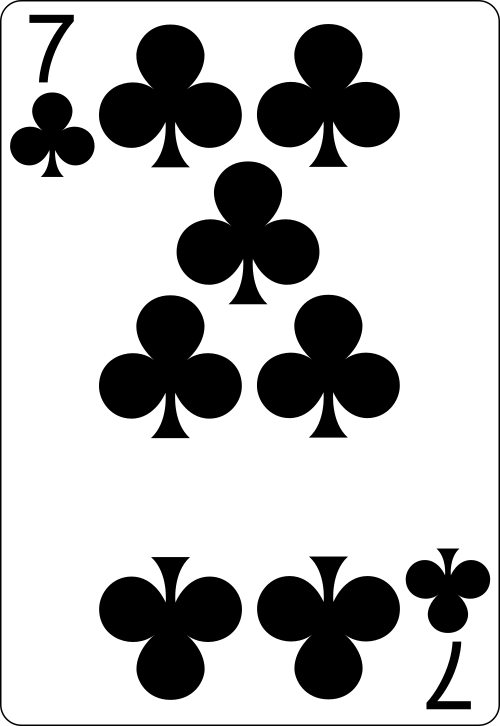}\\\includegraphics[width=0.5cm]{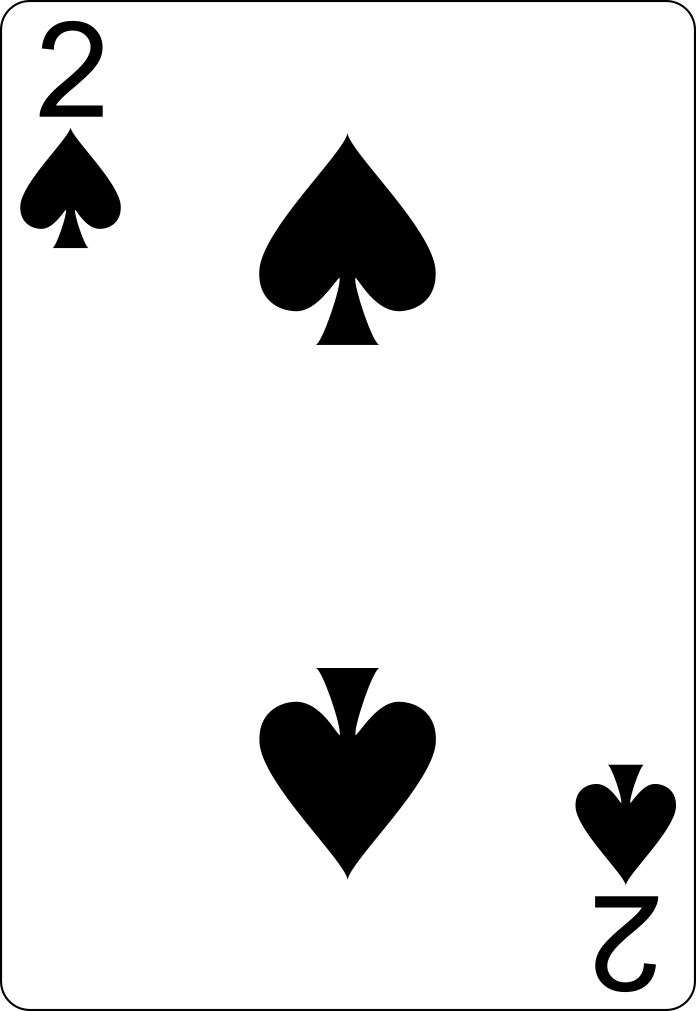}};

\node[gray, text width=2cm, align=center] at (-5.6, 0.35) {Perception network};
\node[block] at (-5.6, -0.5) (perceptionnet0) {\(\mathcal{P}\)};
\path[arrow] (inputs0.east) -- ([xshift=-3]perceptionnet0.west);

\node[text width=0.5cm] at (-6.8, -2) (targets0) {\bf \color{red}\(-\)\$\$};

\node[gray, text width=4cm, align=center] at (-5.6, -3) {Target output encoder\\network};
\node[block] at (-5.6, -2) (targetnet0) {\(\mathcal{O}_e\)};
\path[arrow] ([xshift=3]targets0.east) -- ([xshift=-3]targetnet0.west);

\node[gray, text width=2.5cm, align=center] at (-3.25, 2.3) {Task examples (encoded)};
\node at (-3.25, 1.25) (examples) {
\(\left\{
\begin{matrix}
({\color{bgreen}z_{hand_{1}}}, {\color{bgreen}z_{win_{1}}})\\
$\vdots$
\end{matrix}\right\}\)};

\path[arrow, out=0, in=-90] (perceptionnet0.east) to ([xshift=-11, yshift=20]examples.south);
\path[arrow, out=0, in=-90] (targetnet0.east) to ([xshift=15, yshift=20]examples.south);

\node[gray, text width=2cm, align=center] at (-0.5, 2.1) {Example network};
\node[block] at (-0.5, 1.25) (examplenet) {\(\mathcal{E}\)};
\path[arrow] (examples.east) -- ([xshift=-3]examplenet.west);

%% performing 
\node[bpurp] at (1.25, 1.25) (taskrep) {\(z_{task}\)};
\path[arrow] ([xshift=3]examplenet.east) -- (taskrep.west);

\node[gray, text width=2cm, align=center] at (3, 2.1) {Hyper network};
\node[block] at (3, 1.25) (hypernet) {\(\mathcal{H}\)};
\path[arrow] (taskrep.east) -- ([xshift=-3]hypernet.west);

\node[text width=0.5cm] at (0.7, -1.25) (inputs) {\includegraphics[width=0.5cm]{figures/2_of_spades.png}\\\includegraphics[width=0.5cm]{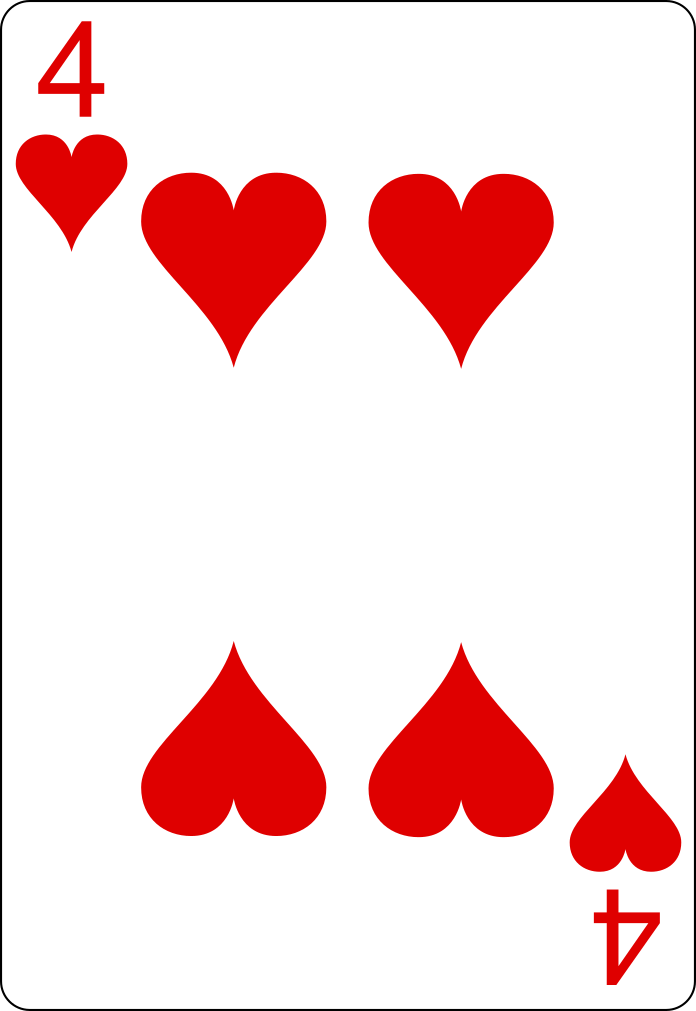}};

\node[gray, text width=2cm, align=center] at (1.9, -0.4) {Perception network};
\node[block] at (1.9, -1.25) (perceptionnet) {\(\mathcal{P}\)};
\path[arrow] (inputs.east) -- ([xshift=-3]perceptionnet.west);

\node[bgreen] at (3.2, -1.25) (handrep) {\(z_{hand}\)};
\path[arrow] ([xshift=3]perceptionnet.east) -- (handrep.west);

\node[bblue, block, dashed] at (4.5, -1.25) (tasknet) {\(\mathcal{T}\)};
\node[bblue, text width=2cm, align=center] at (4.5, -2) {Task network};
\path[arrow] (handrep.east) -- ([xshift=-3]tasknet.west);
\path[arrow, out=0, in=90] ([xshift=3]hypernet.east) to ([yshift=3]tasknet.north);

\node[bgreen] at (5.65, -1.25) (betrep) {\(z_{bet}\)};
\path[arrow] ([xshift=3]tasknet.east) -- (betrep.west);

\node[gray, text width=3cm, align=center] at (6.8, -0.4) {Output decoder\\network};
\node[block] at (6.8, -1.25) (actionnet) {\(\mathcal{O}_d\)};
\path[arrow] (betrep.east) -- ([xshift=-3]actionnet.west);

\node at (7.7, -1.25) (output) {\bf \$};
\path[arrow] ([xshift=3]actionnet.east) -- (output.west);

\node at (8.7, -1.25) (loss) {Loss};
\path[arrow] (output.east) -- (loss.west);

% gradients

\path[arrow, bred, ultra thick] ([yshift=-12, xshift=20]loss.west) to ([yshift=-12, xshift=3]inputs.east);

\path[draw, bred, ultra thick, out=90, in=0] ([yshift=-12, xshift=-10]tasknet.west) to ([yshift=-12, xshift=-10]hypernet.east);

\path[draw, bred, ultra thick] ([yshift=-12, xshift=-10]hypernet.east) to ([yshift=-12, xshift=-40]examples.east);

\path[draw, bred, ultra thick, out=-90, in=0] ([yshift=-12, xshift=-40]examples.east) to ([yshift=-12, xshift=-10]perceptionnet0.east);
\path[arrow, bred, ultra thick] ([yshift=-12, xshift=-10]perceptionnet0.east) to ([yshift=-12]inputs0.east);

\path[draw, bred, ultra thick, out=-90, in=0] ([yshift=-12, xshift=-12]examples.east) to ([yshift=-12, xshift=-10]targetnet0.east);
\path[arrow, bred, ultra thick] ([yshift=-12, xshift=-10]targetnet0.east) to ([yshift=-12]targets0.east);

\end{tikzpicture}
\caption{Basic task inference/training (from examples).} \label{supp_fig:HoMM:gradient_flow:basic_tasks}
\end{subfigure}
\begin{subfigure}{\textwidth}
\begin{tikzpicture}[auto]
%% from examples
\node at (-6.9, 0) {};

\node[gray, text width=3.5cm, align=center] at (-3.25, 2.3) {Mapping examples (input/output tasks)};
\node at (-3.25, 1.25) (examples){
\(\left\{
\begin{matrix}
({\color{bpurp}z_{chess}}, {\color{bpurp}z_{lose chess}})\\
$\vdots$
\end{matrix}\right\}\)};

\node[gray, text width=2cm, align=center] at (-0.5, 2.1) {Example network};
\node[block] at (-0.5, 1.25) (examplenet) {\(\mathcal{E}\)};
\path[arrow] (examples.east) -- ([xshift=-3]examplenet.west);

%% performing 
\node[borange] at (1.25, 1.25) (taskrep) {\(z_{meta}\)};
\path[arrow] ([xshift=3]examplenet.east) -- (taskrep.west);

\node[gray, text width=2cm, align=center] at (3, 2.1) {Hyper network};
\node[block] at (3, 1.25) (hypernet) {\(\mathcal{H}\)};
\path[arrow] (taskrep.east) -- ([xshift=-3]hypernet.west);

\node[bpurp] at (2.66, -1.25) (handrep) {\(z_{poker}\)};

\node[bblue, block, dashed] at (4.5, -1.25) (tasknet) {\(\mathcal{T}\)};
\node[bblue, text width=2cm, align=center] at (4.5, -2) {Task network};
\path[arrow] (handrep.east) -- ([xshift=-3]tasknet.west);
\path[arrow, out=0, in=90] ([xshift=3]hypernet.east) to ([yshift=3]tasknet.north);

\node[bpurp] at (6.5, -1.25) (output) {\(\hat{z}_{lose poker}\)};
\path[arrow] ([xshift=3]tasknet.east) -- (output.west);

\node at (8.7, -1.25) (loss) {Loss};
\path[arrow] (output.east) -- (loss.west);

% gradients

\path[arrow, bred, ultra thick] ([yshift=-12, xshift=20]loss.west) to ([yshift=-12, xshift=5]handrep.east);

\path[draw, bred, ultra thick, out=90, in=0] ([yshift=-12, xshift=-10]tasknet.west) to ([yshift=-12, xshift=-10]hypernet.east);

\path[arrow, bred, ultra thick] ([yshift=-12, xshift=-10]hypernet.east) to ([yshift=-12, xshift=10]examples.east);
\end{tikzpicture}
\caption{Meta-mapping inference/training (from examples).}\label{supp_fig:HoMM:gradient_flow:meta_mappings}
\end{subfigure}
\caption[Schematic of architecture, showing inference and gradient flow through the model on a training step.]{Schematic of architecture, showing inference and gradient flow through the model on a training step. Thin black lines moving rightward represent inference, thick red lines moving leftward represent gradients. (\subref{supp_fig:HoMM:gradient_flow:basic_tasks}) Inference and gradients for the basic tasks. (\subref{supp_fig:HoMM:gradient_flow:meta_mappings}) Inference and gradients for meta-mappings. The gradients end at the examples of the meta-mapping, rather than propagating through to alter how those representations are constructed, due to GPU memory constraints. In the future, it might be useful to explore whether allowing further propagation would improve results for both basic tasks and meta-mappings. (These figures depict the inference/gradient flow when performing tasks and meta-mappings from examples, performing from language is similar, except that the example inputs and example network are replaced with language inputs and the language processing network.)} \label{supp_fig:HoMM:gradient_flow}
\end{figure}

\begin{table*}[p]
\scriptsize
\centering
\begin{tabular}{|p{4cm}||c|c|c|c|}
\hline
& Polynomials & Cards & Visual & RL \\\hline
\hline
$Z$-dimension & \multicolumn{4}{c|}{512} \\\hline
$\mathcal{P}$ num. layers & \multicolumn{4}{c|}{2} \\\hline
$\mathcal{P}$ num. hidden units & \multicolumn{4}{c|}{128} \\\hline
$\mathcal{P}$ conv. layers. (num filters, size, all strides are 2) & \multicolumn{2}{c|}{-} & \multicolumn{1}{p{2.3cm}|}{(64, 5), (128, 4), (256, 4), (512, 2), max pool} & \multicolumn{1}{p{2.3cm}|}{(64, 7), (64, 4), (64, 3)}\\\hline
$\mathcal{L}$ architecture & -  & \multicolumn{3}{c|}{2-layer LSTM + 2 fully-connected} \\\hline
$\mathcal{L}$ num. hidden units & -  & \multicolumn{3}{c|}{512} \\\hline
$\mathcal{O}_e$ num. layers & 1 & 3 & 1 & 3 \\\hline
$\mathcal{O}_e$ num. hidden units & - & 128 & - & 128 \\\hline
$\mathcal{E}$ architecture & \multicolumn{4}{c|}{2 layers per-datum, max pool across, 2 layers} \\\hline
Task, MM representations from & \multicolumn{2}{c|}{Examples} & Language & Examples \\\hline
$\mathcal{H}$ architecture & \multicolumn{4}{c|}{4 layers} \\\hline
$\mathcal{E}$ num. hidden units & \multicolumn{3}{c|}{512} & 1024 \\\hline
$\mathcal{H}$ num. hidden units & \multicolumn{4}{c|}{512} \\\hline
$\mathcal{T}$ num. layers & 3 & 1 & HoMM: 1, Lang: 3 & 3 \\\hline
$\mathcal{T}$ num. hidden units & \multicolumn{3}{c|}{64} & 128 \\\hline
$\mathcal{H}$ output init. scale & 1 & 1 & 30 & 10 \\\hline
$\mathcal{T}$ weight norm. \citep{Salimans2016} & \multicolumn{3}{c|}{No} & Yes \\\hline
$\mathcal{O}_d$ num. layers & \multicolumn{2}{c|}{1} & 2 & 1 \\\hline
$\mathcal{O}_d$ num. hidden units & \multicolumn{2}{c|}{-} & 128 & -  \\\hline
Nonlinearities & \multicolumn{4}{p{11cm}|}{Leaky ReLU in most places, except no non-linearity at final layer of networks outputting to the latent space $Z$, and (where applicable) sigmoid for classification outputs, and softmax over actions.} \\\hline
Base task loss & $\ell_2$ & $\ell_2$ (masked) & Cross-entropy & $\ell_2$ (masked)\\\hline
Meta-mapping loss & \multicolumn{4}{c|}{$\ell_2$}\\\hline
Persistent task representations & \multicolumn{3}{c|}{No} & Yes \\\hline
Persistent embedding match loss weight & \multicolumn{3}{c|}{-} & 0.2 \\\hline
\hline
Optimizer & Adam & \multicolumn{3}{c|}{RMSProp} \\\hline
Learning rate (base) & $3\cdot 10^{-5}$ & $1\cdot 10^{-5}$ & $3\cdot 10^{-5}$ & $1\cdot 10^{-4}$\\\hline
Learning rate (meta) & $1\cdot 10^{-5}$ & $1\cdot 10^{-5}$ & $1\cdot 10^{-5}$ & $1\cdot 10^{-4}$\\\hline
L.R. decay rate (base) & $\times0.85$ & $\times0.85$ & $\times0.8$ & $\times0.8$\\\hline
L.R. decay rate (meta) & $\times0.85$ & $\times0.9$ & $\times0.85$ & $\times0.95$ \\\hline
L.R. min (base) & \multicolumn{2}{c|}{$3 \cdot 10^{-8}$}  & $1 \cdot 10^{-8}$ & $3 \cdot 10^{-8}$\\\hline
L.R. min (meta) & $1 \cdot 10^{-7}$& $3 \cdot 10^{-8}$ &  $1 \cdot 10^{-8}$ & $3 \cdot 10^{-7}$\\\hline
L.R. decays every & 100 epochs & 200 epochs & 400 epochs & 10000 \\\hline
Num. training epochs & 5000 & \multicolumn{1}{p{2.3cm}|}{100000 (optimally stopped)} & \multicolumn{1}{p{2.3cm}|}{10000 for 4 train mappings, 7500 for 8, 5000 for others} & \multicolumn{1}{p{2.3cm}|}{300000 (optimally stopped)} \\\hline
Num. runs & 5 & 5 & 10 & 5 \\ \hline
\hline

Base memory buffer size & \multicolumn{2}{c|}{1024} & 336 & 1000 \\\hline
Base memory buffers refreshed & \multicolumn{2}{c|}{Every 50 epochs} & Every 20 & Every 1500  \\\hline
Target network updated & \multicolumn{3}{c|}{-} & Every 10000 epochs  \\\hline
RL discount & \multicolumn{3}{c|}{-} & 0.85 \\\hline
RL exploration probability (\(\epsilon\)) & \multicolumn{3}{c|}{-} & \multicolumn{1}{p{2.7cm}|}{Initial: 1., decay: -0.03 when LR decays.}\\\hline
Action softmax inv. temp. (\(\beta\)) & - & 8 & - & 8\\\hline
\end{tabular}

\caption{Detailed hyperparameter specification for different experiments. A ``-'' indicates a parameter that does not apply to that experiment. Where only one value is given, it applied to all the experiments discussed. See Table \ref{supp_table:notation_guide} for a guide to the notation for the networks.} \label{supp_hyperparameter_table}
\end{table*}

See Table \ref{supp_hyperparameter_table} for detailed architectural description and hyperparameters for each experiment (note that dataset sizes for the different different domains are specified in Table \ref{supp_dataset_table}). Hyperparameters were generally found by a heuristic search, where mostly only the optimizer, learning rate annealing schedule, and number of training epochs were varied. Architectural parameters were generally chosen based on domain complexity (larger networks for more complex tasks, especially RL), and standard architectural practices.\par
For example, the convolution sizes and strides were generally chosen to result in reasonably even downsampling of the image, while also maintaining sizes divisible by powers of two (which can increase computational efficiency). The activation function chosen for the hidden layers of the MLPs was Leaky ReLU (leaky Rectified Linear Units), which are piecewise defined as 
\[\text{Leaky ReLU}(x) = \begin{cases} 
x & \text{if } x \geq 0\\
0.2x & \text{if } x < 0
\end{cases}\]
This function suppresses negative inputs (but does not completely shut them off). It has been shown to be useful for training deep networks \citep{Xu2015a}. \par
Initialization scales for the HyperNetwork outputs were chosen based on the heuristic that there should be \emph{significant} transmission of signal through the network at initialization to allow for efficient learning \citep{Glorot2010,Saxe2013}, i.e. that when different inputs are presented to the untrained network, its output should vary substantially. Learning rate schedules were chosen by search to be slow enough to give fairly stable learning, but fast enough to not harm generalization \citep[c.f.][]{Li2019b}. \par 
Many of the remaining parameters take the values they do for somewhat arbitrary reasons, e.g. the polynomial experiments were run earlier, before 1-layer task networks were found to be useful in some settings (although the complex tasks and transformations in the polynomial setting may benefit from the more complex task networks). While it would be ideal to fully search the space of parameters for all models, unfortunately our computational resource limitations prohibited it. Thus the results in the paper should be interpreted as a lower bound on what would be possible. \par

\subsubsection{Model training details} \label{supp_model_training_details}

In all experiments, each epoch of training consisted of a single learning step on each task (both base and meta), in a random order. That is, training of the base tasks and meta-mappings was fully interleaved. However, the greater prevalence of base tasks, the learning rate schedules, and the fact that the loss on the meta-mappings is small when the base-task embeddings are small (near initialization) all mean that the base tasks are effectively prioritized earlier in learning. \par
\textbf{Examples \& generalizing:} Where tasks were performed from examples, in each task training step, the meta-learner received only a subset (the ``support set size`` in Table \ref{supp_dataset_table}) of the examples to generate a task representation, and would need to generalize to the remaining probe examples in the batch. In fact, the system was trained to execute the mapping on \emph{both} the support set \emph{and} the probe set. This likely did not substantially alter the learning compared to just training the mapping on the probe set, but may perhaps have made it easier for the model to understand the overall structure of the problem early in learning. Where the task or meta-mapping representations were generated from language, there was no need for a separate support set of examples to generate the task representation. Thus, again, the full batch was used to train the mapping.\par
The representations of the basic tasks for meta-mappings were computed and cached once per epoch, so as the network learned over the course of the epoch, the task representations became ``stale,'' but this did not seem to be too detrimental to learning. In the case of the RL tasks, where there were persistent task representations (see below), they were used instead. \par
\textbf{Gradients:} In Fig. \ref{supp_fig:HoMM:gradient_flow}, we show the flow of inference (forward) and gradients (backward) through our architecture on basic task and meta-mapping training steps. All networks used for performing the base tasks were trained by end-to-end optimization on the appropriate base task loss. That is, the task loss gradients update all networks from the output decoder back through the hyper network, example network, and even the encoding of the task examples and task inputs. \par
During meta-mapping training, the model was trained to match its transformed task representations to target task representations by an \(\ell_2\) loss. Gradients were stopped at the example and inference task representations, rather than updating how those representations were constructed. This simplification was due to memory constraints; it was not possible to fit the construction of all task representations used as examples within GPU memory. An implementation that allowed for this (at least for some task representations, e.g. the source task) might improve learning, and could allow meta-mappings to improve basic meta-learning generalization directly, by shaping the construction of the basic task representations to follow the relationship structure of the task space.\par
\textbf{Multiple runs \& robustness:} The results reported in the figures in this paper are averages across multiple runs, with different trained and held-out basic tasks (in the polynomial and visual concepts domains), different trained and held-out meta-mappings (again in the polynomial and visual concepts domains), and different network initializations and training orders each epoch (in all domains), to ensure the robustness of the findings. \par
\textbf{Classifying task representations:} For classification of task representations, we constructed a representation of the meta-classification, either from examples --- i.e. (task representation, binary classification) tuples --- or language. We constructed these representations using the same example or language network that was used for the basic tasks and meta-mappings. This meta-classification representation then parameterized the task network (via the same hyper network used for the other tasks). Probe task representations were then fed into the task network, and the model was then trained to output appropriate classifications for them through a \emph{separate classification output network} --- it was necessary to have a separate classification output network because in most domains there was not an appropriate classification output. The model was trained on these meta-classifications via a cross-entropy loss. \par 
The idea of this training was that it would help the model identify important features of the task representations that would be relevant for the meta-mappings it needed to perform. However, as we show in Fig. \ref{supp_fig:HoMM_metaclass_lesion}, meta-classification did not prove substantially beneficial in our domains. This may be due to the limited set of classifications we provided. See section \ref{supp_sec:methods_datasets_tasks} for the specific classifications that were used in each domain.\par 
\textbf{Persistent task representations:} In the main approach to performing tasks from examples in our paper, the task representations for basic and meta-mappings were constructed anew on each episode. However, in domains where superficially similar tasks have directly contradicting goals, it can be useful to maintain partly persistent task representations that update more slowly across training steps. Associating each task with a more consistent representation makes it easier for the model to learn the idiosyncrasies of the tasks. We used this approach when performing the RL tasks from examples.\par 
Specifically, the model stored a representation of each task that was updated slowly over learning (persistent), and additionally, on each step constructed a new representation from examples (as in other settings). On each training step, a uniformly random \(t \in [0, 1]\) was chosen, and the representation used for actually performing the task was the convex combination 
\[t \cdot (\text{persistent representation})  + (1-t) \cdot (\text{representation from examples})\]
The model also tried to constrain the persistent and example-constructed representations to match, by minimizing an \(\ell_2\) loss between the two representations. This both updated the persistent task representation to be closer to the representation constructed from examples (thus making the persistent representation essentially a slowly moving average of the example representations), and also updated the representation constructed from examples to be closer to the persistent representation (thus encouraging any useful knowledge contained in the persistent representation to be incorporated in how the example network processed examples). In this way, the knowledge from each representation could support the other. \par 
Note that persistent task representations are not required when performing basic tasks or meta-mappings from language-based representations --- because the language input is consistent across training steps (unlike the examples), the language-based task representations already change relatively slowly between training steps.

\subsubsection{Model \& training modifications for Cards \& RL} \label{supp_sec:model_RL_modifications}

Because in both the Cards and RL domains the system can only take one action, and only receives feedback on that action, we needed to modify the architecture and training slightly. As noted in the main text, we thus replaced the (input, target) examples used to infer a supervised task with (state, (action, reward)) example tuples. These tuples are the basic currency of model-free RL algorithms. To use these tuples, we provide both the action and reward to the target output encoding network, so that it can process them together and produce a single representation. 

The model is trained to output the expected reward of the actions (in the Cards domain), or the Q-value (in the RL domain), via an \(\ell_2\) loss. Again, the fact that the network only receives rewards for the action it takes means that, for any given step in memory, the model can only be trained to better predict the reward (or Q-value) of the single action that it took. 

\textbf{Additional model \& training modifications for RL:} There are a number of additional changes that were necessary for the RL tasks, due to the additional complexity of the temporal structure. These changes generally followed the approach of the original DQN \citep{Mnih2015}. The model received pixel-images as input, and produced \(Q\)-values as output. Target \(Q\)-values were produced by the Bellman equation (that is, the target was the max \(Q\)-value of the subsequent state plus any reward received), but following Mnih and colleagues \citep{Mnih2015}, the target next-state \(Q\)-values were produced by a second (identical) network with frozen weights, that had its weights copied from the main network every 10000 epochs. This helps stabilize learning (by allowing estimates to converge somewhat before the targets change).

We made two additional changes to improve the stability of learning. First, the model maintained persistent representations (see previous section) for each trained task and meta-mapping. The persistent representations helped the model overcome conflicting signals from switched-color tasks, and thereby accelerated learning. (Note that in the experiments performing the RL tasks from language-based task representations, persistent task representations were not used, since the language is already consistent across training steps, unlike examples.) We also incorporated weight normalization \citep{Salimans2016} in the task network, which reparameterizes the weights so that their magnitude and direction are estimated separately. Although learning might have converged without these changes, they seemed to stabilize and accelerate convergence. 

The memory buffers of the system were refreshed every 1500 epochs by allowing the system to play each (training) task for as many episodes as were necessary to generate the 1000 (state, action, reward) tuples necessary to fill the memory buffer. The examples used in any particular network training step were sampled uniformly at random from this buffer, without regard to continuity or epsiode boundaries, as is standard in DQN training. During play to fill the memory buffers, we used both \(\epsilon\)-exploration \emph{and} chose actions from a softmax over \(Q\)-value.

As in all other experiments, the base tasks and meta-mappings were trained simultaneously, but with different learning rate schedules (see Table \ref{supp_hyperparameter_table}).

\textbf{Evaluation for RL:} Evaluation was performed by allowing the system to play each task for a total of 10 randomly generated episodes, with the return assessed as the mean return across this set. While \(\epsilon\)-exploration was turned off during evaluation, the softmax policy was left on. Without the softmax over actions, the model generalized somewhat worse, presumably because its \(Q\)-values are not adapting perfectly and it could easily get stuck in a loop of incorrect actions. The softmax allows some possibilty of breaking out of these loops. Some of the recordings linked in the repository exemplify this, e.g. \url{https://github.com/lampinen/homm_grids/blob/master/recordings/run0_pusher_red_blue_True_False_recording_0.gif}, where the agent gets stuck in the corner after pushing the first three blocks, before eventually breaking out and converging on the correct final block.

We decided when to evaluate the model on each task by: 
\begin{enumerate}
\item Requiring the performance on all trained base tasks to be above 95\% (to ensure that the model had learned both tasks, since the ``push-off'' tasks were slower to learn).
\item Selecting the time when the performance on the \textbf{other} evaluation task was highest (i.e. using the other task as a validation set). 
\end{enumerate}
This means that the performance on each evaluation task may be evaluated at different times during the run. Selection of the stopping point for each task is independent of selecting the stopping point for the other. Note that this optimal stopping approach is not biased, since the task used to decide when to evaluate is always the task that is \textbf{not} being evaluated. To see why this is valid, note that we could have run the model twice for each run, once where we held out one task as a validation set, and the other as the test set, and another run where these were switched. Our evaluation approach is essentially equivalent to this, except applying the two independent evaluations within the same run to save running the entire training process twice as many times.

\subsubsection{Optimizing task representations} \label{supp_sec:model_optimizing_details}

To optimize the task representations on new tasks, we perform gradient descent on those embeddings through the model architecture. We use the same optimizer as was used in the main experiments (i.e. Adam for the polynomials results, RMSProp for the visual concepts), but with a fixed learning rate of \(1 \cdot 10^{-4}\). \par
For the random vector initialization, we sampled the values IID from a normal distribution with variance \(1 / \sqrt{512}\) to give approximately a unit-length vector. The centroid initialization was the centroid of all the trained basic-task representations (i.e. meta-mappings were not included), and the arbitrary trained task representation was likewise an arbitrary trained basic task representation. The untrained model comparison was initialized to exactly the initialization states from which our architectures were trained.

\subsection{Task and dataset details and methods} \label{supp_sec:methods_datasets_tasks}

In this section, we describe the details of basic tasks and meta-mappings in each of our domains. See table \ref{supp_dataset_table} for a summary of the training and hold-out sizes (at the level of support sets and probes for both basic tasks and meta-mappings) for each domain. In the remainder of the section, we describe details of how the tasks were sampled, how they were encoded into language (if applicable), etc.

\begin{table*}[htb]
\scriptsize
\centering
\begin{tabular}{|p{4cm}||c|c|c|c|}
\hline
& Polynomials & Cards & Visual & RL \\\hline
\hline
Base input type & \(\mathbb{R}^4\) & Several-hot vector \(\in \{0,1\}^{12}\) & \(50\!\times\!50\) RGB image & \(91\!\times\! 91\) RGB image \\\hline
Base output type & \(\mathbb{R}\) & Bet values (\(\mathbb{R}^3\)) & Label \(\in \{0, 1\}\) & Action \(Q\)-values \(\in \mathbb{R}^4\)   \\\hline
\hline
Num. base tasks (training) & \multicolumn{1}{p{2.3cm}|}{2260 ( $= 60 + 60 \times 36 + 40$)} & 36 & Varies (\(\sim\)100-300) & 18 \\\hline
Num. base tasks (held out for meta-mapping evaluation) & 1440 ($= 40 \times 36$)  & 4 & Varies & 2 \\\hline
Num. meta classifications & 6 & 8 & 8 & - \\\hline
Num. train meta-mappings & 20 & 3 & Varies (4-32) & 1 \\\hline
Num. held-out meta-mappings & 16 & 0 & 2 & 0  \\\hline
\hline
Base batch size & 1024 & 1024 & 336 & 64 \\\hline
Base support set size & 50 & 768 & - & 32 \\\hline
Meta batch size (train) & 60 & 36 & Varies & 18 \\\hline
Meta support set size (train) & \multicolumn{2}{c|}{Half of train dataset} & - & Half of train dataset \\\hline
Meta support set size (eval) & \multicolumn{2}{c|}{All of train dataset} & - & All of train dataset \\\hline
\end{tabular}
\caption{Dataset compositions and specifications for the different experiments. A ``-'' indicates a parameter that does not apply to that experiment. Batch sizes refer to the total number of data points used per training step (or the number of (s, a, r) tuples for the RL tasks), including both those used as support set examples provided to the example network, and those used as probe examples for generalization. Support set sizes refer to the number of examples presented to the example network in order to construct a task representation. The difference between the batch size and the support set size provides the number used as probes. Note that for the language-based meta-mapping (performed in the visual concepts domain, as well as in later experiments in the RL domain) all the meta-batch is used as probes, since no support set is needed.}
\label{supp_dataset_table}
\end{table*}

\subsubsection{Polynomials} \label{meth_data_poly}
We randomly sampled 100 train polynomials as follows:
\begin{enumerate}
\item Sample the number of relevant variables ($k$) uniformly at random from 0 (i.e. a constant) to the total number of variables.
\item Sample the subset of $k$ variables that are relevant from all the variables.
\item For each term combining the relevant variables (including the intercept), include the term with probability 0.5. If so give it a random coefficient drawn from $\mathcal{N}(0, 2.5)$.
\end{enumerate}
We then split this set of 100 polynomials into 60 that were used to train the meta-mappings, and 40 for which the targets would be held-out to evaluate each meta-mapping. We thus needed to also train the system on the transformed targets for each meta-mapping applied to the 60 polynomials, so the total number of trained polynomials was \(60 + 60 \times 36 + 40 = 2260\). The total number held-out for evaluation was 40 per meta-mapping, i.e. \(40 \times 36 = 1440\). \par
Note that the above means that we trained the system on the transformed polynomials that were in the support set of \emph{even the held-out meta-mappings}. That is, a held-out meta-mapping is held-out in the sense that the meta-mapping itself is not trained, but the supporting polynomials are still in the train set. Of course, in principle the model would be able to perform a meta-mapping supported by polynomials it had never encountered before (using task representations constructed from examples of those polynomials). However, our approach allows more careful evaluation of the meta-mapping generalization of the model, by making the supporting polynomial representations more reliable. This eliminates a confound when comparing held-out meta-mapping generalization to trained meta-mappings, by ensuring base knowledge is matched. \par
The data points on which these polynomials were evaluated were sampled uniformly from $[-1, 1]$ independently for each variable, and an independent set was sampled for each polynomial. Note that although input domain is restricted, the output range can be quite large under this distribution (often around $[-40, 40]$), because of the wide distribution of coefficients and the summing of multiple terms. The datasets were resampled every 50 epochs of training. \par
\textbf{Meta-mappings:} We trained on 20 meta-mapping tasks, and held out 16 related meta-mappings.
\begin{itemize}
\item Squaring polynomials (where applicable, i.e. where degree was \(\leq 1\), so that the squared polynomial wouldn't have degree \(>2\)).
\item Adding a constant (trained constants: -3, -1, 1, 3, held-out: 2, -2).
\item Multiplying by a constant (trained constants: -3, -1, 3, held-out: 2, -2).
\item Permuting inputs (trained on 12 permutations, held-out 12, randomly chosen on each run).
\end{itemize}
\textbf{Meta-classifications:} We also trained the network on 6 task-embedding classification tasks:
\begin{itemize}
\item Classifying polynomials as constant/non-constant.
\item Classifying polynomials as zero/non-zero intercept.
\item For each variable, identifying whether that variable was relevant to the polynomial.
\end{itemize}

\subsubsection{Card games}
\label{meth_data_cards}
Our card games were played with two suits (red and black), and 4 values per suit. In our setup, each hand in a game has a win probability (proportional to how it ranks against all other possible hands). The agent is dealt a hand, and then has to choose to bet 0, 1, or 2 (the three actions it has available). We considered a variety of games which depend on different features of the hand:
\begin{itemize}
\item \textbf{Straight flush:} Most valuable is adjacent numbers in same suit, i.e. 4 and 3 in most valuable suit (royal flush) wins against every other hand. This is the game on which we tested adaptation in the models and human participants.
\item \textbf{High card:} Highest card wins.
\item \textbf{Pairs} Same as high card, except pairs are more valuable, and same suit pairs are even more valuable.
\item \textbf{Match:} The hand with cards that differ least in value (suit counts as 0.5 pt difference) wins.
\item \textbf{Blackjack:} The hand's value increases with the sum of the cards until it crosses 5, at which point the player ``goes bust,'' and the value becomes negative.
\end{itemize}
We also considered three binary attributes that could be altered to produce variants of these games:
\begin{itemize}
\item \textbf{Losers:} Try to lose instead of winning! Reverses the ranking of hands. This is the mapping we evaluated in the models and human participants.
\item \textbf{Suits rule:} Instead of suits being less important than values, they are more important (essentially flipping the role of suit and value in most games).
\item \textbf{Switch suit:} Switches which of the suits is more valuable.
\end{itemize}
Any combination of these options can be applied to any of the 5 games, yielding 40 possible games. We held out all losing variations of the Straight Flush game for evaluation. \par
\textbf{Meta-mappings:} We trained the network on meta-mappings that toggled each of the binary attributes, but evaluated primarily on switching to losing the Straight Flush game (since that corresponded to the human experiment).\par 
\textbf{Meta-classifications:} For meta-tasks, we gave the network 8 task-embedding classification tasks (one-vs-all classification of each of the 5 game types, and of each of the 3 attributes) \par
\textbf{Language:} We encoded the tasks in language by sequences of the form\\
\verb|[``game'', <game_type>, ``losers'', <losers-value>, ``suits rule'', <suits-rule-value>,|\\
\verb|``switch suit'', <switch-suit-value>]|. 

\subsubsection{Visual concepts}
\label{meth_data_visual}
\begin{figure}[!htb]
\centering
\begin{subfigure}{0.24\textwidth}
\includegraphics[width=\textwidth]{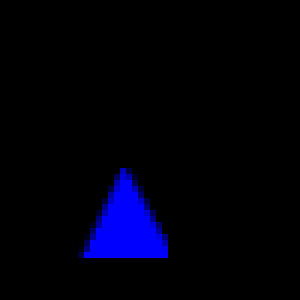}
\end{subfigure}%
\begin{subfigure}{0.24\textwidth}
\includegraphics[width=\textwidth]{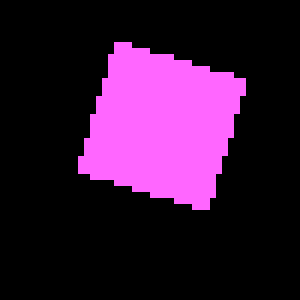}
\end{subfigure}%
\begin{subfigure}{0.24\textwidth}
\includegraphics[width=\textwidth]{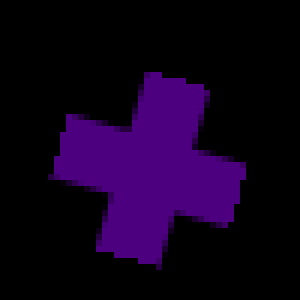}
\end{subfigure}%
\begin{subfigure}{0.24\textwidth}
\includegraphics[width=\textwidth]{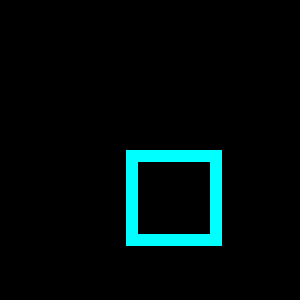}
\end{subfigure}\\
\begin{subfigure}{0.24\textwidth}
\includegraphics[width=\textwidth]{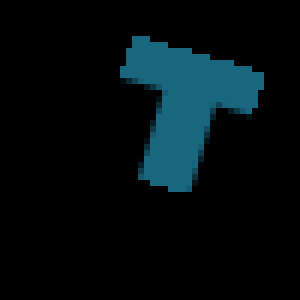}
\end{subfigure}%
\begin{subfigure}{0.24\textwidth}
\includegraphics[width=\textwidth]{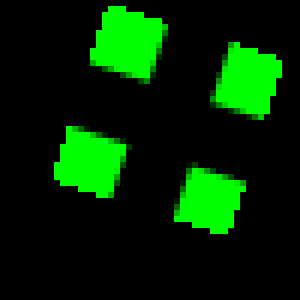}
\end{subfigure}%
\begin{subfigure}{0.24\textwidth}
\includegraphics[width=\textwidth]{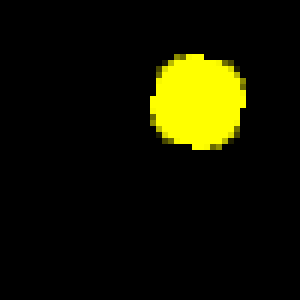}
\end{subfigure}%
\begin{subfigure}{0.24\textwidth}
\includegraphics[width=\textwidth]{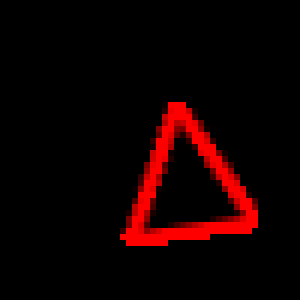}
\end{subfigure}%
\caption{Sample stimuli for visual concept tasks, showing all shapes, colors, and sizes.} \label{supp_fig:HoMM_app_extending_cat_stims}
\end{figure}

In Fig. \ref{supp_fig:HoMM_app_extending_cat_stims} we show all shapes (triangle, square, plus, circle, tee, inverseplus, emptysquare, emptytriangle), colors (blue, pink, purple, yellow, ocean, green, cyan, red), and sizes (16, 24, and 32 pixels) that we used in our experiments. All stimuli were rendered at random positions within a \(50 \times 50\) image (constrained so that the full shape remained within the frame), and at random angles within \(\pm20^{\circ}\) of their canonical orientation.

\textbf{Sampling of meta-mappings:} We sampled an equal number of meta-mappings that switched colors and meta-mappings that switched shape. We held-out one meta-mapping of each type. Within each type, the particular meta-mappings used for training and evaluation on a given run were sampled uniformly at random.  

\textbf{Sampling of basic concepts:} We trained the system on all uni-dimensional concepts as training examples (i.e. one-vs.-all classification of each shape, color, and size), so that it could learn all the basic attributes. We included 6 training example pairs of each mapping (one for each combination of rule type and other attribute). We also included 6 other pairs for evaluation, where the source concept was trained, but the target was held-out for evaluation. Note that our selection criteria mean that each held-out example will have a closely matched trained one. That is, the number of basic concepts the system encounters during training is roughly 18 trained per meta-mapping (roughly because it can be reduced if the meta-mappings have overlapping examples), and the number of evaluation concepts is roughly 6 per meta-mapping. For example, the system might be trained on mappings like ``switch-red-to-blue,'' with corresponding examples like AND(red, triangle) \(\mapsto\) AND(blue, triangle). It would then be evaluated on closely matched examples like AND(red, circle) \(\mapsto\) AND(blue, circle), where the latter is untrained. 

In addition to these sampled pairs, we trained the meta-mapping on any other pairs of concepts which were valid examples of the mapping and happened to be sampled as part of support for other meta-mappings. For example, if AND(red, square) was a train target task for some other mapping, and AND(blue, square) was a trained source task for another, the pair AND(red, square) \(\mapsto\) AND(blue, square) would be used to train the ``switch-red-to-blue'' meta-mapping. 

For a held-out meta-mapping, e.g. ``switch-green-to-blue,'' the same basic concepts instantiating the meta-mapping were trained as would be for a trained mapping, but the meta-mapping itself was not. As in the polynomials domain, matching the training of the supporting basic tasks between trained and held-out meta-mappings makes the comparison between them more precise.

\textbf{Meta-classifications:} In addition to the meta-mappings mentioned in the main text, we trained the system on 9 meta-classifications: classifying whether the task was a basic-level rule on any of the three basic dimensions, classifying whether each dimension was relevant (regardless of whether the task was basic or composite), and classifying the type of composite (if the task was not basic). 

\textbf{Language:} We encoded the tasks in language by sequences from the following grammar: 
\begin{itemize} 
\item \textbf{Basic rules:} encoded as \verb|[<attribute-name>, ``='', <attribute-value>]|, for example\\\verb|[``shape'', ``='', ``triangle'']|
\item \textbf{Composite rules:} encoded as \verb|[<composite-type>, ``('', ``('', <basic-rule>, ``)'',|\\
\verb|``&'', ``('', <basic-rule>, ``)'', ``)'']|, where the \verb|<composite-type>| is one of ``AND'', ``OR'', or ``XOR'', and each \verb|<basic-rule>| is substituted with a sequence as above.
\item \textbf{Meta-mappings:} encoded as \verb|[``switch'', <attribute-name>, <old-attribute-value>, ``~'',|\\
\verb|<new-attribute-value>]|.
\item \textbf{Meta-classifications:} encoded as \verb|[``is'' <composite-type>]| or\\\verb|[``is'', ``basic'', ``rule'', <attribute-name>]| or\\\verb|[``is'', ``relevant'', <attribute-name>]|,
depending on the type of classification.
\end{itemize}

\subsubsection{RL} \label{meth_data_RL}

The RL tasks were implemented using the open-source Pycolab library (\url{https://github.com/deepmind/pycolab}). The tasks were implemented in a \(6 \times 6\) room, surrounded by an impassable varrier. The agent could navigate using four actions, corresponding to moving in the four cardinal directions. If it attemtped an invalid action, the state did not change.\par
Each episode ended after either 150 timesteps elapsed (that is, after the agent took 150 actions, including invalid actions), or after the agent had picked up 4 of the 8 objects (regardless of whether they were good or bad) in the pick-up task, or pushed off 4 of the 8 in the push-off task The agent received a reward of \(+1\) for picking up or pushing off the good-colored objects, and \(-1\) for the bad-colored objects. Selected recordings of the agent playing the games after meta-mapping can be found at \url{https://github.com/lampinen/homm_grids/tree/master/recordings}, which may help clarify any unclear aspects of the tasks. 

\textbf{Meta-classifications:} We did not train any meta-classifications in this setting.

\textbf{Language:} We encoded the tasks in language by sequences of the following form: 
\begin{itemize} 
\item \textbf{Basic task:} encoded as \verb|[<game-type>, <color1>, <color2>, <good-color-position>]|, where \verb|<game-type>| was either ``pusher'' or ``pickup'', colors were names of a color pair, and \verb|<good-color-position>| was ``first'' or ``second'' depending on whether the first color was good, or the second color (after switching).
\item \textbf{Meta-mapping:} encoded as \verb|[``switch'', ``colors'']|.
\end{itemize}

\subsection{Cards behavioral experiment} \label{supp_sec:behavioral}
Here we provide the details of the human experiment for the cards tasks. The human experiment was conducted on Amazon Mechanical Turk.
We tried to design the game that participants played to make it easy for them to learn, without relying on their prior knowledge of card games. The game was a simplified variation of poker, which we denoted ``Straight Flush'' in the card game descriptions above. The participants were dealt hands which consisted of two cards, each with a number (rank) between 1 and 4, and a color (suit) of red or black. The participants played against a computer opponent that was dealt a similar hand. The hands were ranked such that straight flushes (adjacent cards in the same suit) beat adjacent cards in a different suit, which beat non-adjacent cards (including pairs). Ties were broken by the highest card, or by suit if both cards were tied. \par

\begin{figure}[htb]
\centering
\begin{subfigure}[b]{0.5\textwidth}
\includegraphics[width=\textwidth]{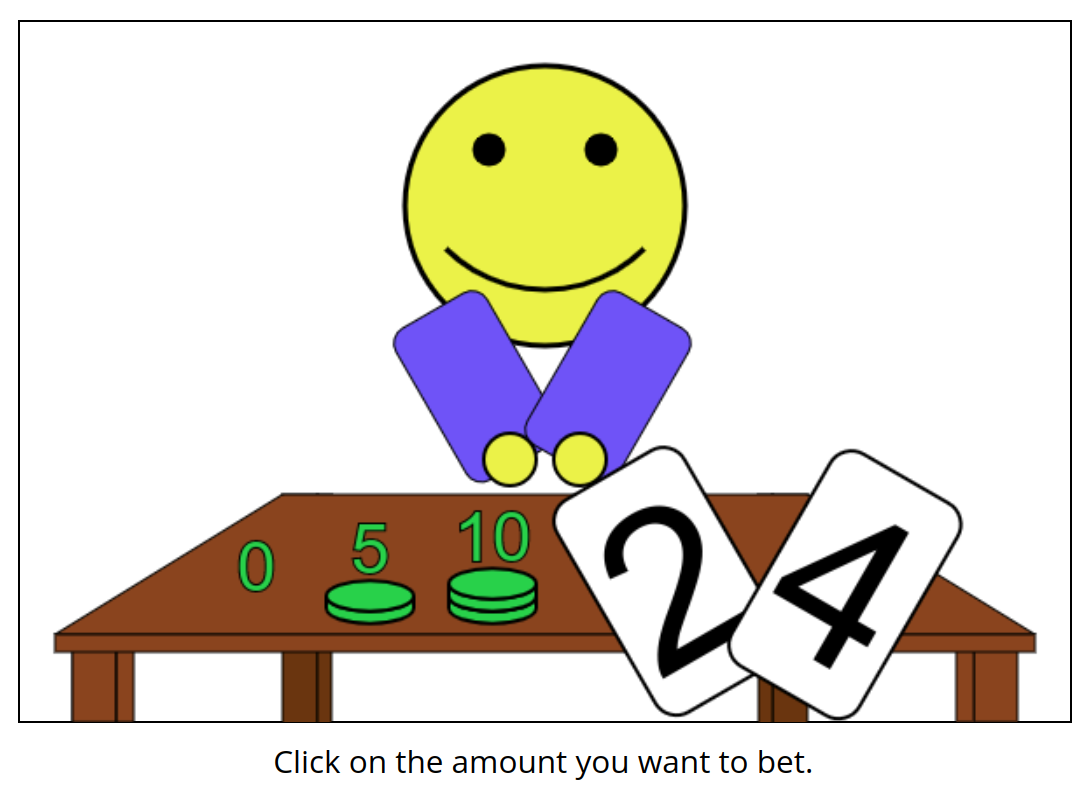}
\caption{Before betting.}\label{fig:human_betting_trial:pre}
\end{subfigure}%
\begin{subfigure}[b]{0.5\textwidth}
\includegraphics[width=\textwidth]{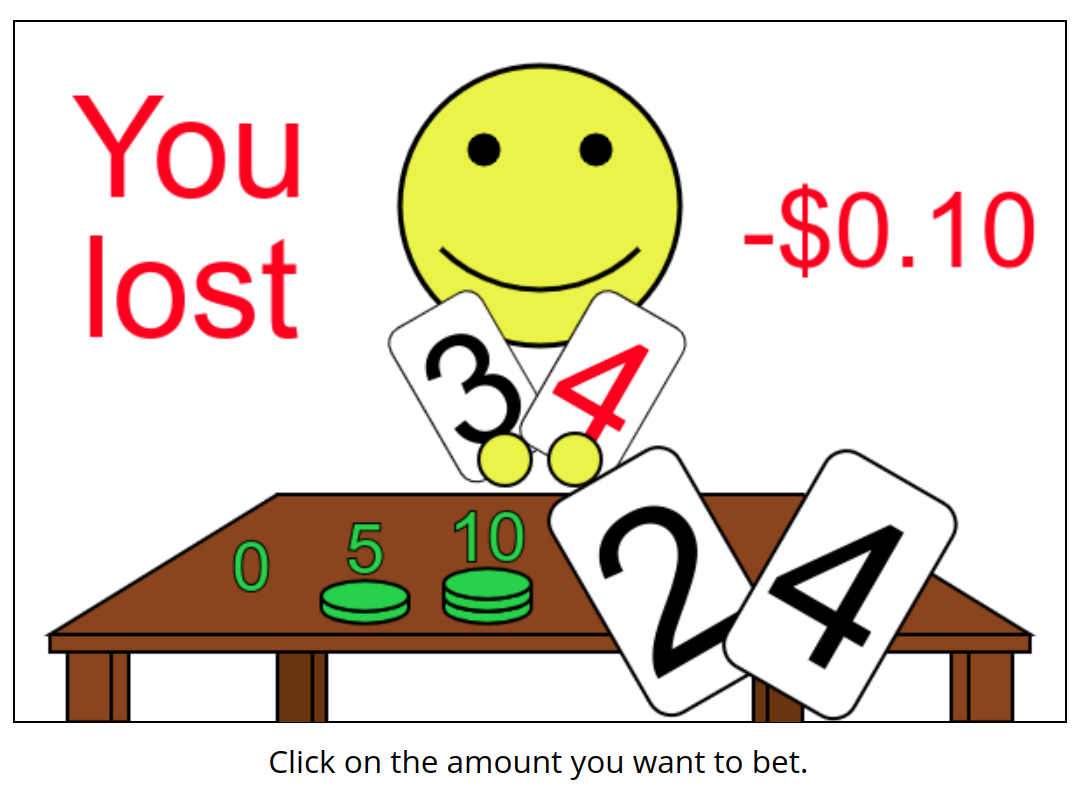}
\caption{Feedback.} \label{fig:human_betting_trial_feedback}
\end{subfigure}%
\caption{The card game experiment trials, as seen by participants. (\subref{fig:human_betting_trial:pre}) The beginning of the trial, in which participants can see their hand, and choose an amount to bet by clicking on it. (\subref{fig:human_betting_trial_feedback}) The feedback phase, where participants saw their opponents hand and the result. In the evaluation trials, where participants did not receive feedback, this phase was replaced with a semi-transparent gray overlay before the next trial.} \label{fig:human_betting_trial}
\end{figure}
On each trial, participants were dealt a hand and asked to make a bet of 0, 5, or 10 cents (see Fig. \ref{fig:human_betting_trial}). If their hand beat the opponent's hand, they won the bet amount. If their hand lost, they lost it. If the hands were tied, they neither won nor lost money. \par
The experiment had several phases. First, participants were instructed in the rules and payment scheme for the experiment. Next, they were instructed on the rules of the game. After this, they were tested with four hand-comparison trials intended to probe their understanding of each of the rules of the game. If they failed more than one of these trials, they were not allowed to continue with the experiment. \par
Following this understanding check, participants played a block of 32 hands (sampled to have a diversity of expected values), where they saw the results of their play (as in Fig. \ref{fig:human_betting_trial_feedback}). After this block, they played a similar block of 24 trials where they did not see the results of their play. The results were replaced with a brief grayed-out screen, and participants were payed the net expected value of their actions over the block (rounded to the nearest 10). The evaluation phase without feedback provides an evaluation with relatively less potential for learning, in order to get a precise estimate of their performance. \par
Finally, participants were told that we wanted them to try to lose for the remaining trials, and that ``for the remainder of the experiment, if you bet and lose, you'll gain the amount you bet, and if you bet and win, you'll lose the amount you bet.'' They were then given an attention check to evaluate whether they had understood this instruction. Subjects who failed this attention check were excluded from the analysis. They then played another block of 24 trials where they were rewarded for losing instead of winning (i.e. the relationship between actions and expected returns was reversed relative to the first phase of the experiment). As in the previous block, they did not see the results of their actions, they were only shown their total earnings at the end of the block. By not providing feedback on each trial, we were able to get many trials of ``zero-shot'' data, to more carefully evaluate their performance. They were finally asked a few demographic questions. \par
Our target comparison was performance in the two blocks without feedback -- were participants able to switch their behavior to lose at the game as well as they won at it? Rather than evaluating on stochastic rewards based on sampled opponent hands, we evaluated them by the expected value of their performance across the hands they played. This is exactly analogous to the experiment performed for the model (except that the performance of the model was evaluated on all possible hands in each condition, which was infeasible for the human participants). \par
Participants were paid \$1 for starting the experiment and completing the instruction section. If they failed the first understanding check, the experiment ended. Otherwise, they were paid an additional \$1.50 to complete the performance phase, and then were bonused based on their winnings to incentivize performance. We recruited 40 participants for the experiment, but only 19 successfully passed the first understanding trials. Of those 19, only 17 passed the try-to-lose attention check, so our analyses were restricted to 17 subjects.\par
Further details of the experiment, including the text of all instructions, can be found in the first author's dissertation \citep[][pp. 112-117, accessible at \url{https://stacks.stanford.edu/file/druid:xj689nb3522/dissertation-augmented.pdf}]{lampinen2020computational}.

%% TODO: remove this if/when it goes down.
%At present, the cards experiment (as seen by the participants) can be experienced at: \url{http://web.stanford.edu/~lampinen/mturk/cards/web/pilot.html}

\subsection{Source repositories} \label{supp_sec:github}
%The full code for all experiments and analyses will be made available via github in the de-anonymized verison.
The full code for the experiments and analyses can be found on github:
\begin{itemize}
\item Meta-mapping library: \url{https://github.com/lampinen/HoMM}
%\item This paper's source: \url{https://github.com/lampinen/metamapping_paper}
\item Polynomials: \url{https://github.com/lampinen/HoMM_polynomial_analysis}
\item Cards (models): \url{https://github.com/lampinen/HoMM_cards}
\item Cards (human experiment): \url{https://github.com/lampinen/cards_for_humans}
\item Concepts: \url{https://github.com/lampinen/categorization_HoMM}
\item RL: \url{https://github.com/lampinen/HoMM_grids}
\item Stroop results (below): \url{https://github.com/lampinen/stroop}
\end{itemize}

\subsection{Other acknowledgements}

The color palettes used in the figures are adapted from ColorBrewer \citep{harrower2003colorbrewer}. The playing card images used in the main text are based on the images at \url{https://commons.wikimedia.org/wiki/Category:Playing_cards_set_by_Byron_Knoll} on WikiCommons, which the creator kindly released for use.

\FloatBarrier
%\newpage
\subsection{Supplemental analyses \& figures} \label{supp_sec:analyses}
 
The analyses are organized as follows. In \ref{supp_sec:analyses:polynomials}, we show additional analyses in the polynomial domain, including evaluation of sample efficiency and several architectural lesions. In \ref{supp_sec:analyses:polynomial_reps}, we analyze the representations of the models in the polynomial domain, showing that they are systematically organized across runs, and presenting further details on the representation transformation results presented in the main text. We also show evidence that the meta-mapping and basic task representations are sharing representational subspaces, and show significant overlap with isomorphisms we know exist between them in the polynomials domain. In \ref{supp_sec:analyses:cards}, we show further analyses of the results of the card game experiments, both from the behavioral and modeling perspective. In \ref{supp_sec:analyses:concepts} we show more detailed results in the visual concepts domain. In \ref{supp_sec:analyses:RL} we show further analyses of the RL experiments. In \ref{supp_sec:analyses:language} we provide details and analyses for the comparisons to generalizing from language alone, and in \ref{supp_sec:analyses:RL_color_shape} we provide additional experiments on generalizing from switching colors to switching shapes in the RL context. In \ref{supp_sec:analyses:timescales}, we provide further analyses of meta-mapping as a starting point for later learning. In \ref{supp:HoMM_cognitive_control}, we demonstrate our model on a simple Stroop-like task, common in cognitive control.

\subsubsection{Polynomials} \label{supp_sec:analyses:polynomials}

\begin{figure}[tbh]
\centering
\includegraphics[width=0.5\textwidth]{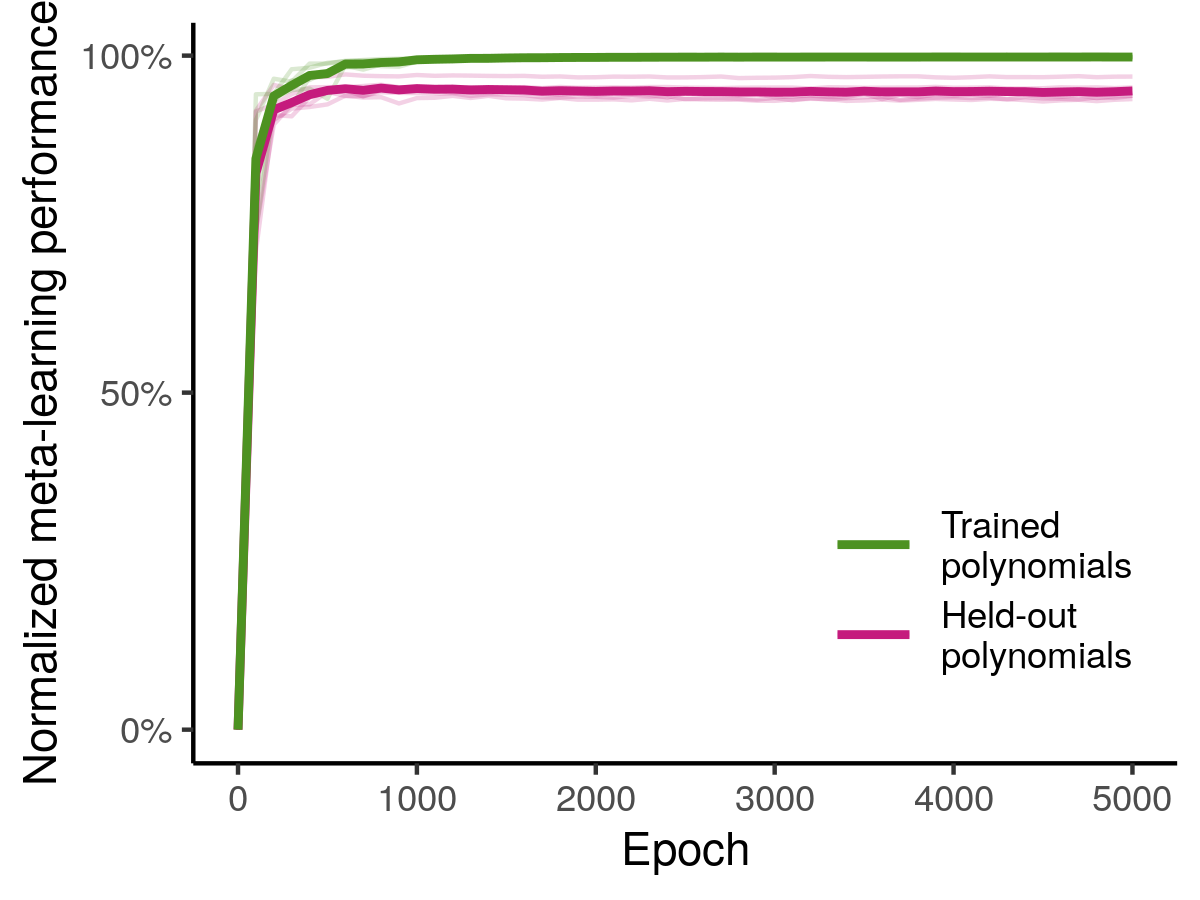}
\caption[Basic task (meta-learning) performance in the polynomials domain over learning.]{Basic task (meta-learning) performance in the polynomials domain over learning. The system is generalizing at the meta-learning level. That is, this graph shows that, after the example network receives a set of (input, output) example tuples, it is generating a sufficiently good representation to regress held-out points from that polynomial. This is true both for polynomials it was trained with (green), and for polynomials that are held-out and never encountered during training (pink). Performance is plotted normalized as 100\% \(\times(1 - \text{loss}/c)\), where \(c\) is the loss for a system outputting all zeros, as in the meta-mapping results. In this case, this measure corresponds exactly to the percentage of variance explained. (Thick dark curves are averages over 5 runs, shown as light curves.)} \label{supp_fig:HoMM:polynomials_basic_meta_learning}
\end{figure}

\begin{table}[tbh]
\centering 
\begin{tabular}{@{\extracolsep{8pt}} ccccccc}
\\[-1.8ex]\hline
\hline \\[-1.8ex]
\multicolumn{2}{c}{Evaluation types} & \multicolumn{3}{c}{MSE loss}  & \multicolumn{2}{c}{Normalized performance} \\\cmidrule{1-2} \cmidrule{3-5}\cmidrule{6-7}
Meta-mapping trained? & Polynomial is trained example? & Meta-mapping & Zeros & No adaptation & Meta-mapping & No adaptation \\
\hline \\[-1.8ex]
Trained & Support (trained) & 0.317 & 18.8 & 18.1 & 98.3\% & 4.18\% \\
Trained & Probe (held-out) & 1.85 & 16.7 & 16 & 89\% & 4.26\% \\
New & Support (trained) & 0.97 & 12.4 & 9.89 & 92.1\% & 20.6\% \\
New & Probe (held-out) & 1.56 & 10.8 & 8.71 & 85.5\% & 19.3\% \\
\hline \\[-1.8ex]
\end{tabular}
\caption{The raw mean-squared-error (MSE) losses and the normalized performance measures after meta-mapping in the polynomials domain. The first column indicates whether the meta-mapping is trained or held-out, and the second indicates whether the polynomial is provided as an example of the mapping (and is therefore used for training the mapping, if the mapping is trained) or if the polynomial is held out for evaluation. The meta-mapping columns provide the MSE/performance for the model after meta-mapping, the zeros column provides the loss for a model that outputs all zeros, and the ``No adaptation'' columns provide the MSE/performance for a model using the unadapted source task representations. The normalized performance measure is calculated as 100\% \(\times (1 - \text{Model MSE} / \text{Zeros MSE})\).} 
\label{supp_tab:analyses:poly:loss_perf} 
\end{table} 

\begin{figure}[tbh]
\centering
\includegraphics[width=0.5\textwidth]{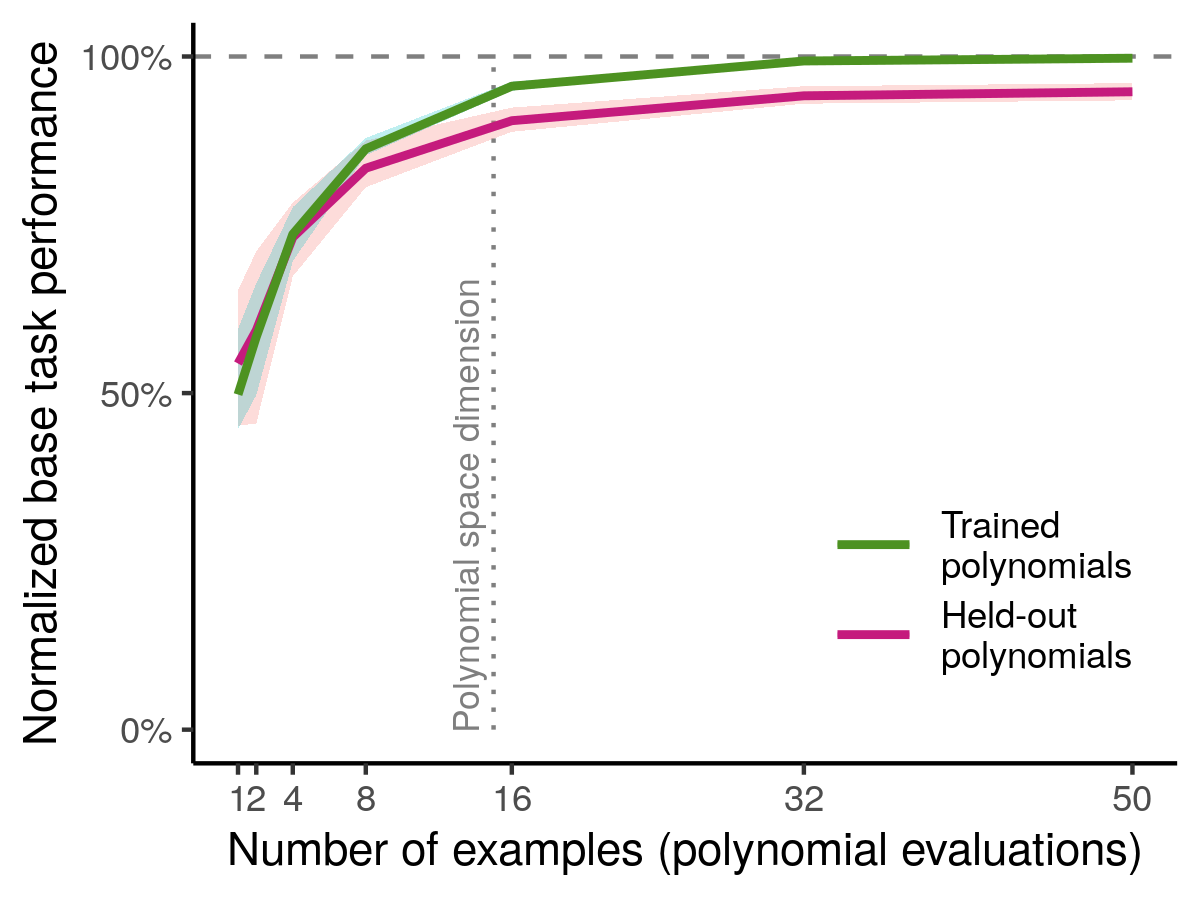}
\caption[The effect of number of examples on basic task performance in the polynomials domain.]{The effect of number of examples on basic task performance in the polynomials domain. The system is relatively sample efficient. Its performance is quite high by the time it has received the minimum number of samples that an optimal least-squares solver with knowledge of the ground-truth task space would need (that is, the dimensionality of the polynomial vectors space, indicated by the vertical dotted line), although the performance continues to improve slowly beyond that point. (Averages across 4 runs, with bootstrap 95\%-CIs across runs.)} \label{supp_fig:HoMM:polynomials_varying_mbs_base}
\end{figure}

\textbf{Basic meta-learning:} In Fig. \ref{supp_fig:HoMM:polynomials_basic_meta_learning}, we show that the basic meta-learning is working well in the polynomials domain. That is, we show that after the example network is presented with a set of example input, output pairs from a polynomial, the system is generalizing well to other points from that polynomial. At the end of training, the mean performance on trained polynomials is 99.78\% (bootstrap 95\%-CI [99.74, 99.84]), and for held-out polynomials it is 94.8 (bootstrap 95\%-CI [93.8, 95.9]).

\textbf{Relationship of raw meta-mapping performance to normalized performance:} In Table \ref{supp_tab:analyses:poly:loss_perf}, we show the relationship between the mean-squared-error (MSE) losses of the model on meta-mapped tasks, and the normalized performance measure we report in the main text. Note that this table includes results for training examples, while the main text only reports the evaluation results. For context, the MSE of the model when performing a trained polynomial from examples is 0.025, and the performance is 99.8\% (see above), so the model is not performing quite as well after meta-mapping even a trained example polynomial as it performs a trained polynomial from examples. This is not particularly surprising, since there are more sources of noise in meta-mapping a task --- first, the representation of the source task, next the representation of the meta-mapping, and finally the transformation itself. 

\textbf{Sample efficiency (base tasks):} In Fig. \ref{supp_fig:HoMM:polynomials_varying_mbs_base}, we explore the sample efficiency of the basic meta-learning system by evaluating how the performance of the system changes depending on the number of examples it is given. Note that because the models were trained with 50 examples per polynomial, performance at smaller sizes would likely improve somewhat beyond these results if it were trained initially with smaller numbers of examples.

\begin{figure}[H]
\centering
\includegraphics[width=0.625\textwidth]{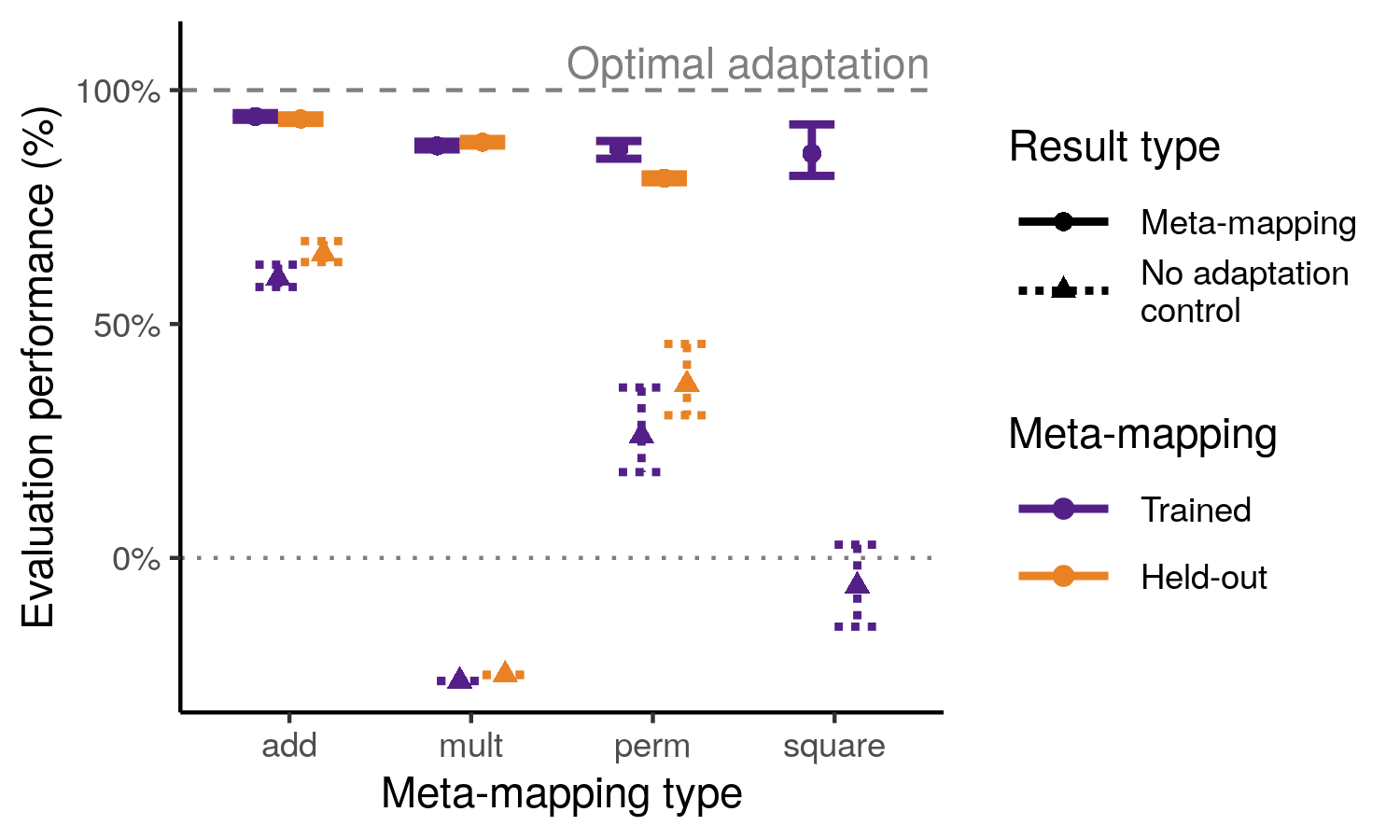}
\caption[Meta-mapping performance in the polynomial domain, broken down by meta-mapping type.]{Meta-mapping performance in the polynomials domain, broken down by meta-mapping type. We plot a normalized performance measure, as in the main text. The system is performing well across all meta-mapping types, although there is some variability. Triangles show performance of a baseline model that does not adapt --- note that some meta-mappings are relatively easier for such a model, while in other cases such a model results in worse performance than outputting all zeros.} \label{supp_fig:HoMM_polynomials_results_by_mapping}
\end{figure}

\textbf{Meta-mapping results by mapping type:} In Fig. \ref{supp_fig:HoMM_polynomials_results_by_mapping} we show the meta-mapping results in the polynomials domain, broken down by the type of mapping. The system performs well across all mapping types. 

\begin{figure}[H]
\centering
\includegraphics[width=0.5\textwidth]{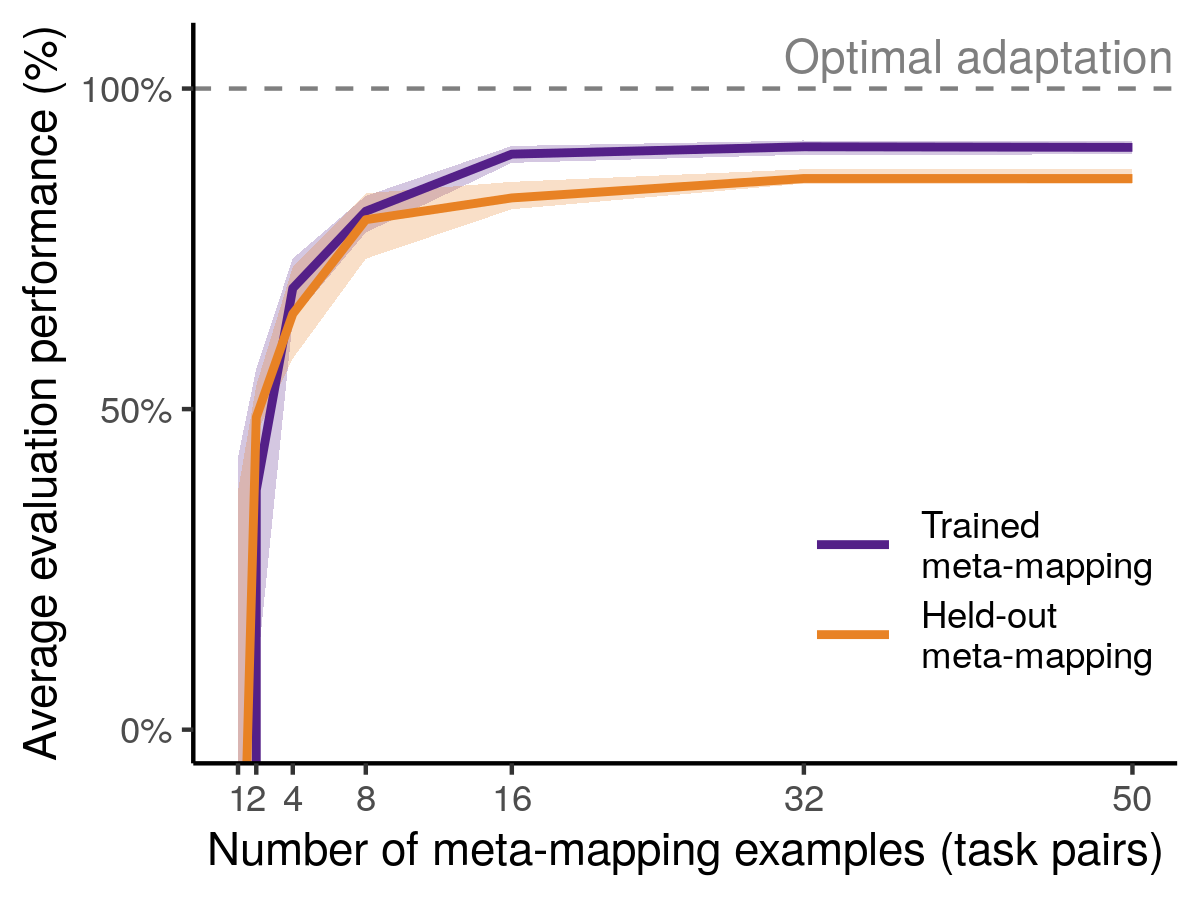}
\caption{The effect of number of examples on meta-mapping performance (for add, multiply, and permute) in the polynomials domain. The system is relatively sample efficient. Although the system was trained with 30 examples of each meta-mapping, performance is relatively stable above 16 examples. (Averages across 4 runs, with bootstrap 95\%-CIs across runs. The square meta-mapping is omitted from the data in this plot because of its unique trajectory, see Fig. \ref{supp_fig:HoMM_polynomials_varying_mbs_meta_by_mm}.)} \label{supp_fig:HoMM_polynomials_varying_mbs_meta}
\end{figure}

\begin{figure}[H]
\centering
\includegraphics[width=\textwidth]{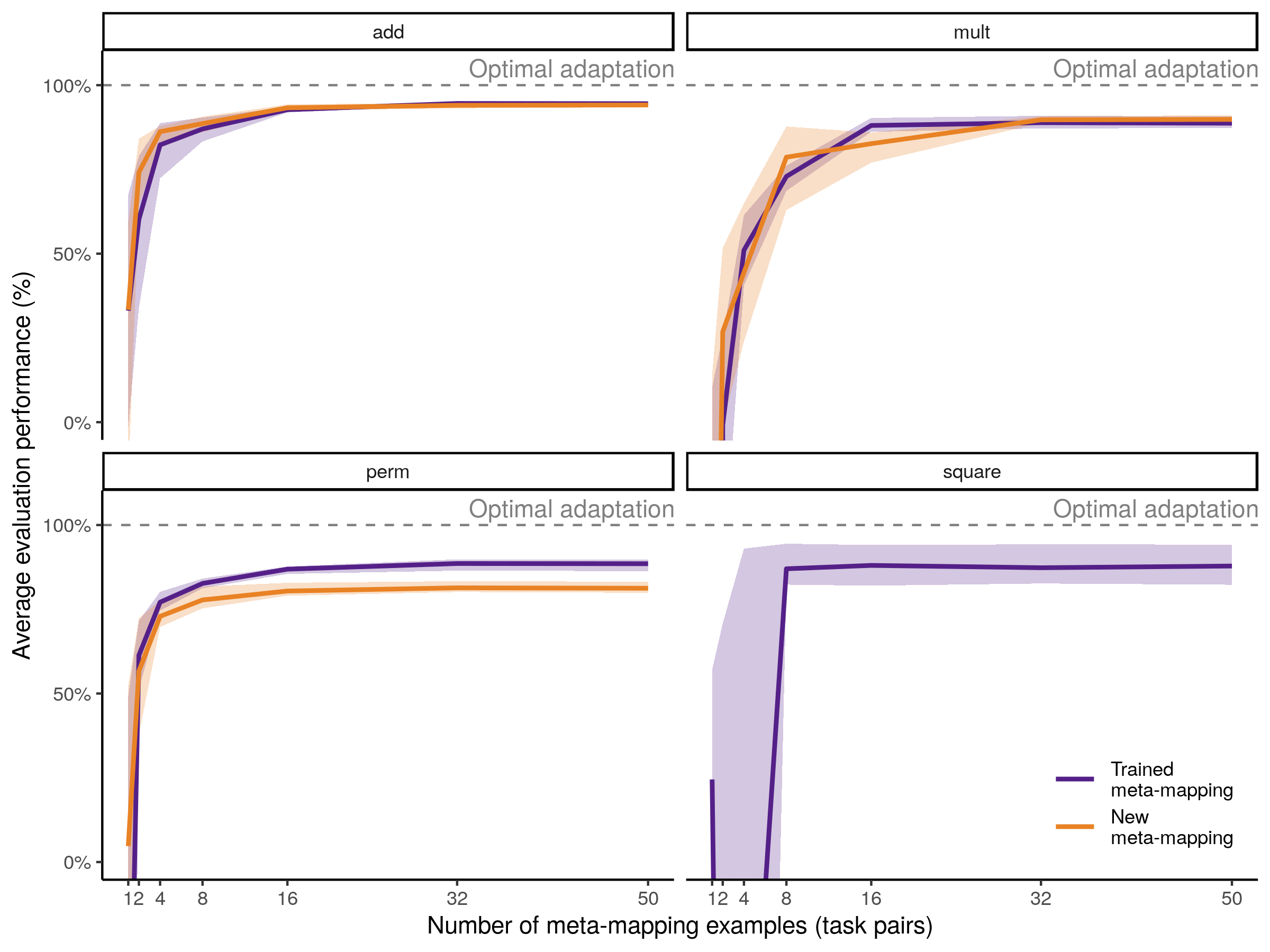}
\caption{The effect of number of examples on meta-mapping performance in the polynomials domain, broken down by meta-mapping type. The sample efficiency of the system depends on the meta-mapping. In particular, the square meta-mapping is difficult to estimate from few examples, and performance on that mapping is quite low with small numbers of examples. (Averages across 4 runs, with bootstrap 95\%-CIs across runs.)} \label{supp_fig:HoMM_polynomials_varying_mbs_meta_by_mm}
\end{figure}

\textbf{Sample efficiency (meta-mappings):} In Fig. \ref{supp_fig:HoMM_polynomials_varying_mbs_meta}, we show how the meta-mapping performance depends on the number of examples --- that is, (input task, output task) tuples --- that the system is given. Performance is unsurprisingly quite low with 1 example, but increases rapidly with a few examples. In Fig. \ref{supp_fig:HoMM_polynomials_varying_mbs_meta_by_mm} we show performance by number of examples for each meta-mapping type. The square meta-mapping in particular is difficult, and performance is actually negative with only a few examples of it, unlike the other mappings. However, once the system receives enough examples, it is able to recognize the square mapping and perform well at it.  

\begin{figure}[H]
\centering
\begin{subfigure}{0.5\textwidth}
\includegraphics[width=\textwidth]{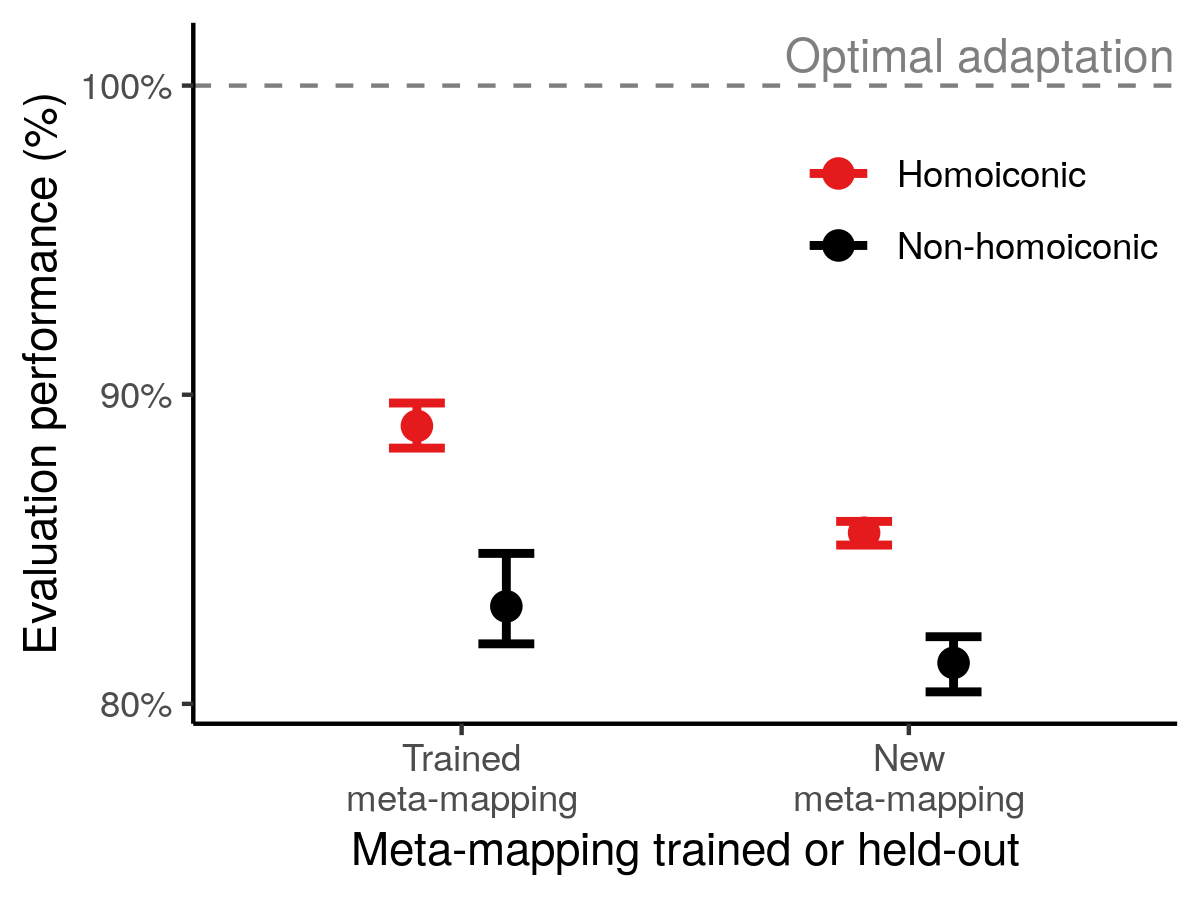}
\caption{The polynomial domain, compare to Fig. \ref{fig:HoMM_polynomials:results}.} \label{supp_fig:HoMM_nonhomoiconic_lesion:polynomial}
\end{subfigure}%
\begin{subfigure}{0.5\textwidth}
\includegraphics[width=\textwidth]{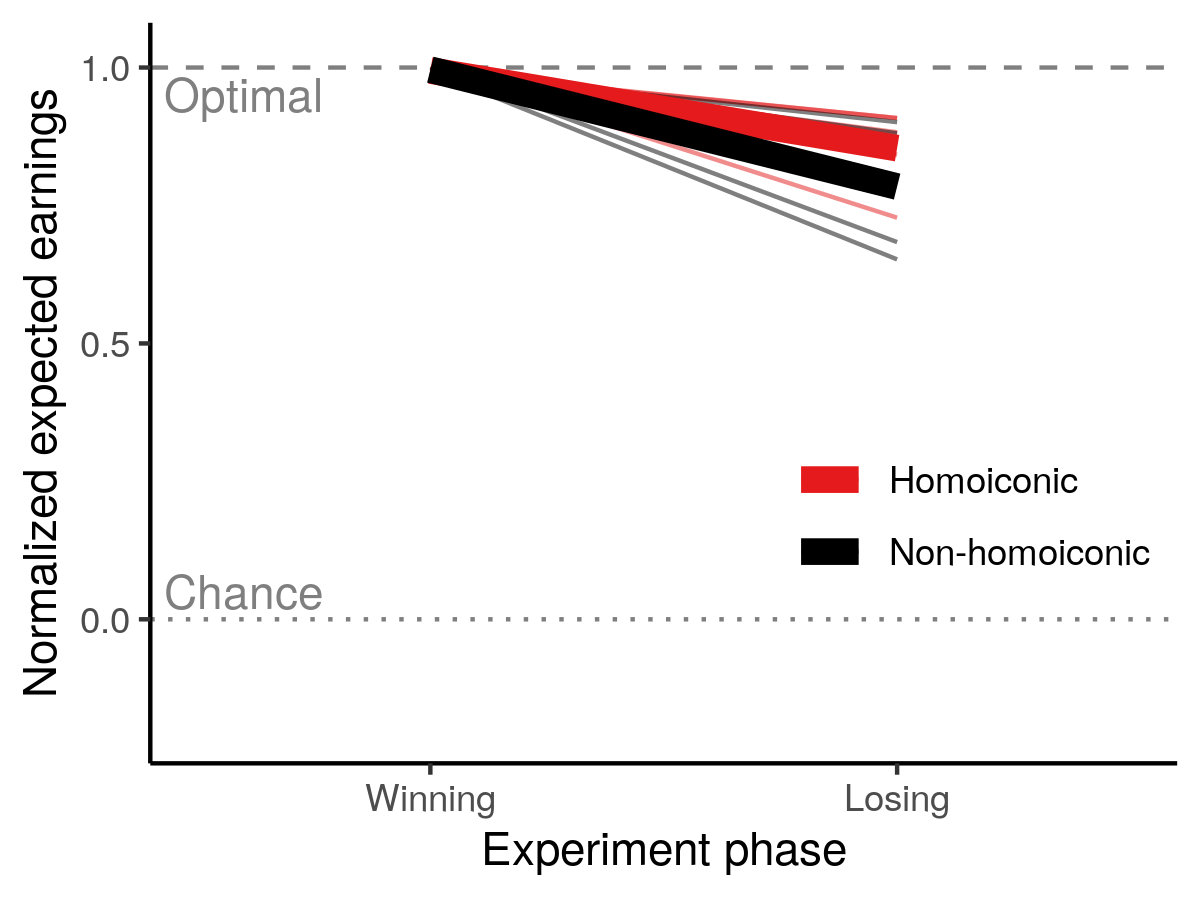}
\caption{The cards domain, compare to Fig. \ref{fig:HoMM_cards}.} \label{supp_fig:HoMM_nonhomoiconic_lesion:cards}
\end{subfigure}
\caption[HoMM outperforms or equals a non-homoiconic baseline in the polynomials and cards domains.]{Our homoiconic model outperforms or equals a non-homoiconic baseline in the polynomials and cards domains. This figure compares the meta-mapping performance of our architecture with that of a nonhomoiconic model that instantiates separate copies of the example network (\(\mathcal{E}_{base},\mathcal{E}_{meta}\)) and hyper network (\(\mathcal{H}_{base},\mathcal{H}_{meta}\)) for the basic tasks and the meta-mappings. In the polynomials domain (\subref{supp_fig:HoMM_nonhomoiconic_lesion:polynomial}), the homoiconic architecture significantly outperforms the nonhomoiconic one, while in the cards domain (\subref{supp_fig:HoMM_nonhomoiconic_lesion:cards}), the difference is not significant. These results suggest that there is sufficient shared structure between the basic tasks and the meta-mappings for the homoiconic approach to improve generalization, at least in the polynomials case, and supports our use of homoiconic architectures.} \label{supp_fig:HoMM:nonhomoiconic_baseline}
\end{figure}

\textbf{Nonhomiconic architectures:} We next consider some architecture lesions. In Fig. \ref{supp_fig:HoMM:nonhomoiconic_baseline}, we compare our homoiconic architecture to a nonhomoiconic architecture -- i.e. one in which there are separate example networks (\(\mathcal{E}_{base},\mathcal{E}_{meta}\)) and hyper networks (\(\mathcal{H}_{base},\mathcal{H}_{meta}\)) for the base tasks and meta-mappings. The nonhomoiconic approach performs substantially worse. Specifically, on trained meta-mappings the HoMM model is achieving a normalized performance of 88.99\% (bootstrap 95\%-CI [88.20, 89.98]), while the non-homoiconic achieving a normalized performance of 83.2\% (bootstrap 95\%-CI [81.9, 84.9]). On new meta-mappings the HoMM model is achieving a normalized performance of 85.54\% (bootstrap 95\%-CI [85.14, 85.94]), while the non-homoiconic model is achieving a normalized performance of 81.3\% (bootstrap 95\%-CI [80.3, 82.2]). (See also Sec. \ref{supp_sec:analyses:polynomial_reps}, in which we show that there is intriguing overlap between the representations of meta-mappings and base tasks in a homoiconic architecture.)

\begin{figure}[htb]
\centering
\begin{subfigure}{0.5\textwidth}
\includegraphics[width=\textwidth]{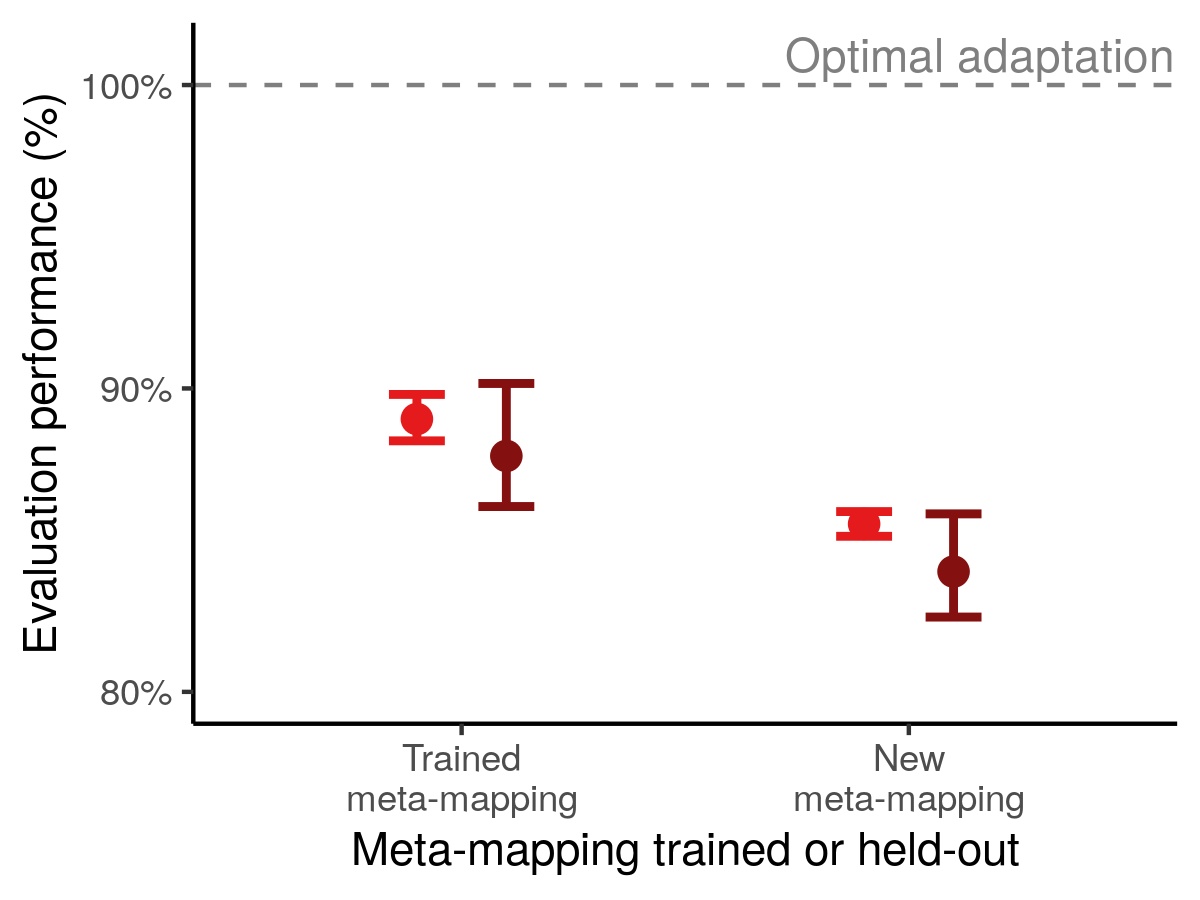}
\caption{The polynomial domain, compare to Fig. \ref{fig:HoMM_polynomials:results}.} \label{supp_fig:HoMM_arch_cond_vs_hyper:polynomial}
\end{subfigure}%
\begin{subfigure}{0.5\textwidth}
\includegraphics[width=\textwidth]{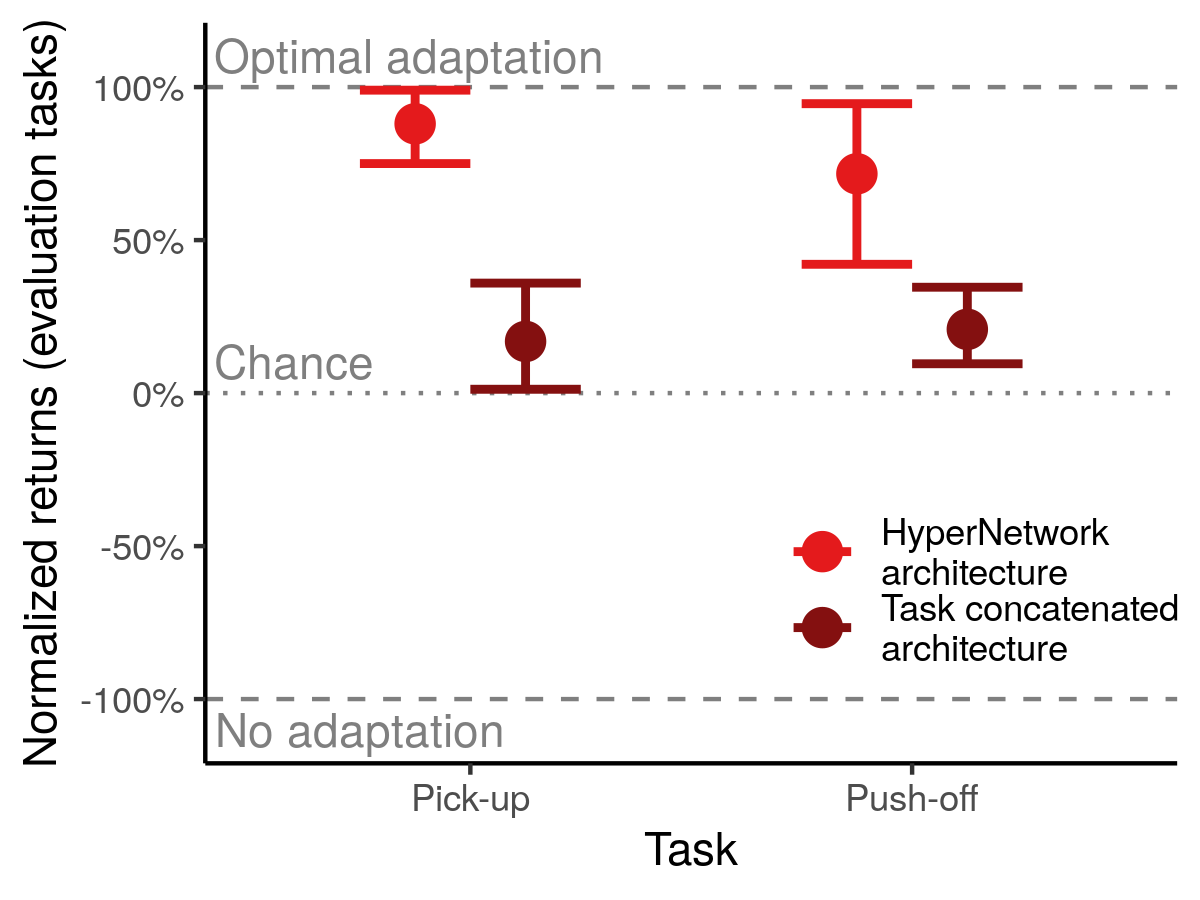}
\caption{The RL domain, compare to Fig. \ref{fig:HoMM_RL_results}.} \label{supp_fig:HoMM_arch_cond_vs_hyper:RL}
\end{subfigure}
\caption{The HyperNetwork-based architecture we propose in the main text performs as well or better on meta-mappings than an architecture that simply concatenates a task representation to the input before passing it through a fixed MLP, at least on the subset of our domains on which we ran a comparison. (See Fig. \ref{supp_fig:human_cards_lang_tcnh_vs_hyper} for a similar comparison for the language generalization baseline.)}\label{supp_fig:HoMM_arch_cond_vs_hyper}
\end{figure}

\textbf{A simpler task architecture:} In Fig. \ref{supp_fig:HoMM_arch_cond_vs_hyper:polynomial} we show that a simpler task network, which just takes a task representation as another input to feed-forward processing, performs perhaps slightly worse than the HyperNetwork-based approach. Specifically, in the simpler architecture, there is a fixed feed-forward task network, and rather than using the task representation to alter the weights of this network, the task-representation is simply concatenated to the input representation and then propagated through the fixed network. Note that the task-concatenated architecture does not perform worse at meta-learning (normalized performance on evaluation tasks 95.7\%, bootstrap 95\%-CI [95.0, 96.6] vs. 94.8\% [93.8, 95.9]), it is adapting via meta-mappings that proves challenging for it. 

\begin{figure}[htb]
\centering
\begin{subfigure}{0.5\textwidth}
\includegraphics[width=\textwidth]{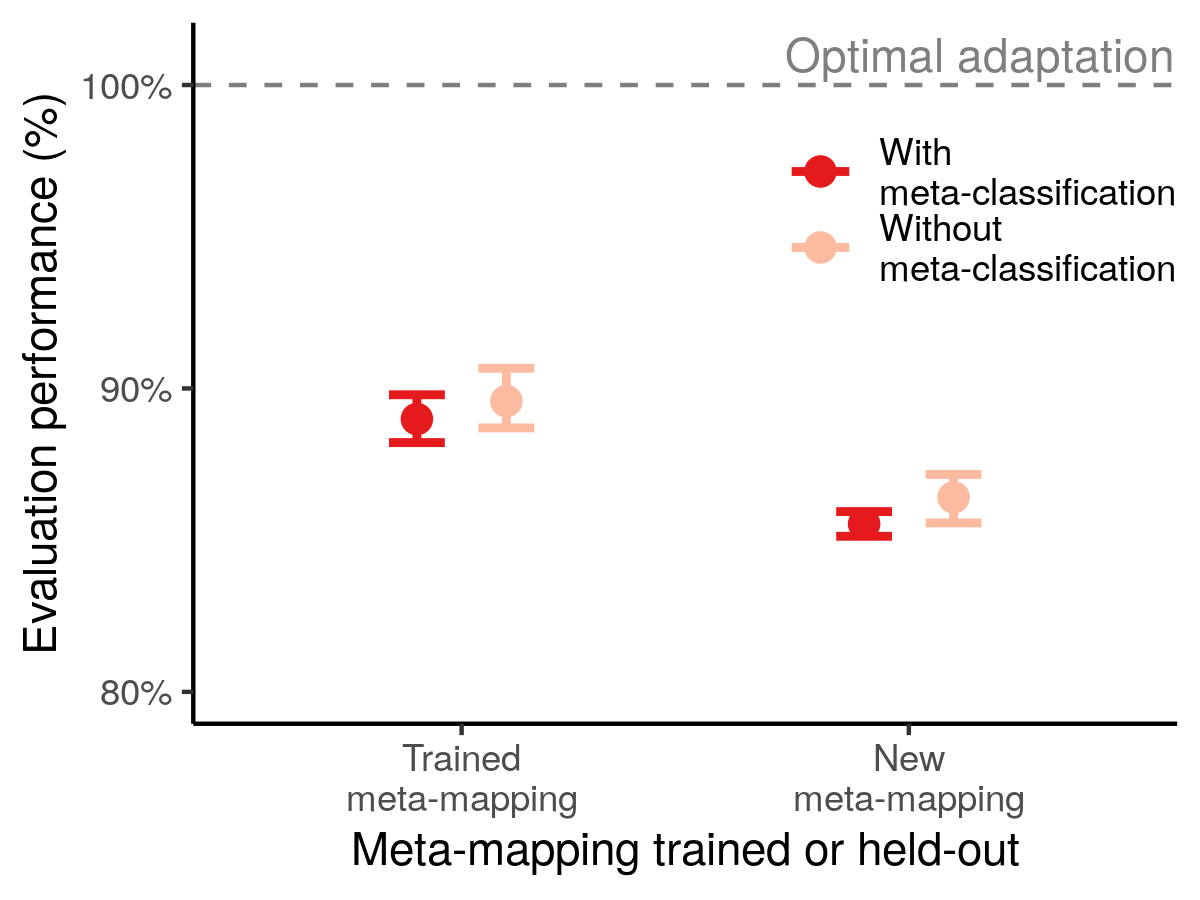}
\caption{The polynomial domain, compare to Fig. \ref{fig:HoMM_polynomials:results}.} \label{supp_fig:HoMM_metaclass_lesion:polynomial}
\end{subfigure}%
\begin{subfigure}{0.5\textwidth}
\includegraphics[width=\textwidth]{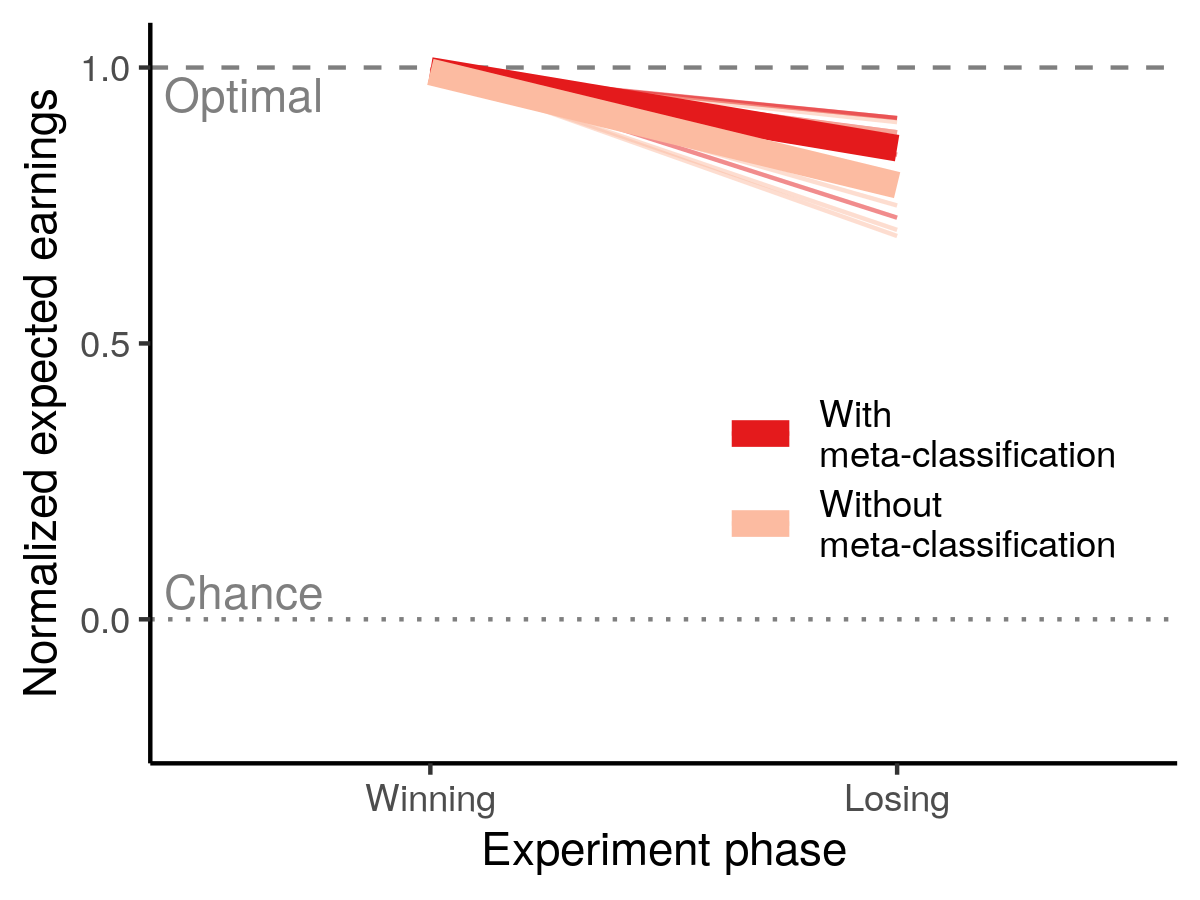}
\caption{The cards domain, compare to Fig. \ref{fig:HoMM_cards}.} \label{supp_fig:HoMM_metaclass_lesion:cards}
\end{subfigure}
\caption[The meta-classifications we trained the model with do not appear to be substantially beneficial.]{The meta-classifications we trained the model with do not appear to be substantially beneficial --- a model trained without them performs slightly better in the polynomials domain, while the model trained with them performs marginally better in the cards domain. This difference may be due to the fact that the model is trained on many more basic tasks in the polynomials domain, perhaps obviating the need for meta-classification to shape the representations.}\label{supp_fig:HoMM_metaclass_lesion}
\end{figure}

\textbf{Meta-classification task lesion:} In Fig. \ref{supp_fig:HoMM_metaclass_lesion:polynomial} we show that the meta-classification training is not beneficial in the polynomials domain. Specifically, on trained meta-mappings the model is achieving a normalized performance of 88.99\% (bootstrap 95\%-CI [88.20, 89.98]), while without meta-classification it is achieving a normalized performance of 89.7\% (bootstrap 95\%-CI [88.87, 90.61]). On new meta-mappings the model is achieving a normalized performance of 85.54\% (bootstrap 95\%-CI [85.14, 85.94]), while without meta-classification it is achieving a normalized performance of 86.29\% (bootstrap 95\%-CI [85.54, 86.79]). However, the effect is small, and in Fig. \ref{supp_fig:HoMM_metaclass_lesion:cards} we show that meta-classification may be helpful in the cards domain, where there are fewer training tasks.

%\newpage
\newpage

%\newpage

%\newpage

\FloatBarrier
\subsubsection{Polynomial representations} \label{supp_sec:analyses:polynomial_reps}
In order to understand the model better, we analyzed its task representations. 

However, we note that there are challenges to interpreting representation analyses, particularly in an architecture like ours. Some recent work \citep{Hermann2020} shows two key challenges of representation analysis. Although that work explored different types of analyses in simpler models, the findings may also apply to this work. First, the representations of a model that performs multiple tasks may be biased towards the simpler tasks, because of the learning dynamics. This may relate to some phenomena we observe below, such as the first principal components of the task representations being driven in large part by the polynomial constant terms, since constant polynomials are the simplest tasks. Second, when the task is non-linear, linear representation analyses can be misleading. The task representations in our model are related to the behavior in a highly non-linear way. Thus, it is not necessary for the representations to be linearly organized for the model to generalize well --- indeed, we show below that the model representations may be organized in a more polar structure. Furthermore, our model's mapping of task representations to tasks may be many-to-one; just as we can write either \((x+1)^2\) or \(x^2 + 2x + 1\) to denote the same function, the model may be able to represent the same task with multiple distinct representations. These issues make finding certainty in the meaning of the model's representations difficult.  

Nevertheless, the representations do show interesting structure that gives some intuitions for how the model may be performing the tasks. This structure is also relatively consistent across runs, suggesting that the underlying dynamics driving the emergence of these representations are fundamental to the interaction of the task space and the architecture --- this merits future investigation. We first examine how the representations of the polynomials are organized, then provide some further details on how they transform under additional meta-mappings, and finally show some relationships between the representations of meta-mappings and basic polynomials.

\begin{figure}[H]
\centering
\includegraphics[width=\textwidth]{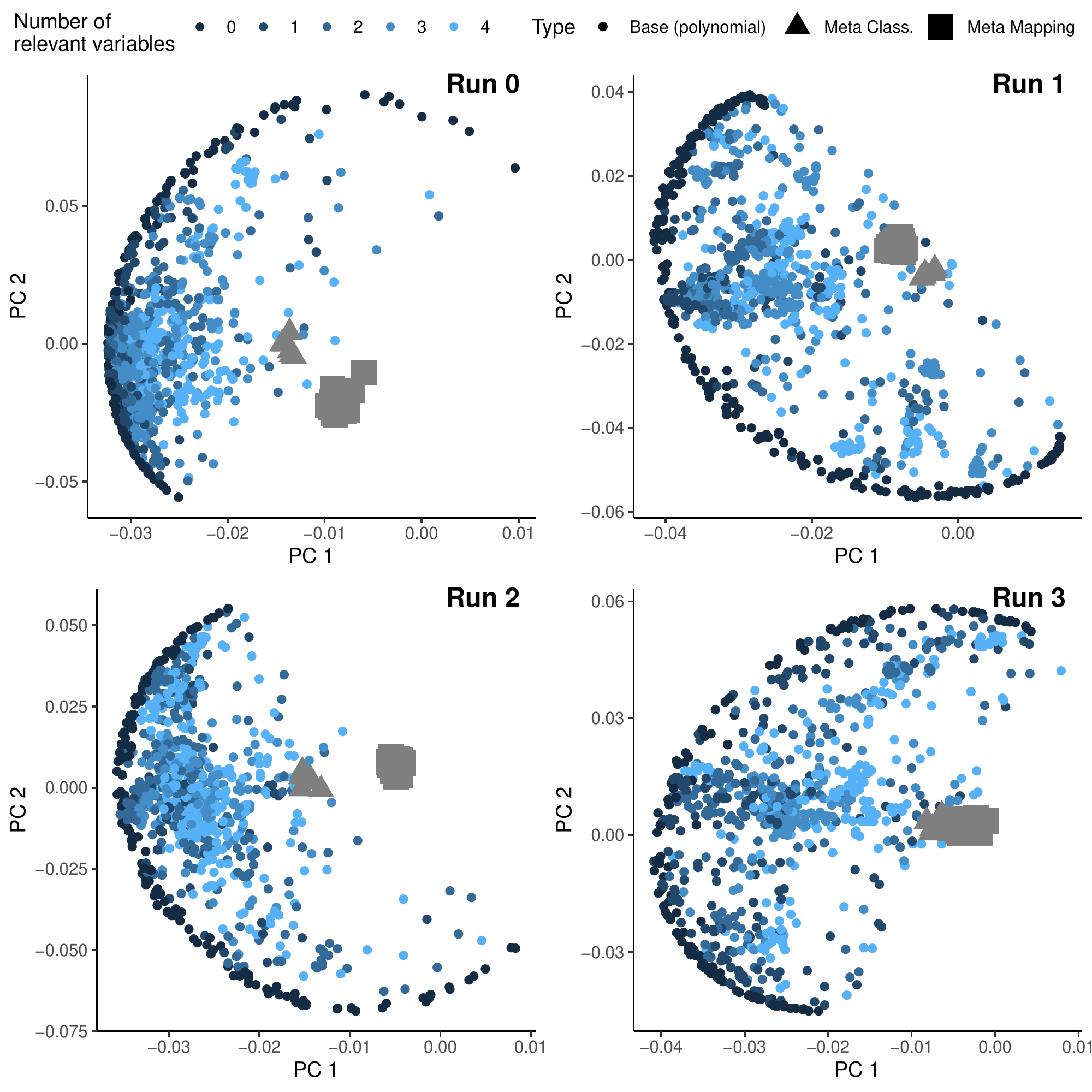}
\caption[Principal components of task and meta-mapping representations of our model after training on the polynomials domain.]{Principal components of task and meta-mapping representations of our model after training on the polynomials domain. The representation space is organized relatively consistently across runs, with constant polynomials pushed to the outside, and meta-mappings and meta-classifications more centrally located.} \label{supp_fig:HoMM_polynomials:reps_overall_PCA}
\end{figure}

\textbf{PCA:} First, we performed principal components analysis on the task and meta-mapping representations in the model after training (Fig. \ref{supp_fig:HoMM_polynomials:reps_overall_PCA}). This analysis reveals strikingly similar organization of the representation space across different training runs, with constant polynomials pushed to the outside in a semi-circle, and more complex polynomials stretching toward the center, where meta-mappings and meta-classifications are located. This may be due to the learning dynamics --- the distance of the task representations from the center appears to be roughly inversely proportional to the complexity of the task, which might imply that the constant polynomials have the largest-magnitude representations because they are easiest to learn, and so their representations receive more consistent updates starting from earlier in the learning process.

\begin{figure}[ptbh]
\centering
\includegraphics[width=\textwidth]{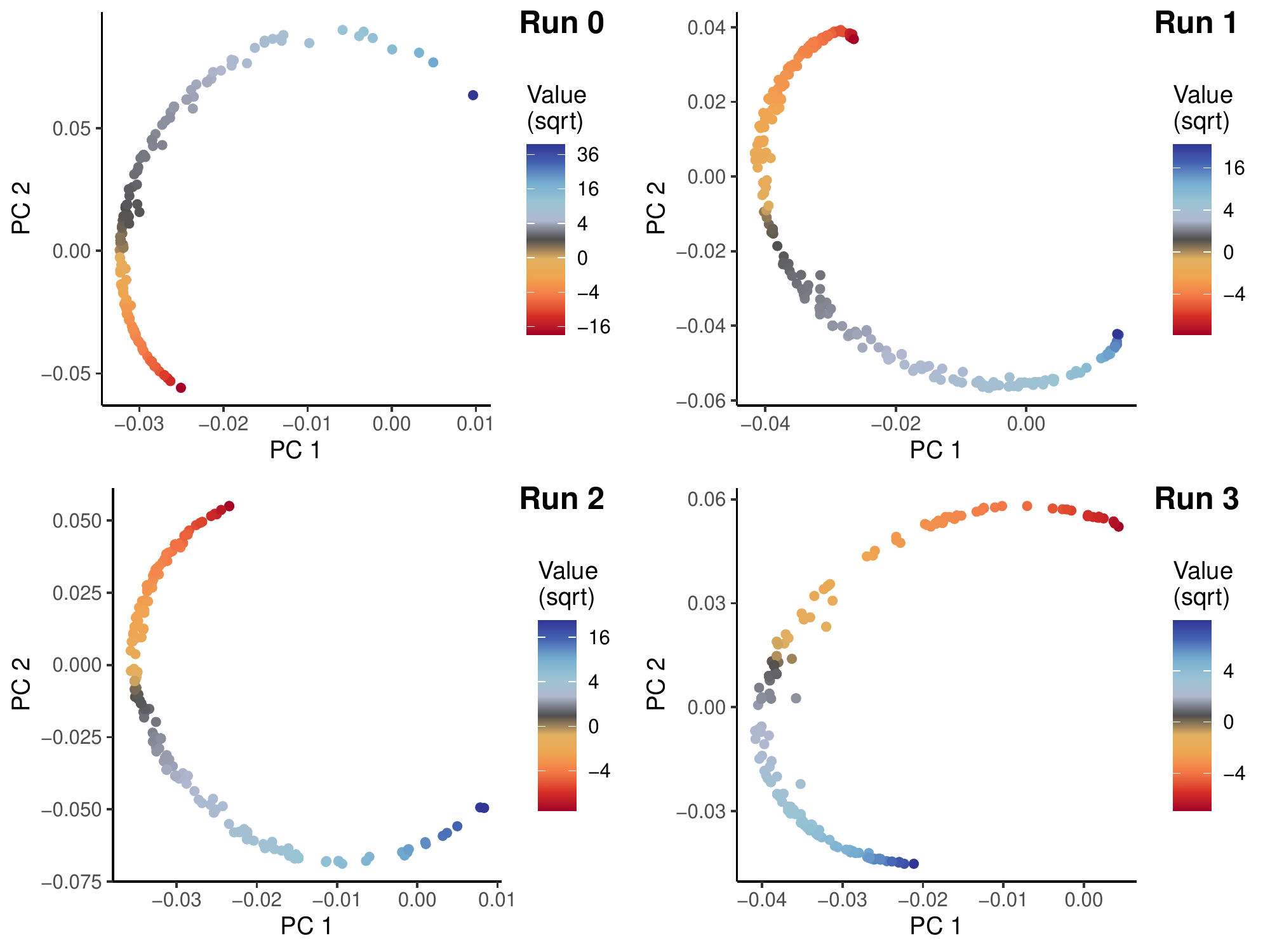}
\caption[Principal components of constant polynomial representations, showing systematic organization by value.]{Principal components of constant polynomial representations, showing systematic organization by value. Intriguingingly, this relationship appears to be systematically non-linear across runs. (PCs computed across all task representations, color scale of values is compressed with a square-root transformation.)} \label{supp_fig:HoMM_polynomials:reps_const_poly_PCA}
\end{figure}

To analyze this further, in Fig. \ref{supp_fig:HoMM_polynomials:reps_const_poly_PCA} we plot the representations for only the constant polynomials, colored by their value (square-root compressed for clarity). This shows that the representations of the constant polynomials are consistently arrayed angularly from lowest to highest value.

\begin{figure}[ptbh]
\centering
\includegraphics[width=\textwidth]{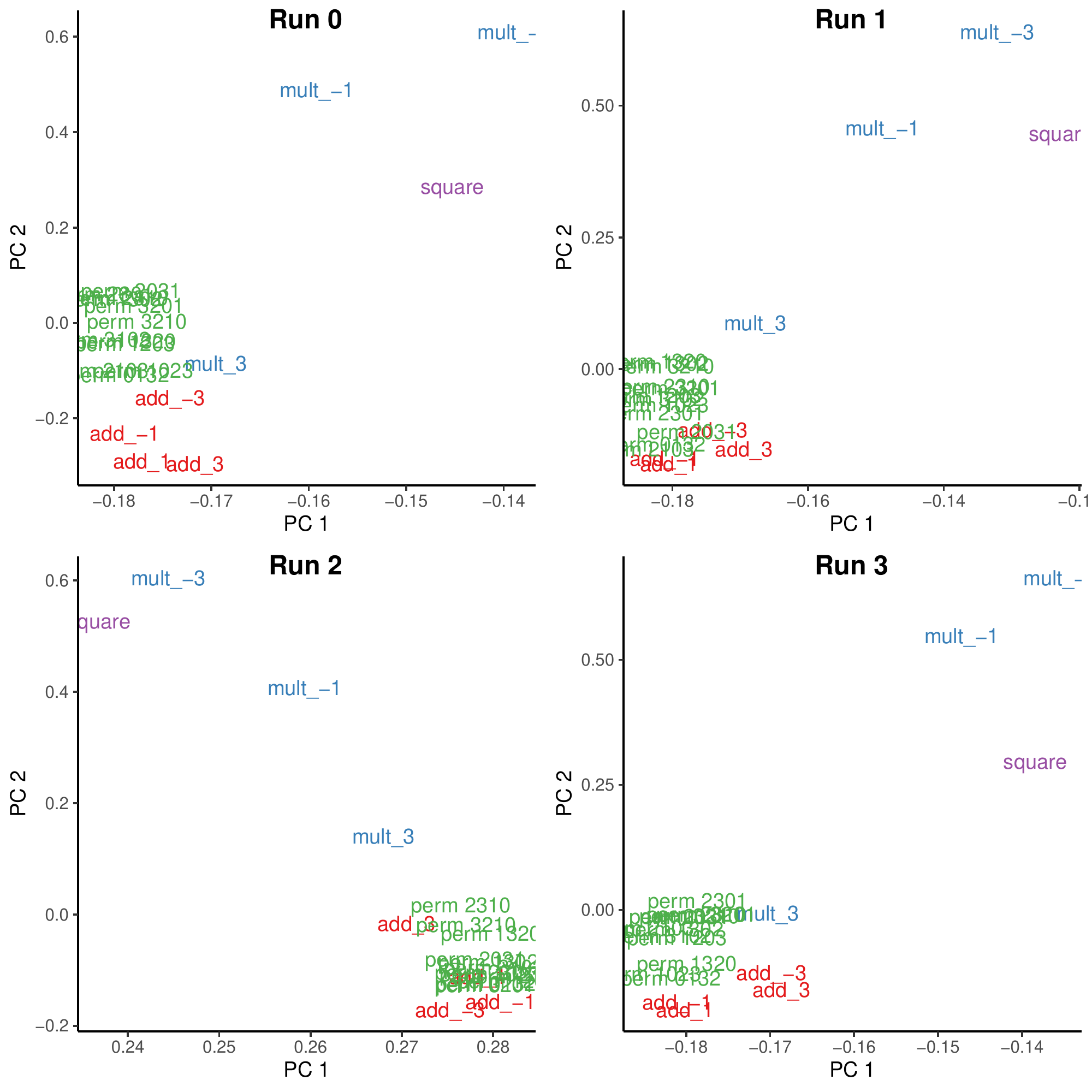}
\caption[Principal components of meta-mapping representations in the polynomial domain, showing systematic organization by type.]{Principal components of meta-mapping representations in the polynomial domain, showing systematic organization by type. Permutation mappings cluster tightly, as do addition, while multiplication and squaring are more dispersed. The addition and multiplication mappings are partially organized by absolute value.} \label{supp_fig:HoMM_polynomials:reps_meta_mapping_PCA}
\end{figure}

Finally, we examined the meta-mapping representations more closely (Fig. \ref{supp_fig:HoMM_polynomials:reps_meta_mapping_PCA}). This analysis shows that the mappings have a consistent organization across runs, with permutations and addition grouping tightly, but multiplication and squaring, which more drastically alter the polynomials, more dispersed. In particular, multiplying by negative numbers and squaring, which can change polynomials signs and therefore cause a more drastic adaptation, are more separated from the remaining meta-mappings. It is also interesting to note that the addition meta-mappings appear to be organized more by absolute value than sign in at least some runs. There is some interesting structure in higher principal components as well, for example the addition mappings appear to be organized linearly by absolute value in principal components 3 and 4. The organization of the permutation mappings is more chaotic --- while mappings that have similar representations appear more likely to differ by only a transposition, because the relationships among the permutations have a much higher-dimensional group structure, they do not project cleanly into two-dimensional plots.

\begin{figure}[H]
\begin{subfigure}[b]{0.458\textwidth}
\includegraphics[width=\textwidth]{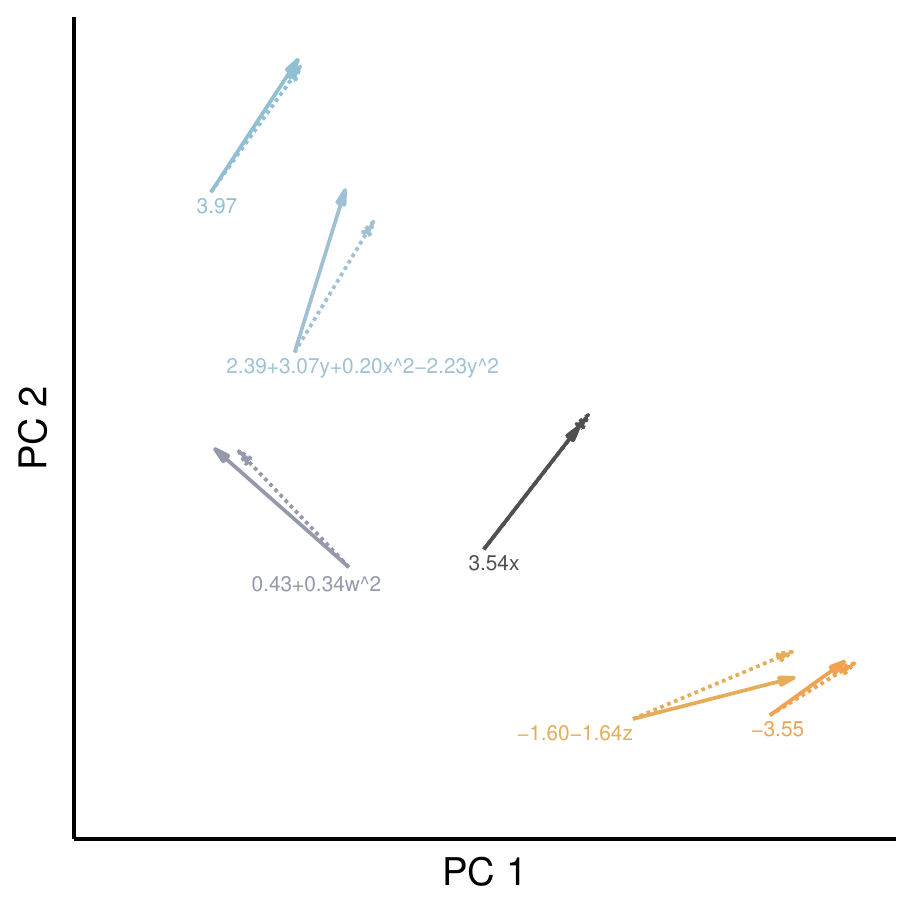}
\caption{Meta-mapping: multiply by 3.}\label{supp_fig:HoMM_polynomial_transforms_insets_detail:mult}
\end{subfigure}%
\begin{subfigure}[b]{0.541\textwidth}
\includegraphics[width=\textwidth]{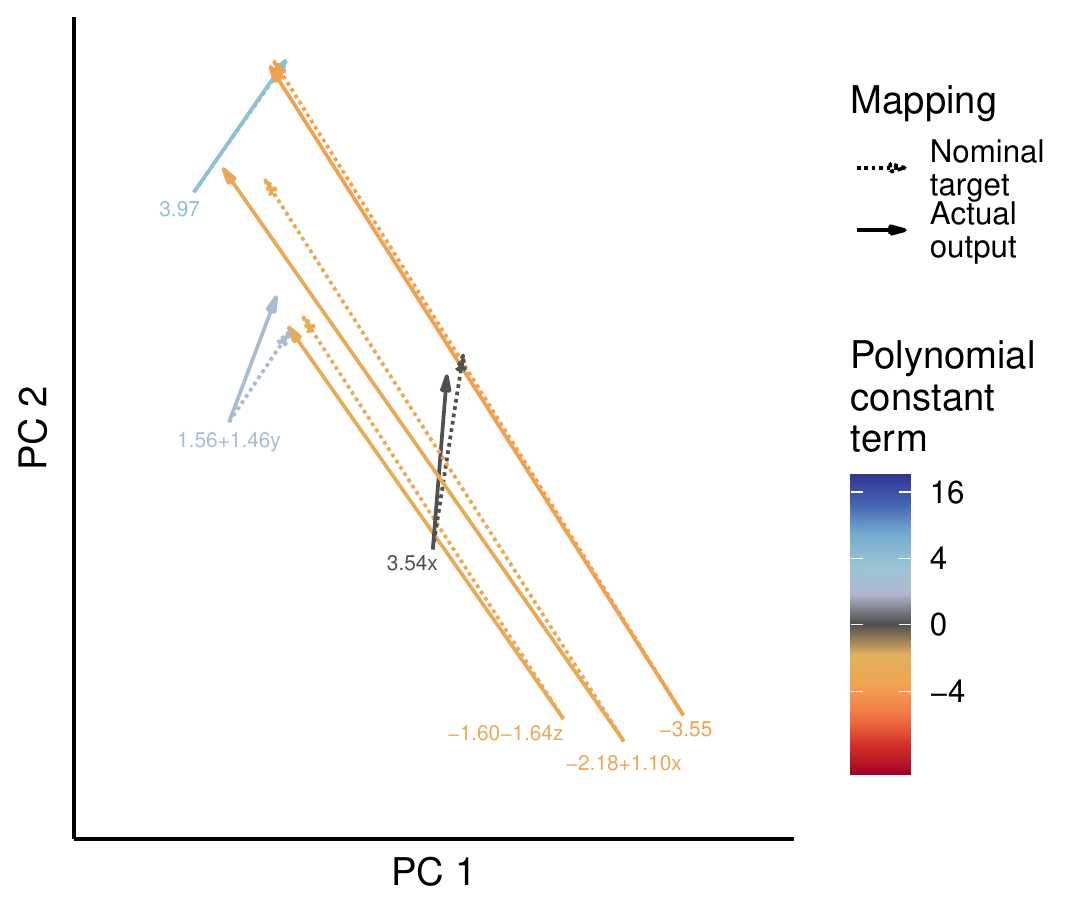}
\caption{Meta-mapping: square.}\label{supp_fig:HoMM_polynomial_transforms_insets_detail:square}
\end{subfigure}
\caption{The match between the meta-mapping outputs and the nominal targets (higher-resolution versions of the inset figures from Fig. \ref{fig:HoMM_polynomial_transforms}). (\subref{supp_fig:HoMM_polynomial_transforms_insets_detail:mult}) The multiply by 3 meta-mapping. (\subref{supp_fig:HoMM_polynomial_transforms_insets_detail:square}) The square meta-mapping. The meta-mapping outputs are generally close to the nominal targets (and note that mismatch does not necessarily indicate a mistake, see main text).}\label{supp_fig:HoMM_polynomial_transforms_insets_detail}
\end{figure}

\begin{figure}[H]
\begin{subfigure}[b]{0.5\textwidth}
\includegraphics[width=\textwidth]{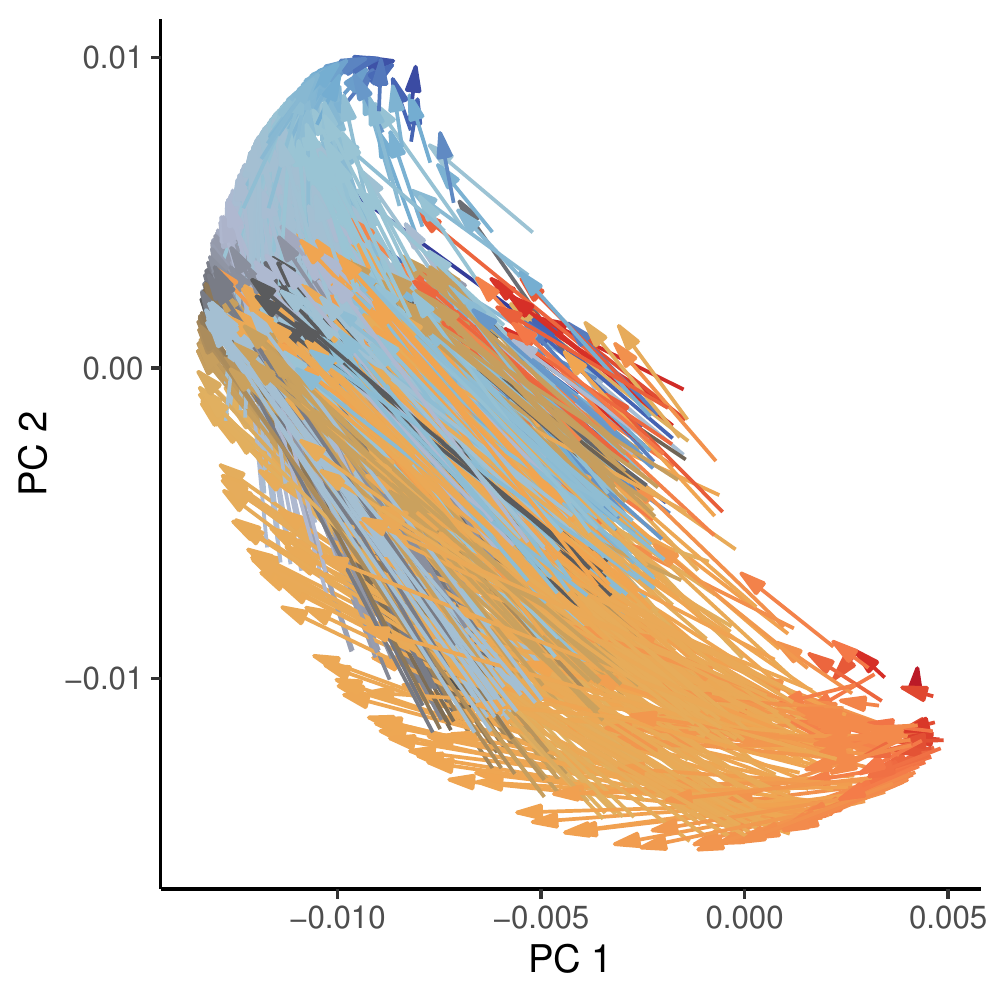}
\caption{Meta-mapping: add 3.}\label{supp_fig:HoMM_polynomial_more_transforms:add}
\end{subfigure}%
\begin{subfigure}[b]{0.5\textwidth}
\includegraphics[width=\textwidth]{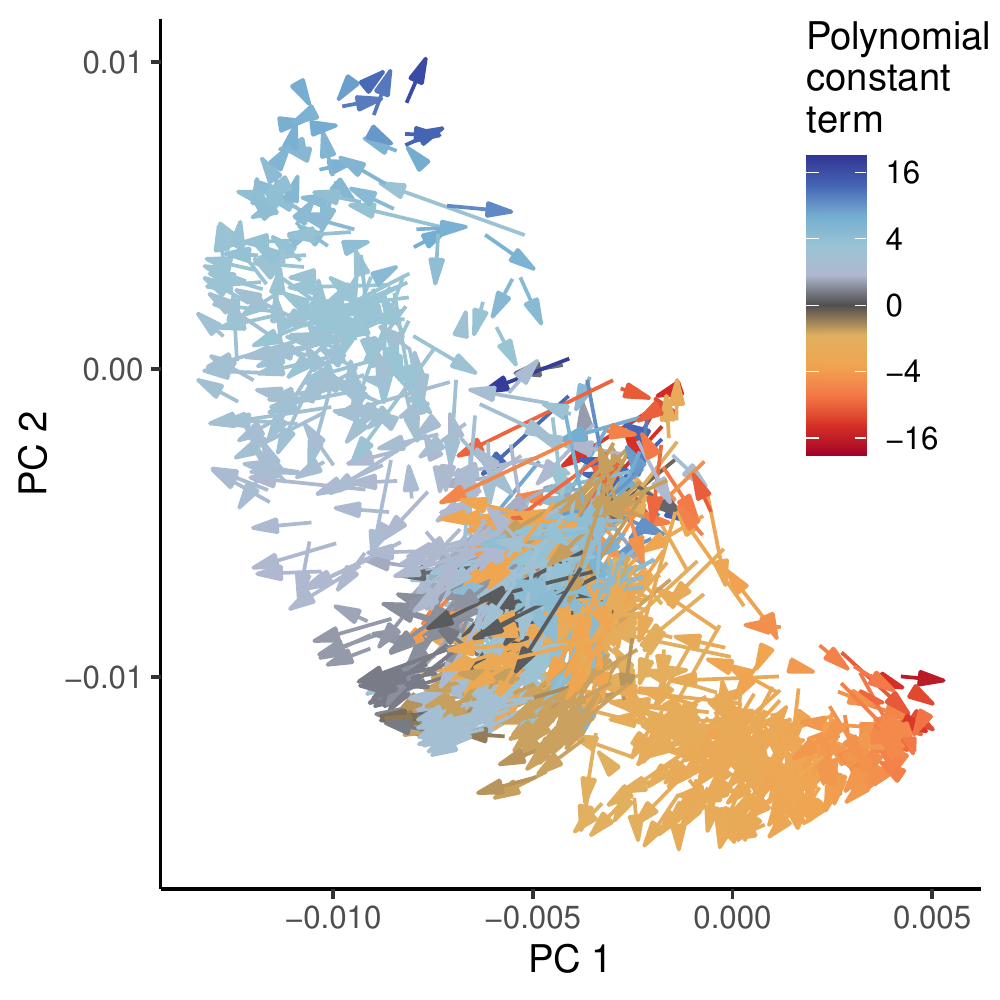}
\caption{Meta-mapping: permute (z, y, x, w).}\label{supp_fig:HoMM_polynomial_more_transforms:permute}
\end{subfigure}
\caption{Visualizing how other meta-mappings transform the polynomial model representations (compare to Fig. \ref{fig:HoMM_polynomial_transforms}). (\subref{supp_fig:HoMM_polynomial_more_transforms:add}) The add 3 meta-mapping. Adding a constant results in rotation of the polynomial representations, and a slight outward expansion (as the polynomials become relatively more dominated by their constant terms). (\subref{supp_fig:HoMM_polynomial_more_transforms:permute}) A permutation meta-mapping which affects all variables (note: only non-constant polynomials are included in this panel). The reorganization of the space under the permutation is difficult to interpret, likely because the structure of the polynomial variables is higher dimensional, and involves many more principal components.}\label{supp_fig:HoMM_polynomial_more_transforms}
\end{figure}

\textbf{How meta-mapping transforms the representations:} Next, we analyzed how meta-mapping transforms the task representations (Fig. \ref{fig:HoMM_polynomial_transforms}). We conducted these analyses (and some of the subsequent ones on homoiconicity and representations) at the suggestion of a reviewer; because of this, there were conducted on a new set of runs, as we had not retained the model parameters for the prior runs. Here, we show some more detailed results. First, in Fig. \ref{supp_fig:HoMM_polynomial_transforms_insets_detail}, we show higher-resolution versions of the inset figures from Fig. \ref{fig:HoMM_polynomial_transforms}, showing the alignment between the meta-mapping outputs and the nominal targets. Second, in Fig. \ref{supp_fig:HoMM_polynomial_more_transforms} we show the transformations induced by two additional meta-mappings, adding 3 and an input permutation. 

\textbf{Homoiconicity and overlap between different representations of different data types in the shared space:} We then explored how homoiconicity contributes to the success of the model, by analyzing the relationship between the representations of basic tasks and meta-mappings. This is motivated by an observation by a reviewer that one possible explanation for our observation (above) that homoiconic architectures yield better performance is that the result is purely due to regularization, and that the basic tasks and meta-mapping representations reside in orthogonal subspaces of the representation space. While it is difficult to completely rule out the possibility that regularization is playing a role, in this section we show at least that there is more overlap between the meta-mapping and base-task subspaces than would be expected by chance, and that at least some sensible isomorphisms between the basic tasks and meta-mappings may be shaping the representations.  

\begin{figure}[H]
\centering
\includegraphics[width=\textwidth]{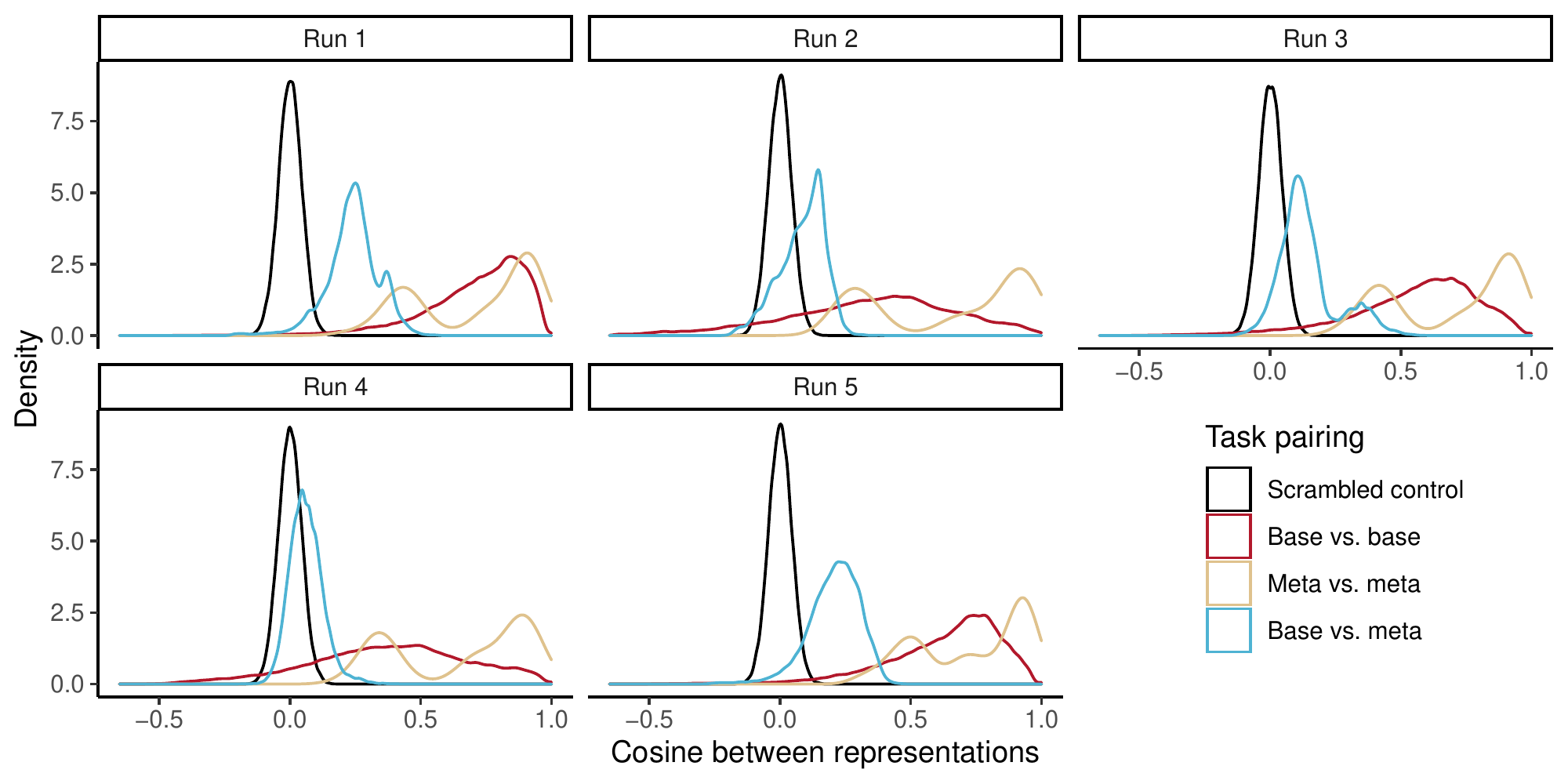}
\caption{There is non-trivial overlap between the representations of meta-mappings and base tasks in the polynomials domain. This figure plots the cosine similarity between different groups of representations, base tasks vs. base tasks. meta tasks vs. meta tasks, base tasks vs. meta tasks, and a control similarity distribution from a scrambled representation matrix. Although base tasks are more similar to other base tasks than to meta tasks, there is more similarity between the base and meta representations than would be expected by chance, though the absolute amount varies from run to run.} \label{supp_fig:HoMM_polynomials:meta_base_rep_simil}
\end{figure}

\begin{figure}[H]
\centering
\includegraphics[width=\textwidth]{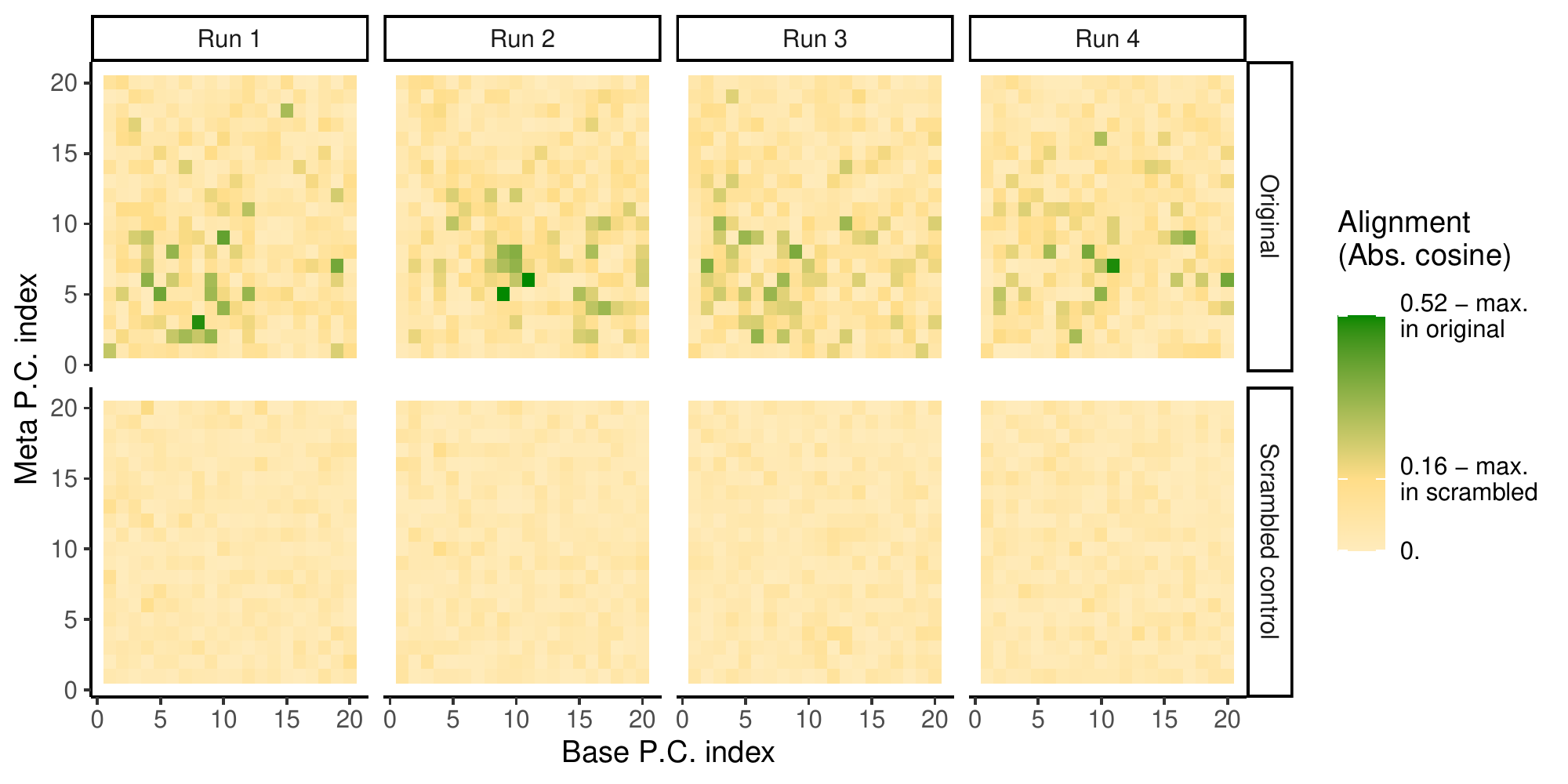}
\caption{There is nontrivial overlap between the top 20 principal components of the base task representations, and the top 20 principal components of the meta task representations, in the polynomials domain. For each run (columns), the top panel shows the alignment (abs. cosine similarity) between base task PCs (x-axis) and meta task PCs (y-axis). The bottom panels show the same results for a matched control (a scrambled representation matrix). The color scale is set so that cells are colored green only if the alignment is larger than any alignment observed in any control matrix. There are strong and relatively sparse alignments between the principal components of the basic- and meta-tasks, showing that the representations are not residing in orthogonal subspaces, and suggesting that homoiconicity is contributing non-trivially to the representation structure.} \label{supp_fig:HoMM_polynomials:meta_base_PC_alignments}
\end{figure}

\begin{figure}[H]
\centering
\includegraphics[width=\textwidth]{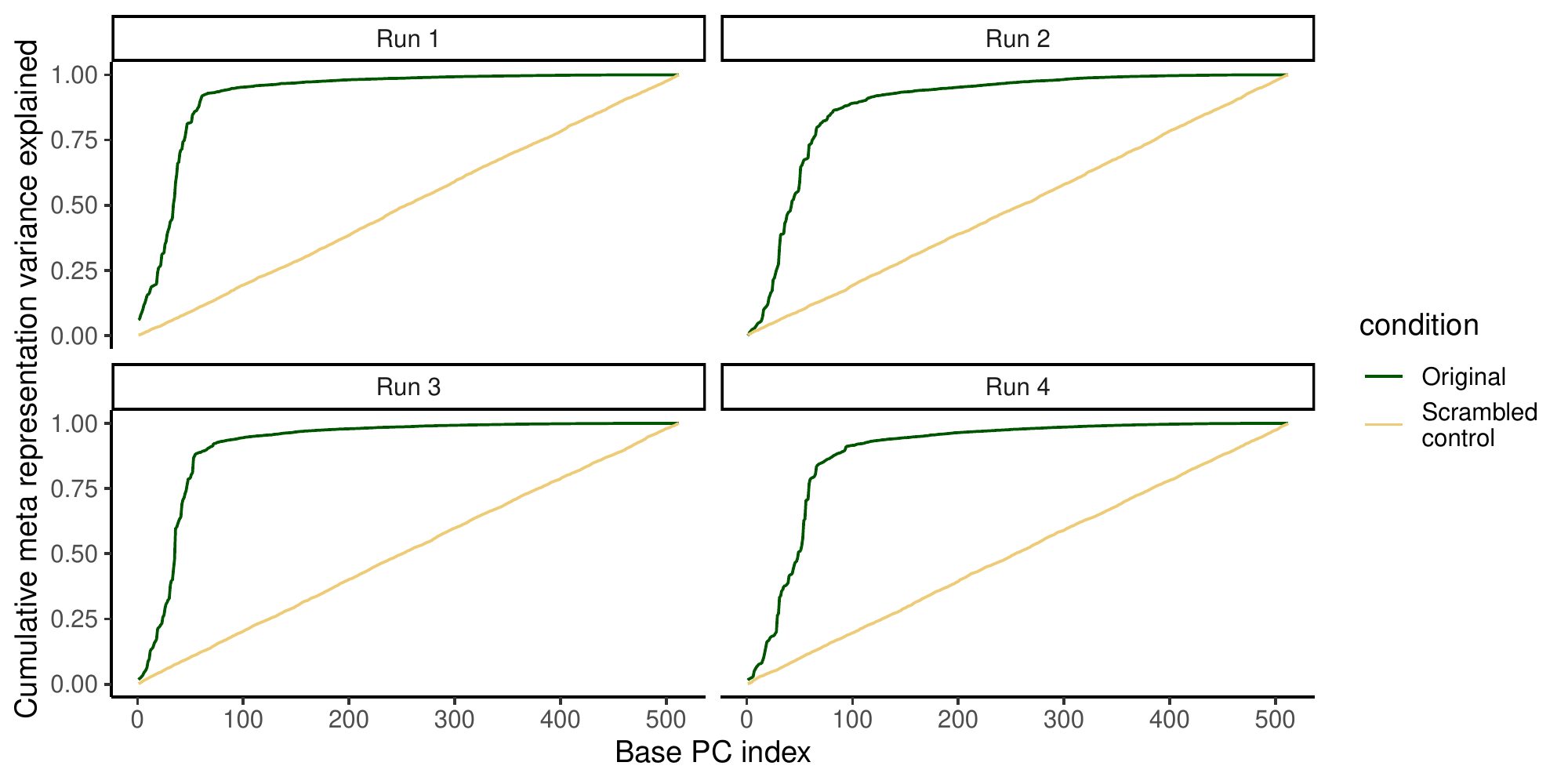}
\caption{The meta-task representation variance is preferentially distributed in the top principal components of the base task representations. For each run (panels), this plot shows the cumulative meta-task representation variance (vertical axis) explained by the base task representation principal components (horizontal axis). The dark green line shows the actual results, while the yellow line shows the results for a matched control (scrambled representation matrix). The meta-task representation variance is mostly contained in the earlier (more important) base principal components, again suggesting that homoiconicity is contributing non-trivially to the representation structure.} \label{supp_fig:HoMM_polynomials:meta_base_PC_variance}
\end{figure}

\begin{figure}[H]
\centering
\includegraphics[width=0.5\textwidth]{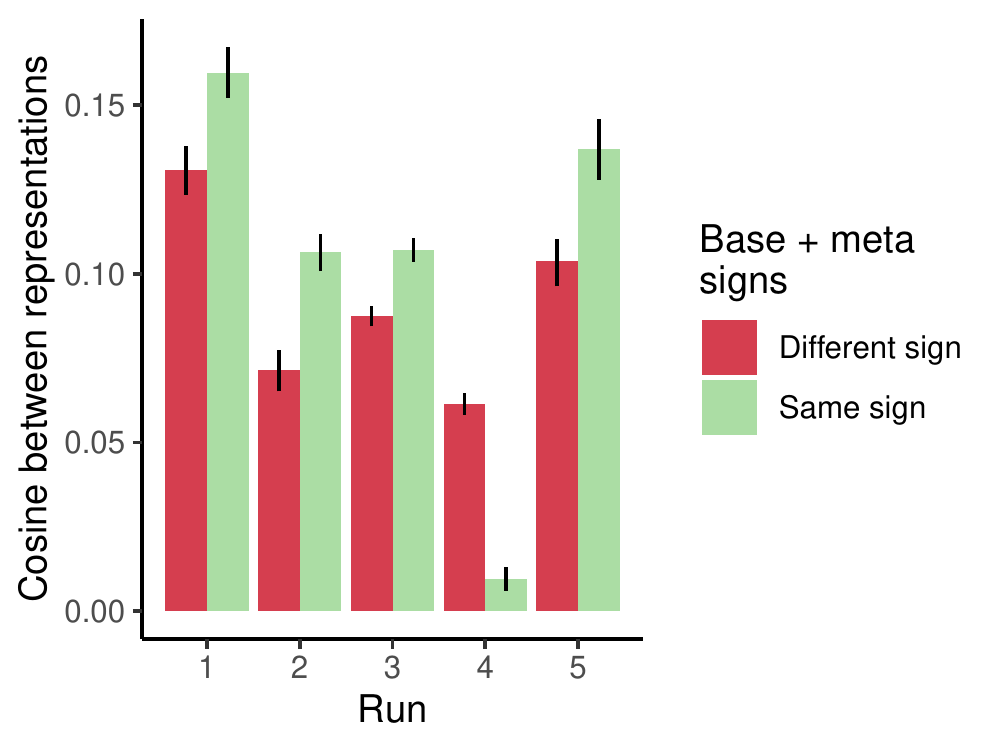}
\caption{There is significant organization of the multiplication meta-mappings by sign, in alignment (or anti-alignment) with the signs of the constant polynomials. This plot shows cosine similarity between representations of meta-mappings for the trained multiplication tasks (multiply by -3, -1,  and 3) and the constant polynomials, depending on whether the multiplication value (for the meta-mappings) and the constant value (for the  basic tasks) have the same sign or different signs. The difference is significant in each run (all \(t\text{s} > 5.3\), all \(p\text{s} < 1 \cdot 10^{-6}\)), with greater similarity when the signs are aligned in all runs except run 4, where the effect goes in the opposite direction. These results suggest that the homoiconic model may be exploiting homomorphisms between scalar values that appear in a constant polynomial, and scalar values that appear in a meta-mapping (note that the non-canonical sign-switching alignment in run 4 may nevertheless capture useful structure).} \label{supp_fig:HoMM_polynomials:meta_mult_base_const_alignment}
\end{figure}

First, in Fig. \ref{supp_fig:HoMM_polynomials:meta_base_rep_simil}, we explore the cosine similarity between base-task and meta-task representations. We observe non-trivial overlap, which we explore in greater detail in Fig. \ref{supp_fig:HoMM_polynomials:meta_base_PC_alignments}, showing that there is strong and sparse alignment between the top principal components of the polynomials and the meta-mappings, and Fig. \ref{supp_fig:HoMM_polynomials:meta_base_PC_variance}, showing that the variance of the meta-mapping representations is mostly contained within lower (more important) principal components of the base task representations. Finally, in Fig. \ref{supp_fig:HoMM_polynomials:meta_mult_base_const_alignment}, we show intriguinging patterns of alignment of the multiplication meta-mappings and constant polynomials depending on whether the signs match, which suggests that the model may be at least partly uncovering the isomorphic numerical structure between these different levels of abstraction. Exploring the alignment between base tasks and meta-mappings further will be an interesting direction for future work.

\begin{figure}[H]
\centering
\includegraphics[width=0.75\textwidth]{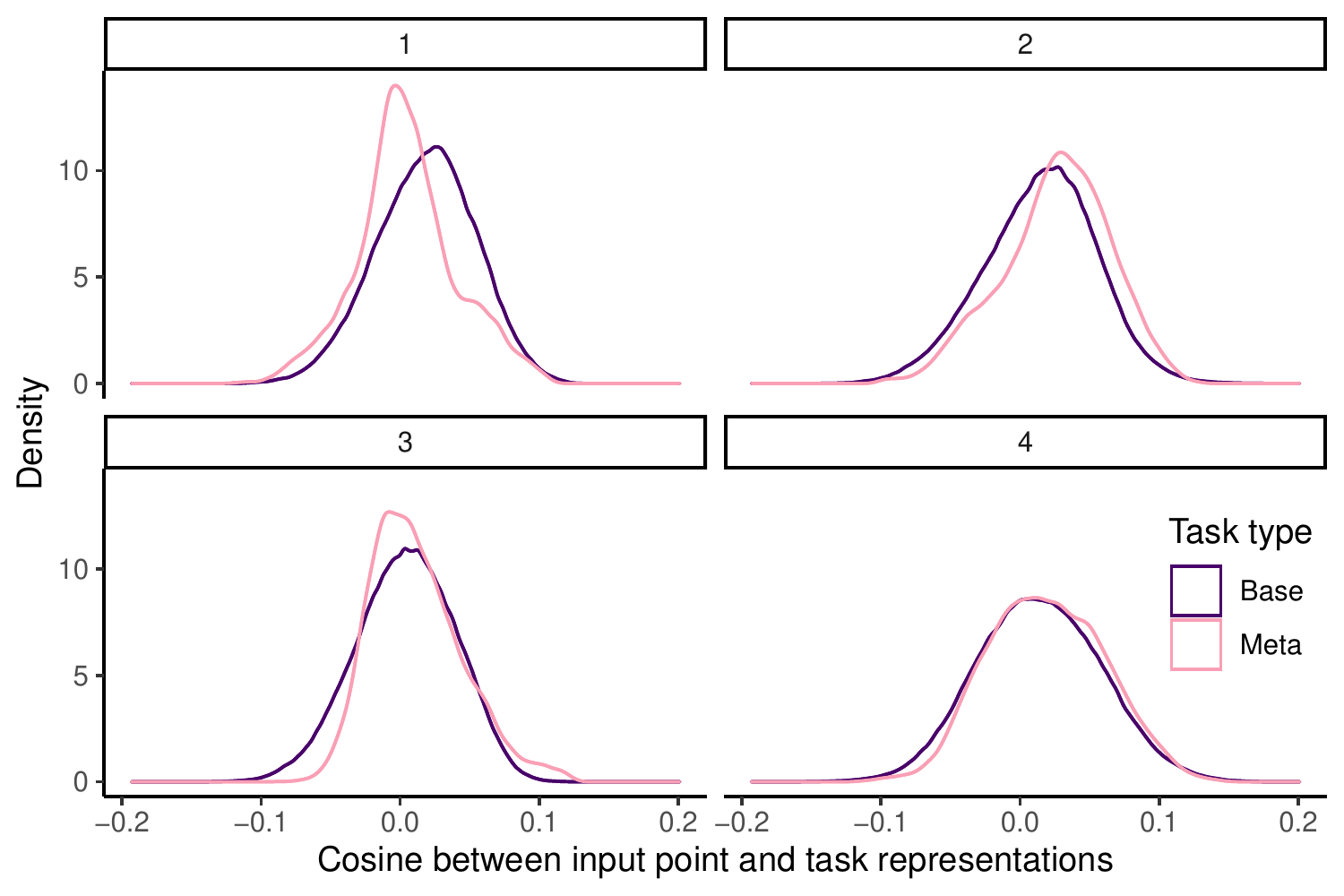}
\caption{The cosine similarity between basic input representations and basic task or meta-mapping representations is fairly small, likely reflecting the smaller amount of shared structure between these different entities, and the weaker constraints on alignment (see text for further discussion).} \label{supp_fig:HoMM_polynomials:input_task_rep_overlap}
\end{figure}

We also explored the relationship between the representations of basic data inputs to the model (that is, \((w, x, y, z)\) tuples at which to evaluate a polynomial), and the representations of tasks and meta-mappings. The magnitude of the similarities was overall quite small (see Fig. \ref{supp_fig:HoMM_polynomials:input_task_rep_overlap}), suggesting that, unlike in the case of meta-mappings and base tasks, the model is not substantially exploiting relationships between tasks and data points. This result is not particularly surprising for several reasons. First, there is more structure in common between basic tasks and meta-mappings than between either category and data points, because both basic tasks and meta-mappings are functions. Second, there are more constraints that encourage basic task and meta-mapping representations to be similar in the homoiconic architecture --- both are output by the same example network, and both are processed by the same hyper network. By contrast, data points and basic tasks only have a one-sided constraint, viz. that they are both processed by the same example network.   
\newpage

\FloatBarrier

\FloatBarrier
%\newpage
\subsubsection{Cards}\label{supp_sec:analyses:cards}

\begin{figure}[H]
\centering
\begin{subfigure}[t]{0.5\textwidth}
\includegraphics[width=\textwidth]{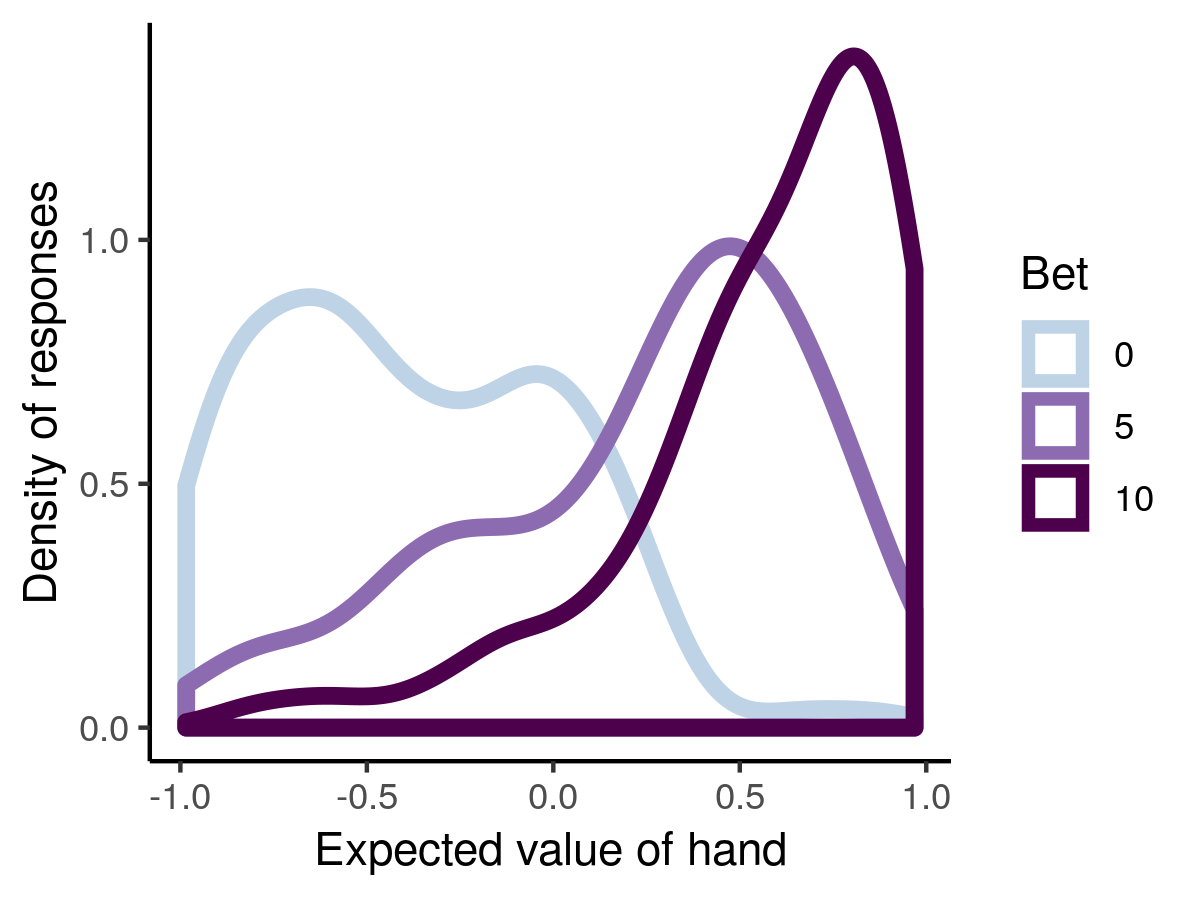}
\caption{Basic game: Bet density by expected value.}
\end{subfigure}%
\begin{subfigure}[t]{0.5\textwidth}
\includegraphics[width=\textwidth]{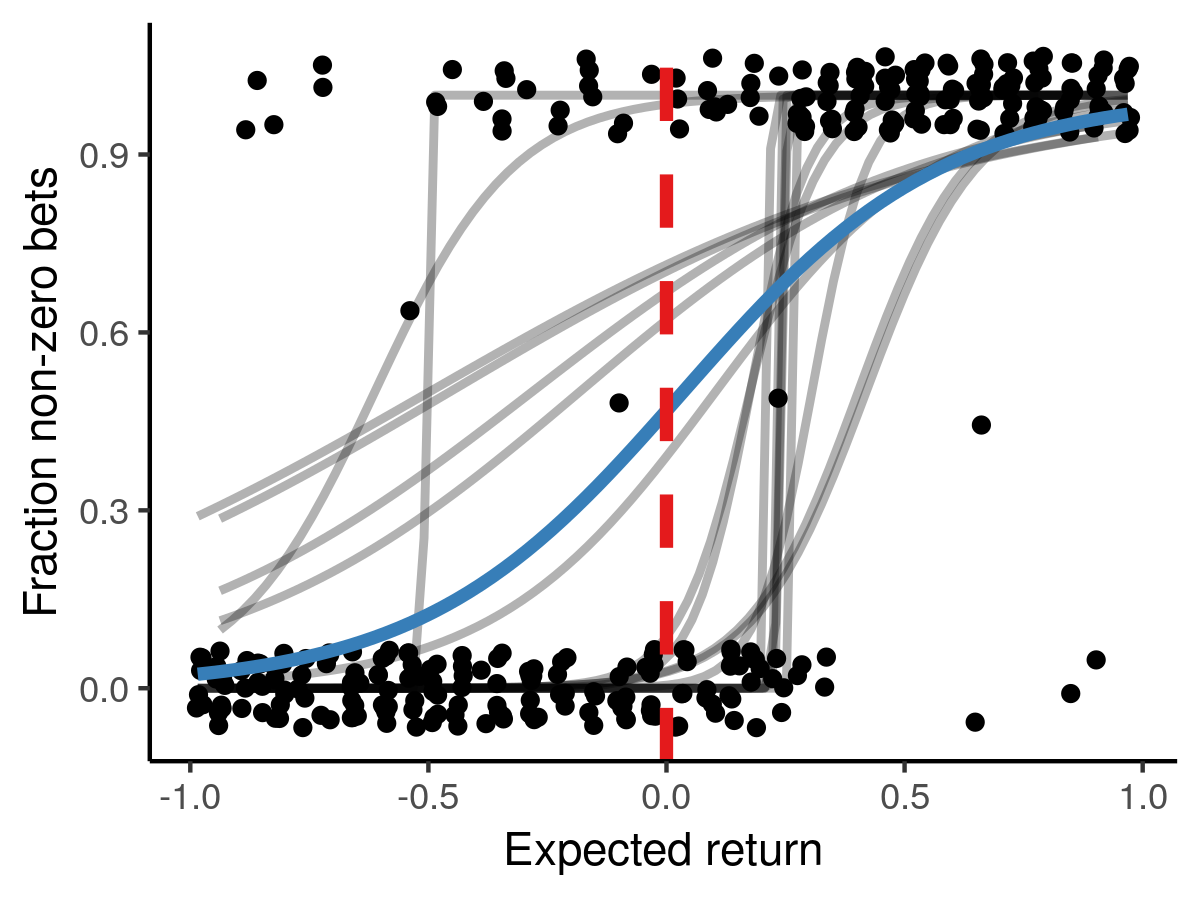}
\caption{Basic game: Probability of non-zero bet by expected value. The red dashed line is the optimal threshold, the grey curves are the individual subject fits.}
\end{subfigure}\\
\begin{subfigure}[t]{0.5\textwidth}
\includegraphics[width=\textwidth]{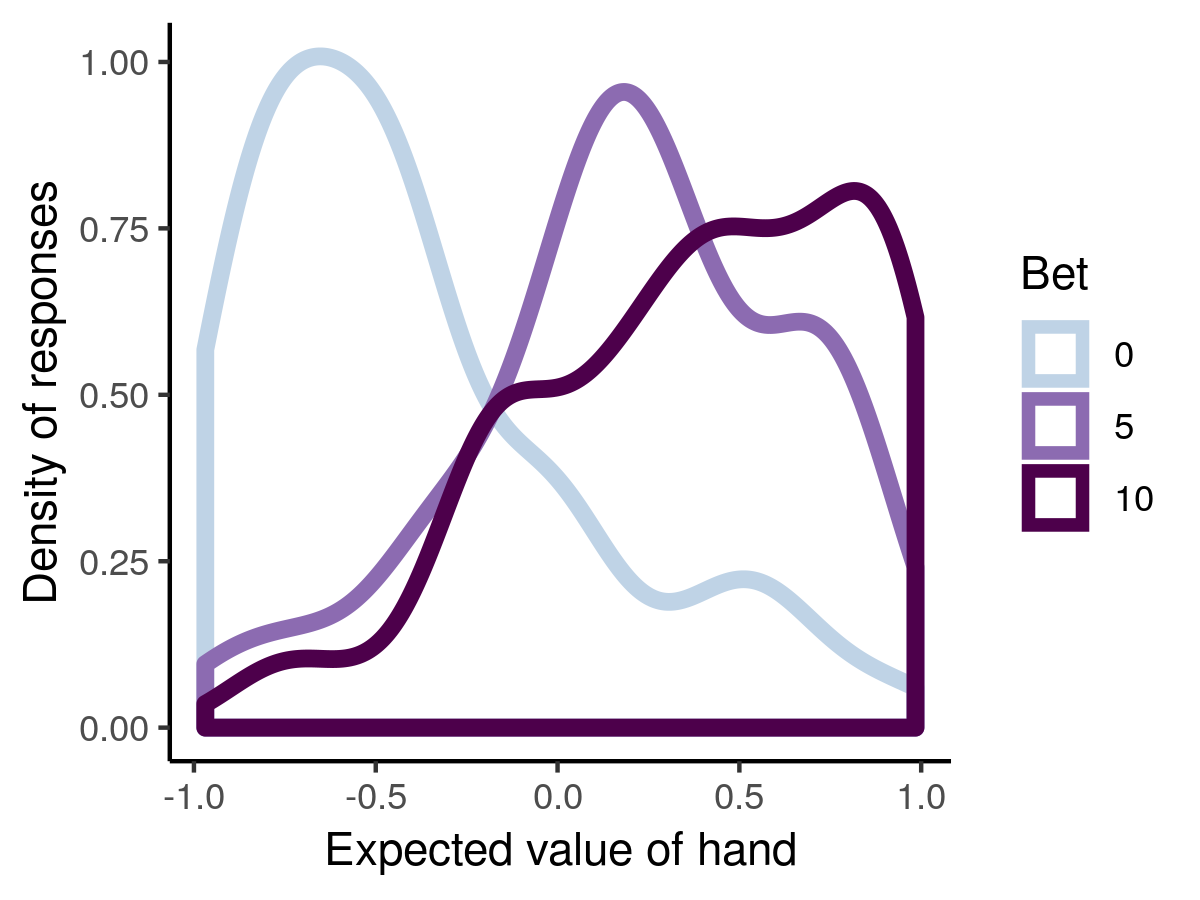}
\caption{Losing variation: Bet density by expected value.}
\end{subfigure}%
\begin{subfigure}[t]{0.5\textwidth}
\includegraphics[width=\textwidth]{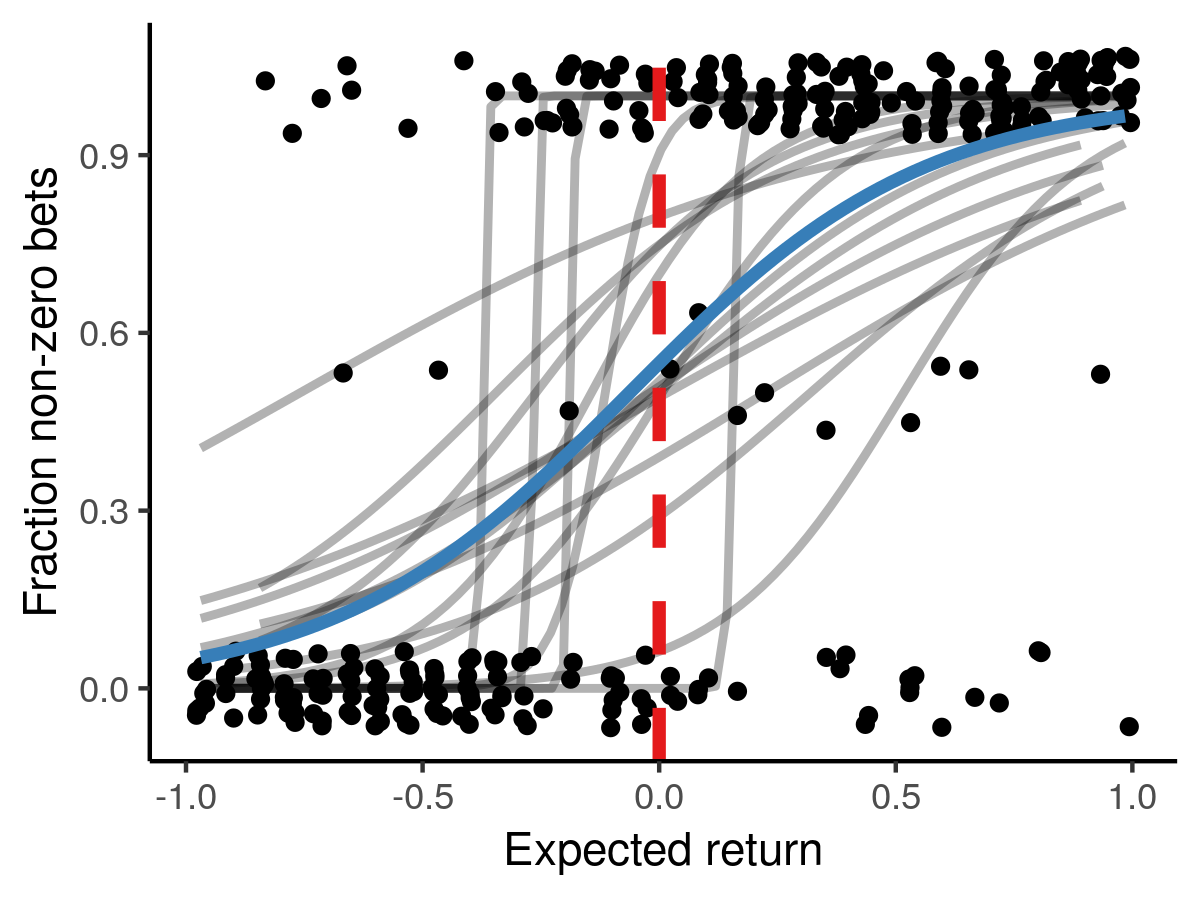}
\caption{Losing variation: Probability of non-zero bet by expected value. The red dashed line is the optimal threshold, the grey curves are the individual subject fits.}
\end{subfigure}%
\caption{Human performance on the card game task. Top row is basic game evaluation (before being told to lose), bottom is after being told to lose. While participants are performing well above chance, they are far from optimal. They make intermediate value bets, and do not switch optimally between betting and not betting depending on hand value. There is also substantial inter-subject variability.} \label{supp_fig:human_cards_detailed_results}
\end{figure}

\textbf{Further analyses of human performance:} In Fig. \ref{supp_fig:human_cards_detailed_results} we show details of human participants performance on the card game tasks, including bet densities and subject-level fits of betting probability by hand value. As noted in the main text, the human subjects are performing far from optimally even in the trained task, and these figures show details on why this is true: subjects are both sub-optimal in finding the threshold at which to switch from betting to not betting, and are betting intermediate values, which an optimal better would not. 

\begin{figure}[H]
\centering
\includegraphics[width=0.5\textwidth]{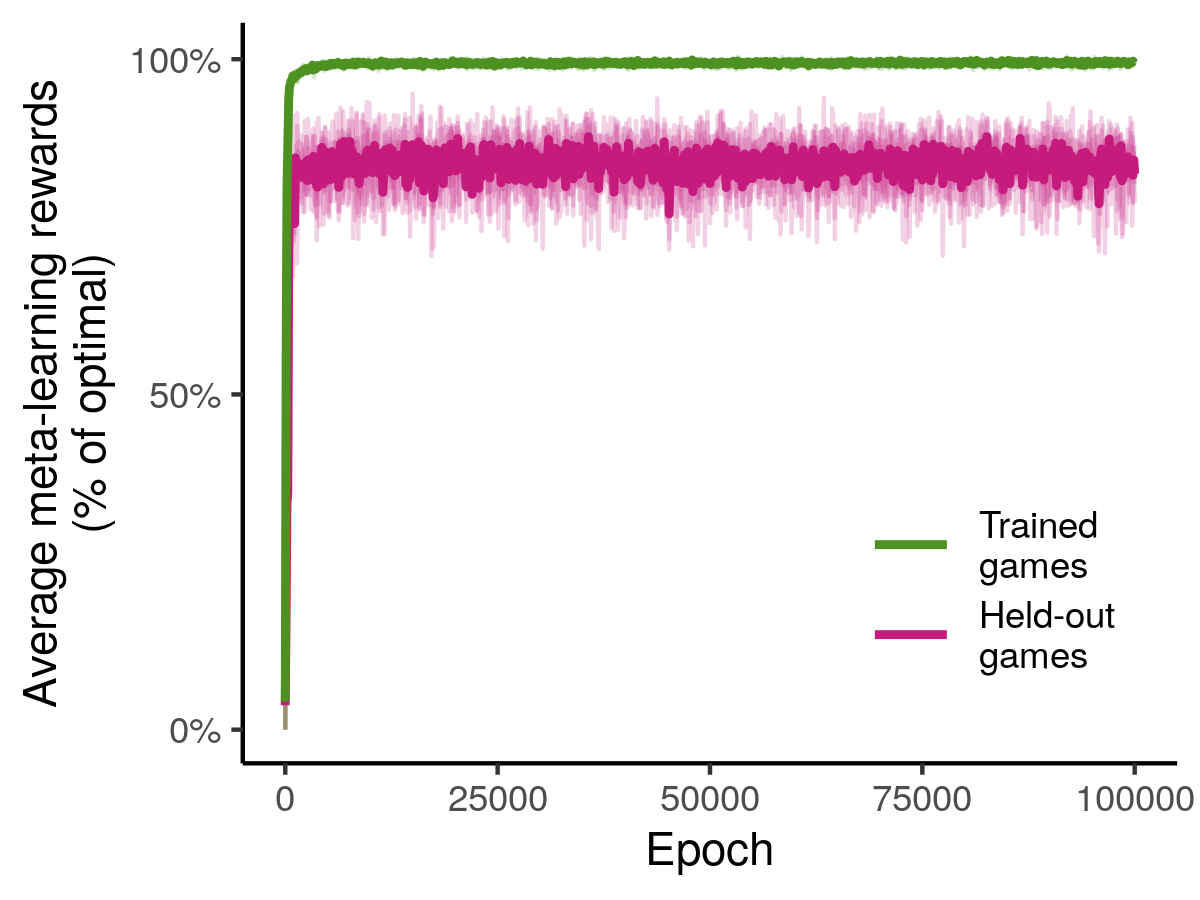}
\caption[Basic meta-learning performance in the cards domain over learning.]{Basic meta-learning performance in the cards domain over learning. The system is generalizing at the meta-learning level. That is, this graph shows that, after the example network receives a set of (hand, bet, reward) example tuples from a game, it is generating a sufficiently good representation of that game to play held-out hands. This is true both for gamess it was trained with (green), and for games that are held-out and never encountered during training (pink). (Thick dark curves are averages over 5 runs, shown as light curves.)} \label{supp_fig:HoMM:cards_basic_meta_learning}
\end{figure}

\textbf{Basic meta-learning:} In Fig. \ref{supp_fig:HoMM:cards_basic_meta_learning}, we show that the basic meta-learning is working well in the cards domain. That is, we show that after the example network is presented with a set of example (hand, bet, reward) tuples, the system is generalizing well to other hands of that game. At the end of training, the mean reward on trained games is 99.20\% of optimal (bootstrap 95\%-CI [98.90, 99.40]), and for held-out games it is 83.82\% (bootstrap 95\%-CI [80.50, 86.00]). 

\textbf{Architectural comparisons:} In Figure \ref{supp_fig:HoMM_nonhomoiconic_lesion:cards} we show that non-homoiconic architectures may perform slightly worse in the cards domain, but the difference is not significant. Specifically, the homiconic model is achieving an average expected reward of 85.38\% (bootstrap 95\%-CI [79.49, 90.32]), while the non-homoiconic model is achieving an average expected reward of 79.49\% (bootstrap 95\%-CI [69.50, 87.34]). 

\textbf{Meta-classification task lesion:} In Figure \ref{supp_fig:HoMM_metaclass_lesion:cards} we show that meta-classification may be slightly beneficial in the cards domain, but the difference is small. Specifically, the model is achieving an average expected reward of 85.38\% (bootstrap 95\%-CI [79.49, 90.32]), while without meta-classification it is achieving an average expected reward of 78.68\% (bootstrap 95\%-CI [71.01, 85.97]). Because the meta-classifications appear to be more useful in this domain than in the polynomials domain, it is possible that they are particularly useful for understanding the structure of the task distribution when there are fewer basic training tasks. However, further work would be needed to verify this.
\newpage

%\newpage
\FloatBarrier
%\clearpage

\subsubsection{Visual concepts}\label{supp_sec:analyses:concepts}
\begin{figure}[htb]
\centering
\includegraphics[width=0.625\textwidth]{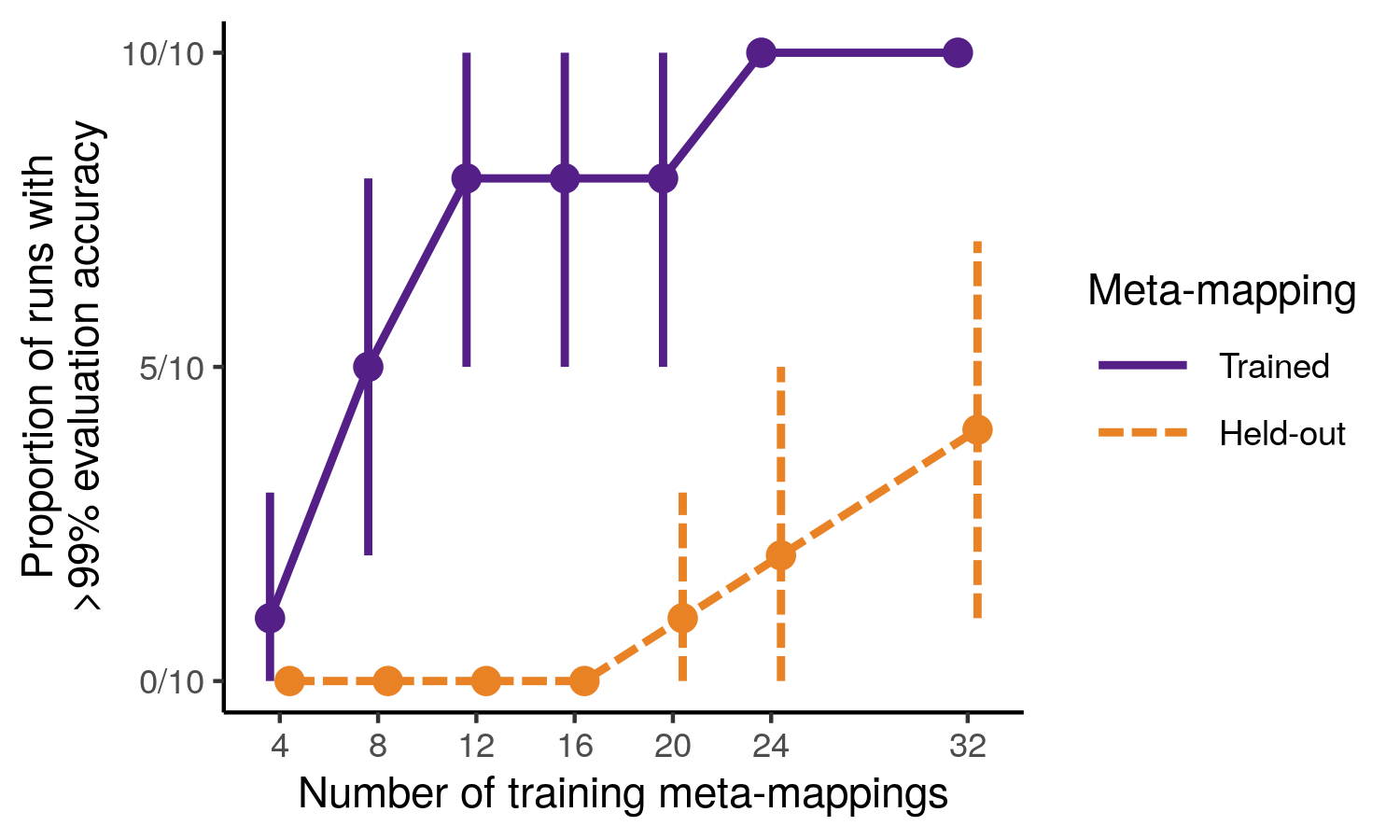}
\caption{In the visual concepts domain, the proportion of runs in which the model attained \(> 99\)\% accuracy across all transformed concepts. The model shows extremely systematic generalization on trained meta-mappings at moderate sample sizes. At the largest sample sizes we considered, the HoMM model is able to adapt near-perfectly to new meta-mappings on many runs. Note that even at this largest sample size, the system is generalizing from only 32 trained meta-mappings.}\label{supp_fig:HoMM_concepts_perfect}
\end{figure}

\begin{figure}[p]
\centering
\begin{subfigure}{\textwidth}
\centering
\includegraphics[width=0.88\textwidth]{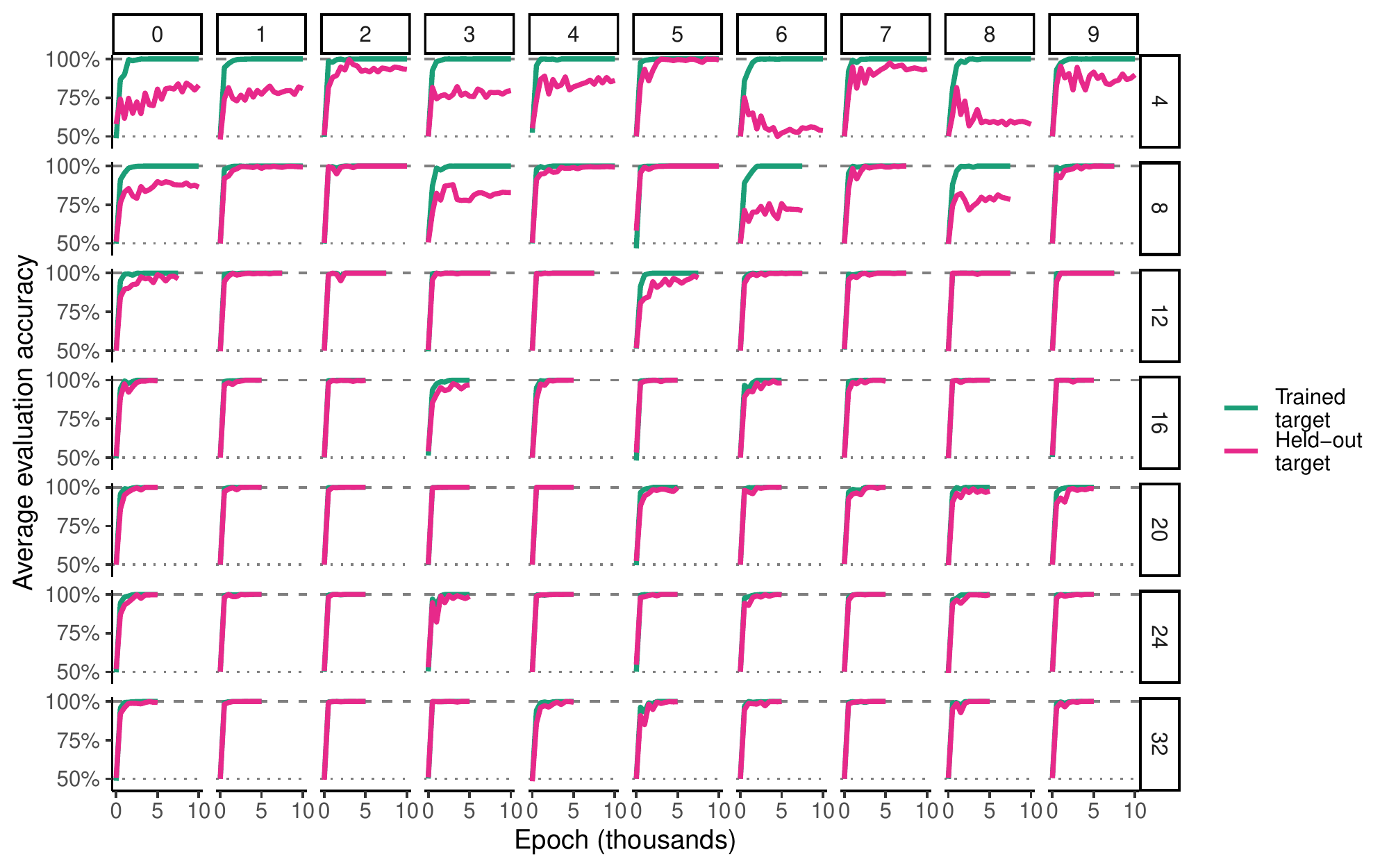}
\caption{Trained meta-mappings.}\label{supp_fig:HoMM_concepts_all_runs:train}
\end{subfigure}\\
\begin{subfigure}{\textwidth}
\centering
\includegraphics[width=0.88\textwidth]{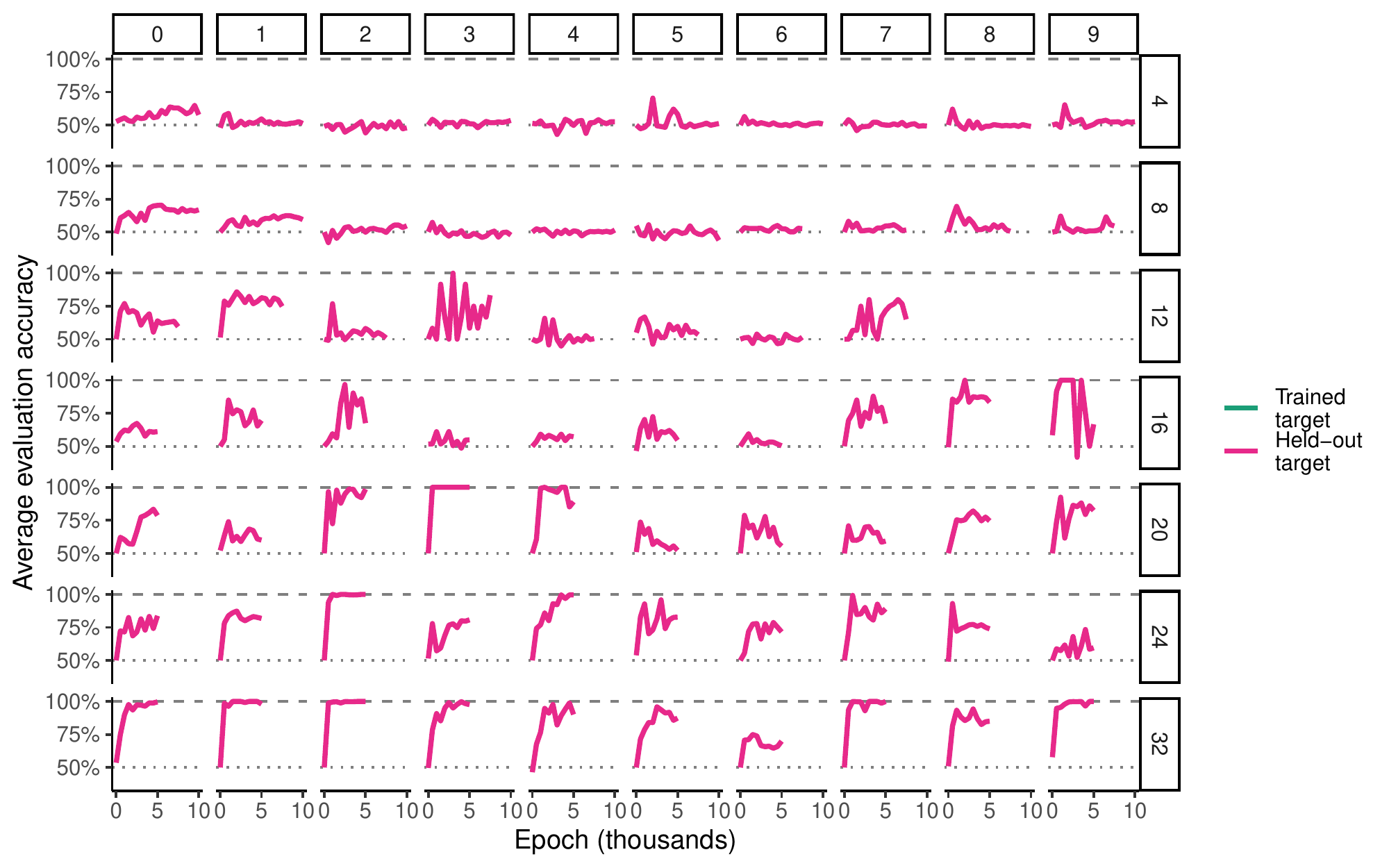}
\caption{Held-out meta-mappings.}\label{supp_fig:HoMM_concepts_all_runs:eval}
\end{subfigure}
\caption{Meta-mapping performance (evaluated as average accuracy on the transformed task) in the visual concepts domain broken down by number of training meta-mappings (rows), and by run (columns). The green lines are performance when the transformed task was encountered during training, the pink lines are performance on transformed tasks that were never encountered during training. Panel (\subref{supp_fig:HoMM_concepts_all_runs:train}) shows the results for trained meta-mappings, and panel (\subref{supp_fig:HoMM_concepts_all_runs:eval}) shows the results for held-out meta-mappings. With more training meta-mappings, generalization is better both when applying the trained meta-mappings to held-out examples (\subref{supp_fig:HoMM_concepts_all_runs:train}), and when applying held-out meta-mappings (\subref{supp_fig:HoMM_concepts_all_runs:eval}). However, even with smaller sample sizes, the model is achieving perfect generalization on the trained meta-mappings on many runs. (The dotted line denotes chance performance, the dashed line optimal.)} \label{supp_fig:HoMM_concepts_all_runs}
\end{figure}

In Fig. \ref{supp_fig:HoMM_concepts_perfect} we show the proportion of runs in which the model achieved \(> 99\%\) performance; systematic generalization is increasingly likely as the number of training meta-mappings increases. In Fig. \ref{supp_fig:HoMM_concepts_all_runs} we show learning curves for all runs of the meta-mapping model on these tasks. 

\FloatBarrier
\subsubsection{RL}\label{supp_sec:analyses:RL}

In Fig. \ref{supp_fig:HoMM_arch_cond_vs_hyper:RL} we also show that the HyperNetwork-based architecture performs better in this domain.

\begin{figure}[!hb]
\centering
\begin{subfigure}{0.5\textwidth}
\centering
\includegraphics[width=\textwidth]{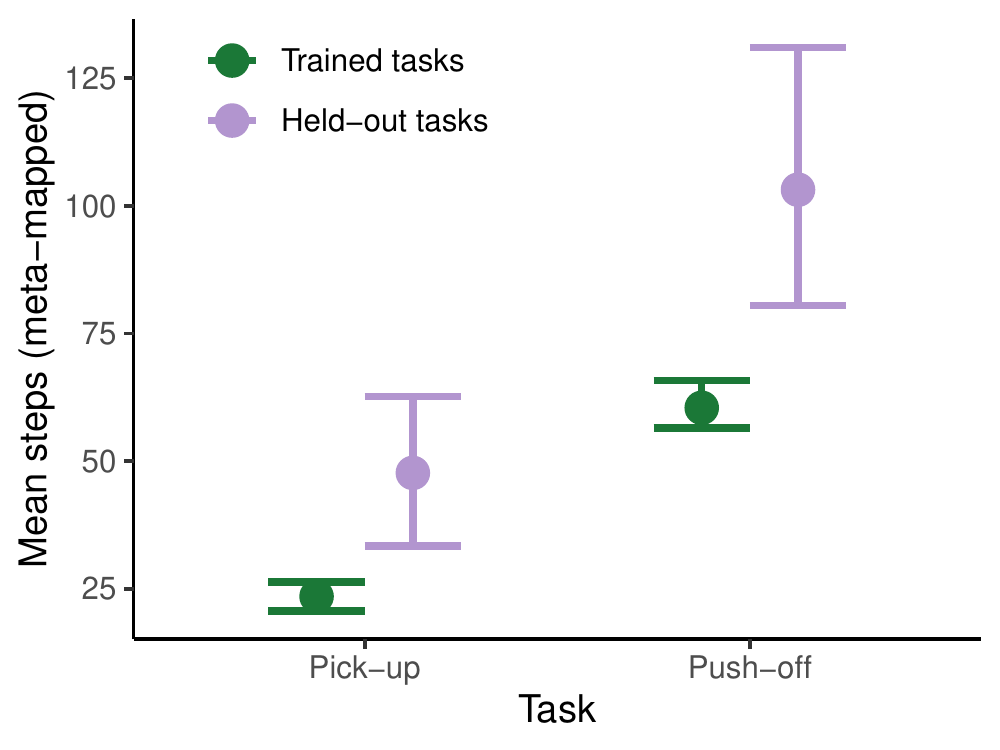}
\caption{Mean step counts.}\label{supp_fig:HoMM_RL_behavioral_uncertainty:main}
\end{subfigure}%
\begin{subfigure}{0.5\textwidth}
\centering
\includegraphics[width=\textwidth]{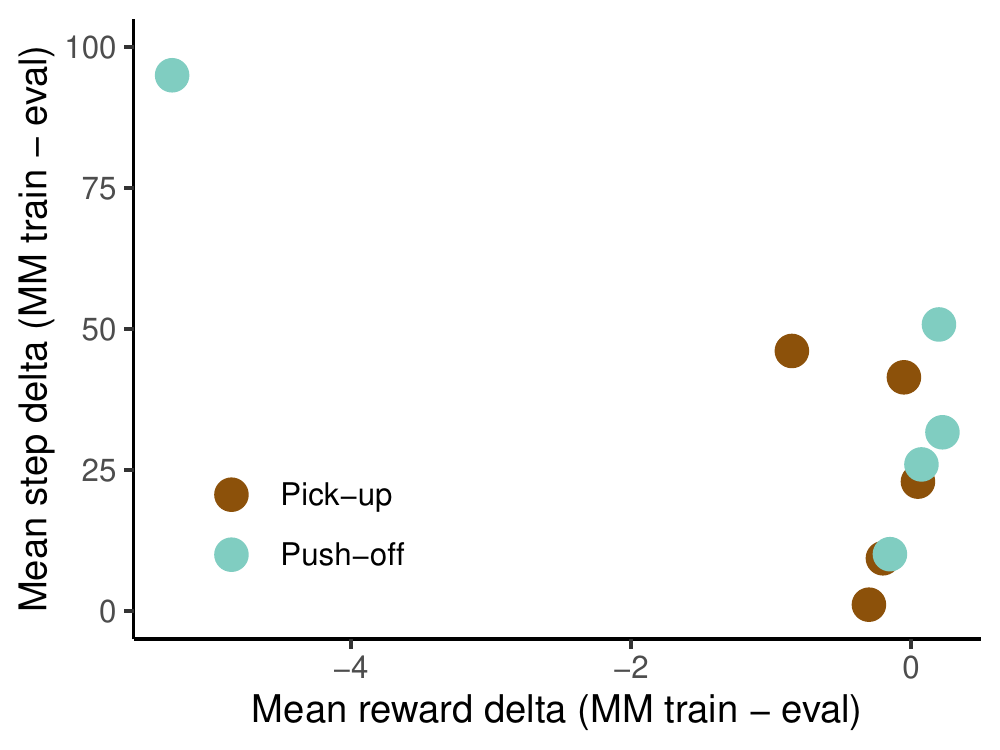}
\caption{Differences in steps vs. differences in rewards.}\label{supp_fig:HoMM_RL_behavioral_uncertainty:deltas}
\end{subfigure}
\caption[The model exhibits behavioral uncertainty in meta-mapping generalization on the RL tasks.]{The model exhibits behavioral uncertainty in meta-mapping generalization on the RL tasks, measured by the steps taken to complete each episode. (\subref{supp_fig:HoMM_RL_behavioral_uncertainty:main}) The model takes more steps to complete episodes from the held-out tasks via a meta-mapping than to complete episodes from tasks used as training targets for the meta-mapping. That is, it appears to be more uncertain about its behavior on the generalization tasks. (\subref{supp_fig:HoMM_RL_behavioral_uncertainty:deltas}) The behavioral uncertainty effect is not solely driven by the model performing more poorly overall; even on the runs where it performs well, it is almost always taking longer to complete the episodes from the tasks it has never seen before. To show this, we plot the difference in average steps vs. difference in average rewards between train and eval. Note that the step difference is almost always positive (evaluation tasks are slower), even where rewards are comparable. (Panel \subref{supp_fig:HoMM_RL_behavioral_uncertainty:main}: means and bootstrap 95\%-CIs across 5 runs. Panel \subref{supp_fig:HoMM_RL_behavioral_uncertainty:deltas}: each point is one game type within one run.)} \label{supp_fig:HoMM_RL_behavioral_uncertainty}
\end{figure}

\textbf{Behavioral uncertainty in generalization:} In Fig. \ref{supp_fig:HoMM_RL_behavioral_uncertainty} we show intriguing behavioral uncertainty in generalization, where the model exhibits more uncertainty (takes longer to solve the task) even when it performs well. Selected recordings of behavior can be found at: \url{https://github.com/lampinen/homm_grids/tree/master/recordings}. 

\FloatBarrier
\subsubsection{Meta-mapping and language} \label{supp_sec:analyses:language}
In this section we show further figures and statistics corresponding for the language comparisons mentioned in the main text, and some supplemental analyses and discussion of the performance of these models.

\begin{figure}[H]
\centering
\includegraphics[width=0.5\textwidth]{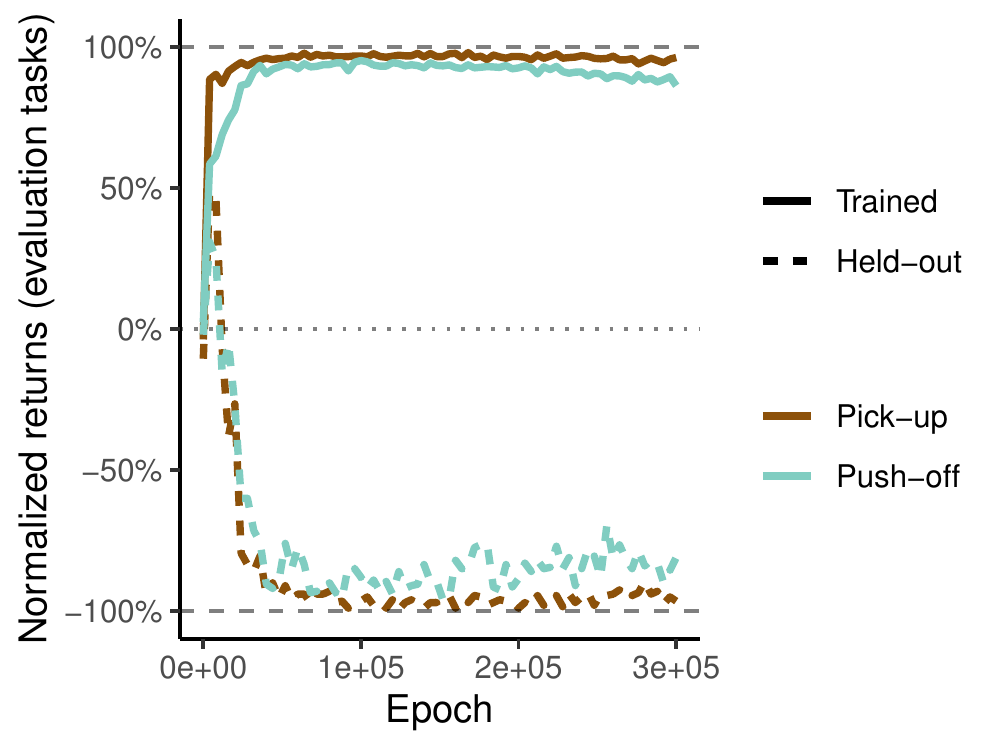}
\caption{Average performance of the language generalization model over training on the RL tasks. The model exhibits intriguing but transient generalization early in learning, before it has understood the full structure of the tasks (especially the more difficult and sequential push-off task), but delays to below-chance generalization as it masters the training tasks.} \label{supp_fig:HoMM_RL_language_learning_curves}
\end{figure}

\textbf{RL:} The language-alone model performs the trained tasks well, but adapts poorly, with generalization performance of -92.8\% (mean, bootstrap 95\%-CI [-96.3, -88.4]) on the pick-up task and -79.7\% (mean, bootstrap 95\%-CI [-92.8, -59.1]) on the pusher task. The difference between the models is significant (\(t (20.6) = -19.515\), \(p < 1\cdot10^{-14}\)) in a mixed linear regression controlling for task type and a random effect of run.\footnote{Degrees of freedom calculated by the Satterthwaite approximation.} Intriguingly, the language model does transiently exhibit slightly positive generalization very early in learning (see Fig. \ref{supp_fig:HoMM_RL_language_learning_curves}), but decays to below chance as the model masters the training tasks. This early generalization is not included in the main results since the train accuracy at this time is below the threshold of having adequately learned the tasks.

By contrast, meta-mapping with task representations constructed from language performs well, with generalization performance of 69.2\% (mean, bootstrap 95\%-CI [49.5, 84.5]) on the pick-up task and 74.9\% (mean, bootstrap 95\%-CI [60.9, 85.5]) on the push-off task. These models were trained separately from the language models whose results are reported below, but the language-alone generalization performance of even the models trained with meta-mapping is poor (respectively -79.6\% [-95.0, -53.8] and -61.0\% [-89.0, -0.195] on the two tasks). That is, meta-mapping at test time is key to generalization. Meta-mapping is not restructuring the basic task representations to allow better generalization from language alone. This is likely due in part to a memory limitation of the models, noted above --- due to GPU memory constraints, meta-mapping training was not able to alter the construction of the basic task representations. If a future implementation of the model allowed this, meta-mapping training might be able to more directly improve basic-task generalization. 

%Meta-mapping from examples also exhibits significantly stronger correlations between its performance on the two tasks than language alone, both within runs at different time-points and across runs (Fig. \ref{supp_fig:HoMM_RL_correlation}, Fig. \ref{supp_fig:HoMM_RL_correlation_details}). Meta-mapping has a correlation of \(r=0.82\) between performance on the two tasks, while the language model only has a correlation \(r=0.10\), and this difference is significant in a mixed linear model predicting push-off performance from pick-up performance, controlling for task type, epoch, and the random effect of run (main effect of meta-mapping model \(t(451.3) = 4.76\), \(p < 1\cdot 10^{-5}\), interaction of meta-mapping model with pick-up performance \(t(452.0) = 3.43\), \(p < 1 \cdot 10^{-3}\)). At a surface level, this means that it is easier to select a good stopping point for the meta-mapping model --- even though the language model is achieving less bad (though still at or below chance) performance at some points in some runs, the lack of correlation between the results on the different tasks means there is no fair way to stop training at that point. More fundamentally, it may suggest that meta-mapping results in more systematic behavior, in the sense that it generalizes well on both tasks, or poorly on both tasks. This may be more like what would be expected from human cognition. 

\begin{figure}[H]
\centering
\includegraphics[width=0.5\textwidth]{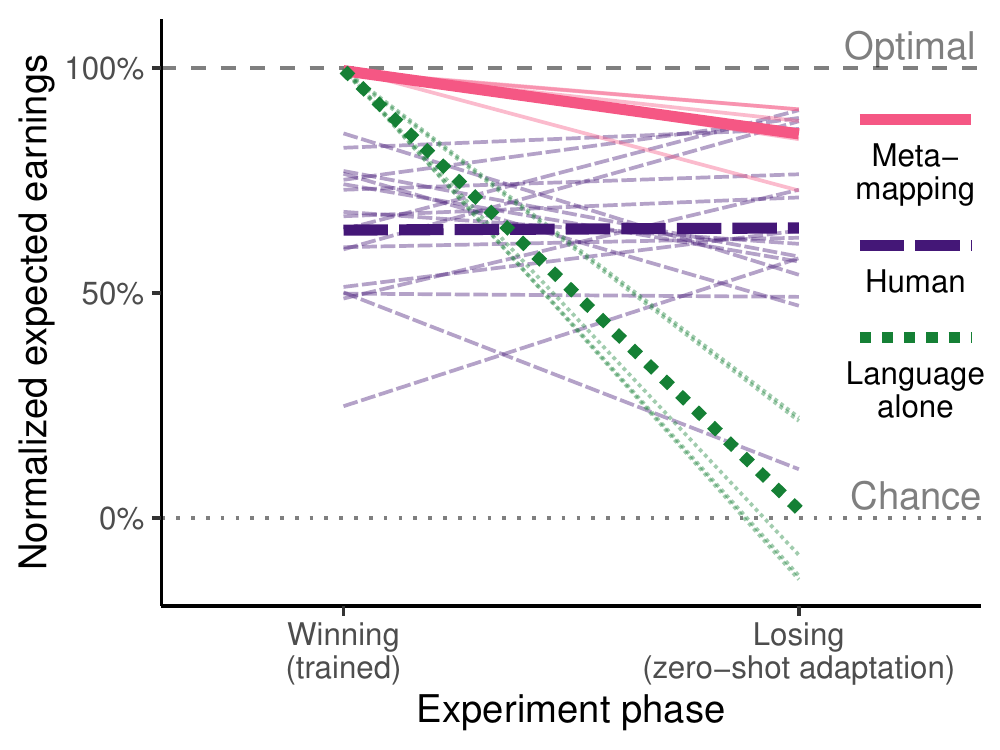}
\caption{Comparing language generalization to meta-mapping and human adaptation in the card games domain. The language-based model performs the trained tasks optimally, but degrades to chance performance on the losing variation. (We plot performance as expected earnings of the actions taken, as a percentage of the earnings of an optimal policy. Thick lines are averages, thin lines are 5 runs of each model, and 19 individual participants who passed attention checks.)} \label{supp_fig:lang_cards}
\end{figure}

\begin{figure}[H]
\centering
\includegraphics[width=0.5\textwidth]{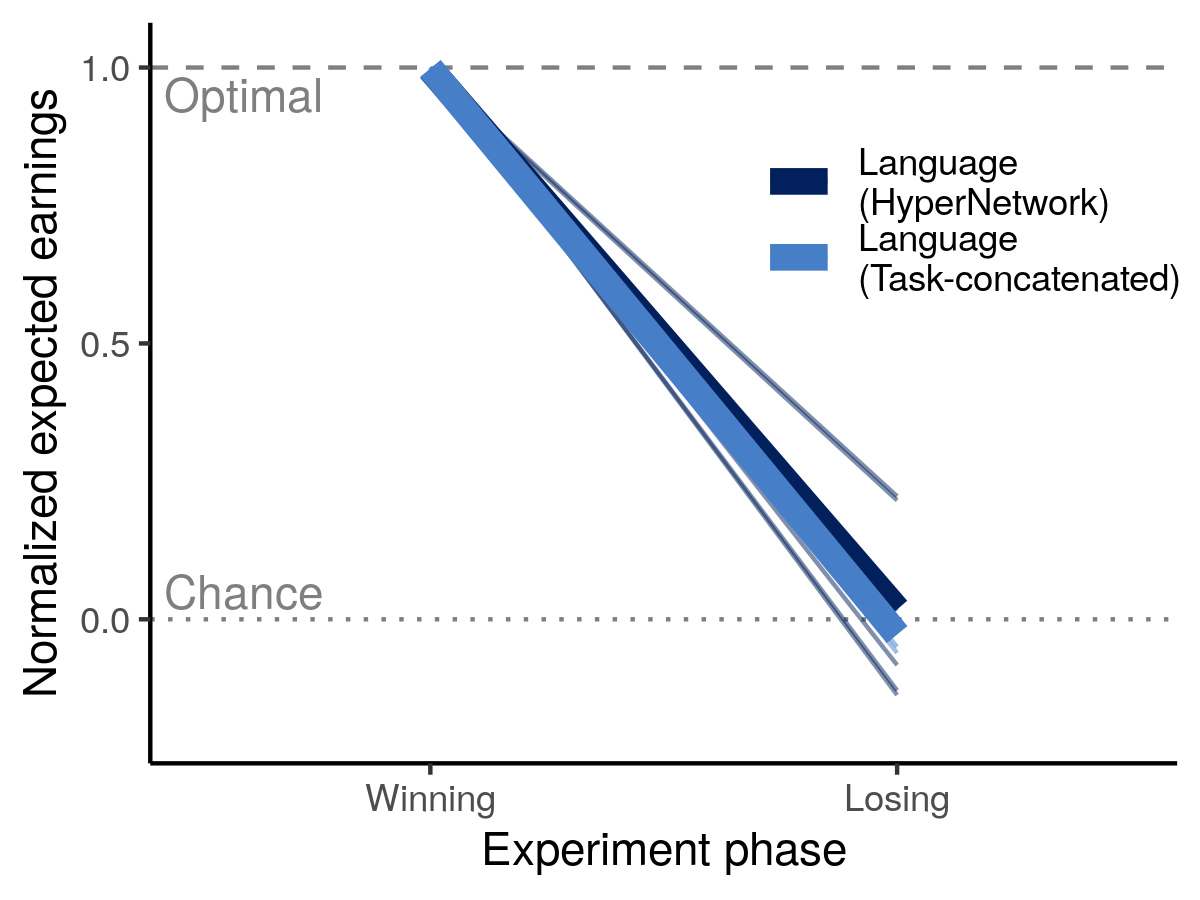}
\caption[Language generalization is similar in the cards domain with either the HyperNetwork architecture used by the meta-mapping appraoch, or a simpler task-concatenated architecture.]{Language generalization is similar in the cards domain with either the HyperNetwork architecture used by the meta-mapping model, or a simpler task-concatenated architecture. See Fig. \ref{supp_fig:HoMM_arch_cond_vs_hyper} above for a similar comparison for meta-mapping itself.} \label{supp_fig:human_cards_lang_tcnh_vs_hyper}
\end{figure}

\textbf{Cards:} The language-alone model performed near-optimally at the trained tasks, but was not able to generalize well to the losing variation from the given dataset (mean performance on losing variation 2\%, bootstrap 95\%-CI \([-12, 16]\)), see Fig. \ref{supp_fig:lang_cards}. Intriguingly, this corresponds to behaving approximately randomly; performance would be worse if the model did not adapt at all. In Fig. \ref{supp_fig:human_cards_lang_tcnh_vs_hyper} we show that the poor language generalization is not simply due to the HyperNetwork architecture, by comparing to a task-concatenated architecture, as we did for meta-mapping in Fig. \ref{supp_fig:HoMM_arch_cond_vs_hyper}.

\begin{figure}[H]
\centering
\includegraphics[width=0.5\textwidth]{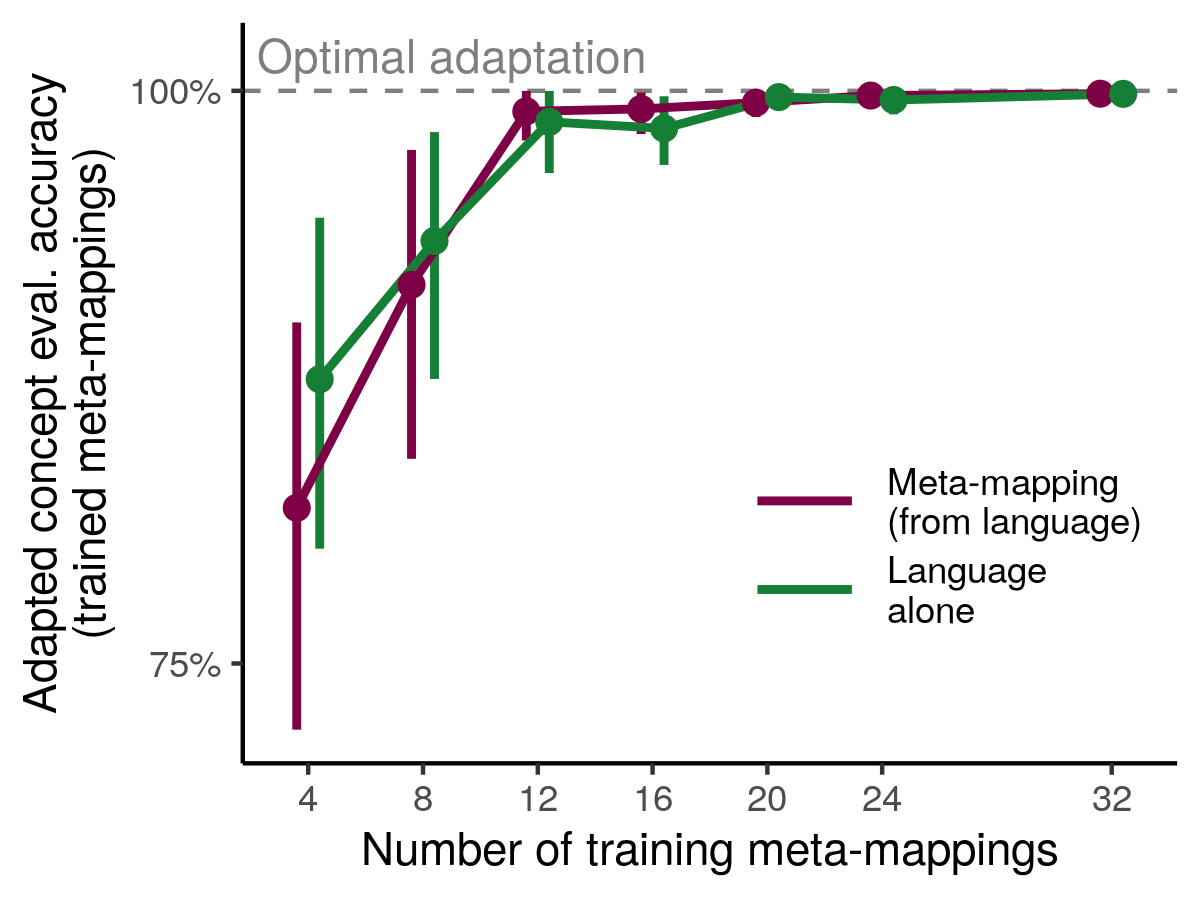}
\caption{Language generalization performs comparably to meta-mapping in the visual concepts domain, across training set sizes. (Results are from 10 runs for each model with each training set size. Errorbars are bootstrap 95\%-CIs across runs.)}\label{supp_fig:lang:visual}
\end{figure}

\begin{figure}[H]
\centering
\includegraphics[width=0.6\textwidth]{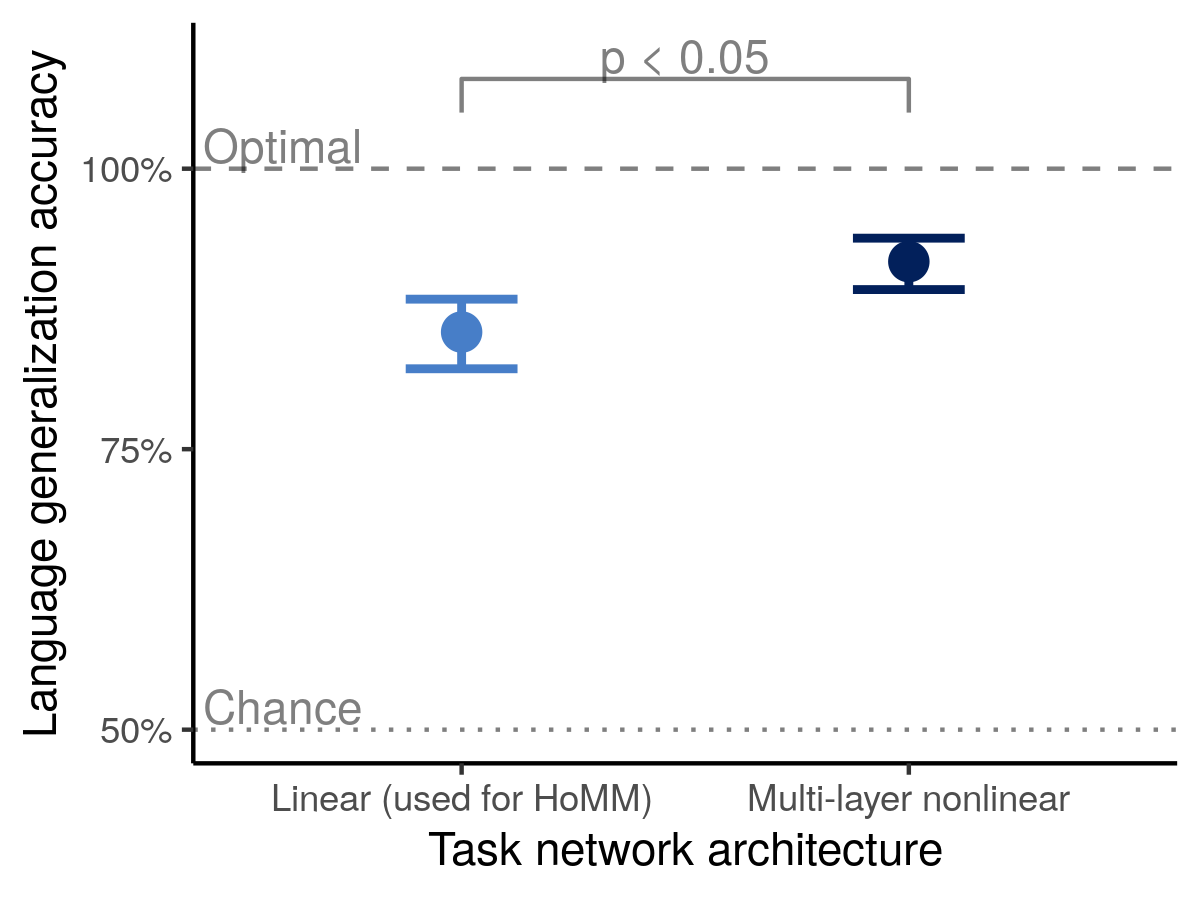}
\caption{Comparing language generalization on the visual concepts tasks between a linear task network architecture and a deep, nonlinear one. The nonlinear task network generalized better to new language instructions (comparisons shown are from the better version).} \label{supp_fig:HoMM_concepts_lang_arch}
\end{figure}

\textbf{Visual concepts:} In this setting the meta-mapping model and the language-alone model perform comparably (Fig. \ref{supp_fig:lang:visual}). %In a mixed linear model, language generalization results in very slightly worse generalization at moderate numbers of training mappings (\(-1.50\)\%, \(t(2612) = -2.775\), \(p =0.006\)), and a small interaction with number of training meta-mappings (\(-0.25\)\% per trained meta-mapping,  \(t(2617) = -4.26\), \(p < 0.001\)). (Effect of one additional trained mapping for HoMM \(1.00\)\%, \(t(6828) = -5.52\), \(p < 0.001\).) The effect is very small, however. 
In Fig. \ref{supp_fig:HoMM_concepts_lang_arch} we show that the language generalization is better with a more complex architecture (deeper \& nonlinear) than we used for the meta-mapping approach. The comparisons in Fig.\ref{supp_fig:lang:visual} use the better-performing architecture for each model. 

\begin{figure}[H]
\centering
\includegraphics[width=0.5\textwidth]{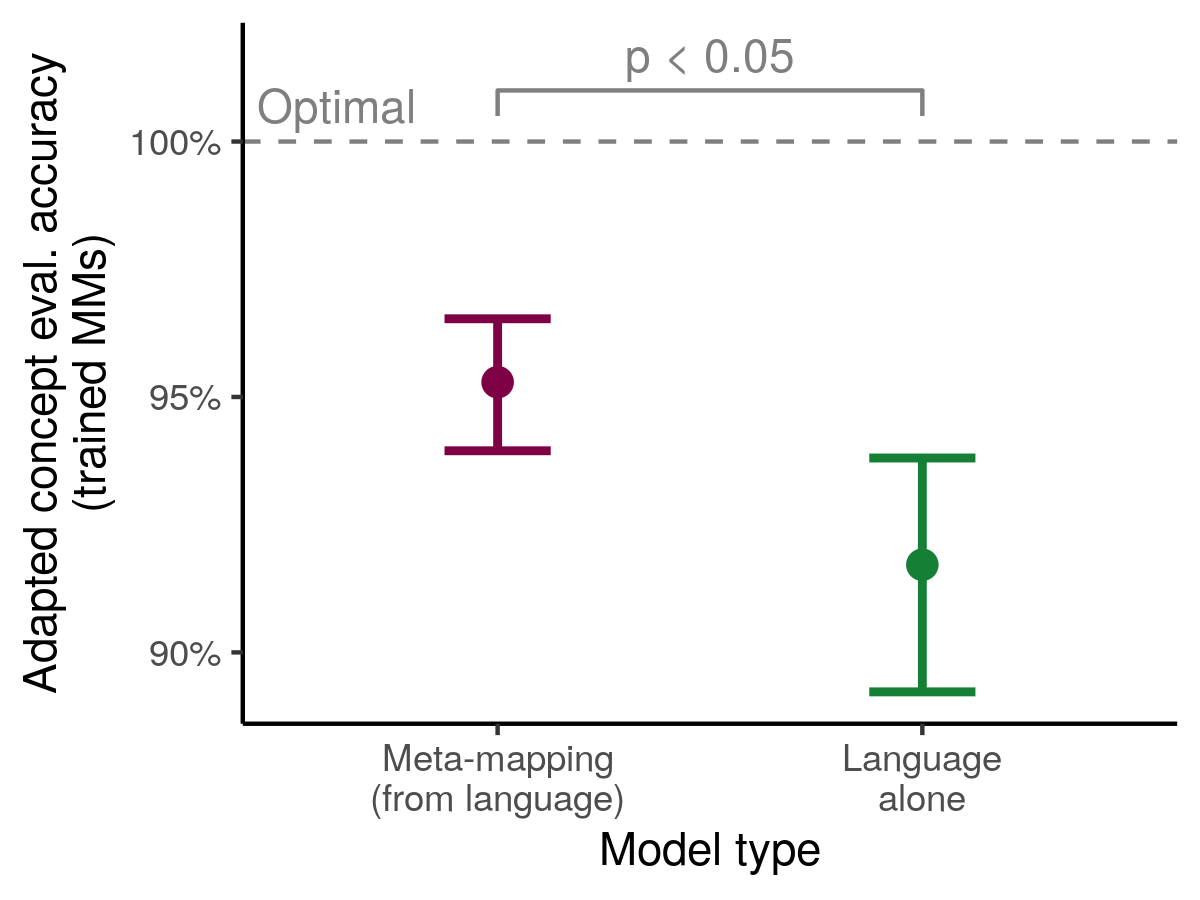}
\caption{Trained meta-mapping results in the visual concepts domain with 150 randomly sampled training concepts, rather than the structured sampling used in the main text. This task sampling scheme means that some evaluation tasks will be farther from the trained tasks. Meta-mapping has a correspondingly larger advantage here. However, the tasks are still likely to be closer to a trained task than in e.g. the RL setting where the evaluation tasks directly contradict the trained ones, and the language model is performing correspondingly better here than on the RL tasks.}\label{supp_fig:HoMM_concepts_random_tasks}
\end{figure}

The comparable performance in this domain may be due in part to the fact that our task sampling guaranteed a training task close to each evaluation task in this setting. This may be because of the structure of the task spaces; there are many more training visual concepts than training tasks in the other domains. Thus, while language-based generalization can be effective, meta-mapping may be especially useful when there are relatively few training tasks --- that is, it may be more sample efficient. However, another factor may be even more critical. The RL and Cards training tasks more directly contradict the evaluation tasks. By contrast, in the visual concepts domain our task sampling guarantees that each held-out concept will have a ``nearby'' training concept, one with the same relation type and same other attribute (see above). With less structured visual concept sampling, meta-mapping's advantage is slightly more clear (Fig. \ref{supp_fig:HoMM_concepts_random_tasks}), even though the meta-mappings have less extensive support sets in that case. 

%\begin{figure}[!h]
%\centering
%\includegraphics[width=0.5\textwidth]{figures/grids_adaptation_correlation_loose.pdf}
%\caption{Correlation of performance on the two hold-out RL tasks, across runs and time points where performance on the training tasks is high. The correlation is much stronger in the meta-mapping model than in the language-generalization model, that is, the meta-mapping model is behaving more systematically in the sense that it is generalizing similarly on both tasks.} \label{supp_fig:HoMM_RL_correlation}
%\end{figure}
%
%\begin{figure}[!h]
%\centering
%\includegraphics[width=\textwidth]{figures/grids_adaptation_correlation_loose_by_run.pdf}
%\caption{Correlation of performance on the two RL tasks, broken down by run. The correlation is higher in the meta-mapping model, both within and across runs.} \label{supp_fig:HoMM_RL_correlation_details}
%\end{figure}

\FloatBarrier
\subsubsection{Generalizing from color to shape in RL} \label{supp_sec:analyses:RL_color_shape}
We next evaluated the generalization capabilities of meta-mapping in a more challenging RL experiment. In this experiment, we trained HoMM on tasks similar to those in the main text experiments, but where the good and bad objects could be discriminated by either color (with shape matched) \textbf{or} shape (with color matched). We trained good-and-bad-switched variations of all color tasks, but did not train any switched variations of the shape-discrimination tasks. Specifically, we used 8 colors, of which we used 4 for the pick-up tasks and 4 for the push-off tasks (so the task type would still be superficially distinguishable. We trained color-discrimination between two pairs of colors in each type, when presented with either both colors appearing on square shapes, or both appearing on diamond shapes. We also trained switched-good-and-bad variations of all those color discrimination tasks. We then trained four shape discrimination tasks for each game type, one in each of that game type's four associated colors. In the shape discrimination tasks, the tee-shaped objects were always good, and triangular objects were always bad. (This results in a total of 24 training tasks, a larger number than were included in the main text experiments.) 

\begin{figure}[h]
\centering
\includegraphics[width=0.5\textwidth]{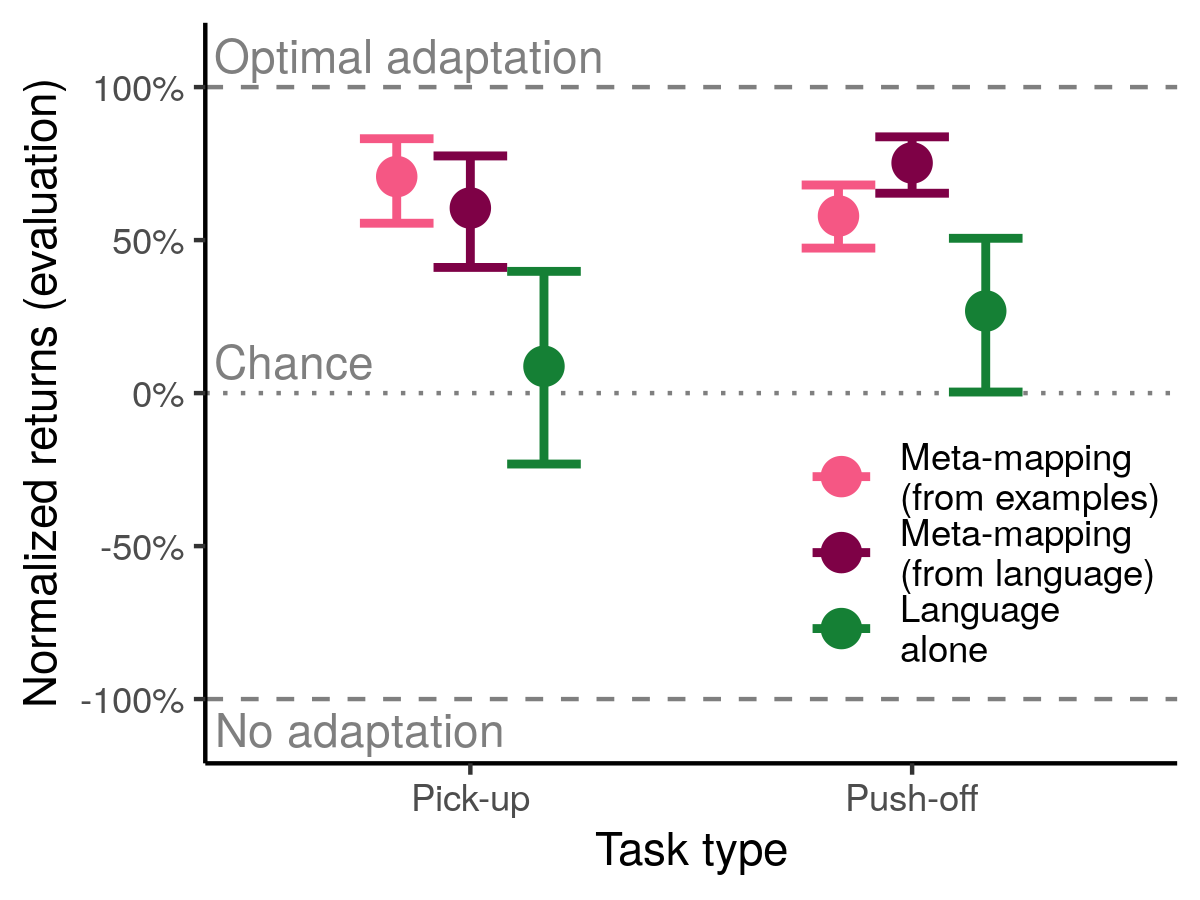}
\caption[HoMM can generalize switching good and bad objects from the color dimension to the shape dimension.]{Meta-mapping can generalize switching good and bad objects from the color dimension to the shape dimension. In this experiment, we trained meta-mapping on tasks similar to those in the main text experiments, but where the good and bad objects could be discriminated by either color (with shape matched) \textbf{or} shape (with color) matched. We trained good-and-bad-switched variations of all color tasks, but did not train any switched variations of the shape-discrimination tasks, to evaluate whether meta-mapping was able to infer how to transfer a mapping from switching colors to switching shapes. Indeed, meta-mapping performs well above chance at this task, though not quite as well as on the simpler generalization in the main text. Intriguingly, the language model also appears to be perfoming somewhat better in this setting, though it is not statistically above chance. (Results from 5 runs, see the text for further details of the experimental setup.)} \label{supp_fig:HoMM:RL:color_to_shape_generalization}
\end{figure}

We trained the ``switch-good-and-bad'' meta-mapping on the color discrimination tasks, and evaluated whether meta-mapping was able to correctly generalize this meta-mapping from switching colors to switching shapes, in order to infer that the triangular objects, which had always been negatively rewarded before, were now beneficial. We found it was useful to increase the initial meta-mapping learning rate to \(3\cdot 10^{-4}\), but otherwise used the same hyperparameters as the main text experiments. See Fig. \ref{supp_fig:HoMM:RL:color_to_shape_generalization} for the results. We found that meta-mapping indeed allowed generalization well above chance. As in the main-text experiments, this is true whether meta-mapping is performed using task and meta-mapping representations constructed from examples (average returns across pick-up and pusher 64.3\% percent of optimal, 95\%-CI [55.1, 72.8]), or task and meta-mapping representations constructed from language (average returns across pick-up and pusher 68.3\% percent of optimal, 95\%-CI [56.6, 78.3]). These experiments show that meta-mapping is able to successfully extrapolate well beyond the training examples of the mapping, to transform behavior along new dimensions.  

Intriguingly, the language-alone baseline model performed less poorly at these experiments than at the main text experiments, although its generalization was not statistically different from chance (average returns 17.8\% of optimal, 95\%-CI [-4.0, 37.4]). Note, however, that there are also 25\% more training tasks in this setting than in the main text experiments. Furthermore, the performance of language alone was still substantially worse than either meta-mapping approach. In a mixed model controlling for game type and its interaction with model and the random effect of run, the difference in performance between meta-mapping from either examples or language and the language-alone performance were both significant (from examples \(t(119.01) = 4.64\), \(p = 8.9 \cdot 10^{-6}\), from language \(t(119.04) = 3.79\), \(p=2.4 \cdot 10^{-4}\)). The effect of game type on generalization in the language model was not significant ( \(t(119.02) = 1.18\), \(p = 0.24\)), nor were the interactions of game-type with either model type (respectively, the interaction of meta-mapping from example by game-type \(t(119.01) = -1.522\), \(p = 0.13\) and from language by game-type \(t(119.03) = -0.08\), \(p = 0.94\)).  

\FloatBarrier
\subsubsection{Meta-mapping as a starting point} \label{supp_sec:analyses:timescales}

\begin{figure}[tbh]
\centering
\includegraphics[width=0.5\textwidth]{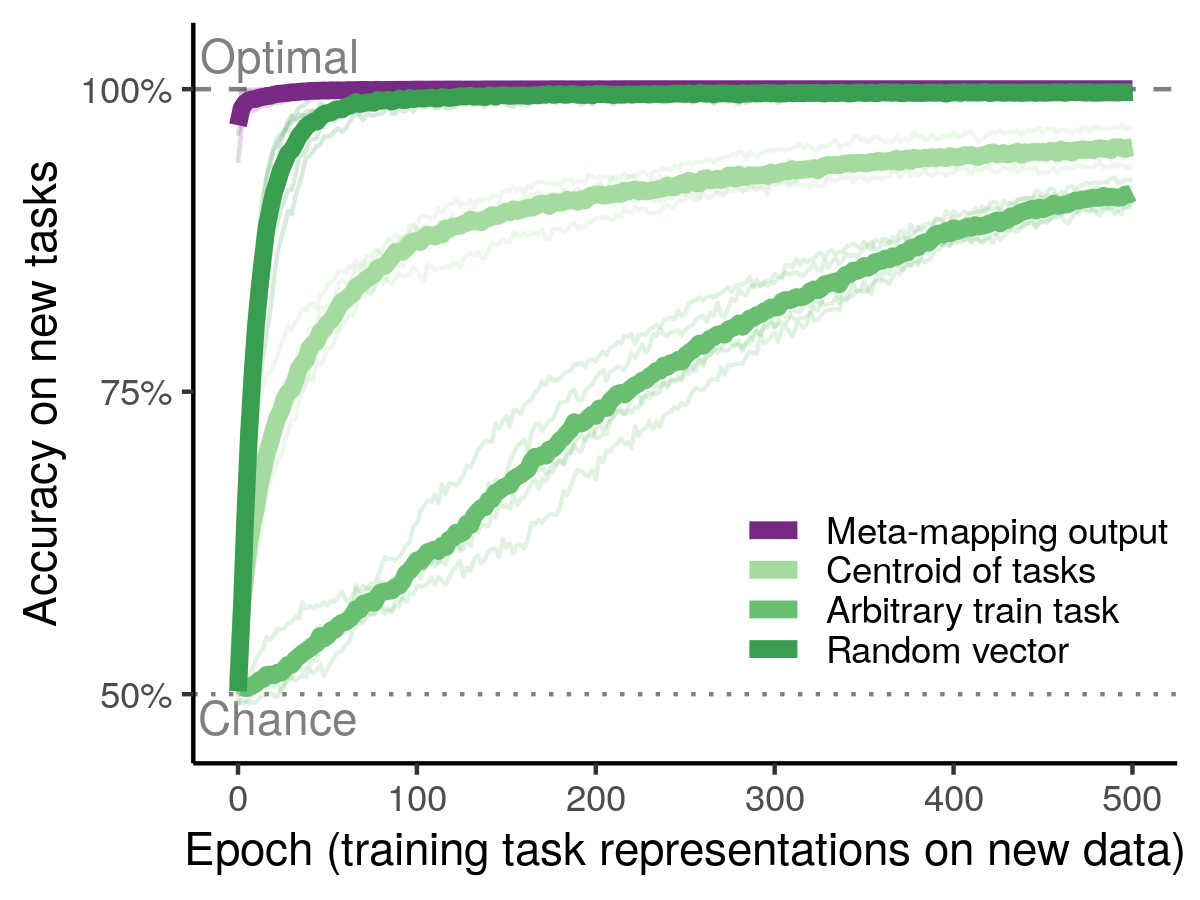}
\caption{Meta-mapping provides a good starting point for later learning in the visual concepts domain. This figure is the visual concepts analog of Fig. \ref{fig:HoMM_timescales_results} in the main text, with 16 training meta-mappings. Using meta-mapping as a starting point offers much lower initial loss, and faster learning than other initializations. (Thick curves are averages over 5 individual runs, shown as light curves.)}\label{supp_fig:optimizing_curves_categories}
\end{figure}

\textbf{Visual concepts:} In Fig. \ref{supp_fig:optimizing_curves_categories} we show that meta-mapping provides a good starting-point for learning in the visual concepts domain as well. In this setting the small random initialization is more competitive, but meta-mapping still yields lower cumulative error over learning than random initialization, and much lower than the centroid (which was better in the polynomials domain). Specifically, initializing with a meta-mapping output results in a mean cumulative error of \(0.33\) (bootstrap 95\%-CI \([0.10, 0.57]\)), while a small random initalization results in a mean cumulative error of \(9.62\) (bootstrap 95\%-CI \([6.63, 13.59]\)). This difference is significant in a mixed linear model (\(t(4) = 4.628\), \(p = 0.01\)).

\textbf{The non-hyper-network architecture makes optimization more difficult:} We have compared our hyper-network-based meta-mapping architecture to the simpler alternative of concatenating a task representation to an input embedding before passing it through a fixed network, in various supplemental analyses (Figs. \ref{supp_fig:HoMM_arch_cond_vs_hyper} and \ref{supp_fig:human_cards_lang_tcnh_vs_hyper}). The hyper network approach generally performs at least as well as, and sometimes substantially better than, the simpler approach. Hyper networks may also be particularly beneficial for continual learning \citep{Oswald2020}. Furthermore, they may also make it easier to optimize the task representation, by giving it more direct control over the computations of the network. Thus, it seems useful to compare these two architectures in this setting. 

\begin{figure}[htb]
\centering
\includegraphics[width=0.5\textwidth]{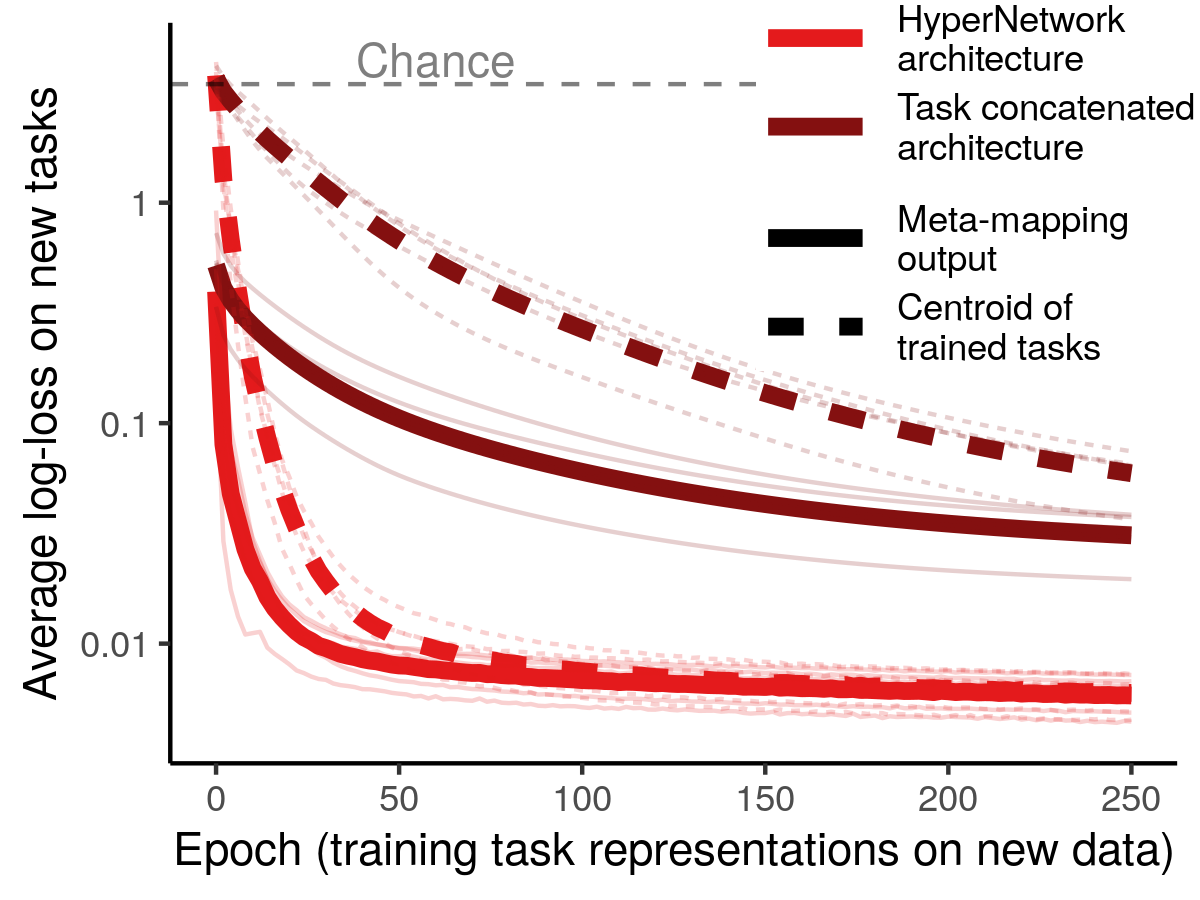}
\caption[Comparing the learning curves of the hyper network architecture and a simpler architecture when optimizing the task representations for new polynomials.]{Comparing the learning curves of the hyper network architecture and a simpler architecture when optimizing the task representations for new polynomials. The simpler architecture improves much more slowly, and appears to plateau at a higher loss. (Note that the y-axis is log-scale. Results are from 5 runs, individual runs are shown as light curves.)} \label{supp_fig:timescales_polynomial_optimization_tcnh_curves}
\end{figure}

We therefore performed the polynomial domain experiments, reported in the main text in the meta-mapping as a starting point section, with the simpler task-network architecture as well. In Fig. \ref{supp_fig:timescales_polynomial_optimization_tcnh_curves}, we show the learning curves for both architectures for the two best initializations (meta-mapping output, and centroid of the trained task representations). The hyper-network architecture learns much more rapidly than the simpler architecture. The initial meta-mapping outputs do not differ so substantially --- most of this effect is due to the slower improvement of the loss when optimizing the task representation in the non-hyper architecture. Indeed, optimization in the non-hyper network architecture appears to be plateauing at a much higher loss value than in the hyper-network architecture. 

\begin{figure}[H]
\centering
\includegraphics[width=0.5\textwidth]{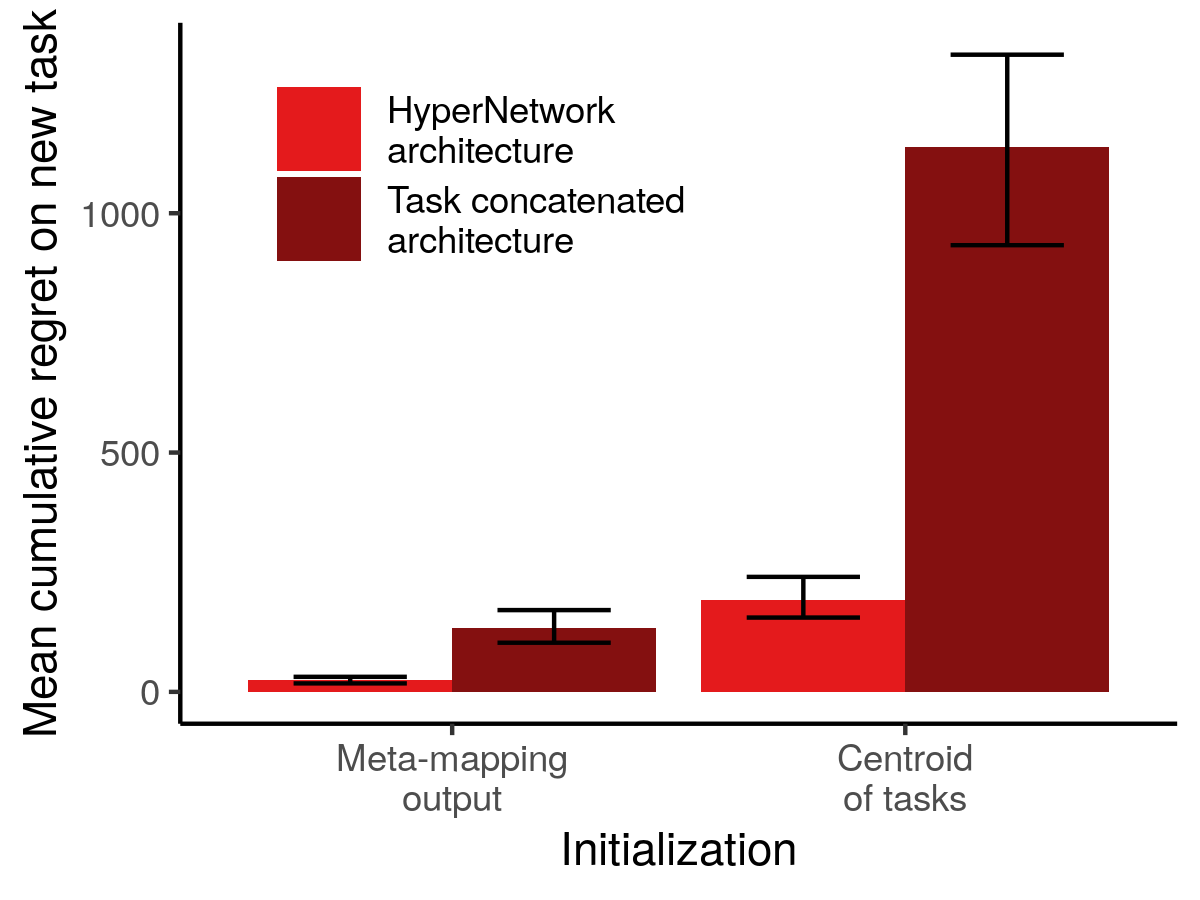}
\caption[Comparing the cumulative losses of the hyper-network architecture and a simpler architecture when optimizing the task representations for new polynomials.]{Comparing the cumulative losses of the hyper-network architecture and a simpler architecture when optimizing the task representations for new polynomials, starting from either the result of a meta-mapping or the centroid of the trained tasks. The simpler architecture results in substantially more cumulative loss. (Results from 5 runs, errorbars are bootstrap 95\%-CIs.)} \label{supp_fig:timescales_polynomial_optimization_tcnh_regret}
\end{figure}

As before, we quantify this by plotting the cumulative loss on the novel tasks in Fig. \ref{supp_fig:timescales_polynomial_optimization_tcnh_regret}. The simpler non-hyper architecture resulted in about five times greater cumulative loss than the hyper network architecture when starting from the meta-mapping output (mean \(= 133.81\), bootstrap 95\%-CI \([102.65, 171.10]\)), and similarly from the centroid of the trained task representations (mean \(= 1139.35\), bootstrap 95\%-CI \([943.60, 1344.52]\)). We therefore conclude that hyper-network-based architectures may be particularly conducive to this perspective on continual learning.

\FloatBarrier
%\clearpage
\subsubsection{Default processing \& cognitive control} \label{supp:HoMM_cognitive_control}
Our architecture could be of interest to researchers in cognitive control, even beyond the idea of meta-mapping as adaptation. The system can perform different tasks based on task examples or language inputs, which is fundamentally the same problems human face when we must adapt our behavior. There are a number of features of the model that offer the opportunity for intriguing investigations based on this idea. For example, the task network in our architecture has a default set of bias weights that are modulated by the HyperNetwork. These can be thought of as the ``automatic'' or ``default'' processing habits of the system, whereas the weight alterations the HyperNetwork imposes can be thought of as the exertion of cognitive control to modulate behavior.  

To explore this, we trained our architecture on a very simple stroop task taken from Cohen et al. \citep{Cohen1990}. The model receives two sets of two inputs, that can be thought of as corresponding to ``word'' and ``color'' domains. One input in each domain is turned on, representing a color word written in a color. The model's task is to report either the color or the word, depending on context.

The context we give the model is in the form of examples of the task as (input, output) pairs. These are used to construct a task representation, which is then used to modulate the parameters in the task network, via the HyperNetwork. We trained the model repeatedly with different proportions of training on the word task vs. the color task, in order to investigate the default vs. controlled behavior in different training regimes. Specifically, we compared training the model to the point that it barely mastered the less frequent task (when it first achieves 100\% performance and cross-entropy loss \(< 0.3\) on both tasks) to the point that it had mastered both tasks (100\% performance and cross-entropy loss \(<0.01\) on both). We then tested the model's default behavior by giving it an all-zeros task representation, and seeing whether its performance was more aligned with the ``word'' or ``color'' task. 

\begin{figure}[htb]
\centering
\includegraphics[width=0.5\textwidth]{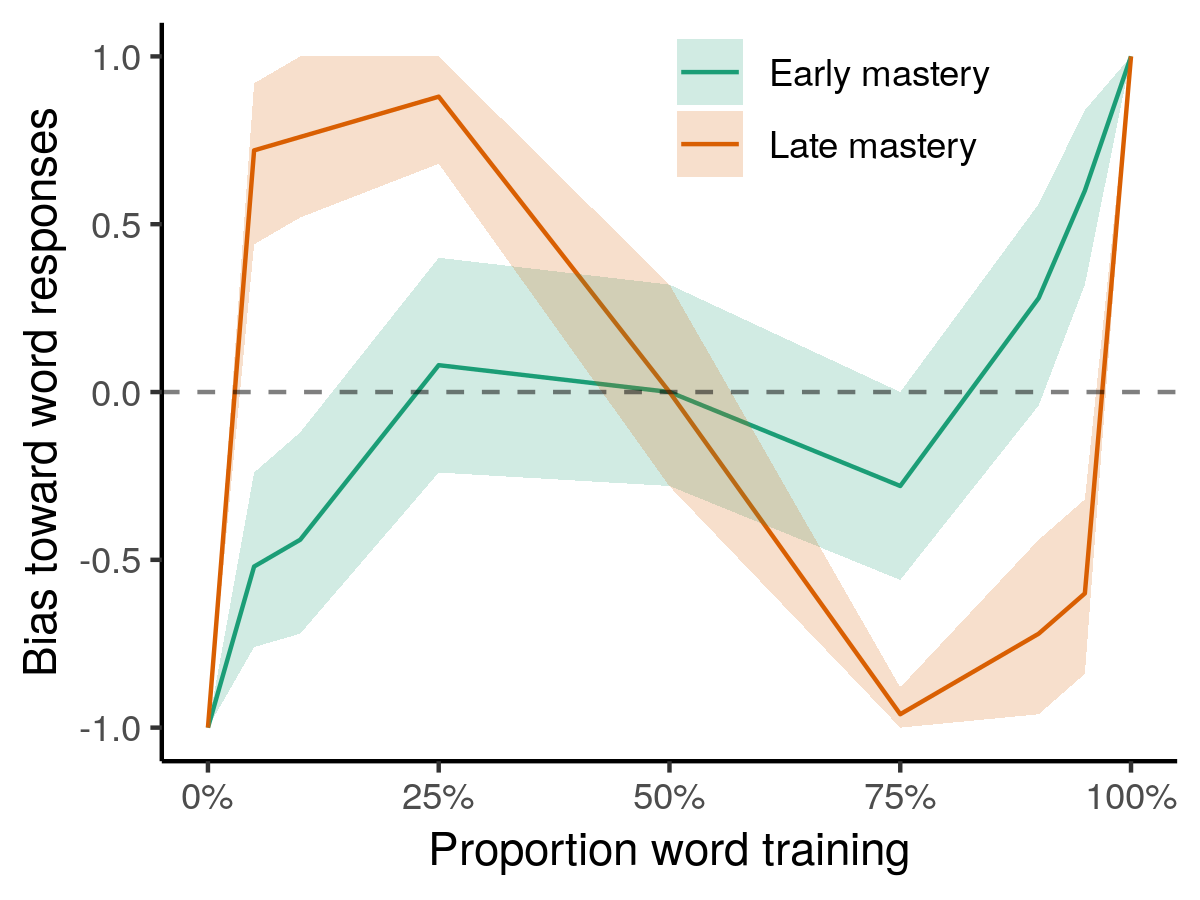}
\caption{Measuring the default behavior of our architecture on a Stroop-like task. We plot the bias of the model towards word or color responses, when given an all-zeros task representation, at different proportions of training on words or colors, and different stages of training. When the model has just mastered the less frequent task, it exhibits a default bias towards the more frequent task. However, later in training, when it has mastered both tasks, it exhibits a paradoxical bias towards the \emph{less} frequent task.} \label{supp_fig:HoMM_cognitive_control}
\end{figure}

In Fig. \ref{supp_fig:HoMM_cognitive_control}, we show the results. We plot the bias as \(2 \times (\text{word accuracy} - \text{color accuracy})\), which is \(-1\) if the model is responding only to color, 1 if the model is responding perfectly to word, and 0 if it is responding equally to each (or otherwise responding randomly). When the model has just barely mastered the less-frequent task, it exhibits a default bias towards the more frequent task. However, once we train it to full master of both tasks, it exhibits a surprising paradoxical bias towards the task that was mastered more recently. This may relate to observations that switching from a less-practiced task back to a more practiced one is difficult \citep{Monsell2003}, possibly because performing the less-practiced task requires strong suppression of the default behavior. It's possible that in the course of achieving full mastery on the less-practiced task, the more practiced task must be so suppressed that it fades away from being the default. These phenomena provide possible inspiration for future investigations in cognitive control. 

For this experiment, we used similar hyperparameters to the polynomials experiments, except we used a much smaller model --- a single-layer task network, a $Z$-dimensionality of 8, and $\mathcal{H}, \mathcal{E}$ had 64 hidden units per layer. We optimized the model via stochastic gradient descent with a learning rate of \(0.01\) to follow more closely the approach taken by Cohen et al., although results are similar with other optimizers.

%\newpage
\FloatBarrier
%\newpage
\subsection{Proofs} \label{supp_sec:proofs}

\subsubsection{Inadequacy of vector analogies for meta-mapping polynomials} \label{supp:HoMM_vector_analogies_inadequate}

One possible implementation of meta-mapping would be to just construct an analogy vector and use that for the mapping. This is motivated by work showing that word vector representations often support vector analogical reasoning, for example if we denote the vector for the word king as \(\vec{v}_{king}\), relationships like \(\vec{v}_{queen} \approx \vec{v}_{king} + \left(\vec{v}_{man} - \vec{v}_{woman} \right)\) often hold \citep{Mikolov2013}. Thus, a plausible approach to meta-mapping would be to take a similar approach, for example in the polynomials domain, the meta-mapping ``Permute \((w, z, x, y)\)'' could be estimated by taking the vector differences between the representations of inputs and targets, computing an average difference vector, and adding that to the held-out examples to produce an output for each one. In this section, we prove that such an approach cannot accurately represent all the meta-mappings in the polynomials domain. Furthermore, we sketch a proof by construction that the linear task network (i.e. an affine transformation, matrix multiplication plus a bias vector) we used in this domain suffices, if it is parameterized separately for each meta-mapping. 

\textbf{Proof that vector analogies are inadequate:} In essence, the proof is simply that many of our meta-mappings are non-commutative, while vector addition is commutative. Consider the mappings for adding 1 to a polynomial, and multiplying by 2. Assume there were vector representations for these mappings, respectively \(\vec{m}_{+1}\) and \(\vec{m}_{\times 2}\). Let \(\vec{f}_{x}\) be the representation for the polynomial \(f(w,x,y,z) = x\). Then \(\vec{f}_{x} + \vec{m}_{+1} = \vec{f}_{x+1}\), \(\vec{f}_{x} + \vec{m}_{\times 2} = \vec{f}_{2x}\). But then:
\[ \vec{f}_{2(x + 1)} = \left(\vec{f}_{x} + \vec{m}_{+1}\right) + \vec{m}_{\times 2} = \vec{f}_{x} + \vec{m}_{+1} + \vec{m}_{\times 2} = \left(\vec{f}_{x} + \vec{m}_{\times 2}\right) + \vec{m}_{+1} = \vec{f}_{2x + 1}\]
Thus such a representation would result in contradictions, such as \(2x + 1 = 2x + 2\). Similar issues occur for input permutation and other non-commutative mappings. 

\textbf{Proof sketch that affine transformations in an appropriate vector space suffice:} Suppose that we have a vector representation for the polynomials, where there is a basis dimension corresponding to each monomial, so that the polynomial can be represented as a vector of its coefficients. (This is the standard vector-space representation for polynomials.) Then permutation corresponds to permuting these monomials, i.e. a permutation of the basis dimensions, which is a linear transformation. Adding a constant corresponds to adding to one dimension, which requires only the vector addition part of the affine transformation. Multiplying by a constant requires multiplying each dimension, i.e. a block-diagonal linear transformation. 

Squaring polynomials is slightly more complex, and requires augmenting the vector space with components whose values are the product of the coefficients of each pair of monomials. In this case, squaring corresponds to a simple linear transformation. However, this augmentation makes the other meta-mappings more complex. Surprisingly, the most complex case in this representational scheme is adding a constant, which requires shifting each pair term containing a constant by the product of the constant and the coefficient of the other monomial, but this again reduces to simply an appropriately parameterized affine transformation --- each pair term containing a constant term simply needs the added constant (from the meta-mapping) as a weight times the component for the other monomial. Thus affine transformations suffice in this setting.

Of course, with a sufficiently complex, deep, recurrent, and non-linear task network, any meta-mapping could be computed in principle, since a sufficiently large such network is Turing-complete \citep{Siegelman1992}. Thus, our approach to meta-mapping is fully general, conditioned on a sufficiently complex task network, while simpler approaches may not be.

%\bibliography{arrr}
%\end{document}

\end{document}